\pdfoutput=1
\documentclass[fleqn,10pt]{wlscirep}
\usepackage[utf8]{inputenc}
\usepackage[T1]{fontenc}
\usepackage{subcaption}
\usepackage[inkscapepath=svgpath]{svg}
\usepackage{multirow}
\usepackage{subcaption}
\usepackage[resetlabels]{multibib}
\usepackage{appendix}
\usepackage{graphicx}
\usepackage{xr}
\usepackage{hyperref}
\newcites{main}{References}

\title{Re-identification of Individuals in Genomic Datasets Using Public Face Images}

\author[1,*]{Rajagopal Venkatesaramani}
\author[2,3,4]{Bradley A. Malin}
\author[1]{Yevgeniy Vorobeychik}
\affil[1]{Washington University in St. Louis, Department of Computer Science and Engineering, St. Louis, 63130, USA}
\affil[2]{Vanderbilt University, Department of Electrical Engineering and Computer Science, Nashville, 37232, USA}
\affil[3]{Vanderbilt University Medical Center, Department of Biomedical Informatics, Nashville, 37203, USA}
\affil[4]{Vanderbilt University Medical Center, Department of Biostatistics, Nashville, 37203, USA}

\affil[*]{rajagopal@wustl.edu}


\begin{document}
\begin{abstract}

  DNA sequencing is becoming increasingly commonplace, both in medical and direct-to-consumer settings.
  To promote discovery, collected genomic data is often de-identified and shared, either in public repositories, such as OpenSNP, or with researchers through access-controlled repositories.
  However, recent studies have suggested that genomic data can be
  effectively matched to high-resolution three-dimensional face images, which raises
  a concern that the increasingly ubiquitous public face images can be
  linked to shared genomic data, thereby re-identifying individuals in
  the genomic data. While these investigations illustrate the possibility of such an attack, they assume that those performing the linkage have access to extremely well-curated data. Given that this is unlikely to be the case in practice, it calls into question the pragmatic nature of the attack. As such, we systematically study this re-identification risk from two
  perspectives: first, we investigate how successful such linkage
  attacks can be when real face images are used, and second, we consider how we
  can empower individuals to have better control over the associated
  re-identification risk.
  We observe that the true risk of re-identification is likely substantially smaller
  for most individuals than prior literature suggests.
  In addition, we demonstrate that the addition of a small amount of
  carefully crafted noise to images can enable a controlled trade-off
  between re-identification success and the quality of shared images,
  with risk typically significantly lowered even with noise that is imperceptible to humans.
\end{abstract}

\flushbottom
\maketitle
\thispagestyle{empty}

\section*{Introduction}

Direct-to-consumer DNA‌ testing has made it possible for people to gain information about their ancestry, traits and susceptibility to various health conditions and diseases. The simplicity of testing services by companies like 23andMe, AncestryDNA and FamilyTree DNA has drawn a consumer base of tens of millions of individuals.
These sequenced genomes are of great use to the medical research community, providing more data for genome-phenome association studies‌, aiding in early disease diagnoses, and personalized medicine.

While genome sequencing data gathered in medical settings is anonymized and its use often restricted, individuals may also choose to share their sequenced genomes in the public domain via services like OpenSNP\citemain{greshake2014opensnp} and Personal Genome Project \citemain{ball2014harvard}.
Moreover, even the sharing of de-identified data for medical research typically faces tension between open sharing within the research community and exposure to privacy risks.
These risks generally stem from the ability of some data recipients to link the genomic data to the identities of the corresponding individuals.
One particularly acute concern raised in recent literature is in the ability to link a genome to the photograph of an individual's face~\citemain{lippert2017identification, crouch2018genetics, qiao2016detecting, caliebe2016more}.
Specifically, these studies have shown that one can effectively match high-quality three-dimensional face maps of individuals with their associated low-noise sequencing data.
However, for a number of reasons, it is unclear whether these demonstrations translate into practical privacy concerns.
First, these studies to date have relied on high-quality, and often proprietary data that is not publicly available. This is a concern because such high-quality data is in fact, quite difficult to obtain in practice. While many people post images of their face in public, these are generally two-dimensional, with varying degrees of quality.
In addition, observed phenotypes in real photographs need not match actual phenotypes thereby making it challenging to correctly infer one's genotype and vice versa. For example, people may color their hair, or eyes (through contact lenses). Finally, increasing population size poses a considerable challenge to the performance of genome-photograph linkage. Given a targeted individual and a fixed feature-space (namely the predicted phenotypes in our case), the chances of encountering an individual that are similar to the target individual in this feature-space increase with population size. 
Another related study by Humbert et al. \citemain{humbert2015anonymizing} investigates the re-identification risk of OpenSNP data, but assumes accurate knowledge of a collection of phenotypes, including many that are not observable from photographs, such as asthma and lactose intolerance. We consider this approach to be a theoretical upper-bound in our study, that is, matching performance when ground-truth phenotypes are known apriori, as opposed to when predicted from face images.

Given these potential confounders in the real world, in this paper we study the risk of re-identification of shared genomic data when it can potentially be linked to publicly posted face images. To this end, we use the OpenSNP~\citemain{greshake2014opensnp} database, along with a new dataset of face images collected from an online setting and paired with a select subset of 126 genomes.
We develop a re-identification method that integrates deep neural networks for face-to-phenotype prediction (e.g., eye color) with probabilistic information about the relationship between these phenotypes and SNPs to score potential image-genome matches.
The first purpose of our study is to assess how significant the \emph{average} risk is, as a function of population size, given the nature of available data as well as current technology.
Our second purpose is to introduce a practical tool to manage \emph{individual} risk that enables either those who post face images online, or social media platforms that manage this data, to trade off risk and utility from posted images according to their preferences.
We find that the overall effectiveness of re-identification and, thus, privacy risk is substantially lower than suggested by the current literature that relies upon high-quality single nucleotide polymorphisms (SNP) and three-dimensional face map data.
While some of this discrepancy can be attributed to the difficulty of inferring certain phenotypes---eye color, in particular---from images, we also observe that the risk is relatively low, especially in larger populations, even when we \emph{know} the true phenotypes that can be observed from commonly posted face images.
Indeed, even using synthetically generated data that makes optimistic assumptions about the nature of SNP to phenotype relationships, we find that the average re-identification success rate is relatively low.

For our second contribution, we propose a method based on adding adversarial perturbations to face images prior to posting them that aims to minimize the score of the correct match.
This framework is tunable in the sense that the user can specify the amount of noise they can tolerate, with greater noise added having greater deliterious effect on re-identification success.
We show that even using imperceptible noise we can often successfully reduce privacy risk, even if we specifically train deep neural networks to be robust to such noise.
Furthermore, adding noise that is mildly perceptible further reduces success rate of re-identification to no better than random guessing.

 \section*{Results}

We investigate the risk of re-identification in genomic datasets "in the wild" based on linkage with publicly posted photos. Using the public OpenSNP dataset, we identified 126 individual genotypes for which we were able to successfully find publicly posted photographs (e.g., some were posted along with genomic data on OpenSNP itself). We used a holistic approach to associate genomes to images as follows. If a user's picture was posted on OpenSNP, higher-quality pictures could often be found under the same username on a different website. When no picture was posted for a certain user on OpenSNP, we found pictures posted on different websites under the same username, and used self-reported phenotypes on OpenSNP to ensure with a reasonable degree of certainty that the image corresponds to the genome.
This resulted in a dataset of SNPs with the corresponding photos of individuals, which we refer to as the \emph{Real} dataset.
To characterize the error rate in phenotype prediction from images, we constructed two synthetic datasets, leveraging a subset of the CelebA face image dataset~\citemain{liu2015faceattributes}, and OpenSNP. We created artificial genotypes for each image (here, genotype refers only to the small subset of SNPs we are interested in - we refer the reader to Supplementary Table~\ref{snps} for a full list) using all available data from OpenSNP where self-reported phenotypes are present. First, we consider an ideal setting where for each individual, we select a genotype from the OpenSNP dataset that corresponds to an individual with the same phenotypes, such that the probability of the selected phenotypes is maximized, given the genotype. In other words, we pick the genotype from the OpenSNP data that is most representative of an individual with a given set of phenotypes.
We refer to this dataset as \emph{Synthetic-Ideal}.
Second, we consider a more realistic scenario where for each individual we select a genotype from the OpenSNP dataset that also corresponds to an individual with the same phenotypes, but this time at random according to the empirical distribution of phenotypes for particular SNPs in our data. Since CelebA does not have labels for all considered phenotypes, 1000 images from this dataset were manually labeled by one of the authors, the results of which were confirmed by another of the authors. After cleaning and removing ambiguous cases, the resulting dataset consisted of 456 records.
We refer to this dataset as \emph{Synthetic-Realistic}.

Our re-identification method works as follows.
First, we learned deep neural network models to predict visible phenotypes from face images, leveraging the CelebA public face image dataset, in the form of 1) sex, 2) hair color, 3) eye color, and 4) skin color.
We learned a model separately for each phenotype by fine-tuning the VGGFace architecture for face classification \citemain{parkhi2015deep}.
The result of each such model is a predicted probability distribution over phenotypes for an input face image.
Second, for each input face image $x_i$, and for each phenotype $p$, we use the associated deep neural network to predict the phenotype $z_{i,p}$, that is the most likely phenotype in the predicted distribution.
Third, we assign a score, based on the log-likelihood, to each genotype-image pair, $(x_i,y_j)$ as follows:
\begin{equation}
\label{E:pred}
p_{i,j} = \sum_{z_{i,p}}^{p \in \{sex,hair,skin,eye\}} \log P(z_{i,p} | y_j )
\end{equation}
This approach is similar to the one introduced by Humbert et al.~\citemain{humbert2015anonymizing}, but differs in that we predict phenotypes from face images as opposed to assuming complete knowledge.
Finally, armed with the predicted log-likelihood scores $p_{i,j}$ for genotype-image pairs, we select the top-$k$-scored genotypes for each face image, where $k$ is a tunable parameter that allows for a trade-off in theprecision and recall of predictions.

The effectiveness of re-identification is strongly related to both the choice of $k$ above, as well as the size of the population that one is trying to match against.
More specifically, as we increase $k$, one would naturally expect recall (and, thus, the number of successful re-identifications) to increase.
On the other hand, a larger population raises the difficulty of the task by increasing the likelihood of spurious matches.
We therefore evaluate the impact of both of these factors empirically.

\subsection*{Average Re-identification Risk is Low in Practice}

We evaluate the effectiveness of re-identification attacks using two complementary measures: 1) the fraction of successful matches and 2) the area under the receiver operating characteristic (ROC) curve (AUC).
The former enables us to study re-identification success (while focusing on recall) as a function of population size, while the latter paints a more complete picture of the tradeoff between precision and recall.

First, we consider the proportion of successful matches as a function of the population size; i.e.,the number of individuals in the genomic database.
To do this, we consider several fixed values of $k$, where a match from a face image $x_i$ to a genome $y_j$ is considered successful if the associated log-likelihood score $p_{i,j}$ is among the top-$k$ for the image $x_i$.

\begin{figure}[ht!]
\centering
\begin{subfigure}[t]{0.35\textwidth}
    \includegraphics[width=6.5cm]{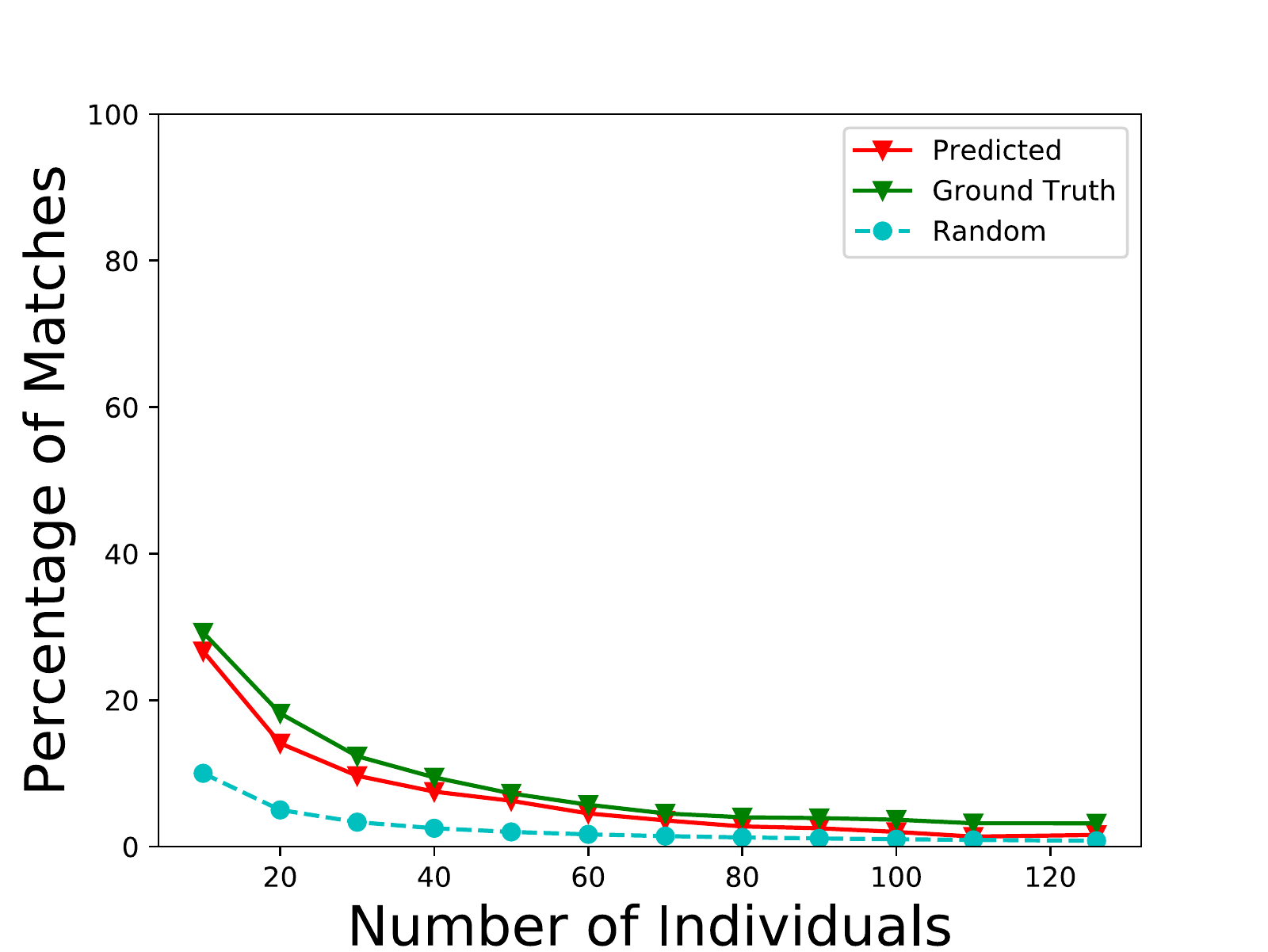}
    \caption{}
\end{subfigure}
\begin{subfigure}[t]{0.35\textwidth}
    \includegraphics[width=6.5cm]{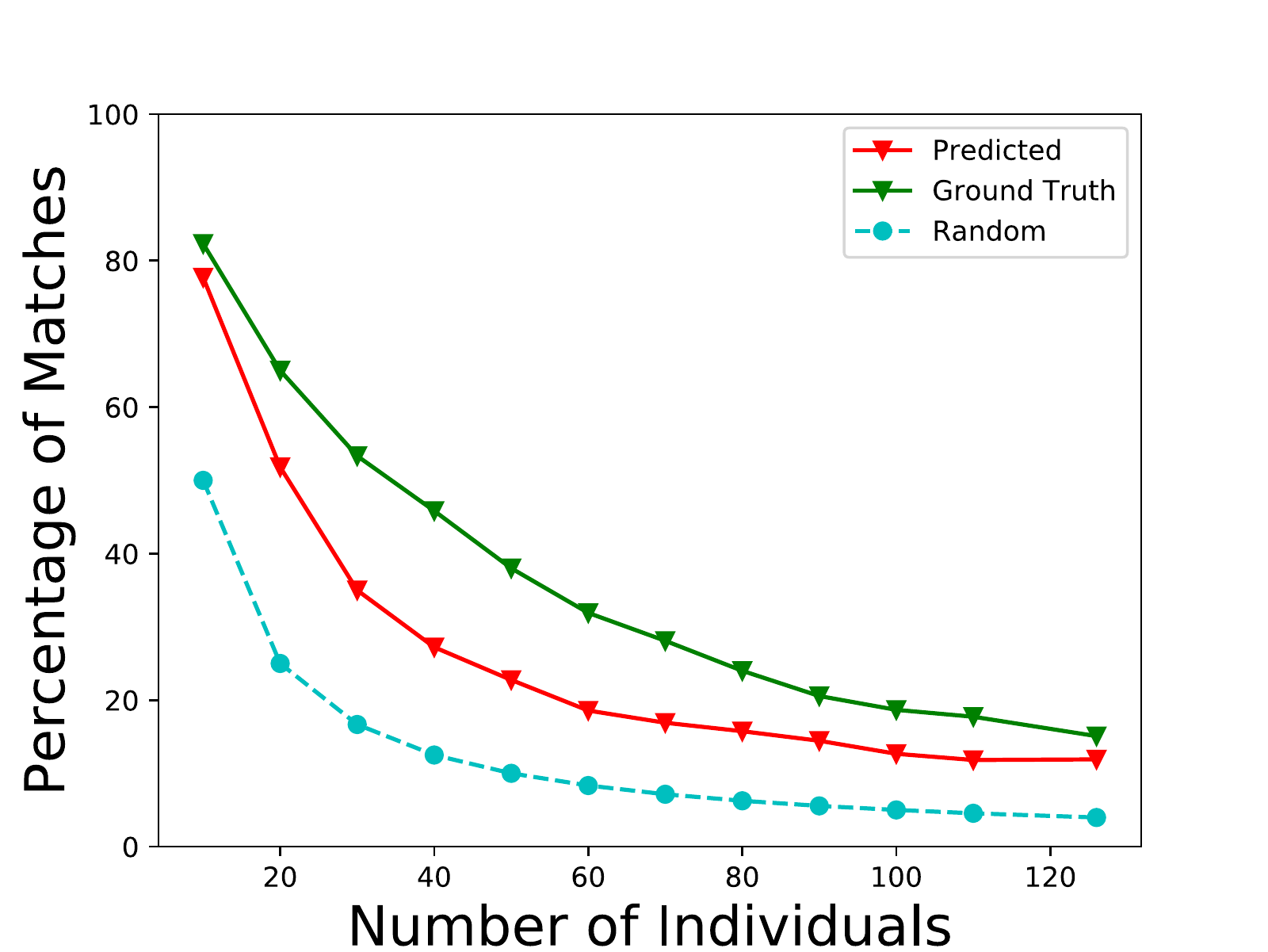}
    \caption{}
\end{subfigure}
\begin{subfigure}[t]{0.35\textwidth}
    \includegraphics[width=6.5cm]{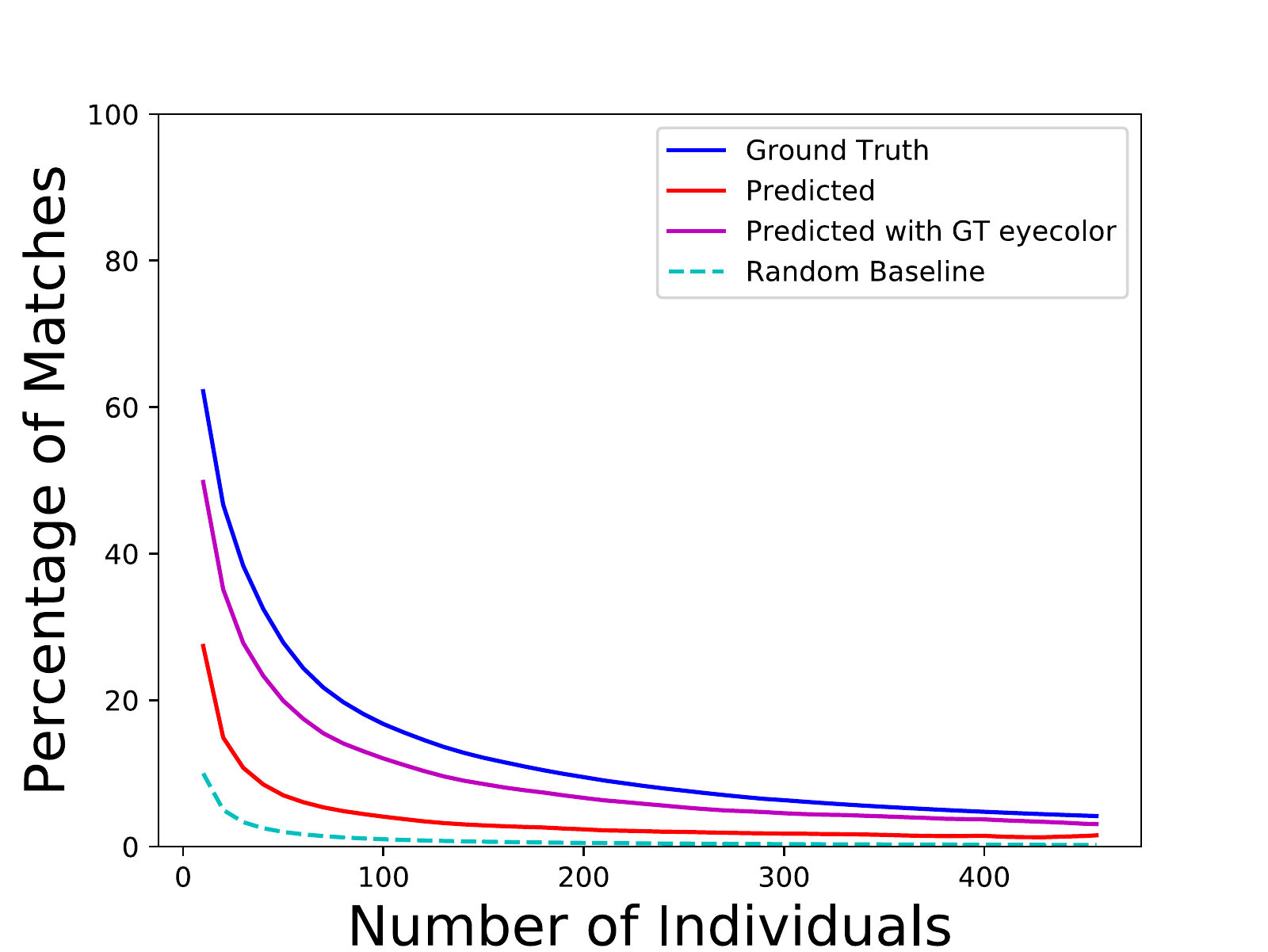}
    \caption{Ideal}
\end{subfigure}
\begin{subfigure}[t]{0.35\textwidth}
    \includegraphics[width=6.5cm]{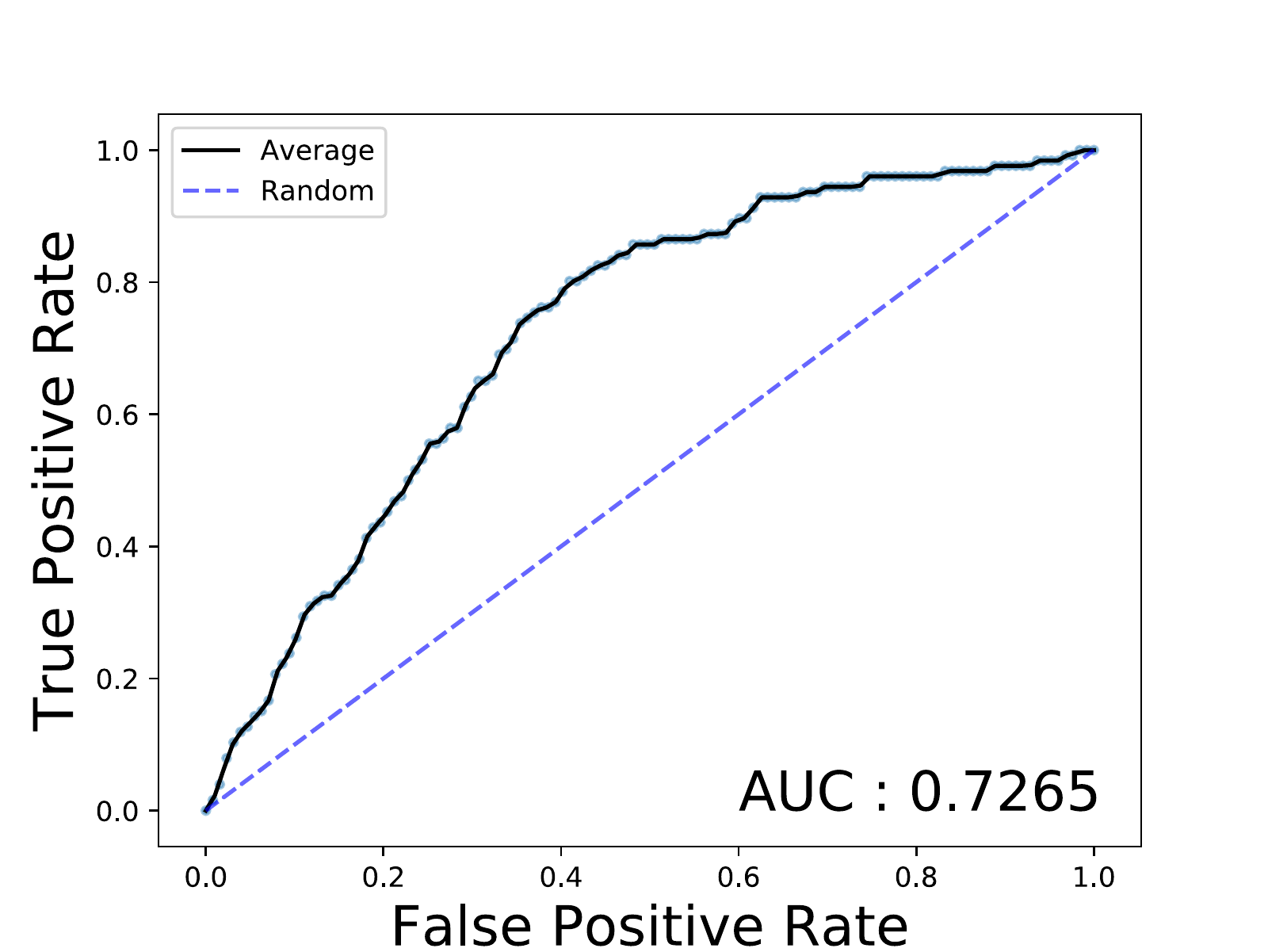}
    \caption{}
\end{subfigure}
\caption{\textbf{Effectiveness of matching individuals' photos to their DNA sequences in OpenSNP.} (a) Success rate for top-1 matching for the \emph{Real} dataset; (b) success rate for top-5 matching for the \emph{Real} dataset; (c) Success rate for top-1 matching in the \emph{Synthetic-Ideal} dataset; 
(d) ROC curve for 126 individuals.
(a), (b), and (c) present matching success results as a function of the population size (the number of individual genomes to match a face image to) for a fixed $k$.
}
\label{fig:matching}
\end{figure}

The results for the \emph{Real} dataset for $k=1$ and $k=5$ are shown in Fig.~\ref{fig:matching} (a) and \ref{fig:matching} (b) respectively.
We compare the success of our re-identification approach to two baselines: 1) when matches are made randomly (a lower bound) and 2) when matches use \emph{actual}, rather than predicted, phenotypes (an upper bound).
Based on Fig.~\ref{fig:matching} (a), it can be seen that the matching success (where we solely take the top-scoring match) is relatively low \emph{even for the upper bound}, where we actually know the phenotypes (and, consequently, do not need the images). However, the top-1 matching success rate is close to the upper bound (which assumes perfect knowledge of phenotypes) and is considerably better than random. As expected \citemain{lippert2017identification}, prediction efficacy reduces as the population size grows. Fig.~\ref{fig:matching} (c) shows that in an idealized setting, re-identification accuracy can be considerably higher; however, effectively predicting eye color is crucial, and this is also a major limitation of existing techniques. Fig.~\ref{fig:matching} (d) shows that when we treat matching as a binary prediction problem, the effectiveness is well above that achieved by randomly guessing.
Nevertheless, re-identification risk "in the wild" does not appear to be especially high. While we observe success rate as high as 25\%, this is only achieved when the genomic dataset is extremely small - on the order of 10 individuals. By contrast, the success rate for top-1 matching drops quickly and is negligible for populations of over 100 individuals. And, it should be kept in mind that this result assumes we can predict the phenotypes perfectly.

The overall pattern does not substantially change when $k=5$. However, in this case the matching success rates naturally increase, approaching 80\% for small populations, and slightly below 20\% for populations above 100 individuals.
In this case, we do observe that our re-identification approach, while significantly better than random, is also considerably below the theoretical upper bound. This suggests that, when more than a single re-identification claim is permitted for each image, the error in phenotype prediction from face images has a greater influence.

Next, we delve more deeply into the nature of re-identification risk using the larger synthetic datasets.
We present the results for \emph{Synthetic-Ideal} in Fig.~\ref{fig:matching} (c). Additional results for both the \emph{Synthetic-Ideal} and \emph{Synthetic-Real} datasets when the top $1, 3$, and $5$ matches are predicted to be true matches are provided in Supplementary Fig.~\ref{fig:adv_synth} (a)-(f).
These results offer two insights.
First, if an attacker has access to particularly high-quality data, re-identification risk can be relatively high for small populations.
For example, the upper bound is now over 60\% in some cases.
However, it can also be seen that of the phenotypes we aim to predict, eye color is both the most difficult and highly influential in matching. If we assume that we \emph{know} this phenotype, and only have to predict the others, re-identification risk is near its upper bound (which assumes that we know the true phenotypes).
This is even more striking in the case of the \emph{Synthetic-Real} data, as shown in Supplementary Fig.~\ref{fig:swapeye} (a)-(d).
To determine if this result was an artifact of the specific method we selected for eye color prediction, we considered several alternative methods\citemain{7553523} for eye color prediction, ranging from conventional computer vision techniques to deep learning (see Supplementary Fig.~\ref{fig:eyecomp} (e)-(g)). None of these methods were particularly efficacious.

\subsection*{Precision-Recall Tradeoff}
We then turned our attention to a different evaluation of prediction efficacy: the tradeoff between false positives and false negatives that obtains as we vary $k$.
The results, shown in Fig.~\ref{fig:matching} (d) for a population size of 126 individuals, suggest that the overall re-identification method is relatively effective (AUC $>$70\%), particularly when compared to random matching when viewed as a binary predictor (match vs.~non-match) for given genome-image pairs. ROC curves when thresholding on $k$ for various population sizes are presented in Supplementary Fig.~\ref{fig:indeps_roc} (a)-(l). In Supplementary Fig.~\ref{fig:indeps_roc} (m)-(x), we also consider a common alternative where we use a tunable threshold $\theta$ on the predicted log-likelihood to claim when a match occurs.

Overall, our results suggest that there is some relationship between face images and public genomic data, but the success rates are well below what other prior literature appears to suggest, even in idealized settings.
We believe there are several contributing factors behind this observation.  First, the quality of face images in the wild is much lower than the high-definition 3D images obtained in highly controlled settings in prior studies~\citemain{lippert2017identification, crouch2018genetics, qiao2016detecting, caliebe2016more}.
Second, there is a relative scarcity of high-quality training and validation data for this particular task. While there are large, well-labeled datasets for face classification~\citemain{parkhi2015deep, panetta2018comprehensive, kazemi2014one, grgic2011scface, GeBeKr01, weyrauch2004component}, the data required for re-identification requires paired instances of genomes and images, which is far more challenging to obtain at scale.
Third, visible phenotypes are influenced by factors other than just the SNPs that are known to have a relationship with them, particularly when you add artificial factors, such as by dying one's hair or wearing tinted contact lenses, introducing considerable noise in the matching.
Finally, our analysisassumed (as did all prior studies) that we already know that there is, in fact, a match in the genomic dataset corresponding to each face.
In reality, success rates would be even lower, since a malicious actor is unlikely to be certain about this~\citemain{10.1145/2806416.2806580}.

 \subsection*{Achieving Privacy through Adversarial Image Perturbations}

While our assessment of re-identification risk above suggests that the risk to an average individual exhibited in prior literature are likely somewhat inflated in prior literature, we are, indeed, able to successfully re-identify a subset of individuals.
Moreover, some of the settings, such as situations where there is sufficient prior knowledge that can narrow down the size of the population to which a face can be matched, are indeed quite risky.
This led us to investigate the natural question: how can we most effectively mitigate re-identification risks associated with the joint public release of both genomic data and face images.
Our specific goal is to provide tools that can reduce re-identification risks \emph{to individuals} who publicly post their photos.
Such tools can then be used either directly by individuals to manage their own risks, or by platforms where photos are posted to manage risks to their subscribers.
In particular, we show that this can be accomplished by adding small adversarial image perturbations (via adversarial examples) to reduce the effectiveness of genomic linkage attacks.

The idea behind adversarial examples is to inject a small amount of noise into an image, where ``small'' is quantified using an $l_p$ norm, in order to cause a misprediction.
In our case, however, we do not have a single deep neural network making predictions, but rather a collection of independent phenotype predictors \emph{using the same image as input}.
One direct application of adversarial examples in our setting would be to arbitrarily choose a phenotype (say, sex, which is the most informative) and target this phenotype for misprediction, with the anticipation that this would cause re-identification attacks to fail.
However, we can actually do much better at protecting privacy by tailoring adversarial examples to our specific task.
This is because our ultimate goal is not to cause mispredictions of phenotypes per se, but rather to cause an attacker to fail to link the image with the correct genome.
Leveraging the scoring function in Equation~\eqref{E:pred}, we aim to minimize the score $p_{ij}$ for image $x_i$ and the correct corresponding genome $y_j$.
However, this is a non-trivial task because the scoring function has a discontinuous dependence on predicted phenotypes (since we use only the most likely phenotype given image in computing it).
To address this issue, we augment the score with the log of the predicted probabilities.
More precisely, let $g_p(v_p, x_i)$ denote the probability that the neural network predicts a variant $v_p$ for a phenotype $p$ (e.g., eye color) given the input face image $x_i$.
The problem we aim to solve is to find adversarial perturbation to the input image $\delta^*$ that solves the following problem
\[
  \min_{-\epsilon \le \delta \le \epsilon} \sum_{p}^{p \in \{sex,hair,skin,eye\}} \sum_{v_p} \log g_p(v_p, x_i+\delta) \log P(v_p | y_j).
\]
Since this expression is differentiable with respect to the adversarial noise $\delta$, we can solve it using standard gradient-based methods (see Materials and Methods for further details).

\begin{figure}[h!]
\centering
\begin{tabular}{p{0.35\textwidth}c}
\begin{subfigure}[t]{0.35\textwidth}
  \includegraphics[width=6.5cm]{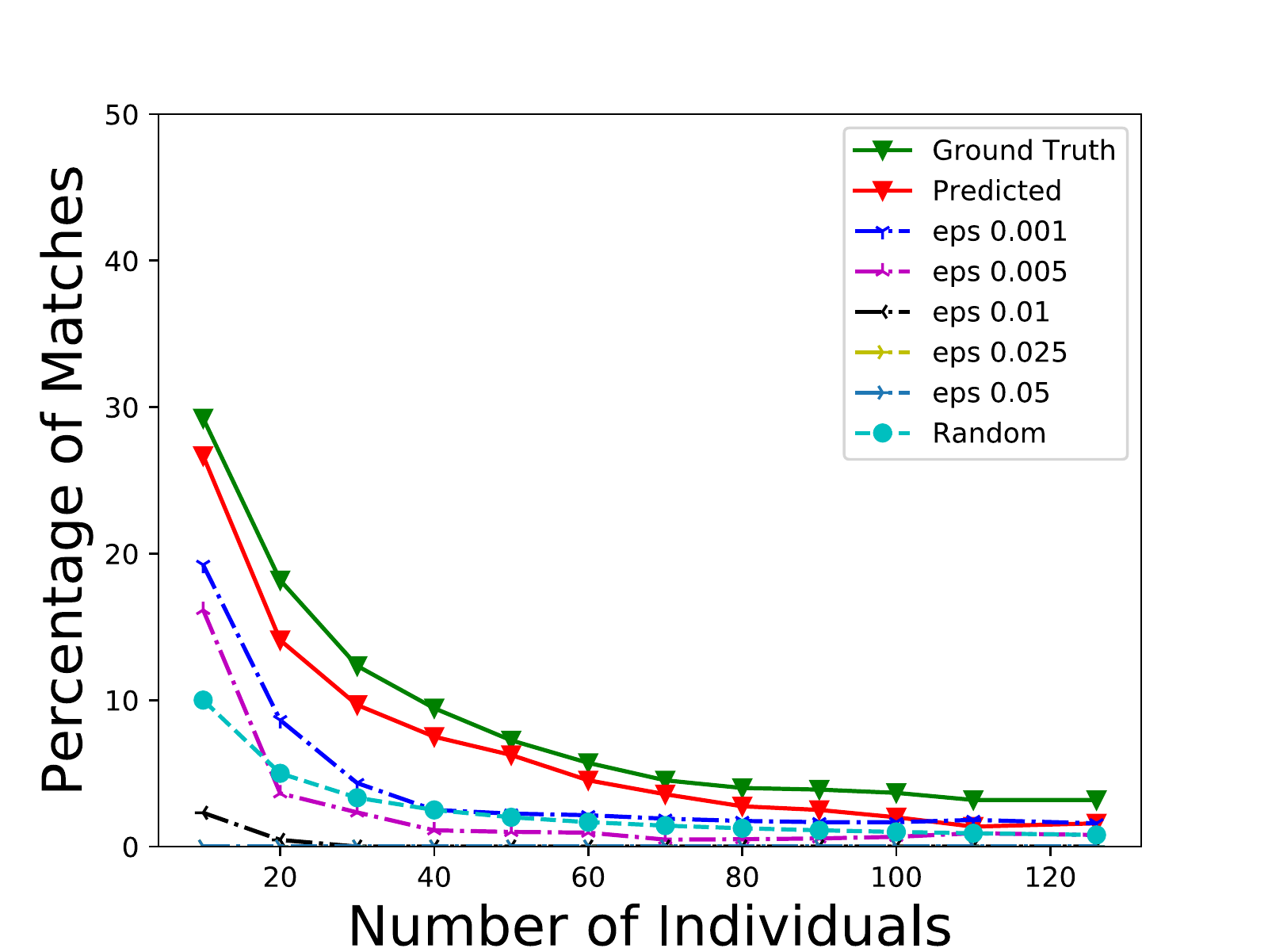}
    \caption{}
\end{subfigure}
\begin{subfigure}[t]{0.35\textwidth}
      \includegraphics[width=6.5cm]{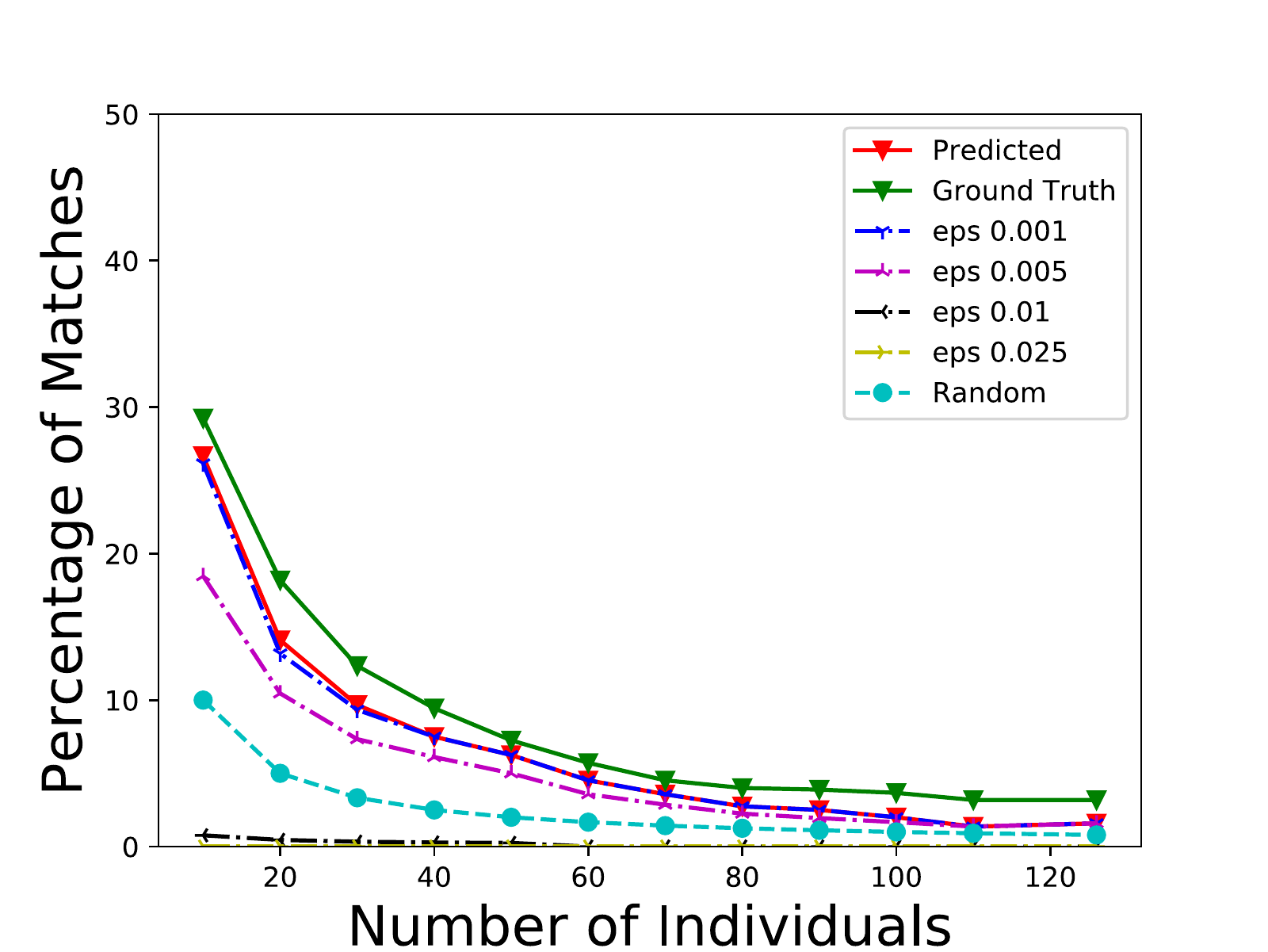}
    \caption{}
\end{subfigure}
& 
\begin{tabular}{c}
\\
\\
$\epsilon=0.001$\\
    \includegraphics[height=1.8cm, width=4.9cm]{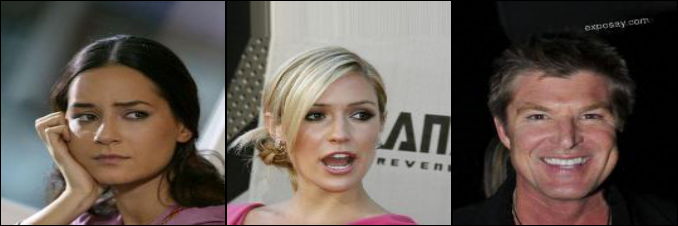}\\
    $\epsilon=0.005$\\
    \includegraphics[height=1.8cm, width=4.9cm]{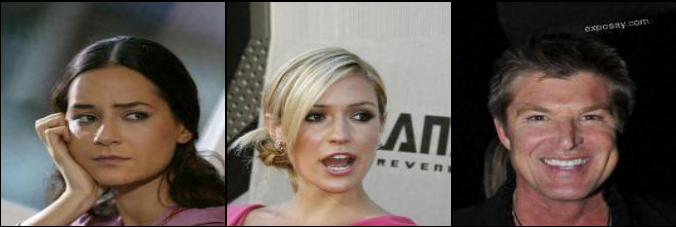}\\
        $\epsilon=0.01$\\
    \includegraphics[height=1.8cm, width=4.9cm]{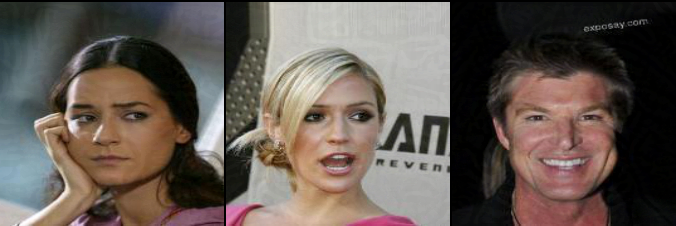}\\
        $\epsilon=0.025$\\
    \includegraphics[height=1.8cm, width=4.9cm]{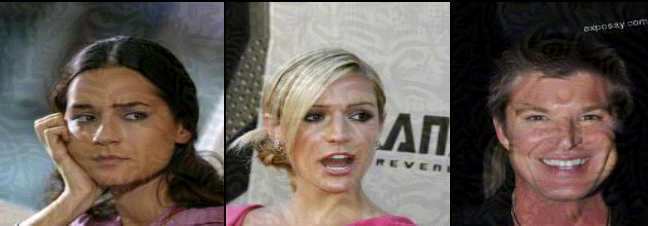}
    \end{tabular}
\end{tabular}
\caption{\textbf{Evaluating adversarial perturbations as a defense.} (a) Effectiveness of adversarial perturbation as a defense against re-identification for $k=1$ (i.e., the attacker considers only the top match).  Pixel values are normalized to a $[0,1]$ interval, and perturbation strengths $\epsilon$ are with respect to these normalized pixel values.
  It can be seen that prediction accuracy is near zero at a perturbation strength $\epsilon \geq 0.01$.  Moreover, even for very small amounts of adversarial noise, such as $\epsilon = 0.001$, matching success is nearly indistinguishable from random matching if we have at least 20 individuals in a consideration pool.
  (b) Effectiveness of perturbations that only target sex prediction from a face image.  The effect of larger perturbations ($\epsilon \ge 0.01$ is similar to Fig.~\ref{fig:nonrobust_attacked} (a).  However, smaller perturbations are considerably less effective.
  Example images on the right (drawn from the public celebrity face image dataset) illustrate the extent of visible effect of introduced adversarial perturbations to images.
  The perturbations are essentially imperceptible to a human until $\epsilon > 0.01$, when the effect becomes clearly visible.}
\label{fig:nonrobust_attacked}
\end{figure}

Our first evaluation, shown in Fig.~\ref{fig:nonrobust_attacked}, presents the effectiveness of our method for preserving privacy in public face images.
Fig.~\ref{fig:nonrobust_attacked}(a), in particular, demonstrates that when we take deep neural networks for phenotype prediction as a given, the effectiveness of the re-identification attack described above declines significantly even for very small levels of noise added to images.
For sufficiently large noise (e.g., $\epsilon=0.01$), the success rate is close to zero, which is considerably lower than random matching.
Moreover, by comparing Fig.~\ref{fig:nonrobust_attacked} (a) to Fig.~\ref{fig:nonrobust_attacked} (b), it can be seen that our approach is also more effective than designing adversarial perturbations that target a single sex phenotype. The effectiveness of targeting other phenotypes is provided in Supplementary Fig.~\ref{fig:hair_eye_skin_1} (d)-(f), where it can be seen that attacking only hair color, eye color or skin color alone is insufficient to induce  a significant level of misclassification.
While the presented results are only for $k=1$ (i.e., the attacker only considers the top-scoring match), results for $k=3$ and $k=5$ offer similar qualitative insights (as shown in Supplementary Fig.~\ref{fig:nonrobust_attacked_5}).

The visual effect of the designed image perturbations is illustrated on the right in Fig.~\ref{fig:nonrobust_attacked} using images drawn from the public celebrity face image dataset.
As can be seen, most of the levels of added noise have negligible visual impact. It is only when we add noise at $\epsilon=0.025$ that we begin to clearly discern the perturbations.
However, to address the re-identification problem, it appears sufficient for privacy to use levels of noise of $\epsilon = 0.01$ or smaller, which leads to near-zero matching-score.

While introducing adversarial noise appears to be sufficiently reduce the risk of re-identification introduced by publicly shared photographs, this still supposes that re-identification makes use of deep neural network models trained in the regular fashion, such as on the CelebA dataset.
However, there are various techniques for enabling the training of such models that yield significantly better resistance to adversarial noise, the most effective of which is adversarial training~\citemain{madry2017towards,Vorobeychik18book}.
The adversarial training approach learns a model by augmenting each iteration with training images that are adversarially perturbed, using the model from the previous iteration as a reference.
The main limitation of adversarial training is that it also results in lower accuracy on data that is not adversarially perturbed.
Given the relatively limited effectiveness of the re-identification approach above and all of the practical limitations of that exercise, we now investigate whether, in practice, adversarial training sufficiently handicaps re-identification, such that its increasing robustness to added perturbations offers little value.

\begin{figure}[h!]
\centering
\begin{subfigure}[t]{0.45\textwidth}
    \centering
    \includegraphics[width=6.5cm]{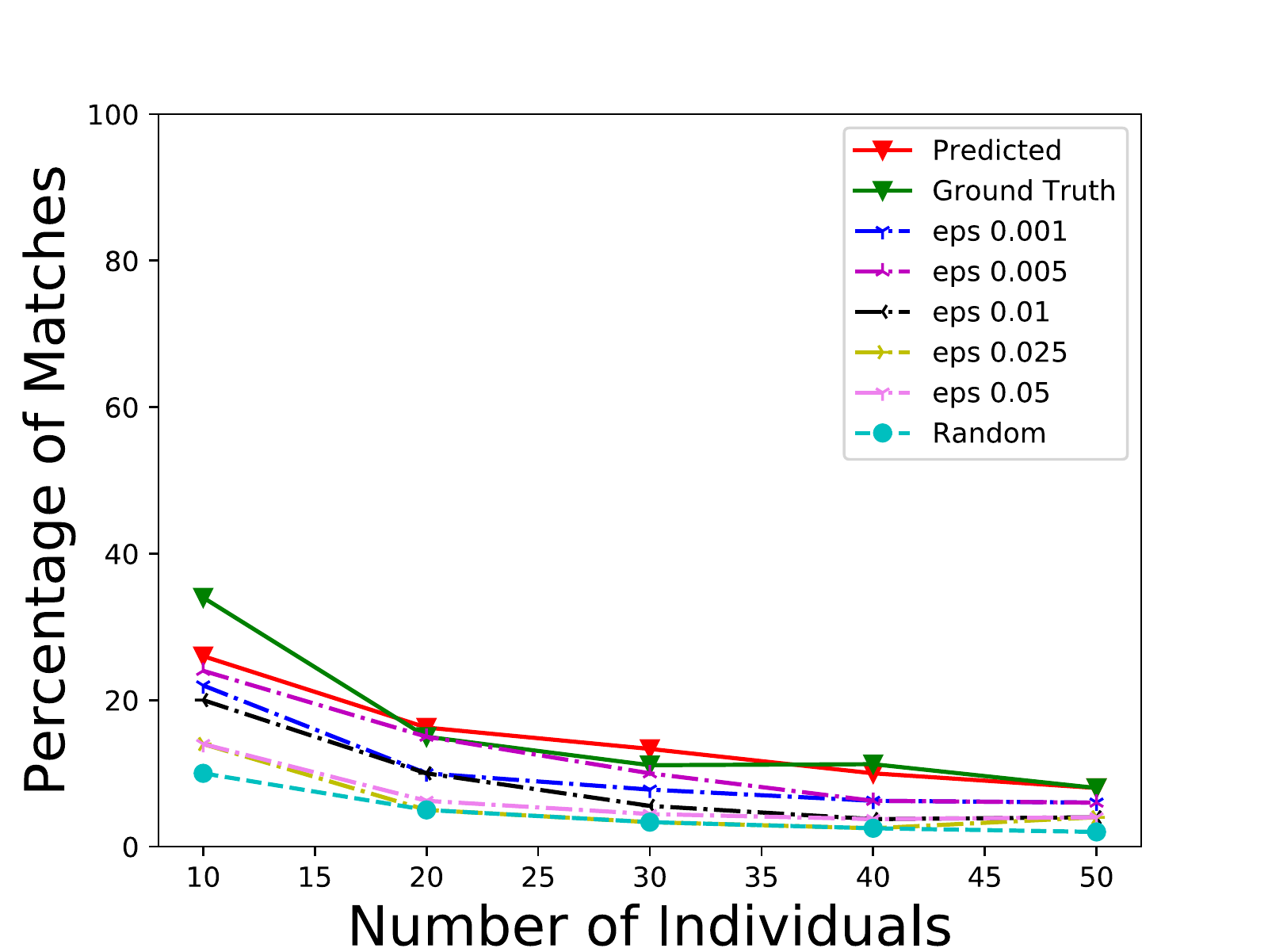}
    \caption{}
\end{subfigure}
\begin{subfigure}[t]{0.45\textwidth}
    \centering
    \includegraphics[width=6.5cm]{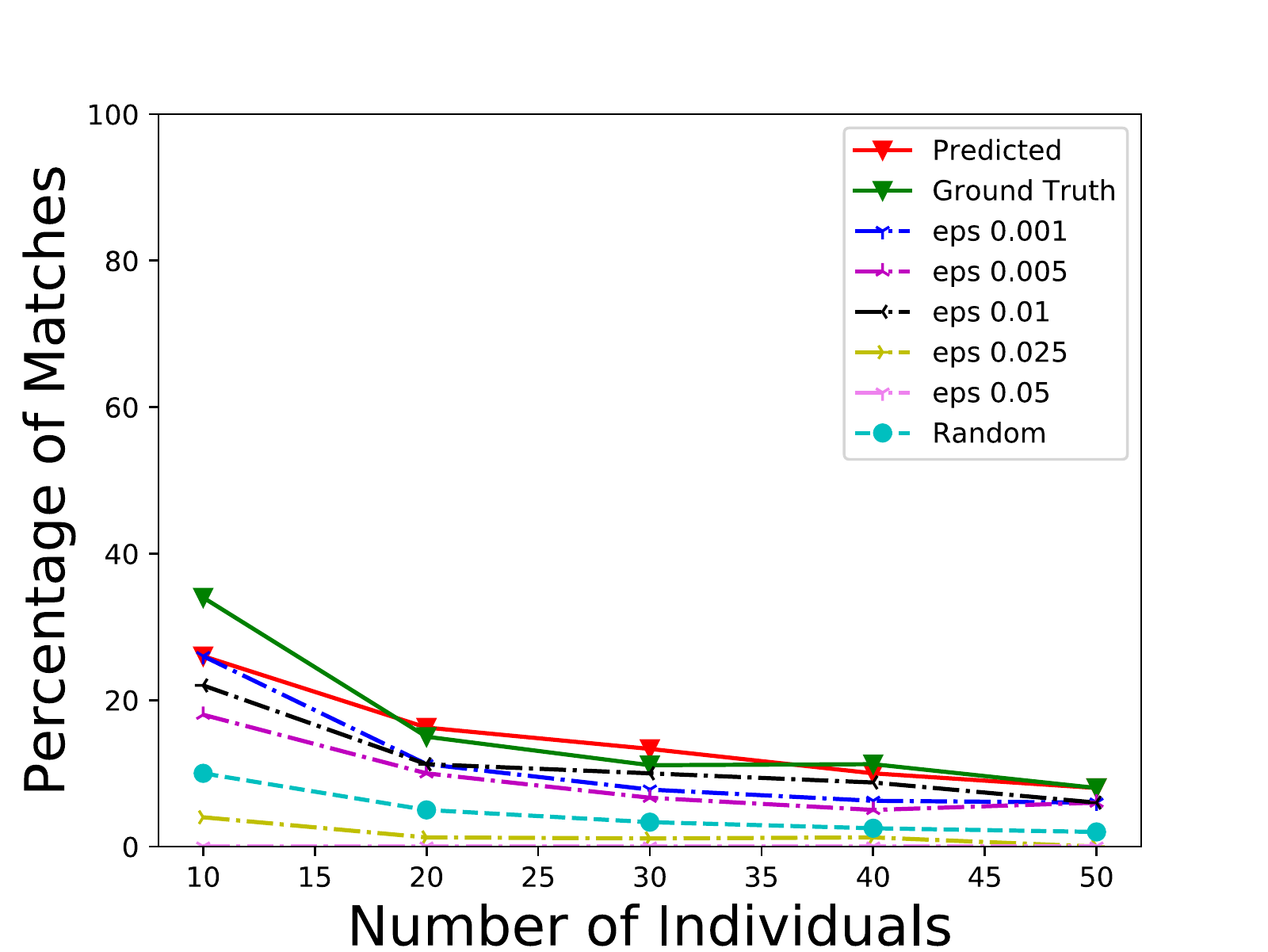}
    \caption{}
\end{subfigure}
\caption{\textbf{Evaluation of models that are trained to increase robustness to perturbations through adversarial training when only the top match is considered in re-identification.}  a) Matching accuracy of ``robust'' models trained by adding adversaries with varying adversarial noise levels $\epsilon$ when unperturbed face images are used as inputs.  Using $\epsilon > 0.01$ causes matching accuracy to be effectively equivalent to random. b) Matching accuracy of ``robust'' models trained by adding adversaries with $\epsilon = 0.01$ when input face images are perturbed with varying levels of adversarial noise.  Using $\epsilon > 0.01$ is sufficient to cause sub-random matching accuracy.  For noise with $\epsilon = 0.01$, matching accuracy degrades from original, but remains higher than random.}
\label{fig:robust_noattack}
\end{figure}

To evaluate the effect of adversarial training on re-identification, we run further training iterations on the phenotype-prediction models with adversarial examples generated over subsets of the original training sets. We make five passes over adversarial examples, each time using the model from the previous iteration to generate these examples.
Since the match score depends on images as well as corresponding genomes, we use paired genome-image datasets for adversarial training (the most optimistic setting from the re-identification perspective). 
Specifically. we use a random subset of 77 image-DNA pairs (approximately $60\%$) from our OpenSNP dataset for training, and the remaining 49 for testing the matching accuracy.
We construct five sets of adversarially robust phenotype prediction models using this procedure, each set adversarially trained using a different amount of added adversarial noise, from $\epsilon = 0.001$ to $\epsilon = 0.05$.

As expected, Fig.~\ref{fig:robust_noattack} (a) illustrates that baseline prediction accuracy (i.e., using original face images without perturbations) reduces as strength of the perturbation used for adversarial training increases.
Indeed, once $\epsilon > 0.01$, the effectiveness of matching is essentially equivalent to random, suggesting that the most robust model that holds any utility is the one with $\epsilon = 0.01$.
Next, we consider how robust this model is to images that now have adversarial perturbations of varying magnitudes.
The results are shown in Fig.~\ref{fig:robust_noattack}(b).
Notably, adversarial noise with $\epsilon = 0.025$ again yields near zero matching success.
A smaller amount of noise that preserves imperceptibility, such as $\epsilon = 0.01$, is still effective in reducing the accuracy of the robust model, although the resulting re-identification success rate is still above random matching.
Nevertheless, re-identification success even in this case is now $\sim$ 20\% even in the most optimistic case.
Moreover, our framework is sufficiently flexible that a particularly risk-averse individual may simply raise the noise level to 0.025, accepting some visible corruption to the image, but effectively eliminating privacy risk.
  \section*{Discussion}

Our findings suggest that the concerns about privacy risks to shared genomic data stemming from the attacks matching genomes to publicly published face photographs are low, and relatively easy to manage to allay even the diverse privacy concerns of individuals.
Of course, our results do not imply that shared genomic data is free of concern. There are certainly other potential privacy risks, such as membership attacks on genomic summary statistics~\citemain{hagestedt2020membership, liu2018detecting, Chen2020.08.03.235416, wan2017controlling, wan2017expanding, zerhouni2008protecting, raisaro2017addressing, shringarpure2015privacy, craig2011assessing}, which would allow the receipient of the data to determine the presence of an individual in the underlying dataset. This type of attack is of concern because it would allow an attacker to associate the targeted individual with the semantics inherent in the dataset. For instance, if the dataset was solely composed of individuals diagnosed with a sexually transmitted disease, then membership detection would permit the attacker to learn that the target had the disease in question.
Moreover, we emphasize that our results are based on \emph{current} technology; it is possible that improvements in either the quality of data, such as broad availability of high-definition 3D photography, or the quality of AI, such as highly effective approaches for inferring eye color from images, will indeed significantly elevate risks or re-identification.
However, through several studies which include synthetic variants controlling for the quality of data, as well as evaluations that assume we can infer observable face phenotypes with perfect accuracy (see, for example, the results in Fig.~\ref{fig:matching}, as well as in the Supplementary Figures~\ref{fig:adv_synth} and~\ref{fig:swapeye}), we show that even with advances in technology the risk is likely to remain limited.

 \section*{Materials and Methods}

\subsection*{Re-Identification Attack with Public Face Images} In our attack, we consider the following phenotypes to be readily visible in face images: eye color, hair color, sex, and skin color.
To this end, we first collected genomes uploaded to OpenSNP that were sequenced by 23andMe.
We then filtered the users who have self-reported all four phenotypes we are interested in.
Some of these users also uploaded their face images directly on OpenSNP.
For others, we found their faces through a Google reverse search using their OpenSNP usernames, which we then manually verified these using the self-reported phenotypes.
This process yielded 126 pairs of face images and DNA profiles of individuals that were carefully curated.
The full study was approved by the Washington University in St.~Louis Institutional Review Board, with the condition that we will not publicly release this curated dataset (in order to mitigate any possible harm to these individuals).
However, we are able to share it privately with other researchers for validation purposes, subject to an agreement preventing public release.

\subsubsection*{Predicting Phenotypes from Images} To predict phenotypes from facial images, we leveraged the VGGFace \citemain{parkhi2015deep} convolutional neural network architecture.
Due to the relative scarcity of labeled training data for phenotypes of interest, we employed transfer learning~\citemain{pratt1993discriminability}, which has been extensively and successfully used for various classification tasks where labeled data is scarce \citemain{Cao_2013_ICCV,10.1145/2818346.2830593,rajagopal2014exploring,7378076,7139098,xia2011kinship,dornaika2019age, Rossler_2019_ICCV}.
Specifically, we started with a model previously trained on a the CelebA dataset~\citemain{liu2015faceattributes} for a face recognition task.
We then fine-tuned these models on subsets of the CelebA dataset.
For sex and hair color, the CelebA dataset already contains labels for all $\sim 200,000$ images, and we thus used the entire dataset to fine-tune the sex prediction classifier.
For hair color, we find that fine-tuning on a subset of 10000 images with equal number of blonde, brown, and black hair color images outperforms a model trained on the entire dataset; we thus use the latter.
For skin color, 1000 images were labeled on a 5-point scale by Amazon Mechanical Turk (AMT) workers, and then manually sorted into 3 classes.
For eye color prediction, 1000 images were labeled by AMT workers; however, after manual verification of these labels, we retained $\sim$ 850 images, dropping the rest because the eye color was indeterminate.

\subsubsection*{Matching Faces to DNA}
Once we learned phenotype classifiers for each visible phenotype, we then for each face image in test data predicted the most likely variant (i.e., the one with the largest predicted probability) for each face image in test data.
We then used this prediction for matching the face image to a DNA record as follows.
Let an image be denoted by $x_i$ and a genome by $y_j$.
We use the following matching score, where phenotype variant $z_{i,p}$ is the most likely predicted variant:
    
\[P_{ij} = \prod_{p}^{p \in \{sex,hair,skin,eye\}} P(z_{i,p}|y_j).  \]

To ensure numerical stability, we transform it into log-space, resulting in
    
    \[ \log P_{ij} = \sum_{p}^{p \in \{sex,hair,skin,eye\}} \log P(z_{i,p}|y_j) \]
    
The variant $z_{i,p}$ is predicted from \{F, M\} for sex, \{blue, brown, intermediate\} for eye color, \{black, blonde, brown\} for hair color and \{pale, intermediate, dark\} for skin color.
    The probability of each phenotype variant given a genome, $P(z_{i,p}|y_j)$, is in turn expressed as a product of probabilities over relevant SNPs (see Supplementary Table~\ref{snps} for these lists).
    The probability of a specific phenotypic variant given a certain SNP is calculated empirically from the available data, with priors of the phenotypic variant used where SNPs are missing. 

Having calculated the likelihood of a match between image $X_i$ and DNA sequence $y_j$ for all images, for all DNA sequences, we rank the DNA sequences in decreasing order of matching likelihood for each image. The presented results correspond to when the correct match resides in the top scored $1, 3$ or $5$ entries in this sorted list.

\subsection*{Protecting Privacy by Adding Noise to Face Images}

Recall that our goal is to minimize the score $p_{ij}$ where $x_i$ is the image and $y_j$ the correct corresponding genome.
Since the score function has a discontinuous dependence on phenotype predictions, we augment the score with the log of the predicted phenotype probabilities.
Specifically, let $g_p(v_p, x_i)$ denote the probability that the neural network predicts a variant $v_p$ for a phenotype $p$ (e.g., eye color) given the input face image $x_i$; these are just the outputs of the corresponding softmax layers of the neural networks.
The problem we aim to solve is to find adversarial perturbation to the input image $\delta^*$ that solves the following problem:
\[
  \min_{-\epsilon \le \delta \le \epsilon} \sum_{p}^{p \in \{sex,hair,skin,eye\}} \sum_{v_p} \log g_p(v_p, x_i+\delta) \log P(v_p | y_j).
\]

We use \emph{projected gradient descent} to solve this problem, invoking a combination of automated differentiation in pytorch~\citemain{NEURIPS2019_9015}, and the Adam optimizer~\citemain{kingma2014adam}.
After each gradient descent step, we simply clip the noise to be in the $[-\epsilon, \epsilon]$ range, and to ensure that the resulting pixel values are valid.
We use the original image as the starting point in this procedure (i.e., initializing $\delta = 0$).

\subsubsection*{Training Robust Phenotype Classification Models}

While the idea of adding noise as privacy-protection works well when we use regularly trained phenotype prediction models, one can make such models more robust, albeit at a cost to accuracy on noise-free data.
As such, we investigate how effective adversarial training, a state-of-the-art approach for making predictions robust to adversarial noise, is at overcoming our noise injection approach.

The main premise of adversarial training is as follows.
The broad goal is to solve the following optimization problem:
\[ \min_\theta \sum_{x,y \in \mathcal{D}} L(\theta, x+\delta^*, y), \]
where $\delta^*$ is adversarially induced noise and $L(\cdot)$ is the loss function.
In practice, computing an optimal noise to add is difficult, and we instead use the approaches such as gradient descent described above.
However, adversarial training proceeds like regular training (using gradient descent), except that each training input is $x+\delta^*$ rather than $x$, that is, we use adversarially perturbed inputs in place of regular inputs.

where $||\delta|| \leq \epsilon$. A standard way of achieving empirically robust models is to simply augment training data with adversarial examples generated over a subset of the training data.
In generating these instances, we use random starting points for generating adversarial examples.
The downside of adversarial training, however, is that robustness to adversarial images often comes at the cost of accuracy on unaltered images, and a careful balance must be achieved between adversarial robustness and baseline accuracy.

\bibliographystylemain{naturemag-doi}
\bibliographymain{sample}

\section*{Acknowledgements}
We acknowledge support of this work by the National Human Genome Research Institute, National Institutes of Health under grant number RM1HG009034.

\section*{Author Contributions Statement}
R.V., B.A.M. and Y.V. jointly designed the study. R.V. implemented the study, and all authors analysed the results. B.A.M. and Y.V. jointly supervised this work. All authors contributed to writing, editing and reviewing the manuscript.

\section*{Competing Interests}
The authors declare no competing interests.

\section*{Data Availability}
The \emph{Synthetic-Real} and \emph{Synthetic-Ideal} datasets generated as part of the study are available along with our code in a publicly available GitHub repository. As this study was approved by Washington University in St. Louis Institutional Review Board under the condition that the \emph{Real} dataset will not be publicly released to protect the identity of the individuals involved, this dataset is made available privately upon reasonable request to \href{mailto:rajagopal@wustl.edu}{rajagopal@wustl.edu}, subject to a data-sharing agreement prohibiting public release.

\section*{Code Availability}
The code used in this work, the synthetic data generated in the study, and instructions to set up a similar Python environment are available in a publicly accessible GitHub repository: \url{https://github.com/rajagopalvenkat/GenomicReID}.

\clearpage
\appendix
\renewcommand{\appendixpagename}{Supplementary Information}
\appendixpage
\renewcommand\thesection{\arabic{section}}
\renewcommand\thesubsection{\thesection.\arabic{subsection}}
\renewcommand{\figurename}{Supplementary Figure}
\renewcommand{\tablename}{Supplementary Table}

\section*{Introduction to the Supplement}
This supplement provides additional methods, descriptions and results for our study, Re-identification of Individuals in Genomic Datasets Using Public Face Images. Section 1 outlines the SNPs and corresponding phenotypes considered in our study. Section 2 presents a brief overview of the deep-learning architecture used for phenotype extraction from face images. We present results when treating the re-identification process as a binary prediction problem - Receiver Operating Characteristic curves for various population sizes when the prediction threshold is varied - in Section 3, and Section 4 illustrates the use of a Support Vector Machine to make binary matching predictions. While the main body focuses primarily on re-identification risk, and steps taken to mitigate it in the top-$1$ case, i.e. only the genome with the maximum calculated log-likelihood is predicted to be a true match, in the supplement, we also present results in the top-$3$ and top-$5$ cases. Sections 5 and 6 contain results for adding adversarial noise, and the effects of adversarial training. Additionally, when attacking a single-phenotype, the main body focuses on the prediction of sex from face images; results for the other phenotypes - namely skin color, hair color and eye color - as well as top-$3$ and top-$5$ results when attacking sex prediction are explored here. Finally, Section 7 elaborates upon our experiments with the two synthetic datasets referred to in the main body, with particular emphasis on the impact of eye color prediction on the matching pipeline.

\section{SNP-Phenotype Associations}
Single Nucleotide Polymorphisms (SNPs) are variations in an individual's DNA that are probabilistically linked to various phenotypes. In the context of our study, the phenotypes we are interested in are the individual's sex, skin color, eye color and hair color. To associate phenotypes predicted from images to genomes, we consider the same set of SNPs as used in the study\cite{humbert2015anonymizing} used to produce what we refer to as the upper-bound `ground-truth' baseline for the relevant phenotypes. The SNPs considered in the study are presented in Supplementary Table~\ref{snps}.

\begin{table}[hb!]
\centering
\begin{tabular}{|l|l|l|}
\hline
\textbf{Phenotype} & \textbf{Traits}                                                & \textbf{SNPs}                                                                                                                                                                         \\ \hline
Sex                & \begin{tabular}[c]{@{}l@{}}M\\ F\end{tabular}                  & Sex Chromosome                                                                                                                                                                        \\ \hline
Skin Color         & \begin{tabular}[c]{@{}l@{}}Pale\\ Int\\ Dark\end{tabular}      & \begin{tabular}[c]{@{}l@{}}rs26722\\ rs1667394\\ rs16891982\end{tabular}                                                                                                              \\ \hline
Hair Color         & \begin{tabular}[c]{@{}l@{}}Blonde\\ Brown\\ Black\end{tabular} & \begin{tabular}[c]{@{}l@{}}rs12821256\\ rs35264875\end{tabular}                                                                                                                       \\ \hline
Eye Color          & \begin{tabular}[c]{@{}l@{}}Blue\\ Brown\\ Int\end{tabular}     & \begin{tabular}[c]{@{}l@{}}rs916977\\ rs1129038\\ rs1800401\\ rs2238289\\ rs2240203\\ rs3935591\\ rs4778241\\ rs7183877\\ rs8028689\\ rs12593929\\ rs1800407\\ rs7495174\end{tabular} \\ \hline
\end{tabular}
\caption{Phenotypes considered for matching images to DNA, and corresponding SNPs .}
\label{snps}
\end{table}



\section{Matching accuracy with varying prediction thresholds - Receiver Operating Characteristic (ROC) curves}
\begin{figure}[]
\centering
\begin{subfigure}[t]{0.22\textwidth}
    \includegraphics[width=4.0cm]{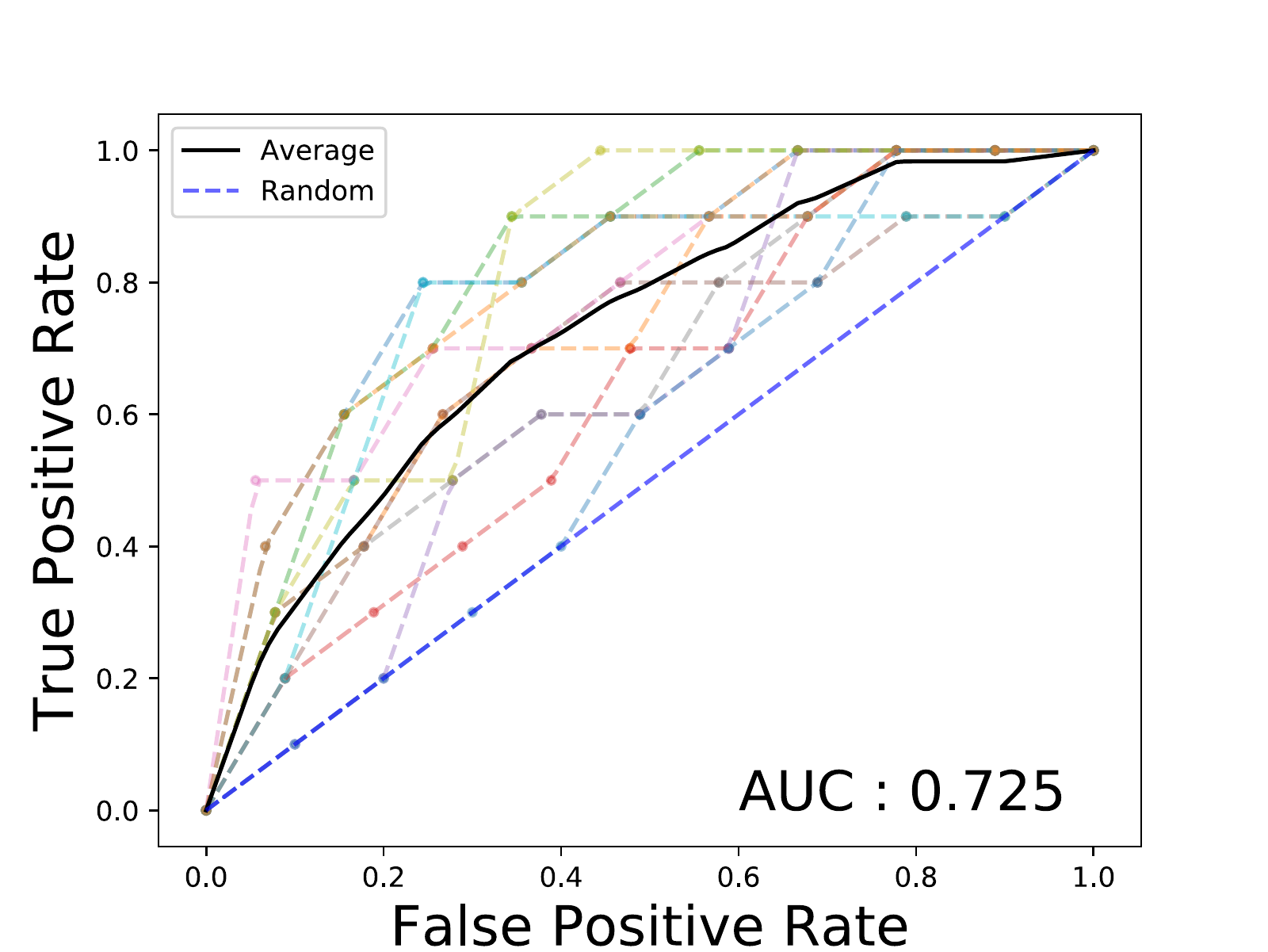}
    \caption{Population : 10}
\end{subfigure}
\begin{subfigure}[t]{0.22\textwidth}
    \includegraphics[width=4.0cm]{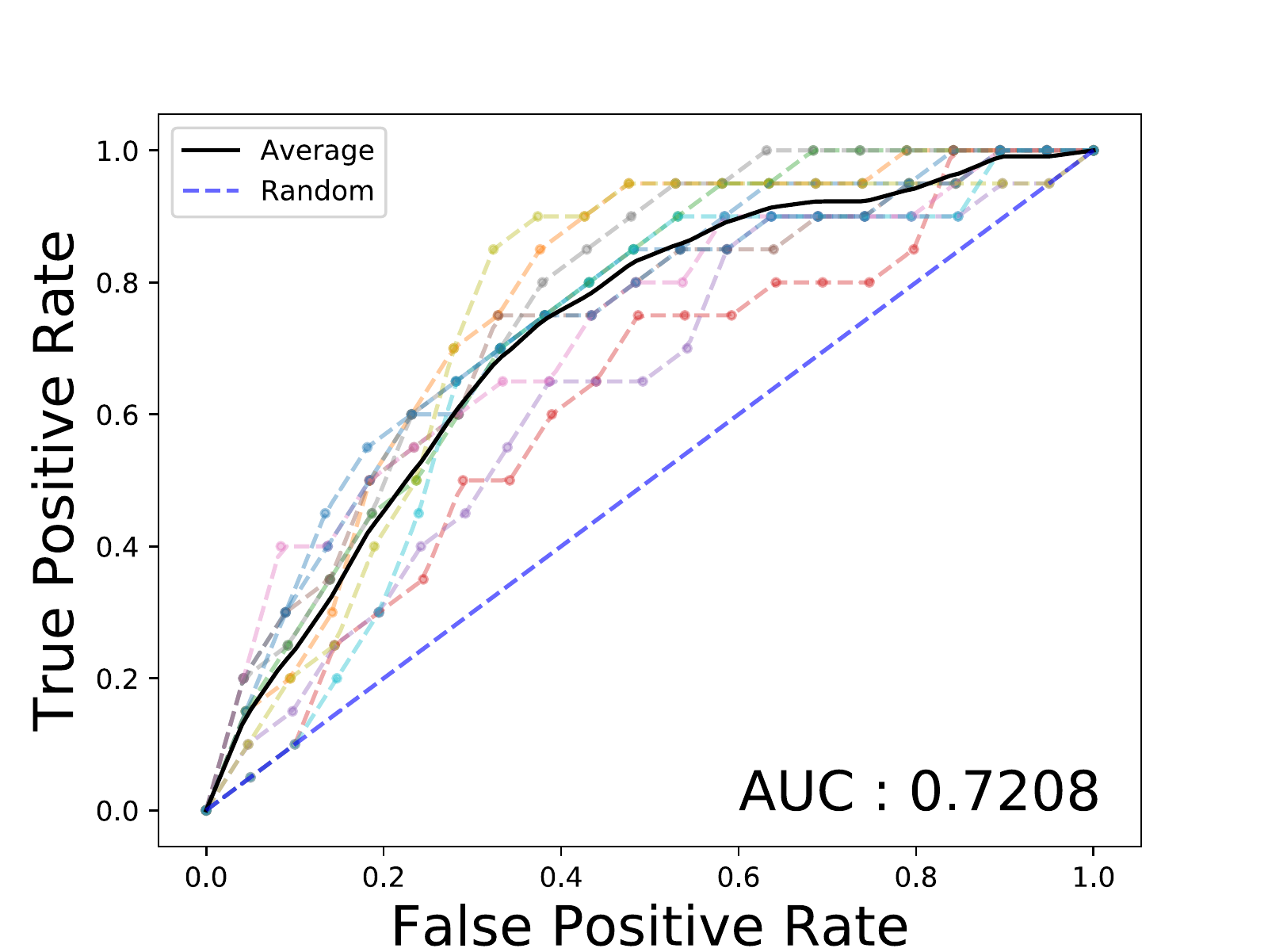}
    \caption{Population : 20}
\end{subfigure}
\begin{subfigure}[t]{0.22\textwidth}
    \includegraphics[width=4.0cm]{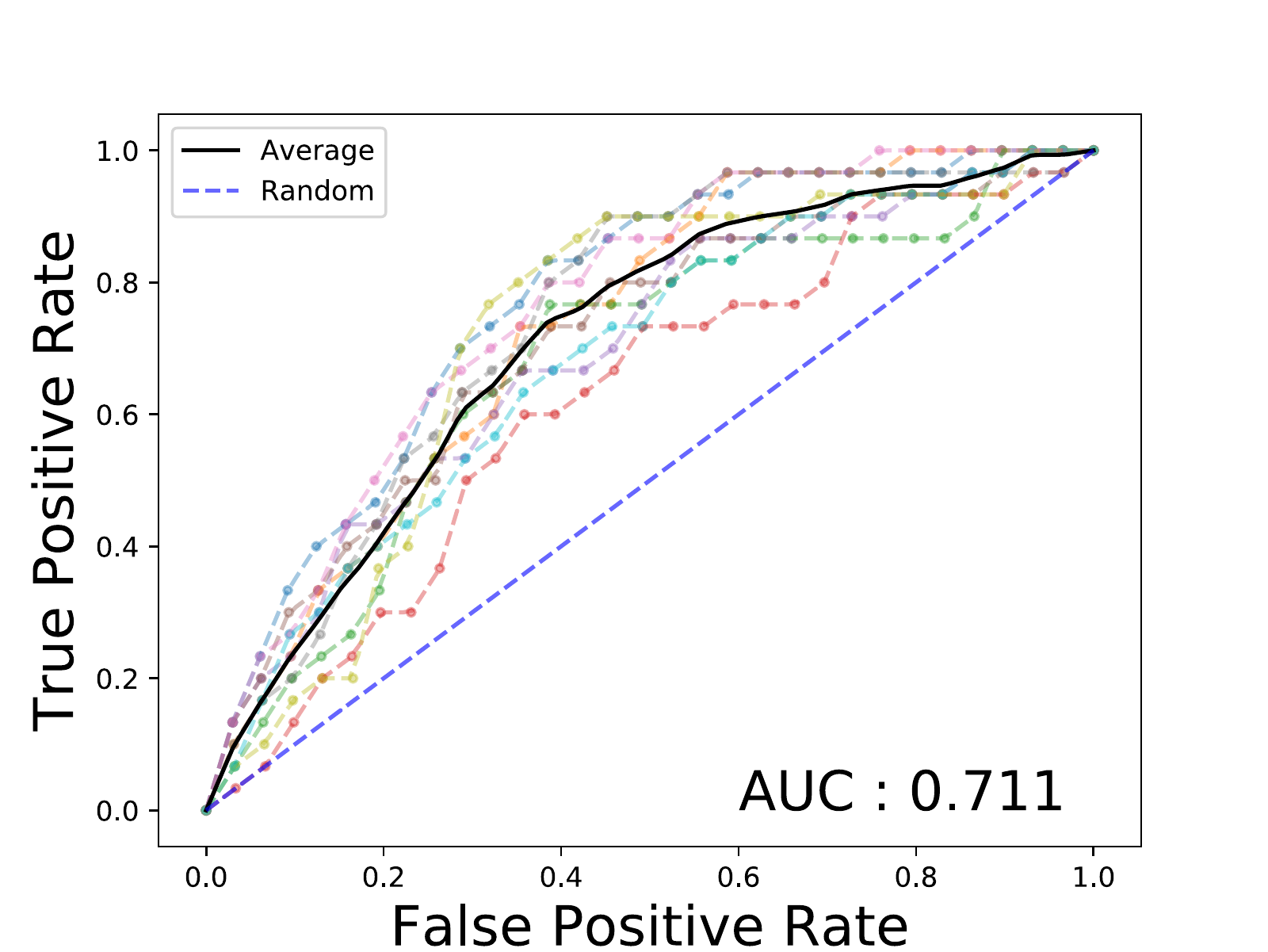}
    \caption{Population : 30}
\end{subfigure}
\begin{subfigure}[t]{0.22\textwidth}
    \includegraphics[width=4.0cm]{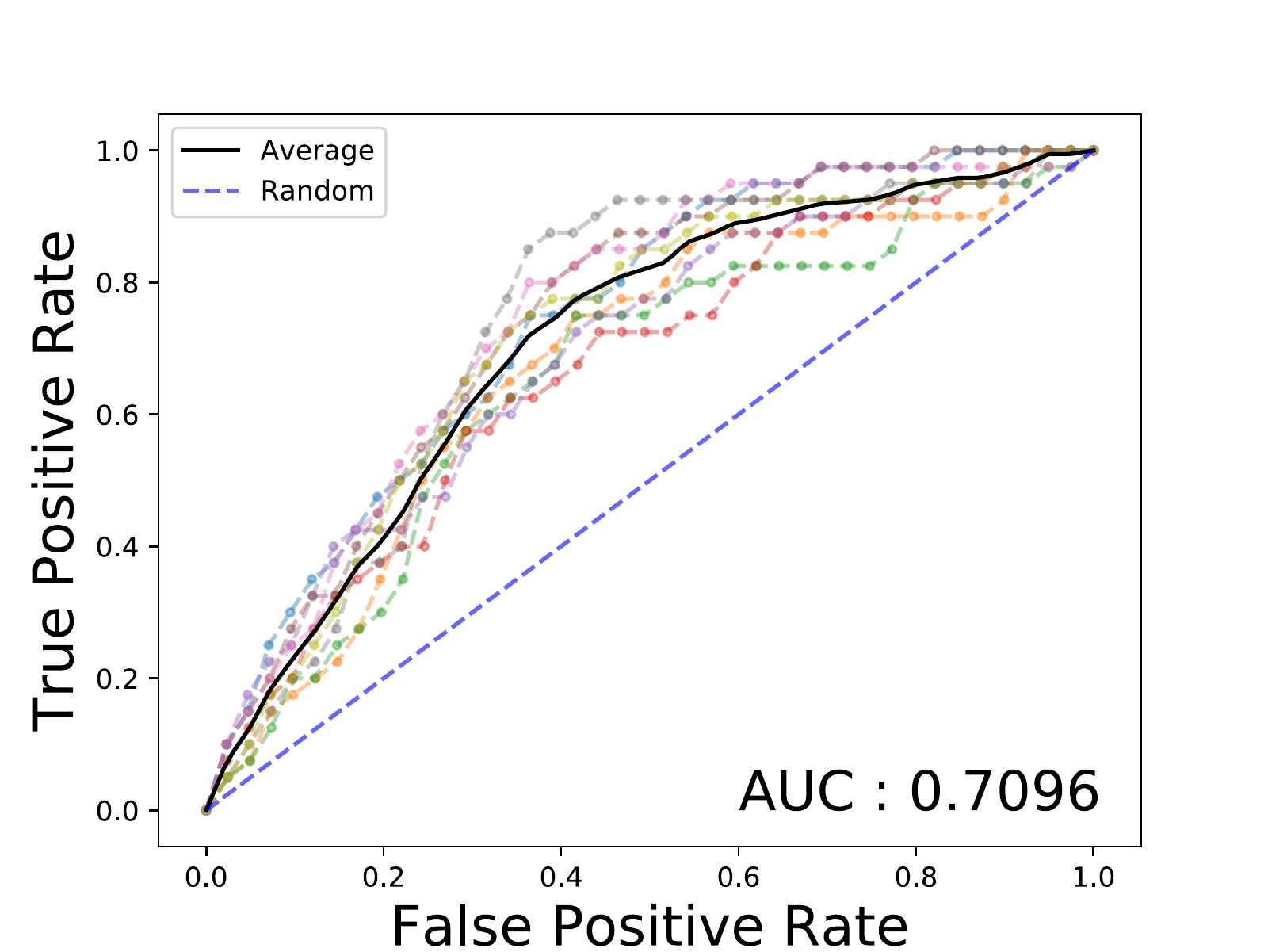}
    \caption{Population : 40}
\end{subfigure}
\begin{subfigure}[t]{0.22\textwidth}
    \includegraphics[width=4.0cm]{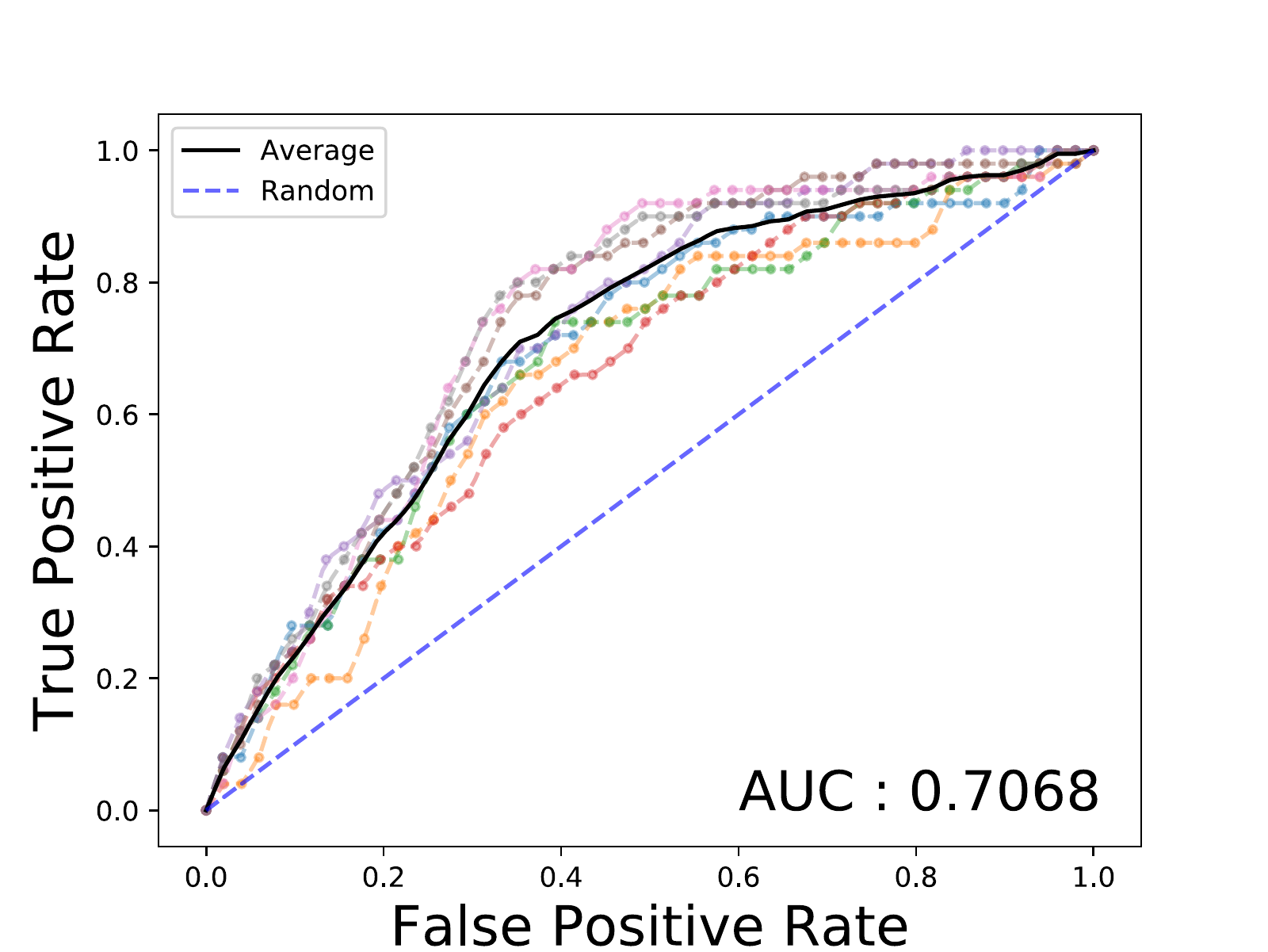}
    \caption{Population : 50}
\end{subfigure}
\begin{subfigure}[t]{0.22\textwidth}
    \includegraphics[width=4.0cm]{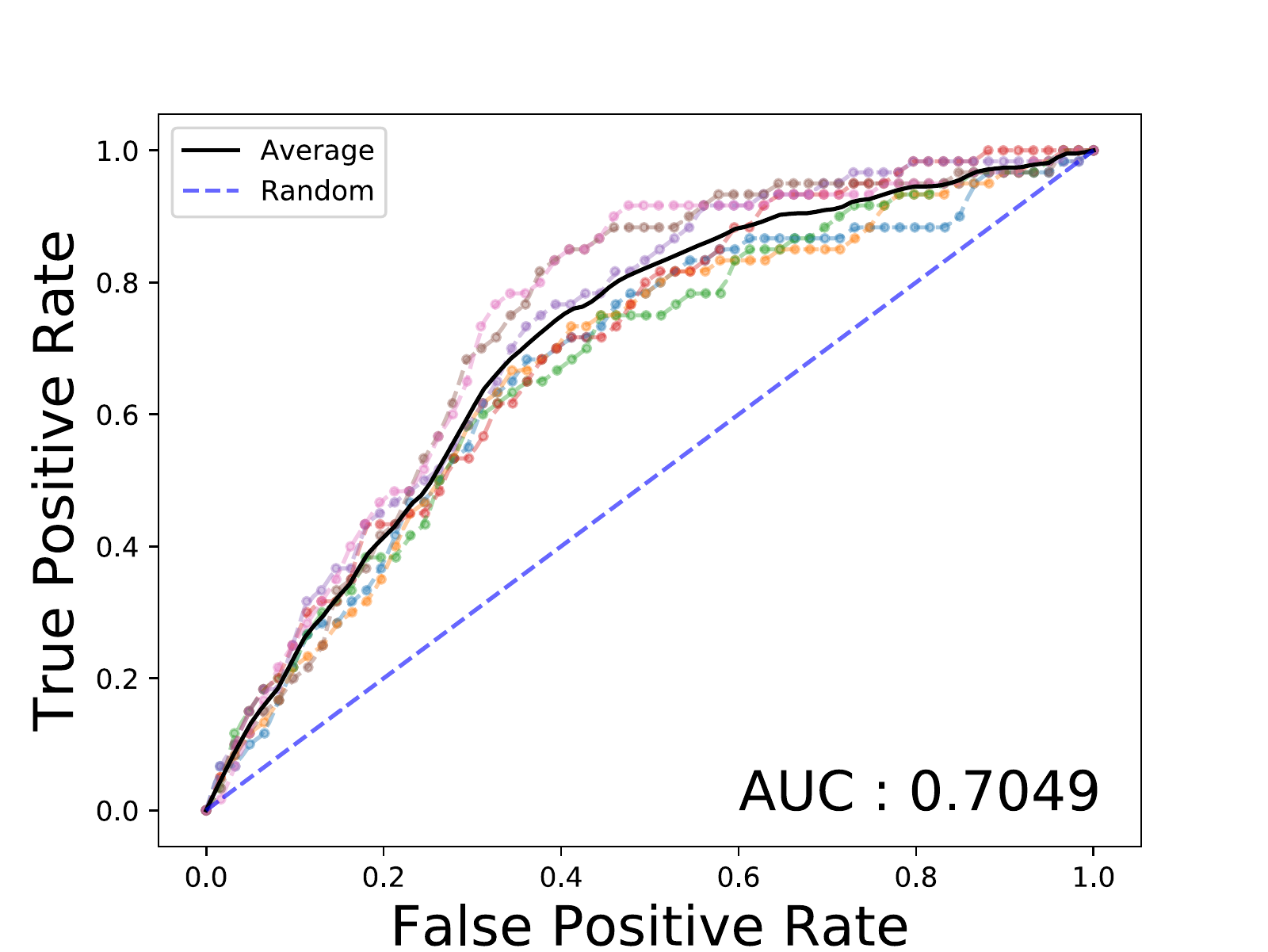}
    \caption{Population : 60}
\end{subfigure}
\begin{subfigure}[t]{0.22\textwidth}
    \includegraphics[width=4.0cm]{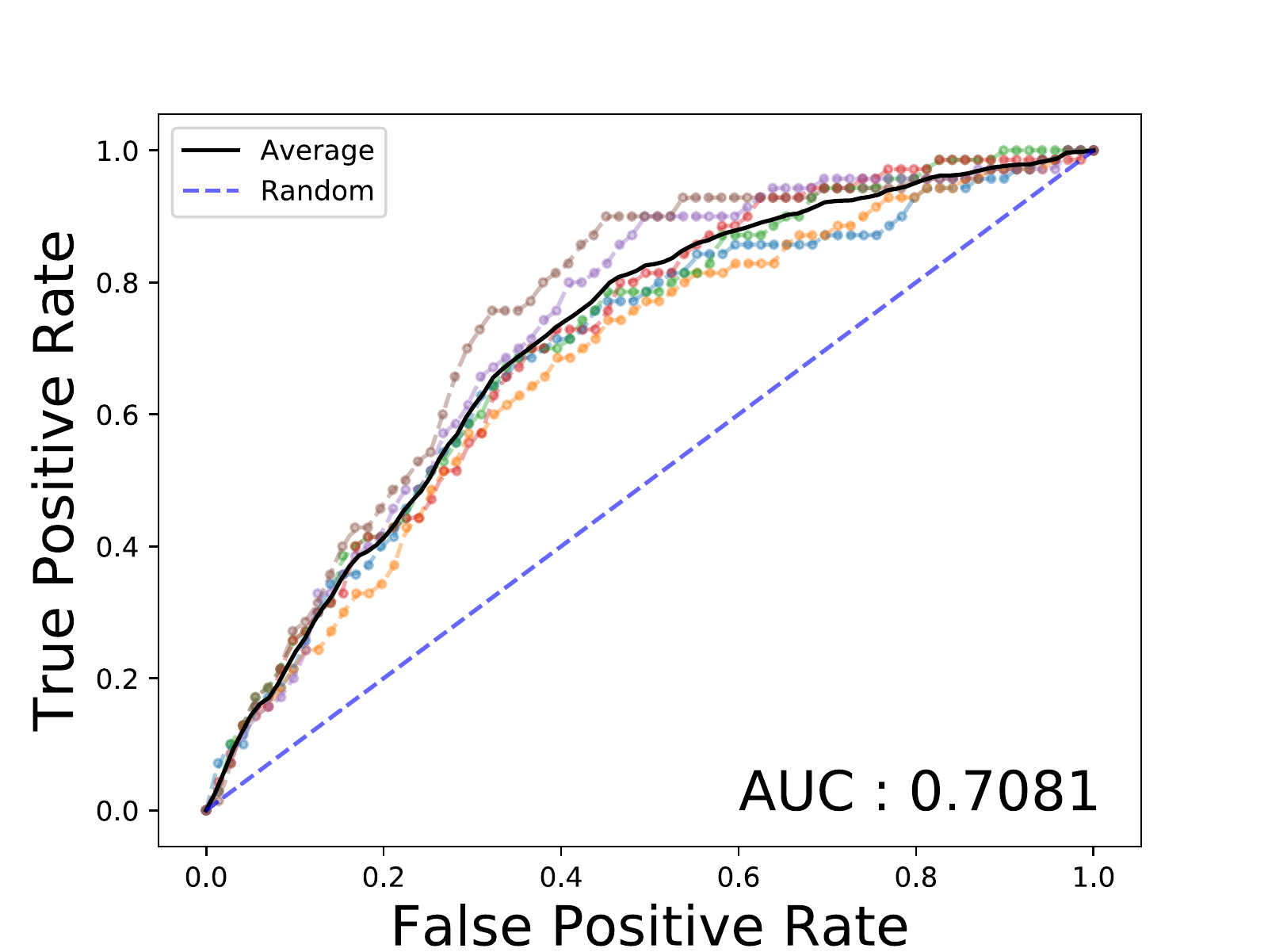}
    \caption{Population : 70}
\end{subfigure}
\begin{subfigure}[t]{0.22\textwidth}
    \includegraphics[width=4.0cm]{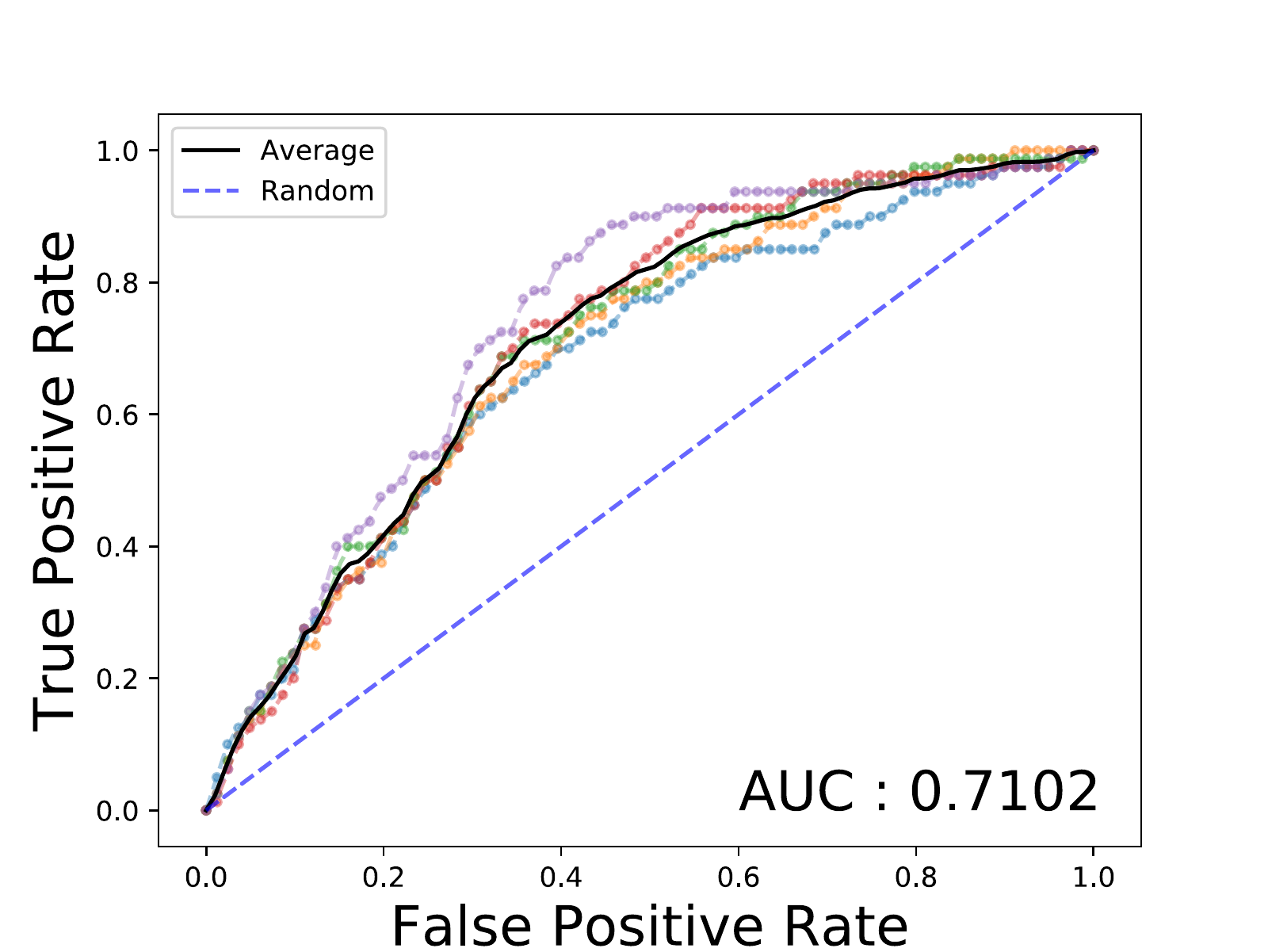}
    \caption{Population : 80}
\end{subfigure}
\begin{subfigure}[t]{0.22\textwidth}
    \includegraphics[width=4.0cm]{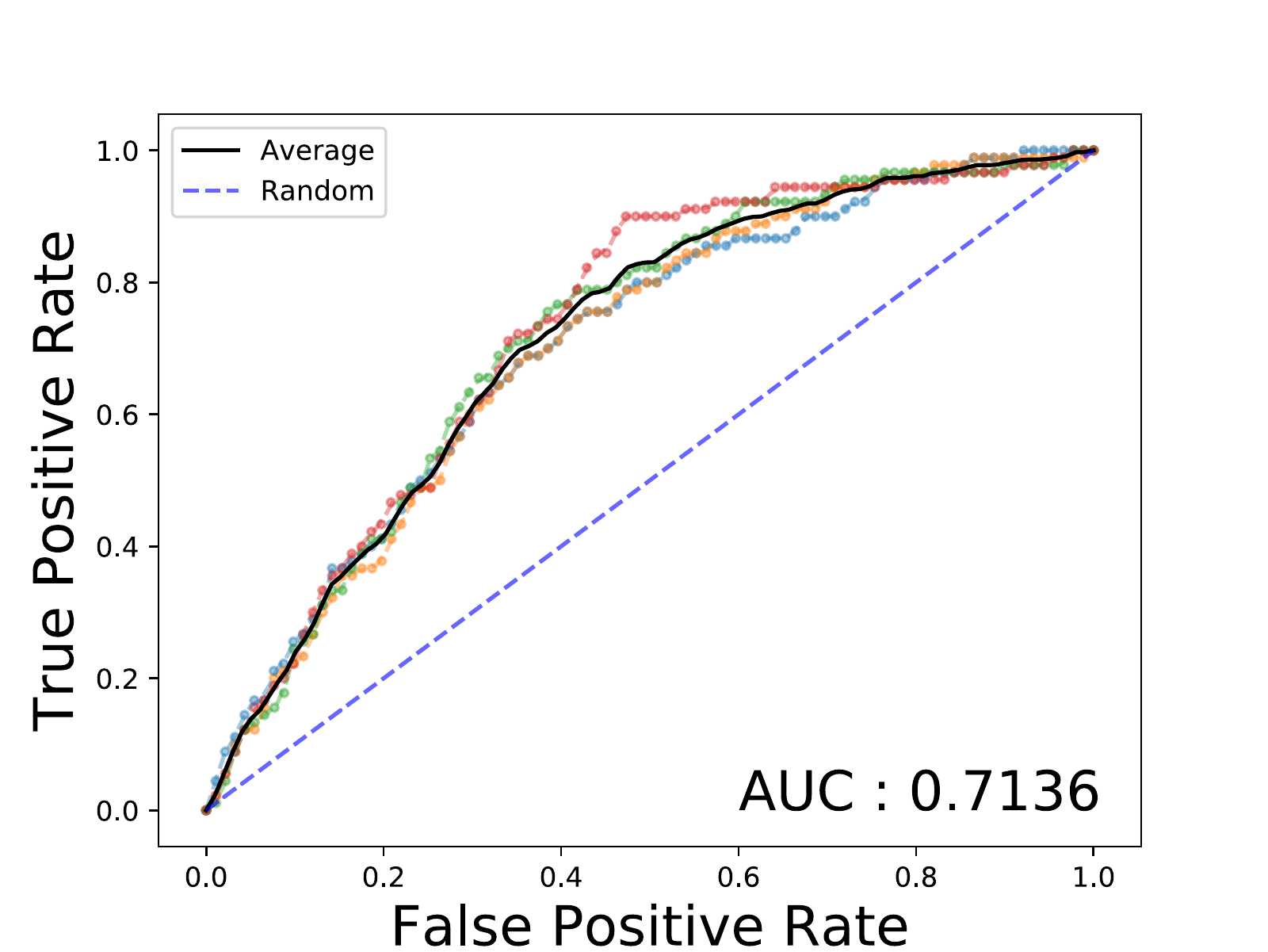}
    \caption{Population : 90}
\end{subfigure}
\begin{subfigure}[t]{0.22\textwidth}
    \includegraphics[width=4.0cm]{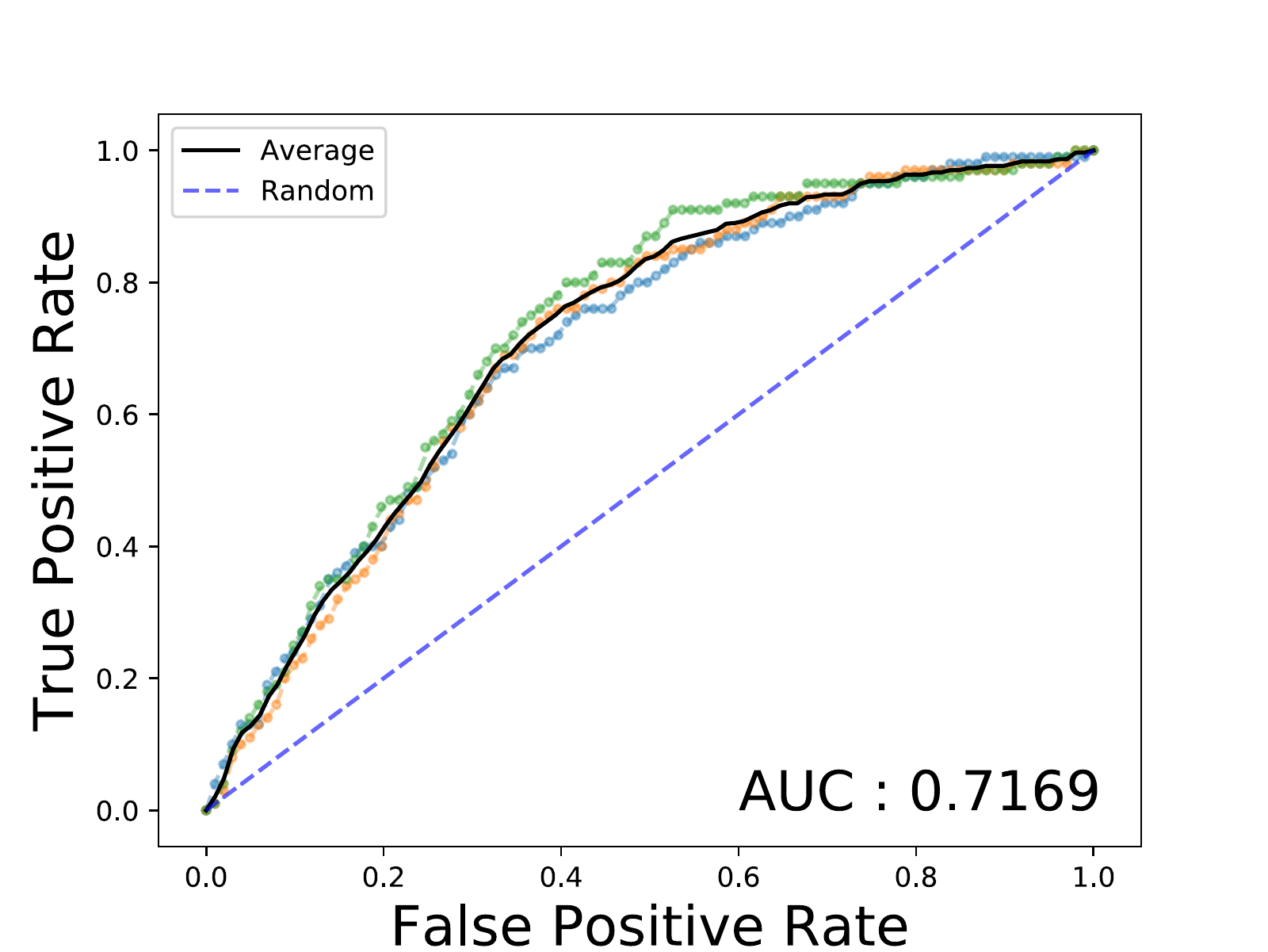}
    \caption{Population : 100}
\end{subfigure}
\begin{subfigure}[t]{0.22\textwidth}
    \includegraphics[width=4.0cm]{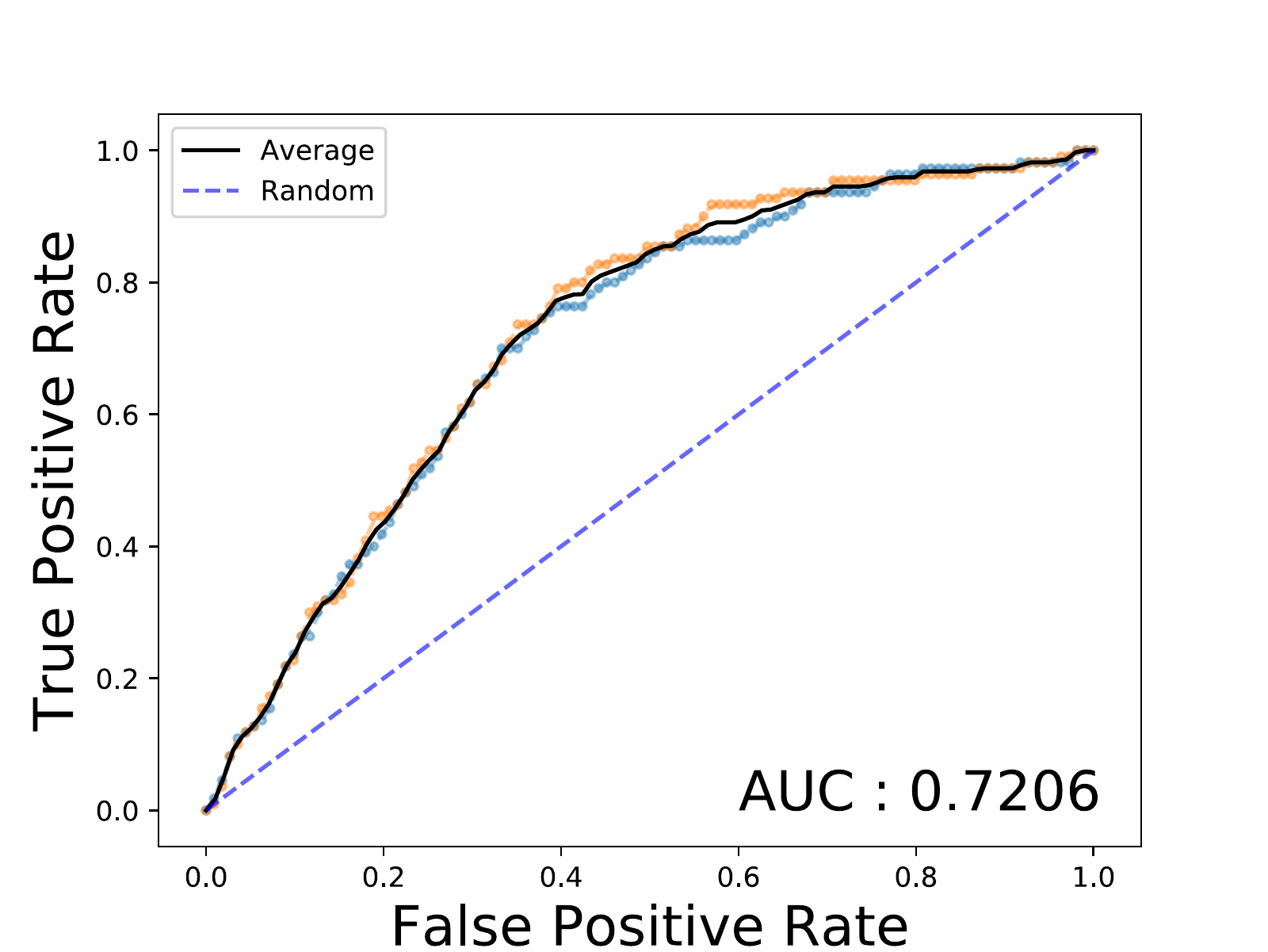}
    \caption{Population : 110}
\end{subfigure}
\begin{subfigure}[t]{0.22\textwidth}
    \includegraphics[width=4.0cm]{topk_auc_curves/topks_126_svg-tex.pdf}
    \caption{Population : 126}
\end{subfigure}
\begin{subfigure}[t]{0.22\textwidth}
    \includegraphics[width=4.0cm]{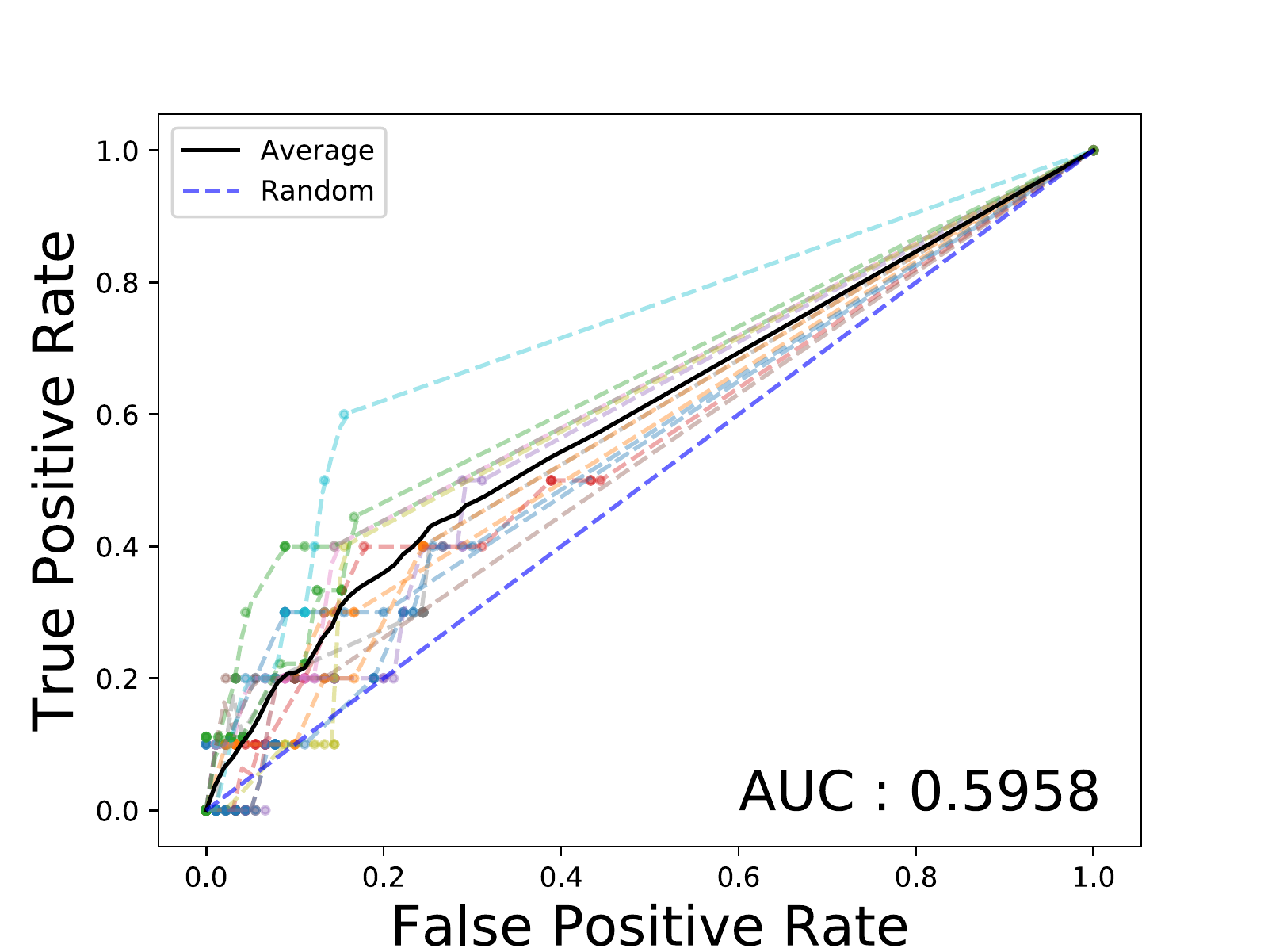}
    \caption{Population : 10}
\end{subfigure}
\begin{subfigure}[t]{0.22\textwidth}
    \includegraphics[width=4.0cm]{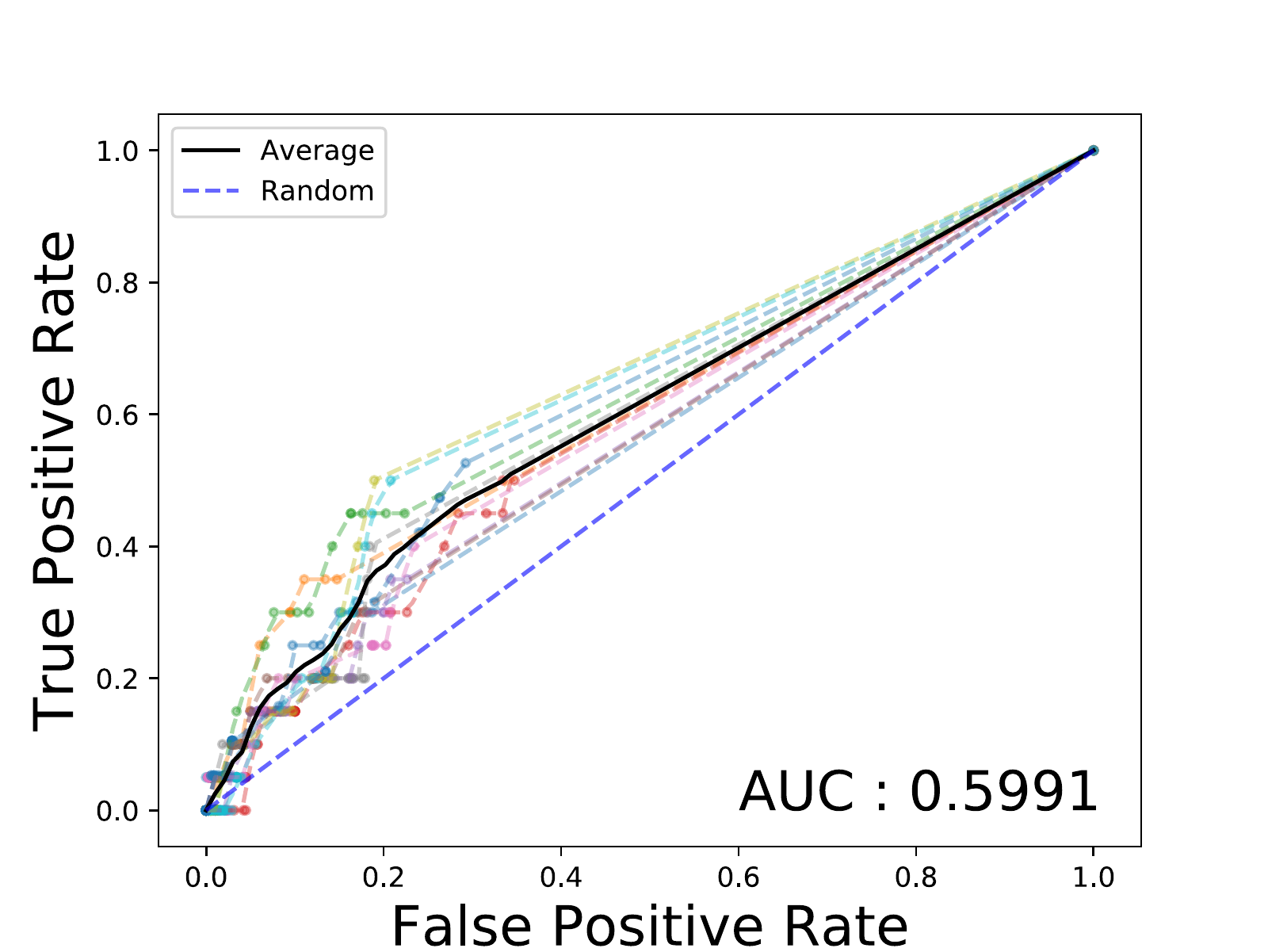}
    \caption{Population : 20}
\end{subfigure}
\begin{subfigure}[t]{0.22\textwidth}
    \includegraphics[width=4.0cm]{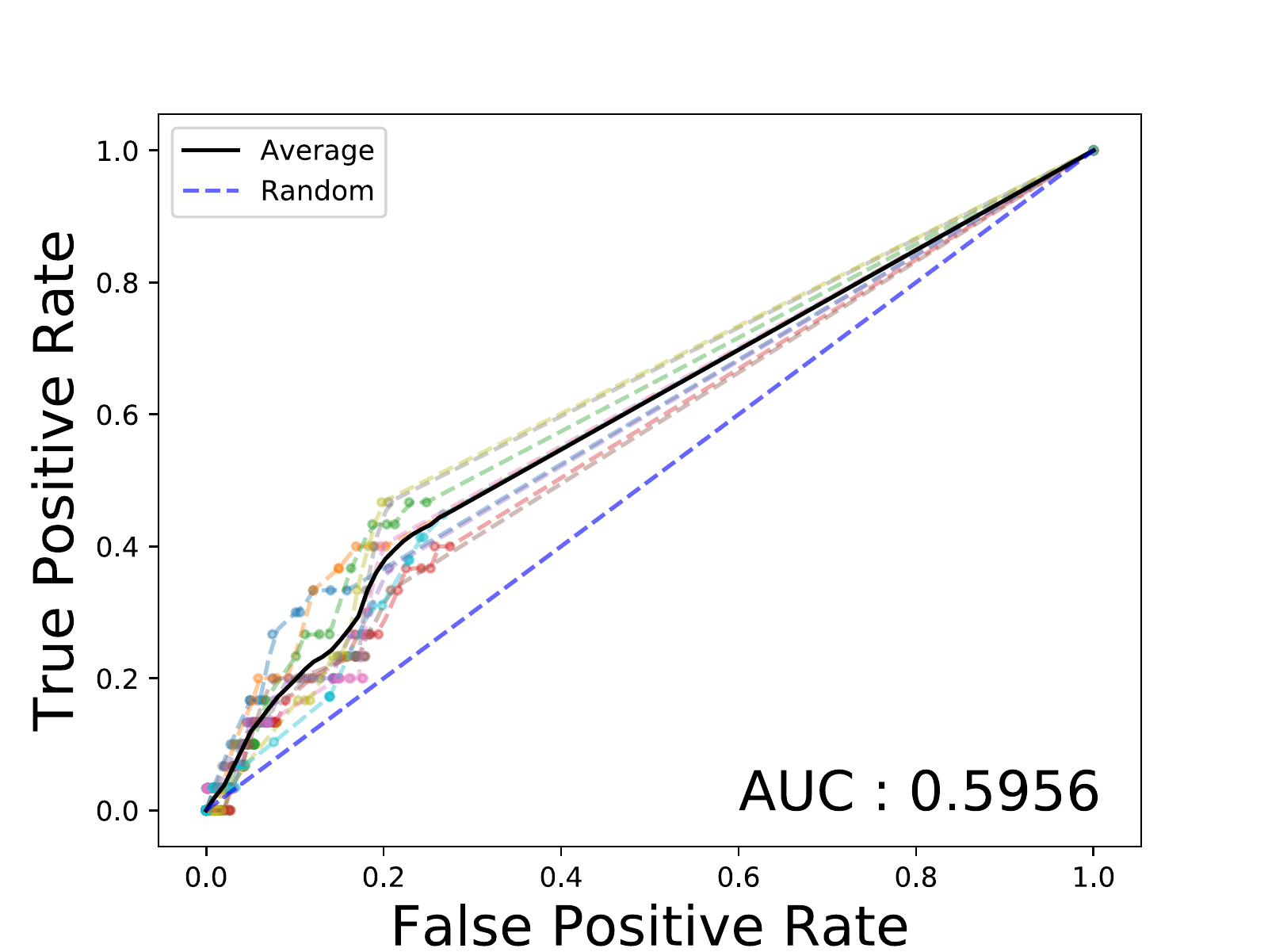}
    \caption{Population : 30}
\end{subfigure}
\begin{subfigure}[t]{0.22\textwidth}
    \includegraphics[width=4.0cm]{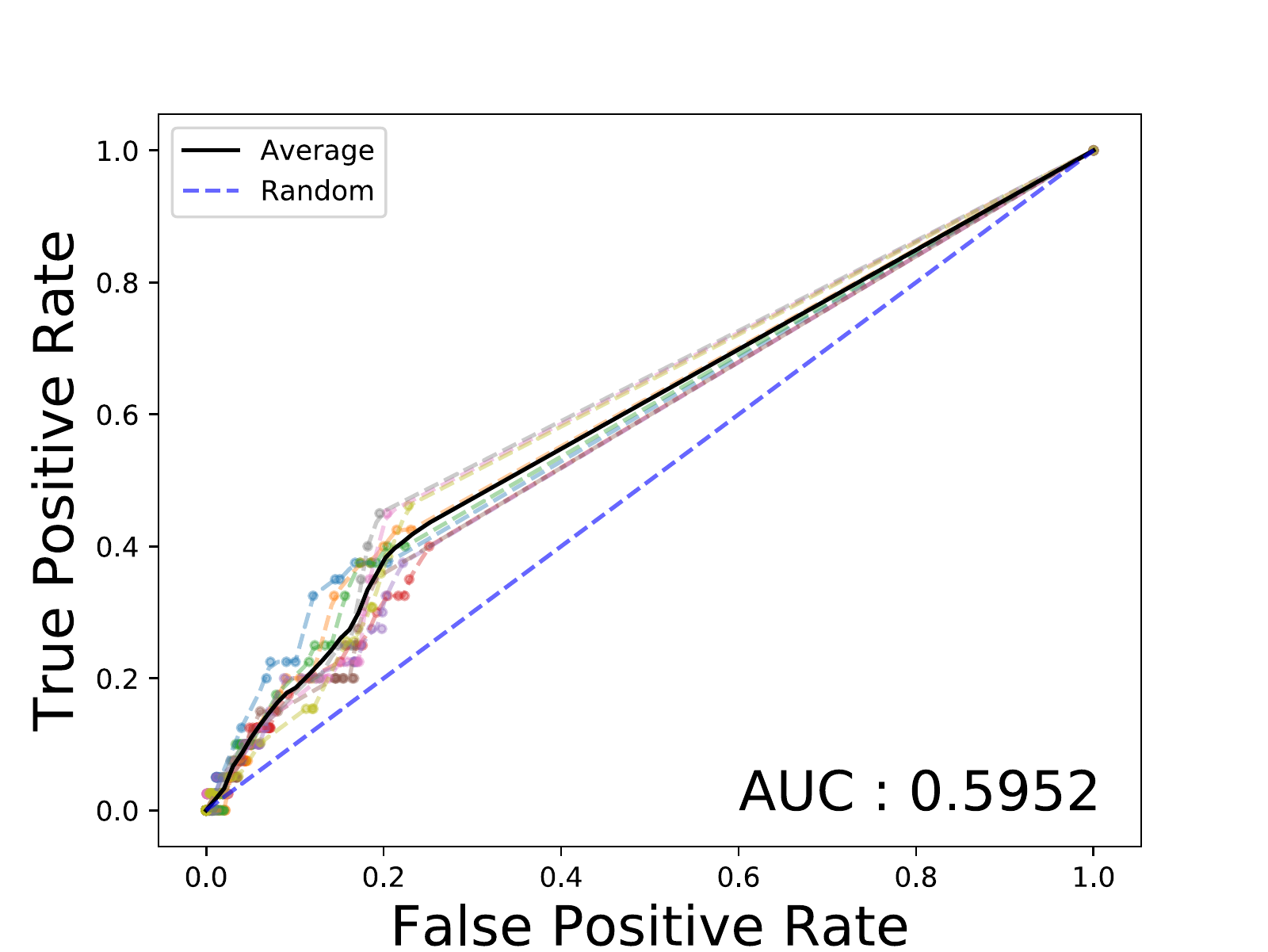}
    \caption{Population : 40}
\end{subfigure}
\begin{subfigure}[t]{0.22\textwidth}
    \includegraphics[width=4.0cm]{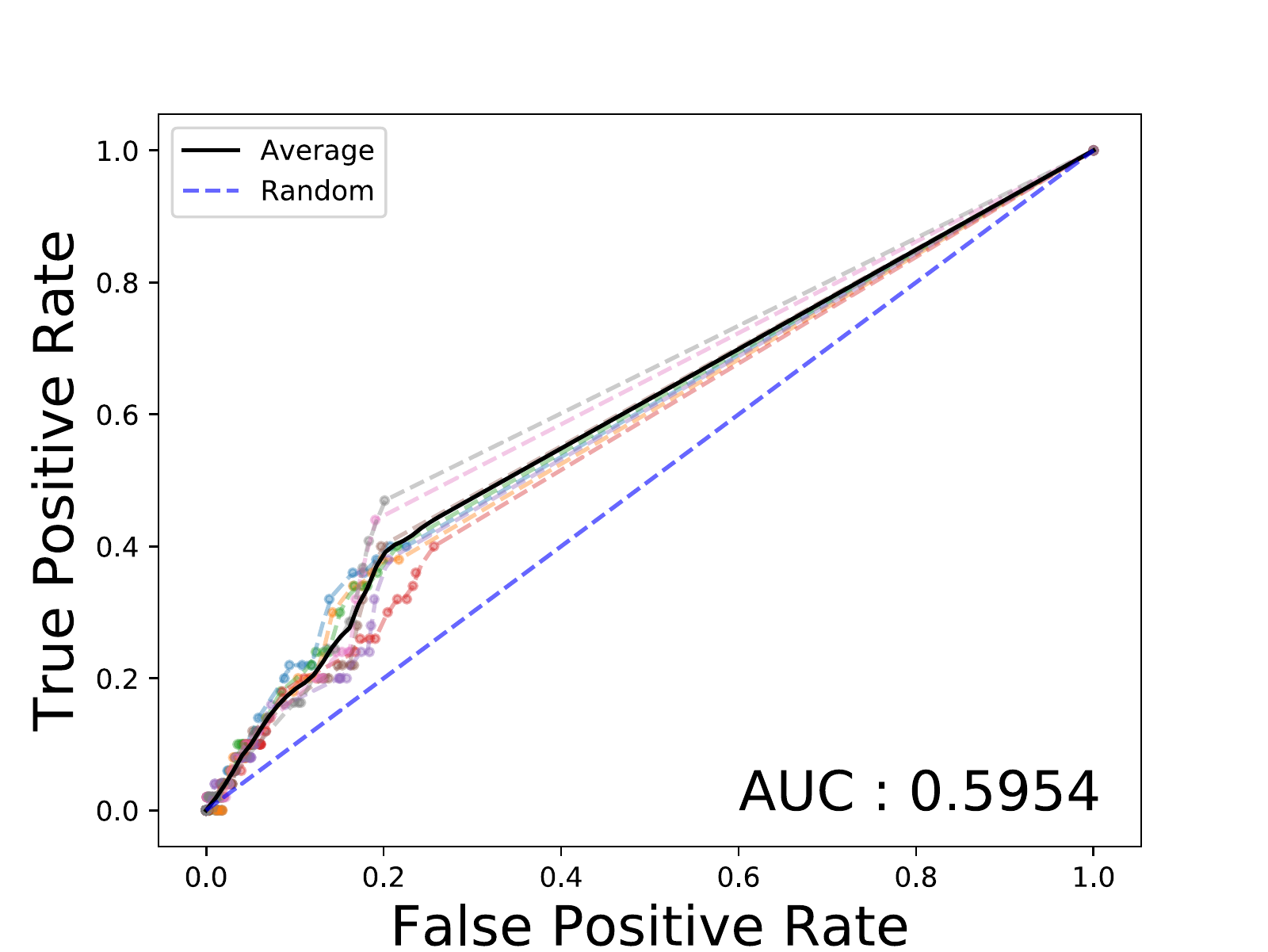}
    \caption{Population : 50}
\end{subfigure}
\begin{subfigure}[t]{0.22\textwidth}
    \includegraphics[width=4.0cm]{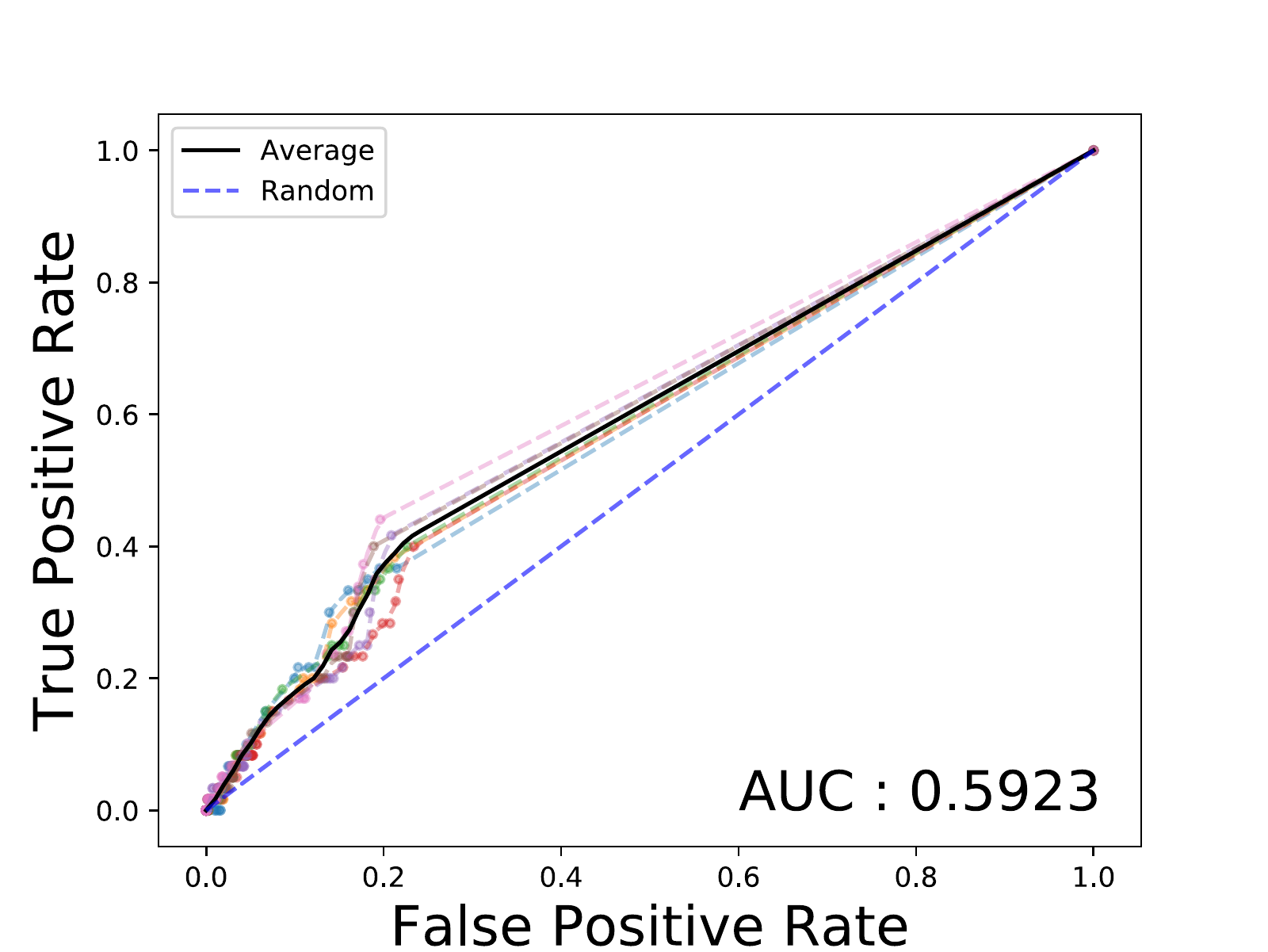}
    \caption{Population : 60}
\end{subfigure}
\begin{subfigure}[t]{0.22\textwidth}
    \includegraphics[width=4.0cm]{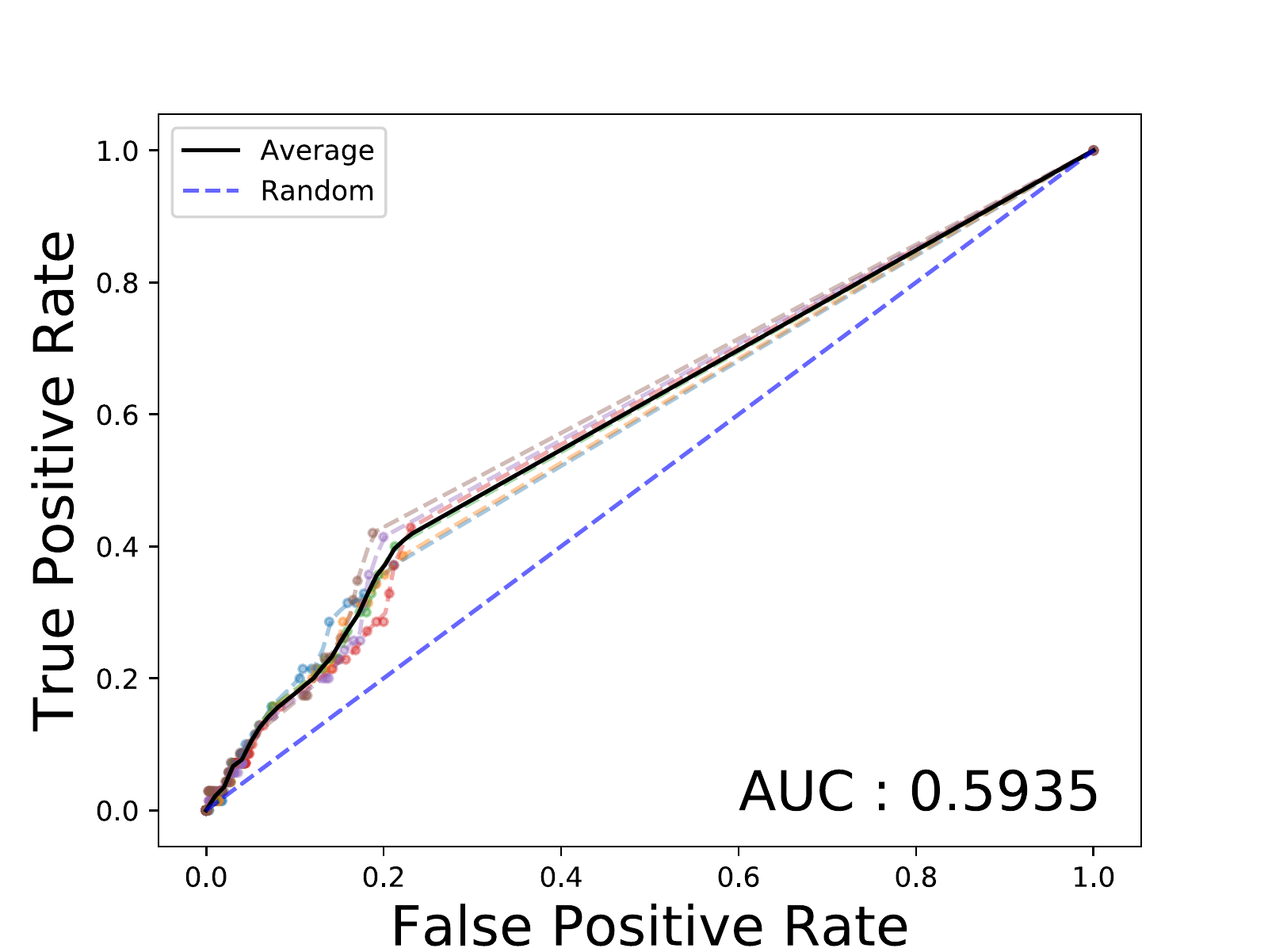}
    \caption{Population : 70}
\end{subfigure}
\begin{subfigure}[t]{0.22\textwidth}
    \includegraphics[width=4.0cm]{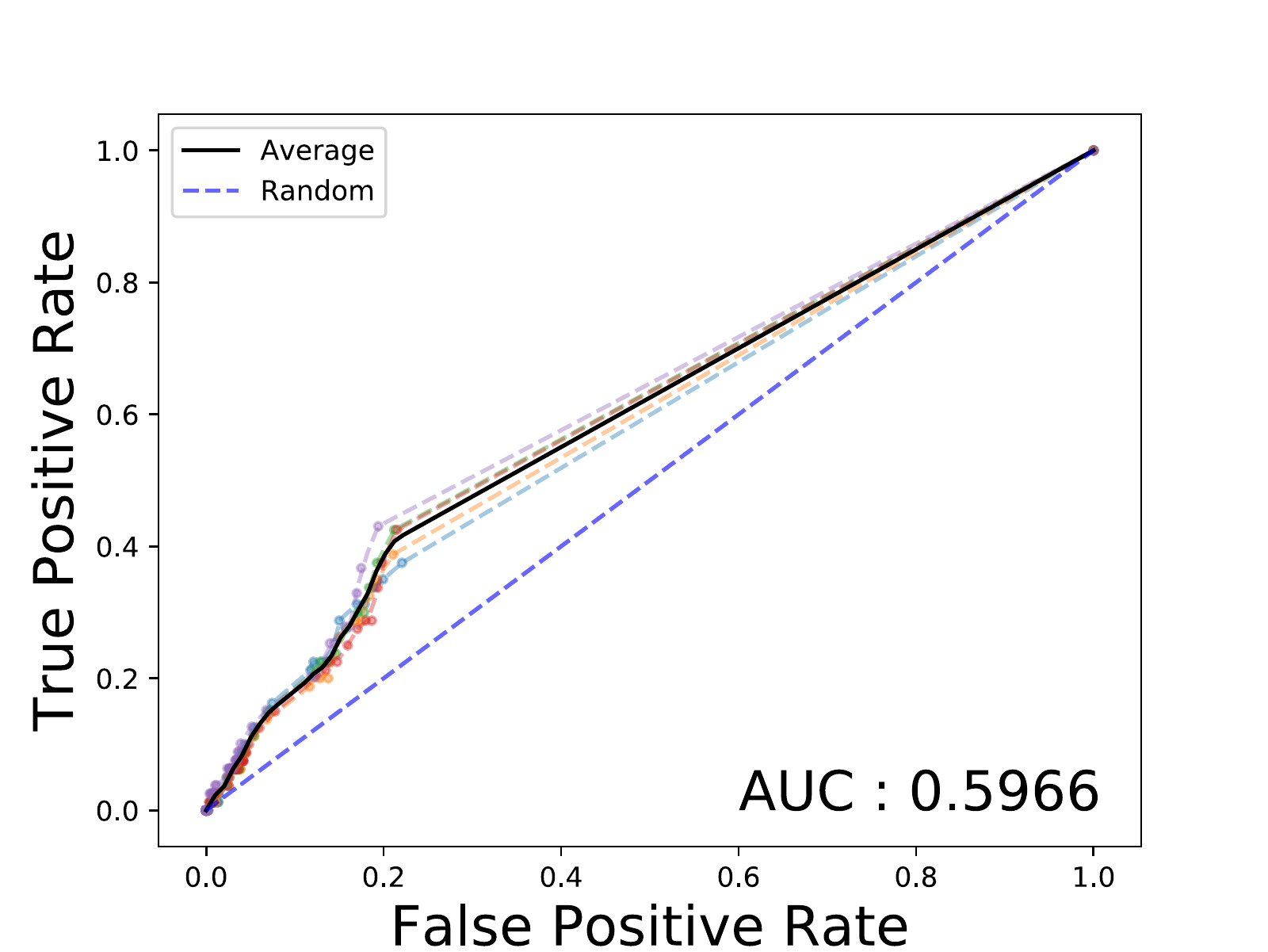}
    \caption{Population : 80}
\end{subfigure}
\begin{subfigure}[t]{0.22\textwidth}
    \includegraphics[width=4.0cm]{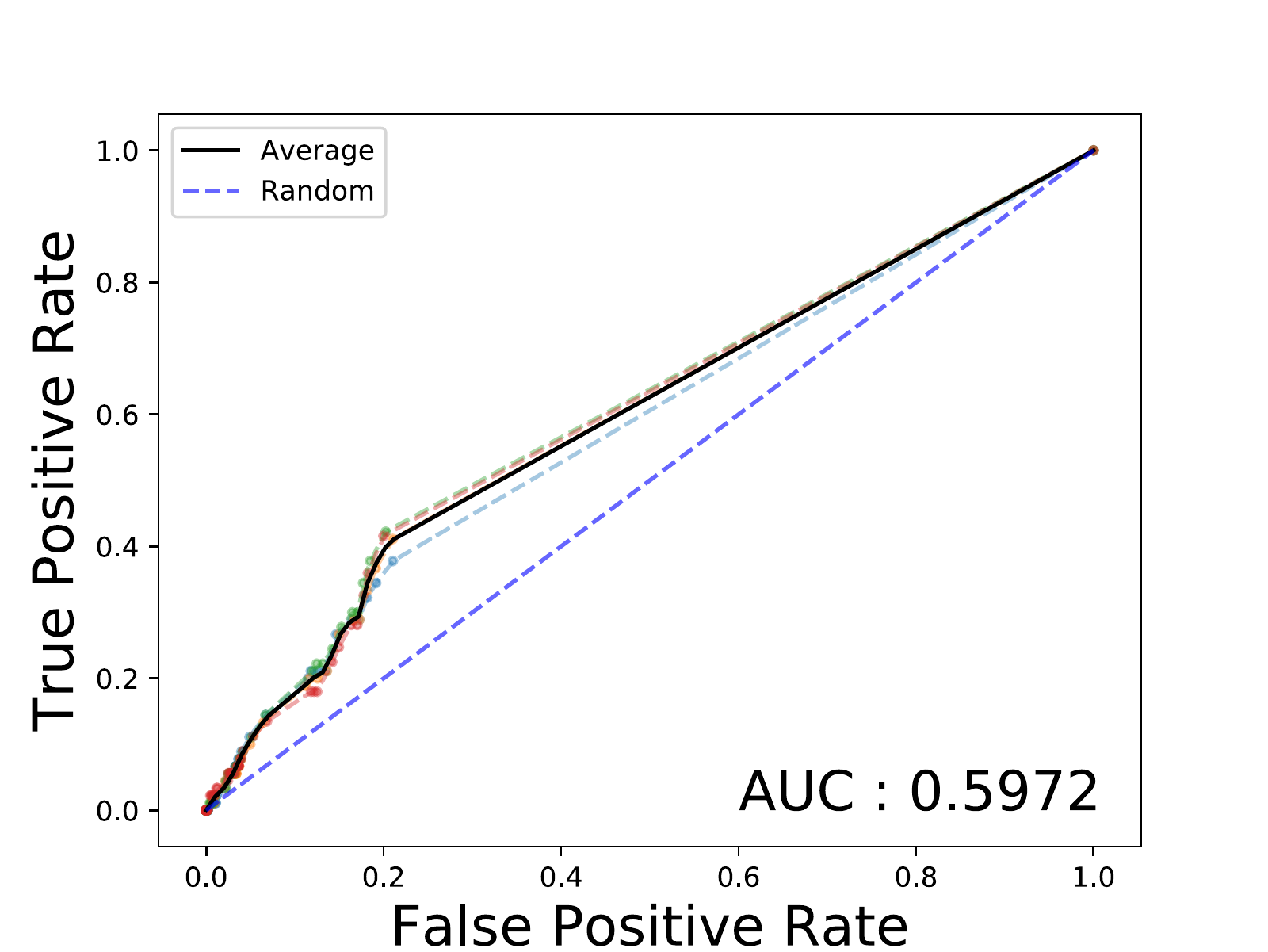}
    \caption{Population : 90}
\end{subfigure}
\begin{subfigure}[t]{0.22\textwidth}
    \includegraphics[width=4.0cm]{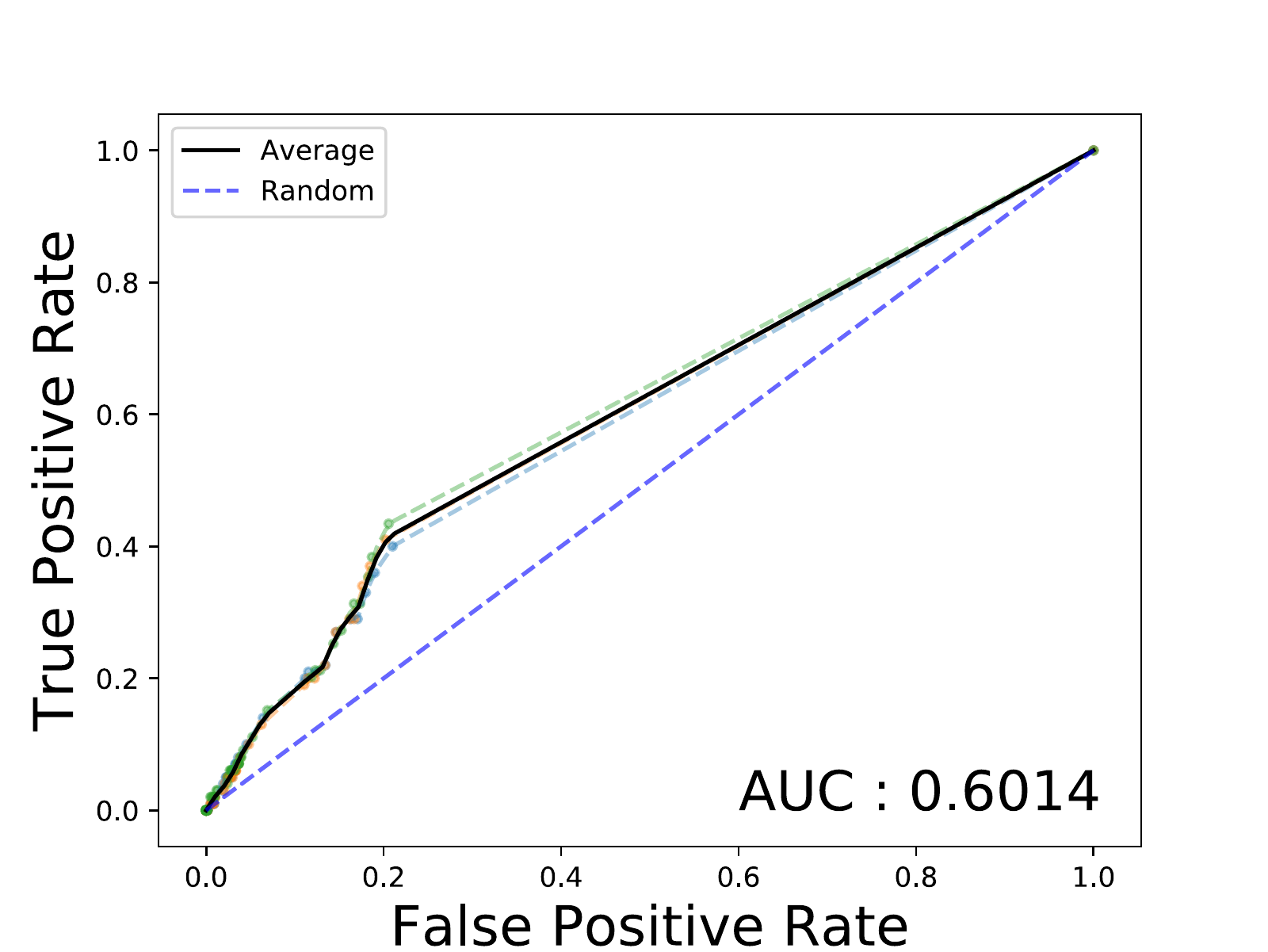}
    \caption{Population : 100}
\end{subfigure}
\begin{subfigure}[t]{0.22\textwidth}
    \includegraphics[width=4.0cm]{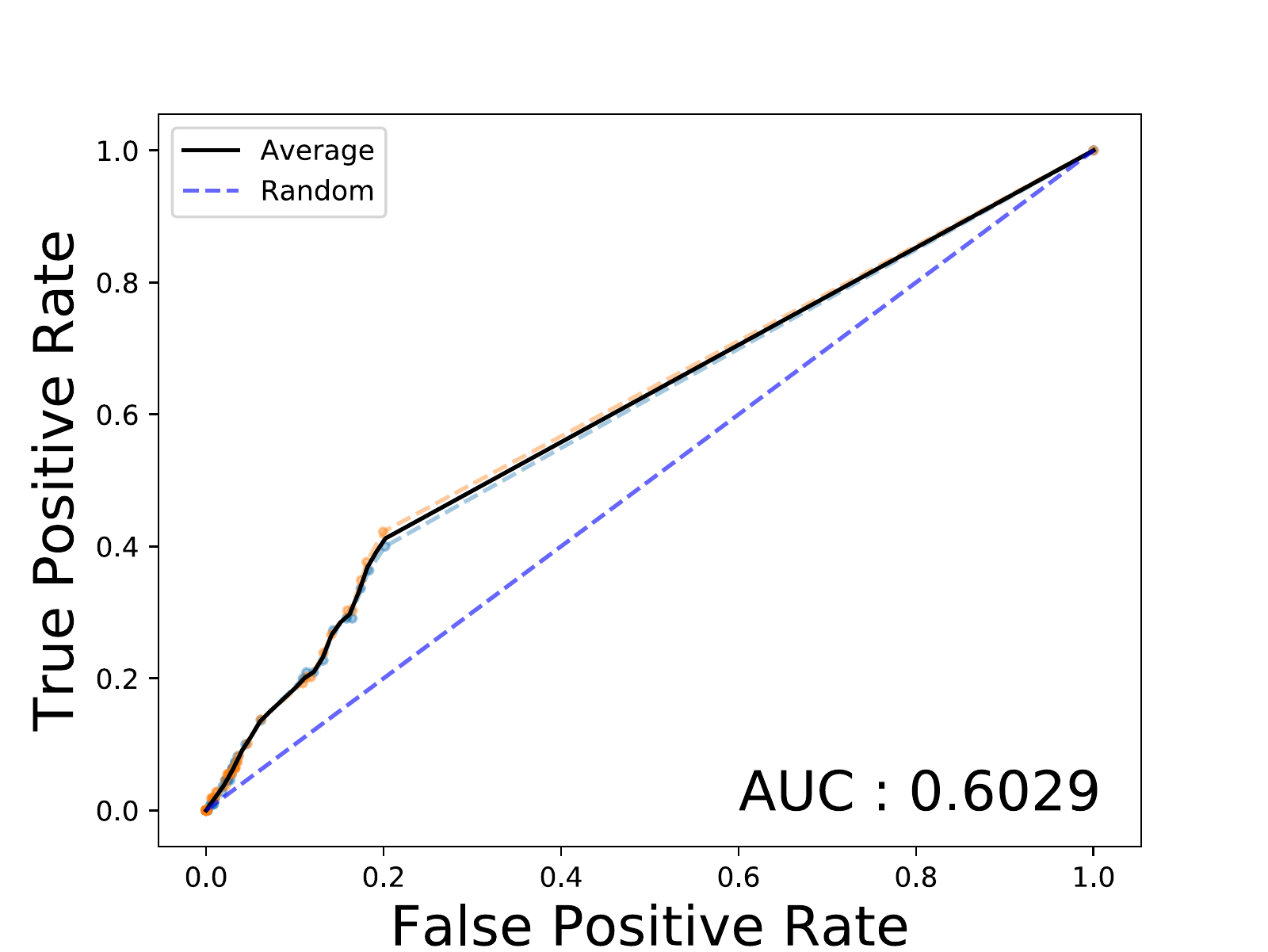}
    \caption{Population : 110}
\end{subfigure}
\begin{subfigure}[t]{0.22\textwidth}
    \includegraphics[width=4.0cm]{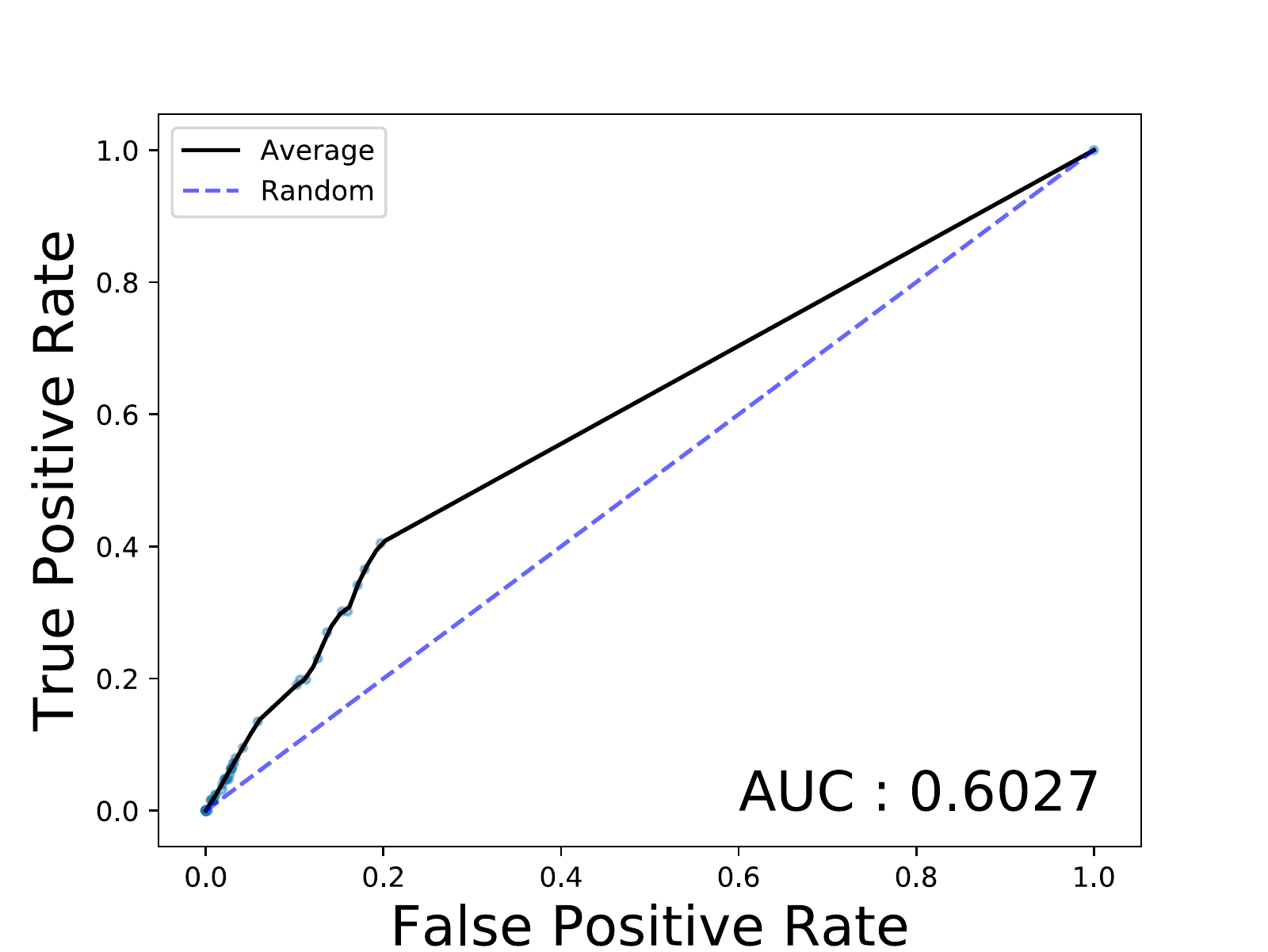}
    \caption{Population : 126}
\end{subfigure}
\caption{\textbf{(a)-(l)}: Receiver Operating Characteristic curves for various population sizes, when top $k$ entries in the sorted list of DNA sequences per image are predicted to be matches. \textbf{(m)-(x)}: Receiver Operating Characteristic curves for various population sizes, when predictions are made independently for each image-DNA pair.}
\label{fig:indeps_roc}
\end{figure}

When predicting that an image matches with a particular genome, we can use thresholds to make predictions in two different ways. The first method is by thresholding with an integral value $k$, where we predict a positive match when a selected genome is in the top-$k$ potential matches by likelihood, given an image. Alternatively, we can predict a positive match for each image-DNA pair if the matching likelihood is above a real valued threshold $\theta$. These approaches complement each other, in the sense that in the former case, we understand precision-recall tradeoff in terms of narrowing potential matches down to a likely subpopulation, whereas the latter allows us to understand the performance of making independent predictions for image-DNA pairs. Supplementary Fig.~\ref{fig:indeps_roc} (a)-(l) show the precision recall curves for various population sizes, when $k$ is increased from $1$, where a single match is predicted, to the population size where everyone is predicted to be a match. Similarly, Supplementary Fig.~\ref{fig:indeps_roc} (m)-(x) show ROC curves for various population sizes, when the threshold $\theta$ is increased from $0$ to $1$ in real-valued increments. In both cases, we observe that the classifier performs better than a random-guessing baseline, although the area under the curve remains relatively low, around $0.7$ for the top-$k$ method, and $0.6$ when making independent predictions.

\section{SVM - Matching as Binary Classification}
For the sake of completeness, we also study the use of classical machine learning methods to predict matches between images and DNA, given a vector representing the likelihood of phenotypes detected in an image corresponding to a selected genome. We train a linear SVM with equal number of true and false matches, selected from subsets of the 126 individuals (10-fold cross-validated). From results in Supplementary Fig.~\ref{fig:hair_eye_skin_1} (a)-(c), we see that this approach does not contribute much to matching accuracy, likely arising from limited signal present in the small dataset. Linear SVM was found to outperform SVM with a non-linear (rbf) kernel, as well as several other learning methods.

\begin{figure}[]
\centering
\begin{subfigure}[t]{0.32\textwidth}
    \includegraphics[width=5.6cm]{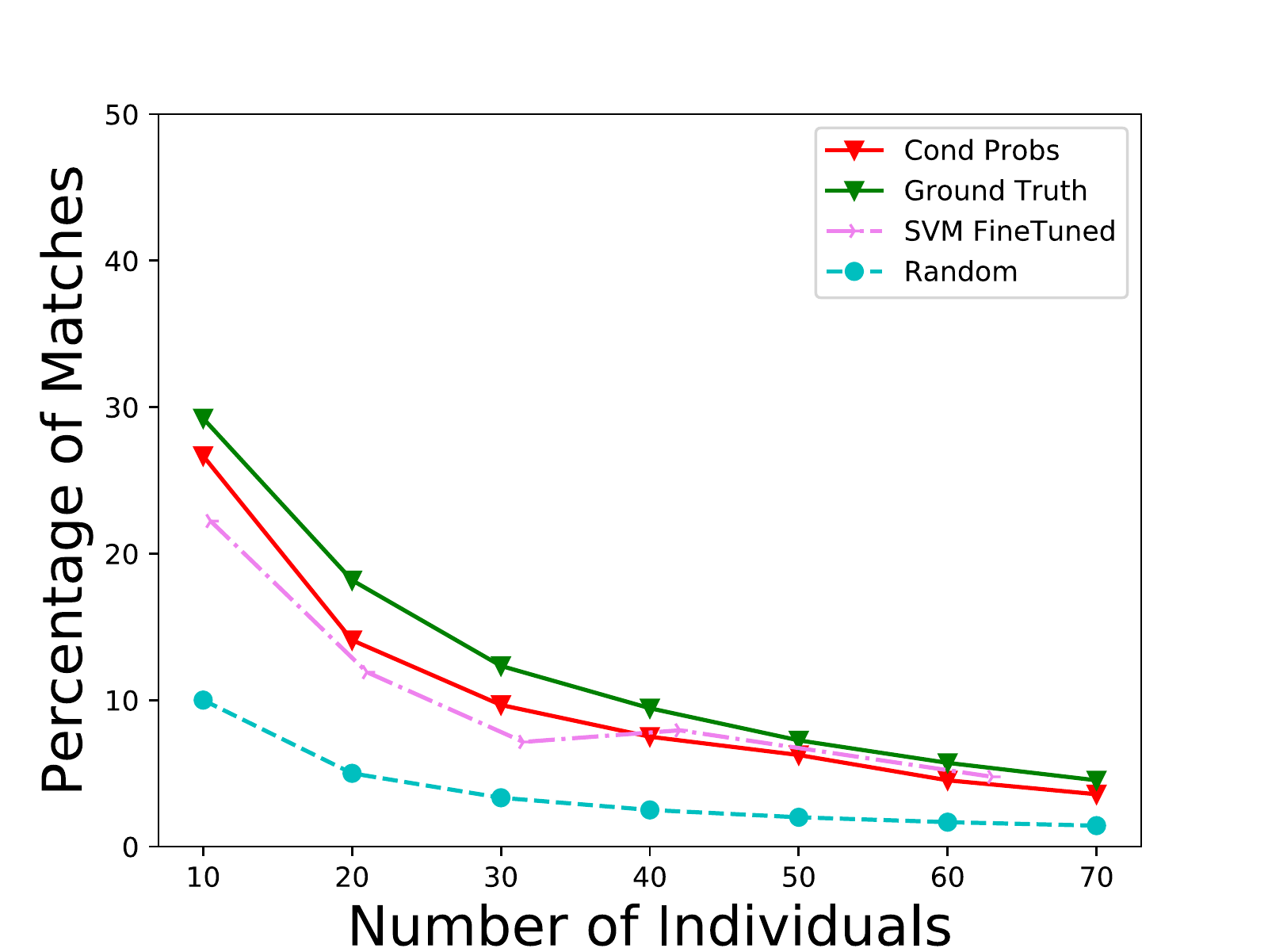}
    \caption{Top 1}
\end{subfigure}
\begin{subfigure}[t]{0.32\textwidth}
    \includegraphics[width=5.6cm]{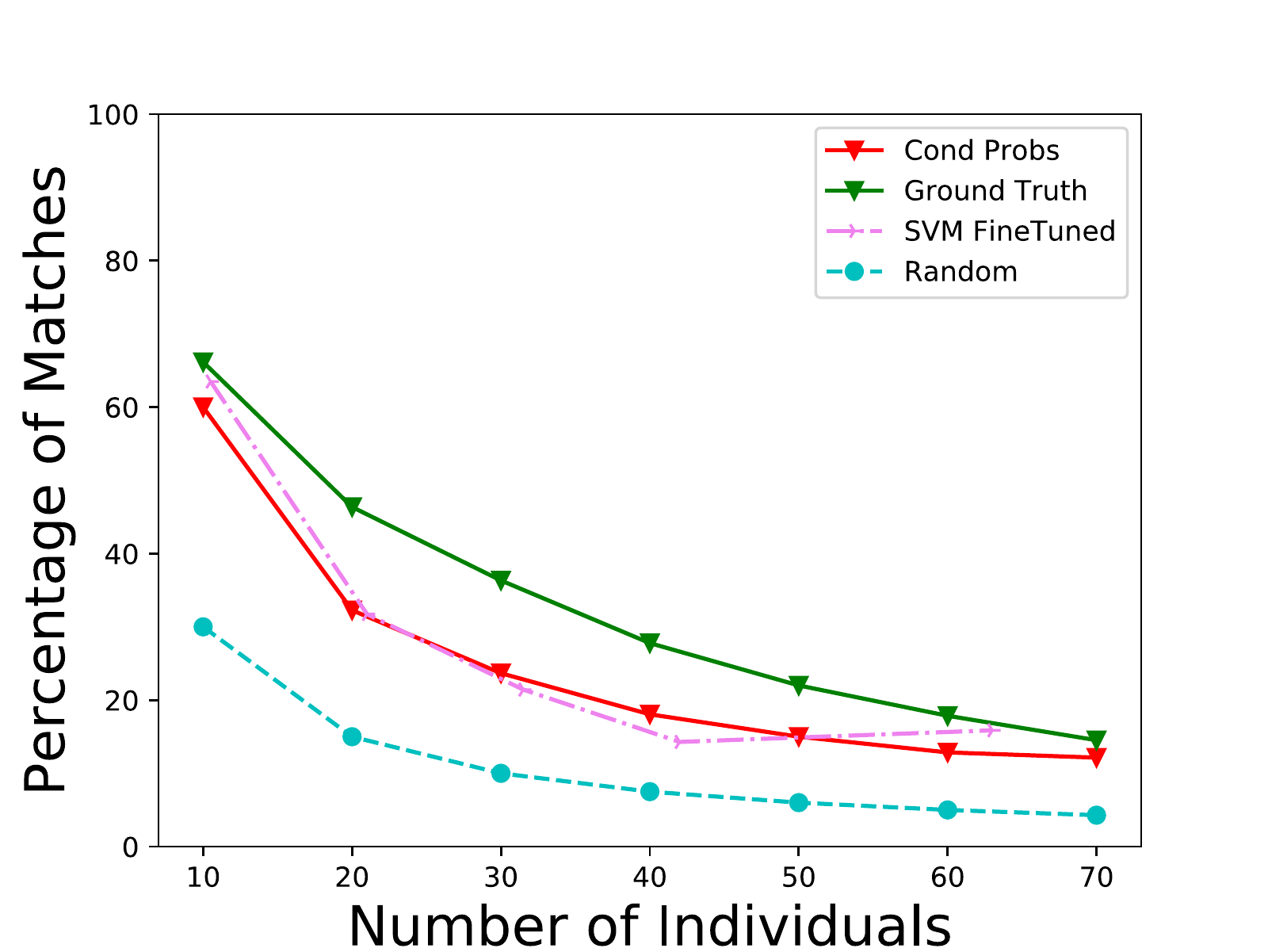}
    \caption{Top 3}
\end{subfigure}
\begin{subfigure}[t]{0.32\textwidth}
    \includegraphics[width=5.6cm]{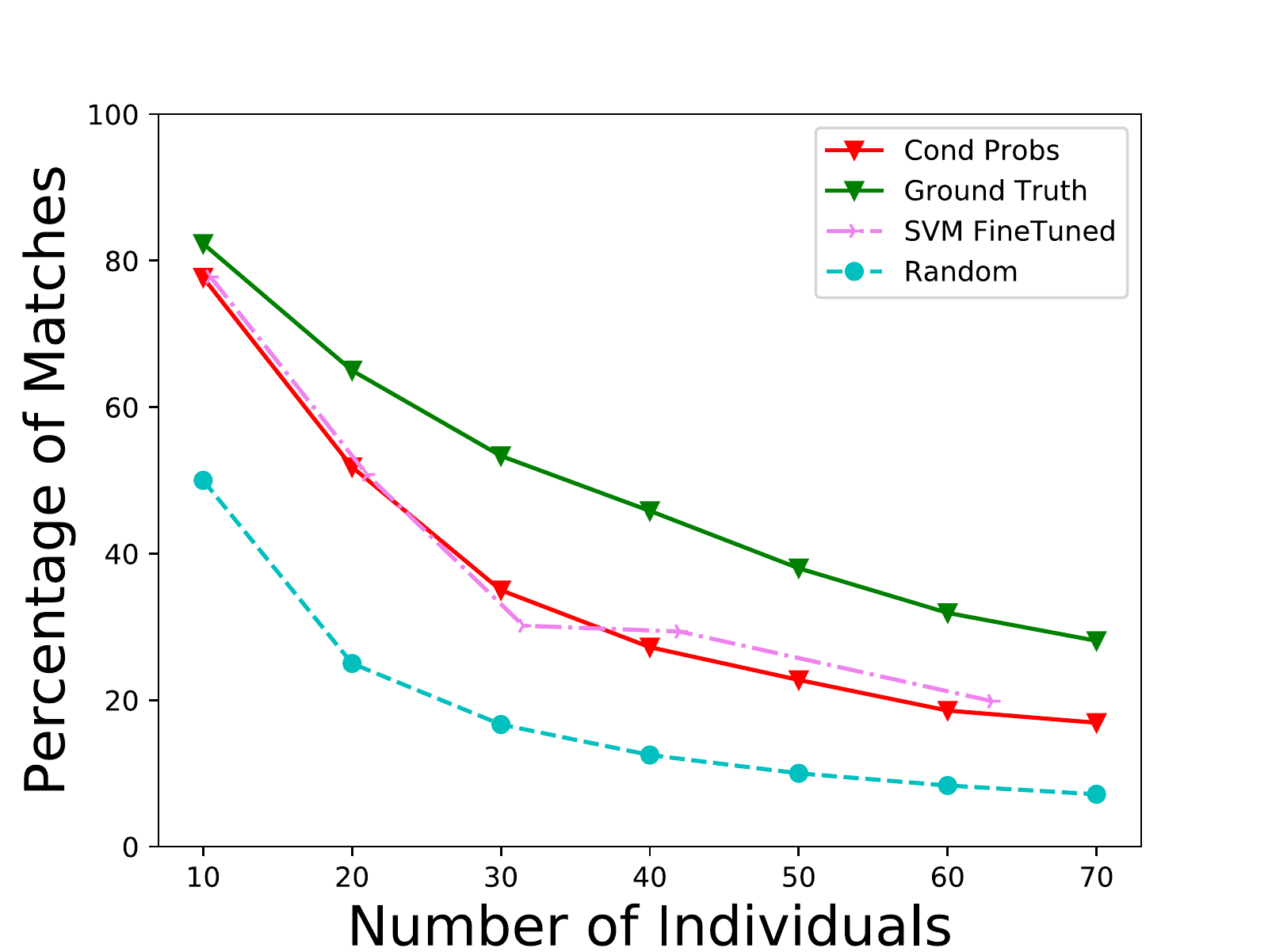}
    \caption{Top 5}
\end{subfigure}

\begin{subfigure}[t]{0.32\textwidth}
    \includegraphics[width=5.6cm]{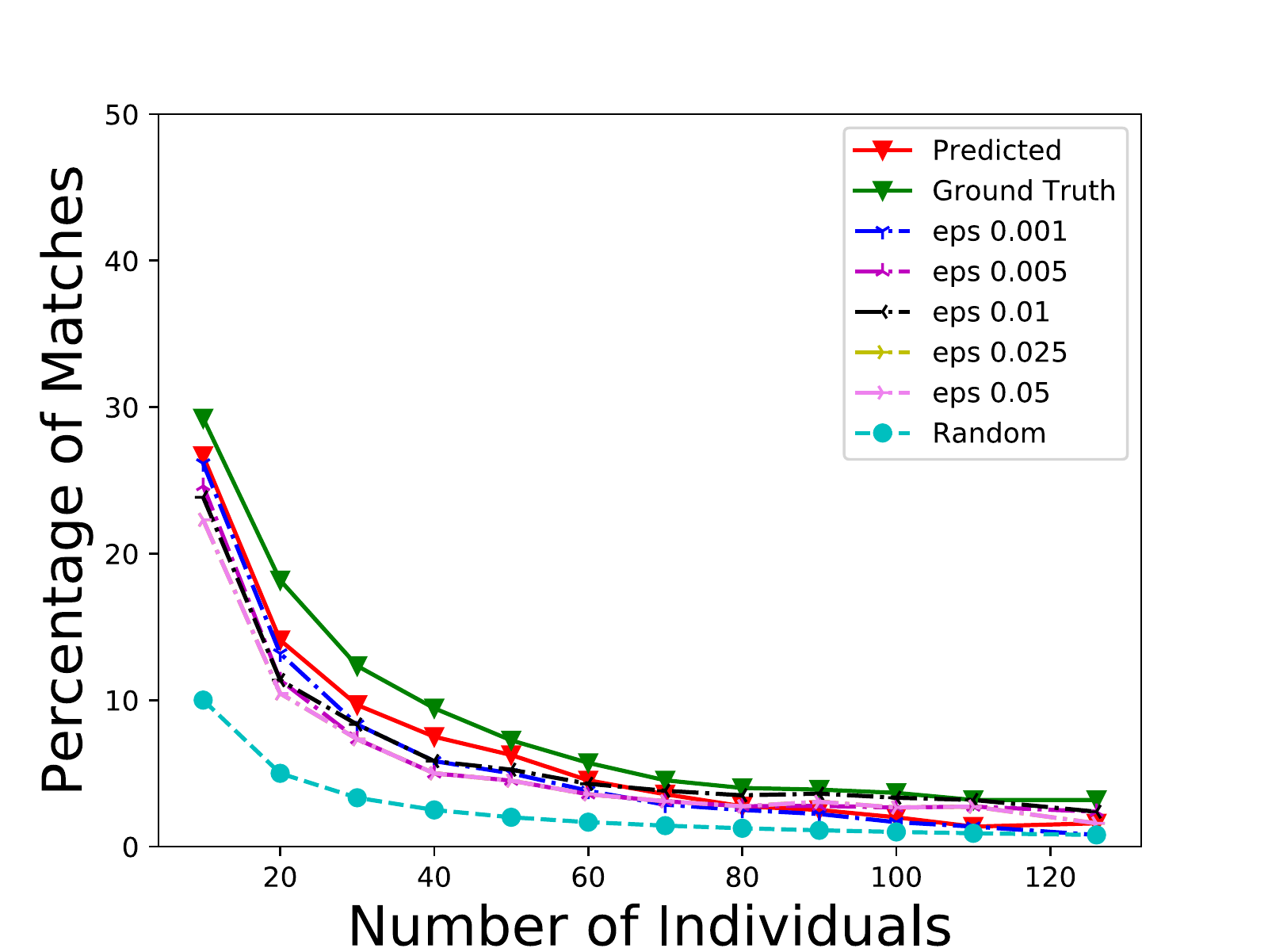}
    \caption{Eye Color}
\end{subfigure}
\begin{subfigure}[t]{0.32\textwidth}
    \includegraphics[width=5.6cm]{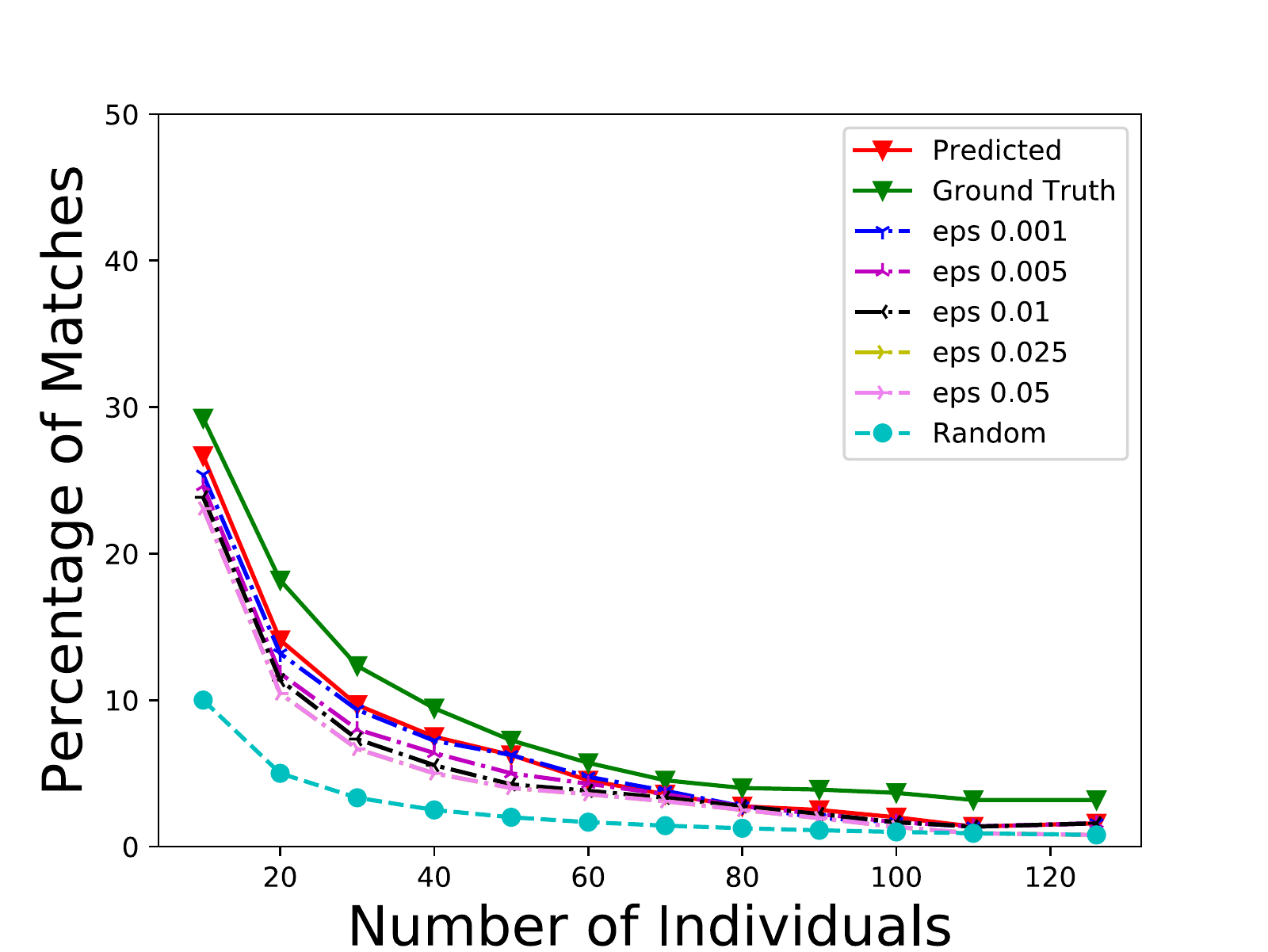}
    \caption{Skin Color}
\end{subfigure}
\begin{subfigure}[t]{0.32\textwidth}
    \includegraphics[width=5.6cm]{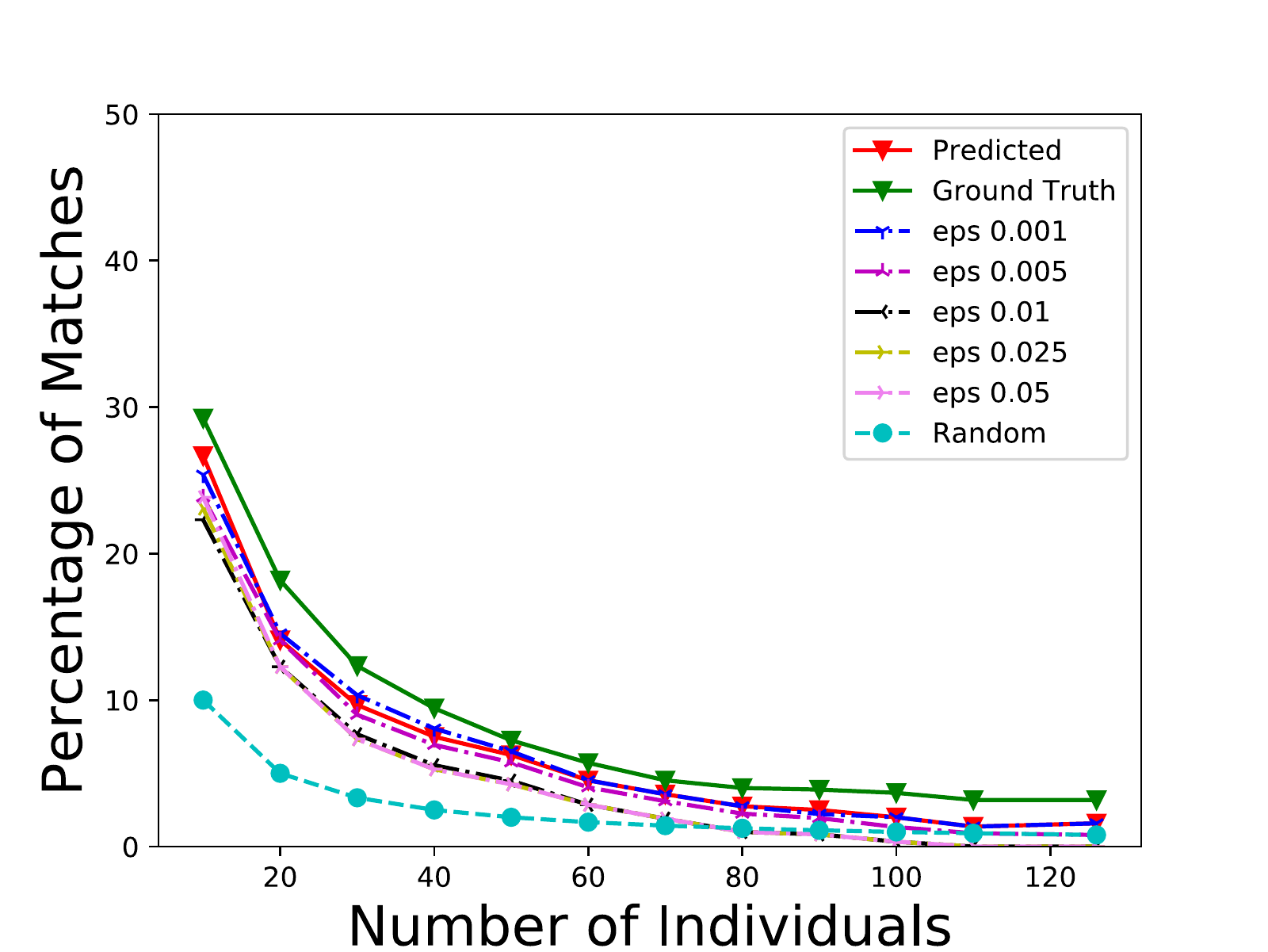}
    \caption{Hair Color}
\end{subfigure}
\caption{\textbf{(a)-(c)}: Matching accuracy when SVMs are used on the 126 individuals to fine-tune. From left to right, a) Top 1, b) Top 3, c) Top 5 matching accuracy with SVMs. The input to the SVM for an image-DNA pair is a vector of probabilities of phenotypes, where phenotypes are extracted from the image, and conditional probabilites are calculated from the DNA sequence, and the labels are binary indicating a match or otherwise. For each image in the test set, we rank all DNA sequences in order of their distance from the separating hyperplane, from most likely match to least likely match, and consider the top $k$ as true matches. We note that this approach does not significantly improve matching accuracy, which we believe to be the result of low signal-noise ratio. \textbf{(d)-(f)}: Top-$1$ matching performance when attacking phenotypes other than sex independently. While effective relative to clean images, attacking hair color, skin color or eye color alone does not reduce accuracy to below random except for fairly large populations, where matching accuracy is low to begin with.}
\label{fig:hair_eye_skin_1}
\end{figure}

\section{Protecting Privacy with Adversarial Noise}
To defend against potential re-identification by a malicious actor, we
propose the use of adversarial examples as a tool to preserve
privacy. Here, adversarial noise is calculated using gradient based
methods to minimize the matching log-likelihood.
We present results for both directly solving this minimization using
projected gradient descent where the noise is calculated over
all phenotype classifiers (\emph{Universal Noise} attack), as well as
using PGD to attack one phenotype classifier at a time by maximizing
the corresponding neural network's cross entropy loss.
The former was described in the main body.
We now briefly review the PGD attack targeting a single phenotype at a
time (e.g., sex phenotype prediction); see Madry et
al.~\cite{madry2017towards} for further details.



Recall that $g_p(v_p, x_i)$ denotes the probability that a phenotype
variant $v_p$ is predicted from image $x_i$.
Slightly abusing notation, let $g_p(x_i)$ be the probability distribution over variants given an
input $x_i$.
Let $L(g_p(x_i),y_p)$ be the loss associated with the true
variant $y_p$ and predicted variant distribution $g_p(x_i)$.
The goal of the PGD approach for generating adversarial noise is to
maximize loss:
\[
  \max_{-\epsilon \le \delta \le \epsilon} L(g_p(x_i+\delta),y_p).
\]
We can do this by using a form of gradient descent.
Specifically, let $\delta_k$ be the value of noise in iteration $k$
(starting with $\delta_0 = 0$, or a small random noise).
Then
\[
\delta_{k+1} = \delta_k +  \alpha\ sgn \nabla\mathcal{L}(g_p(x_i+\delta_k),y_p),
\]
where $\alpha$ is the learning rate.
This process is run for a fixed number of iterations, or until convergence.



Results for top-$1$ matching when attacking sex are explored in the main body of the paper, while here in the supplement, we first explore the scenario when other phenotypes are attacked independently, and when the correct match lies within the top-$3$ or top-$5$ most likely genomes. Consider first, the top-$1$ matching accuracy when attacking eye-color, hair-color and skin-color predictions using PGD in Supplementary Fig.~\ref{fig:hair_eye_skin_1} (d)-(f). While attacking with increasing values of $\epsilon$ does degrade performance compared to clean images, these attacks are not as effective as fooling the sex-prediction model. Independently, these phenotypes do not seem to contribute greatly to matching accuracy, in contrast to sex, where accuracy drops below random for small populations for $\epsilon=0.01$. This problem is addressed in the \emph{Universal Noise} setting where all phenotypes are attacked in parallel by directly minimizing the matching log-likelihood.

\begin{figure}[]
\centering
\begin{subfigure}[t]{0.32\textwidth}
    \includegraphics[width=5.6cm]{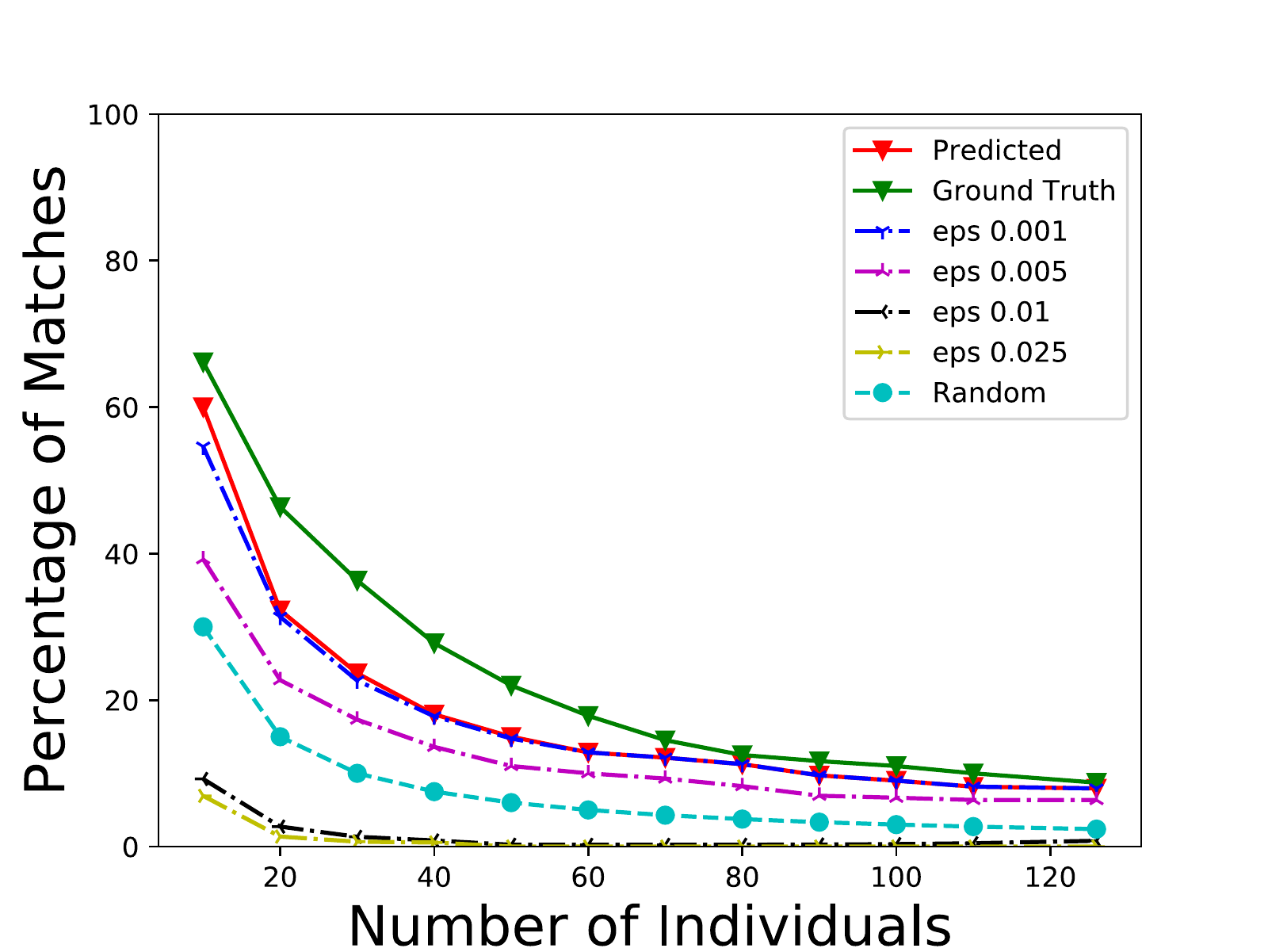}
    \caption{Sex - Top 3}
\end{subfigure}
\begin{subfigure}[t]{0.32\textwidth}
    \includegraphics[width=5.6cm]{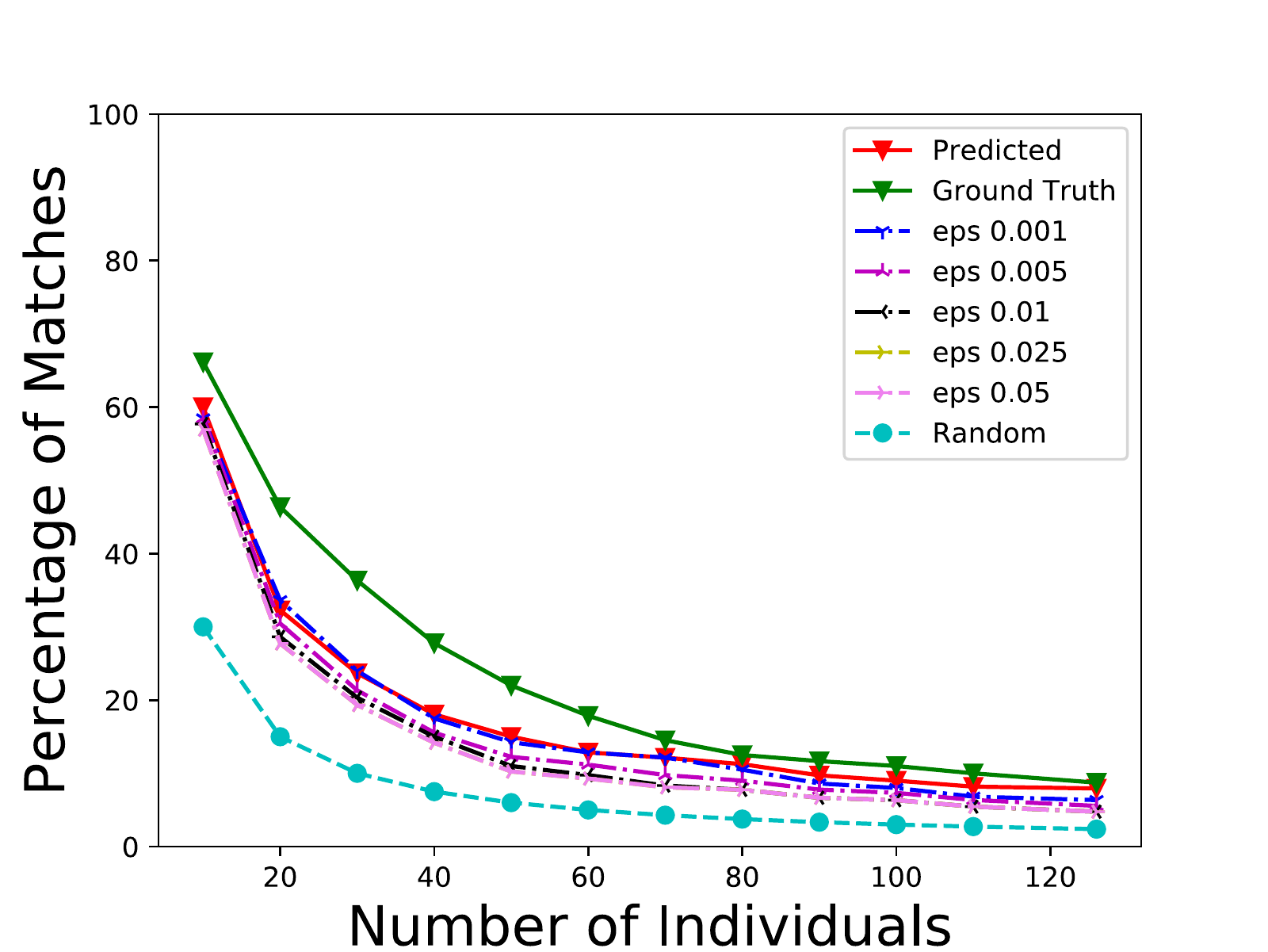}
    \caption{Skin Color - Top 3}
\end{subfigure}
\begin{subfigure}[t]{0.32\textwidth}
    \includegraphics[width=5.6cm]{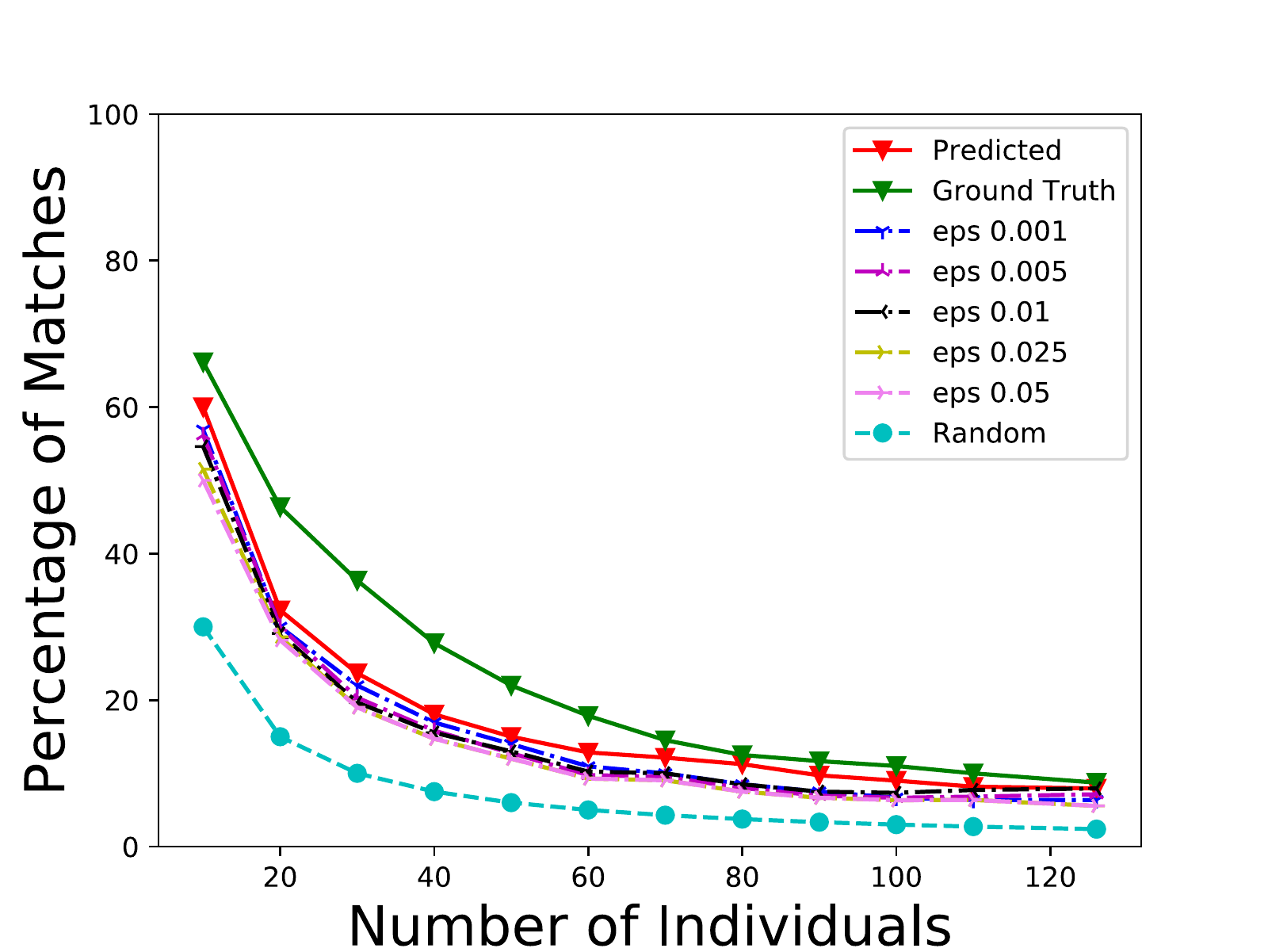}
    \caption{Eye Color - Top 3}
\end{subfigure}
\begin{subfigure}[t]{0.32\textwidth}
    \includegraphics[width=5.6cm]{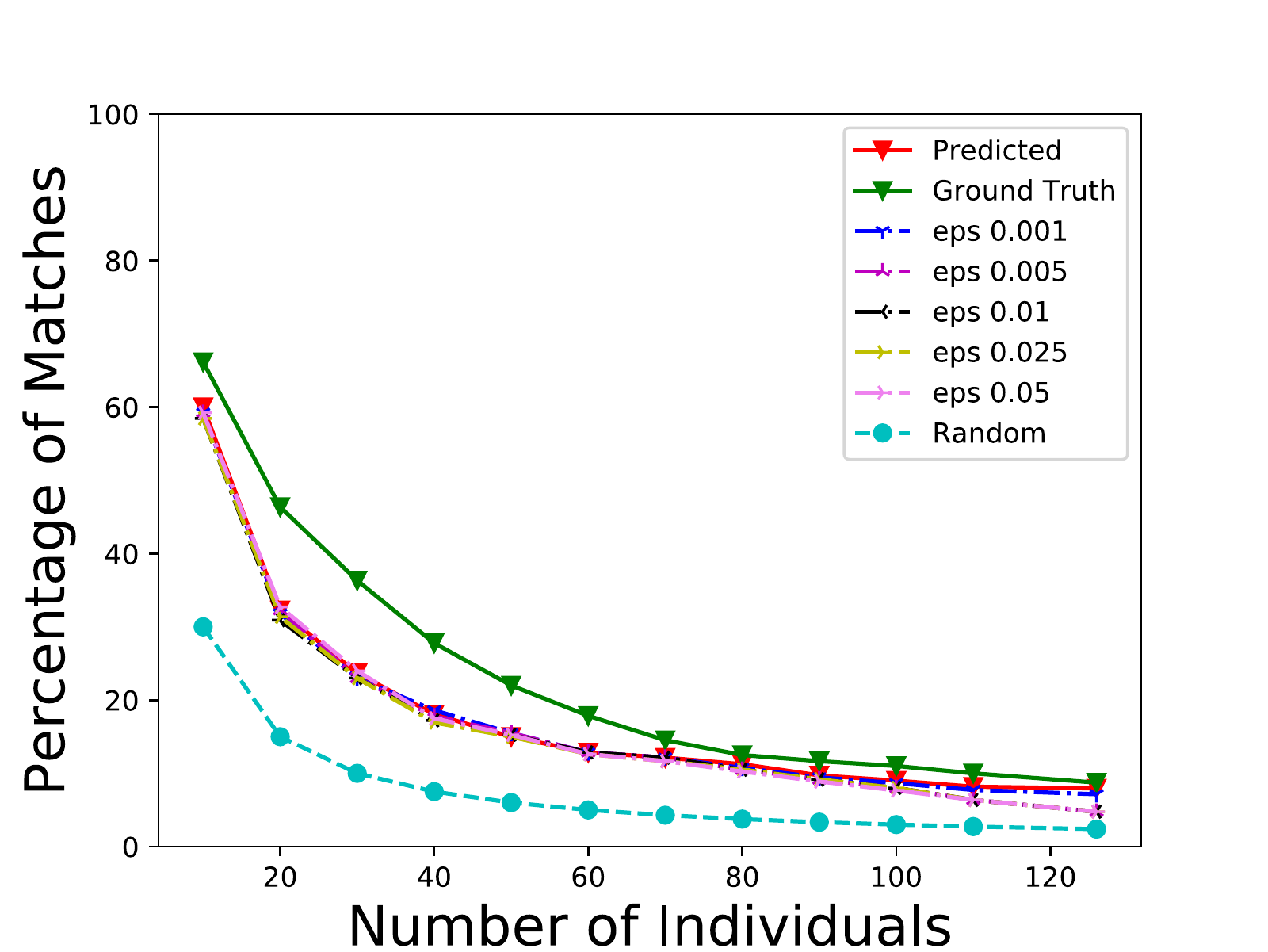}
    \caption{Hair Color - Top 3}
\end{subfigure}
\begin{subfigure}[t]{0.32\textwidth}
    \includegraphics[width=5.6cm]{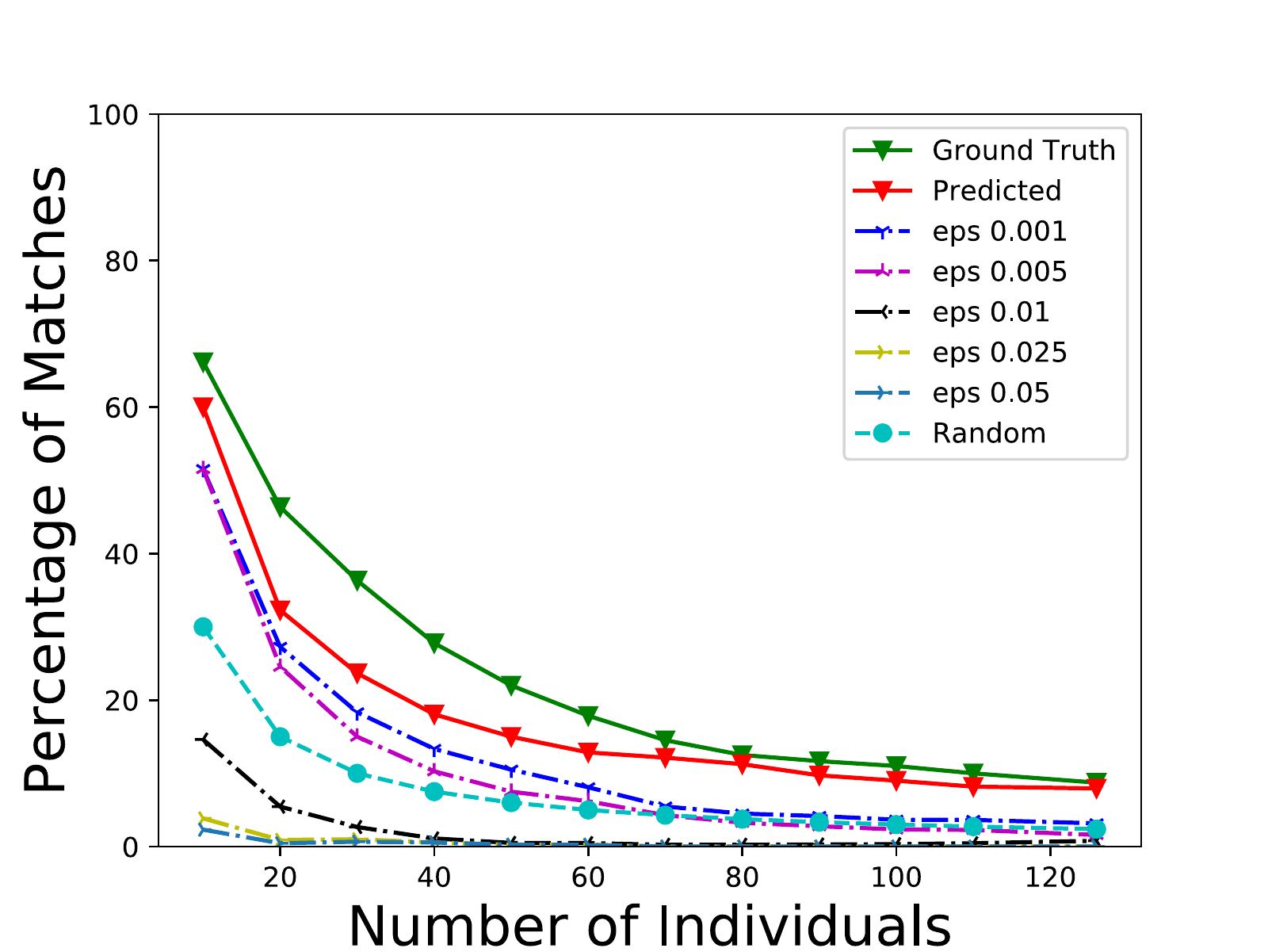}
    \caption{\emph{Universal Noise} attack - Top 3}
\end{subfigure}

\begin{subfigure}[t]{0.32\textwidth}
    \includegraphics[width=5.6cm]{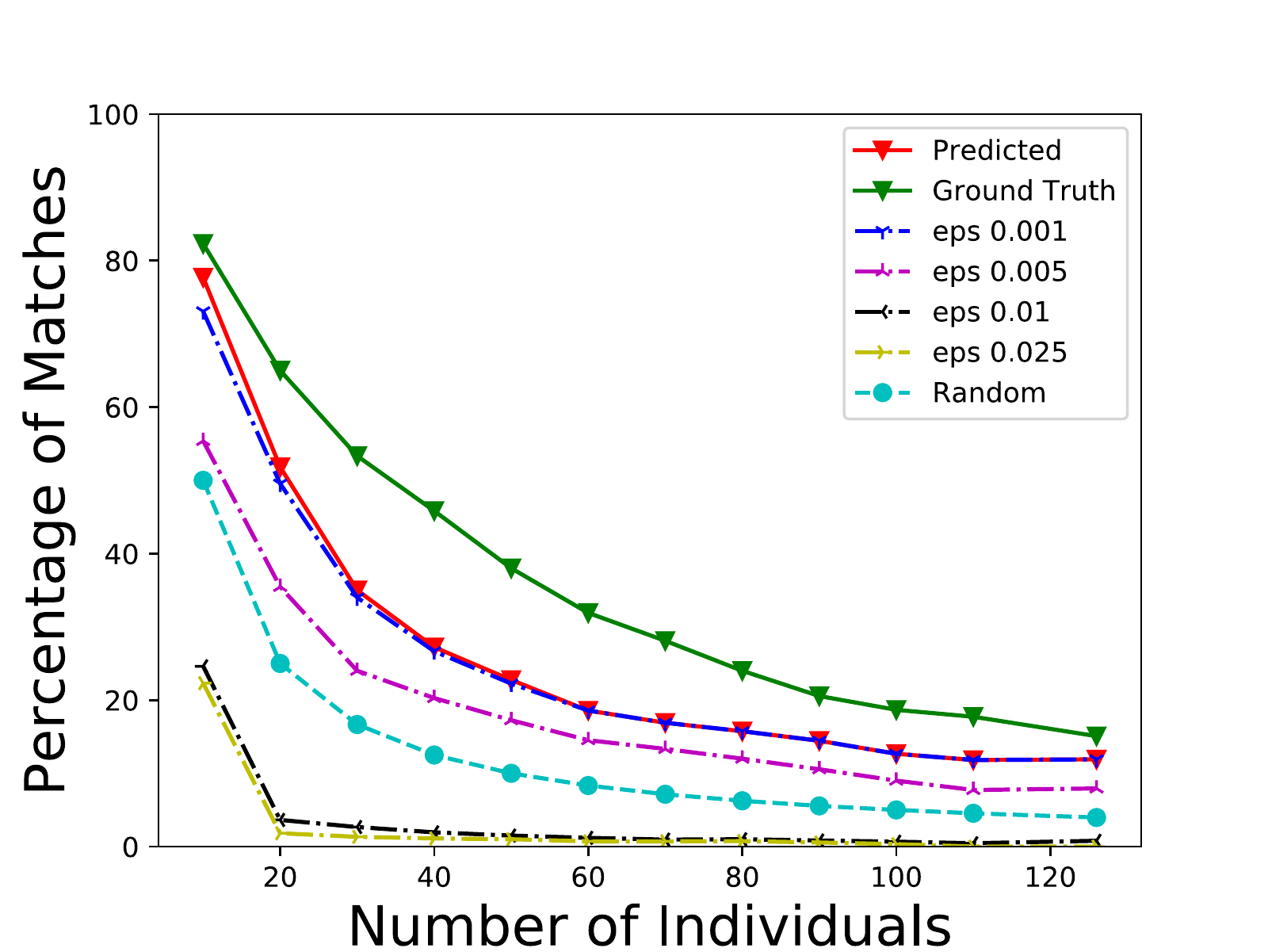}
    \caption{Sex - Top 5}
\end{subfigure}
\begin{subfigure}[t]{0.32\textwidth}
    \includegraphics[width=5.6cm]{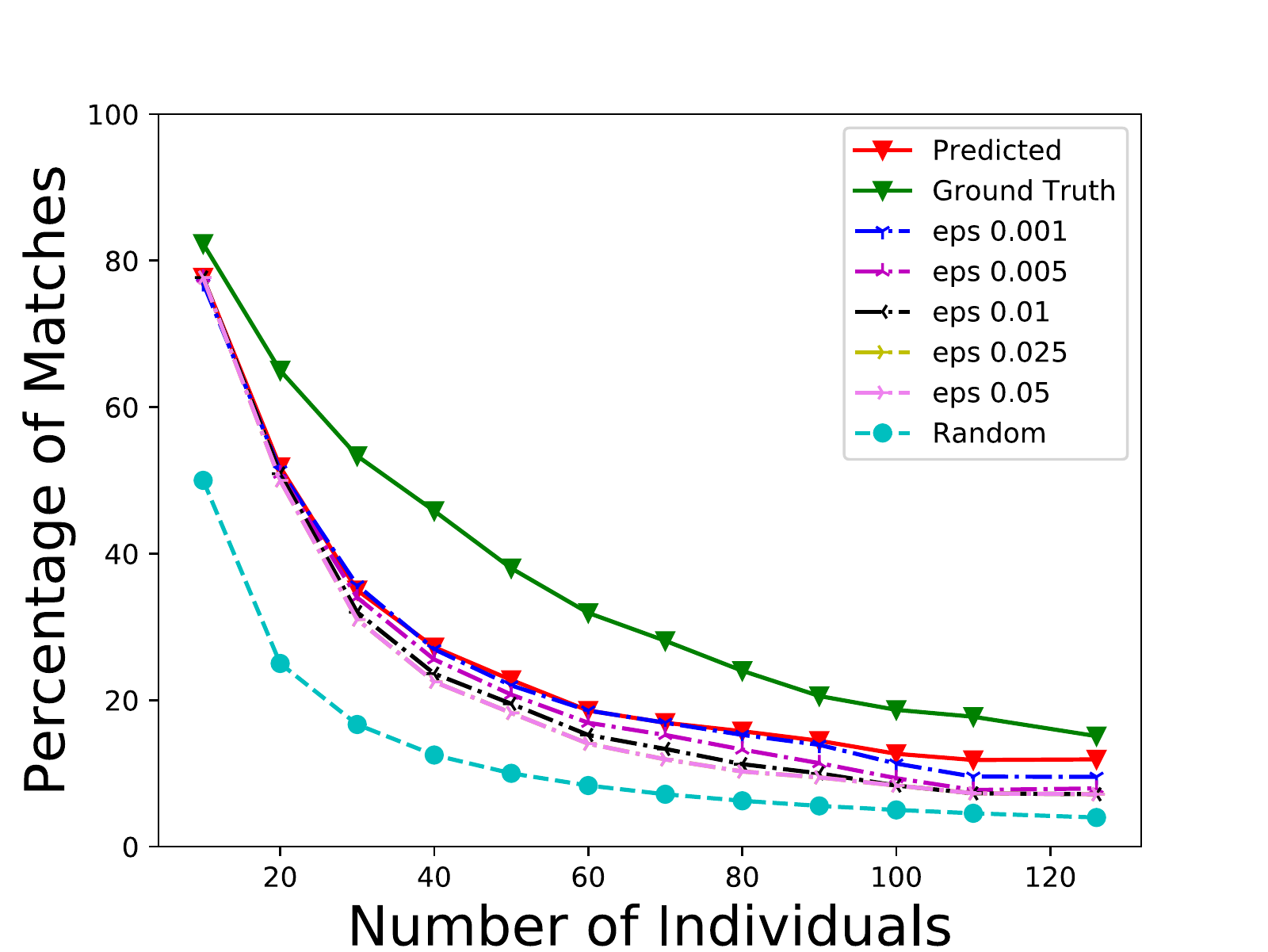}
    \caption{Skin Color - Top 5}
\end{subfigure}
\begin{subfigure}[t]{0.32\textwidth}
    \includegraphics[width=5.6cm]{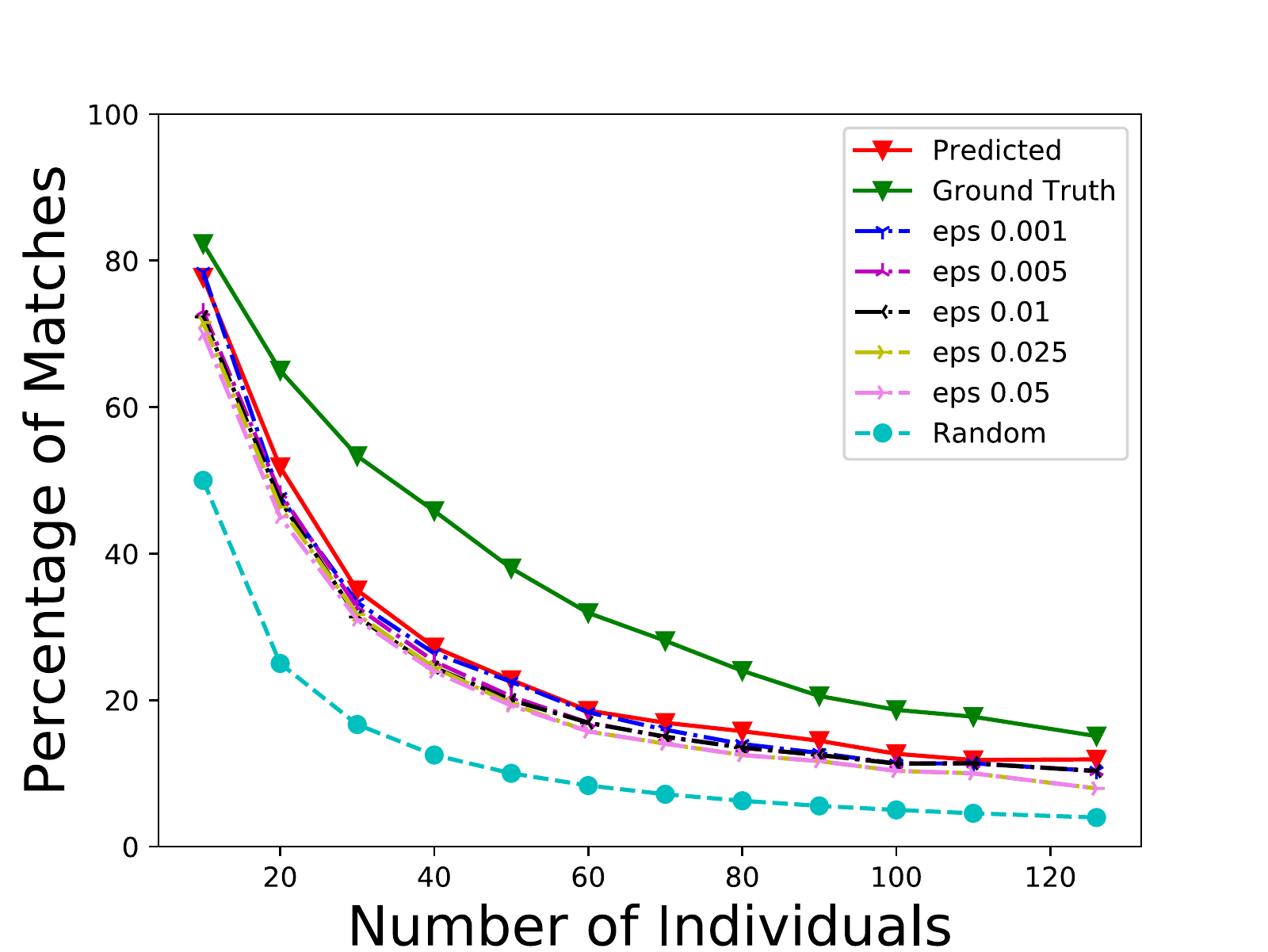}
    \caption{Eye Color - Top 5}
\end{subfigure}
\begin{subfigure}[t]{0.32\textwidth}
    \includegraphics[width=5.6cm]{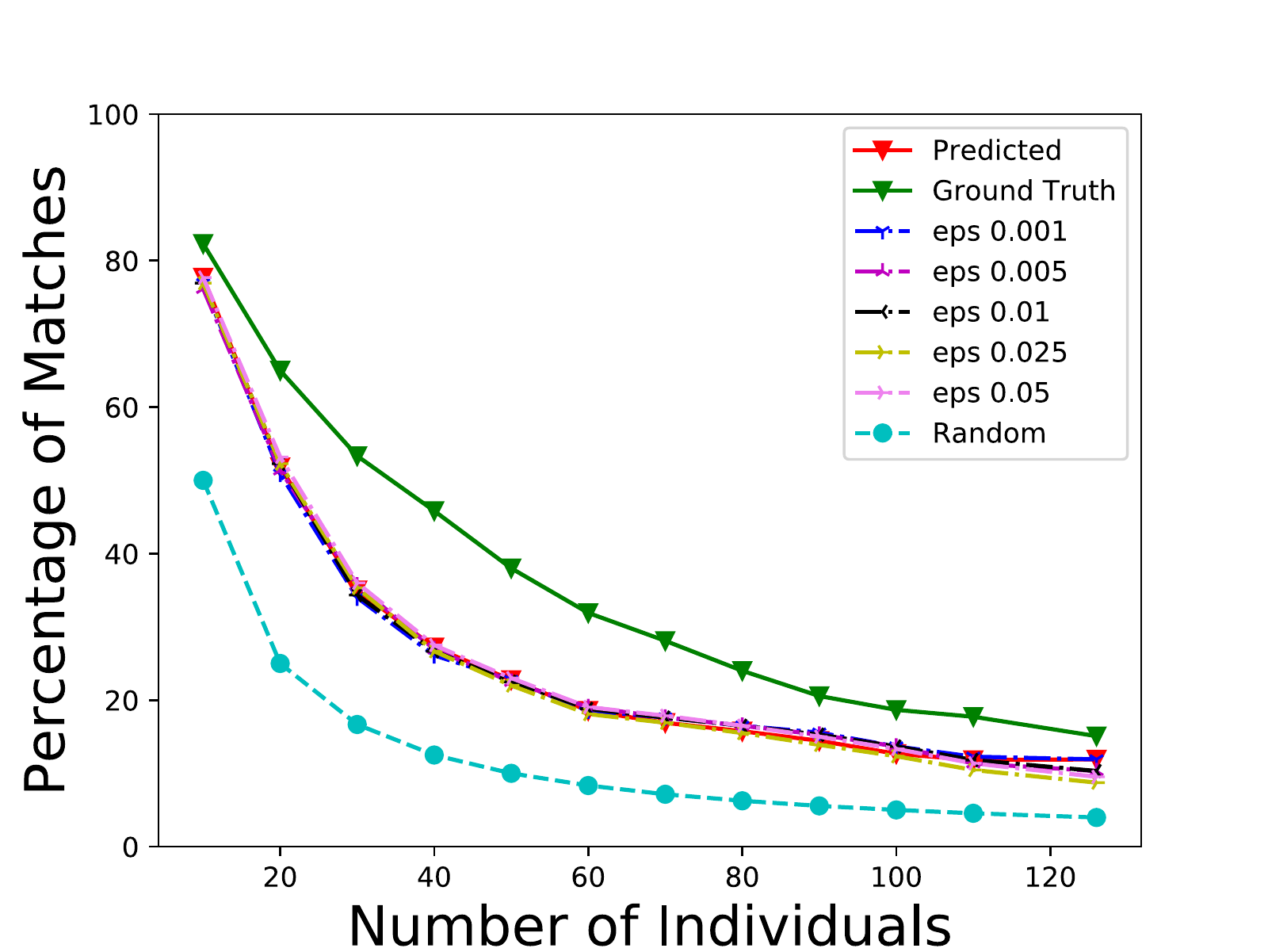}
    \caption{Hair Color - Top 5}
\end{subfigure}
\begin{subfigure}[t]{0.32\textwidth}
    \includegraphics[width=5.6cm]{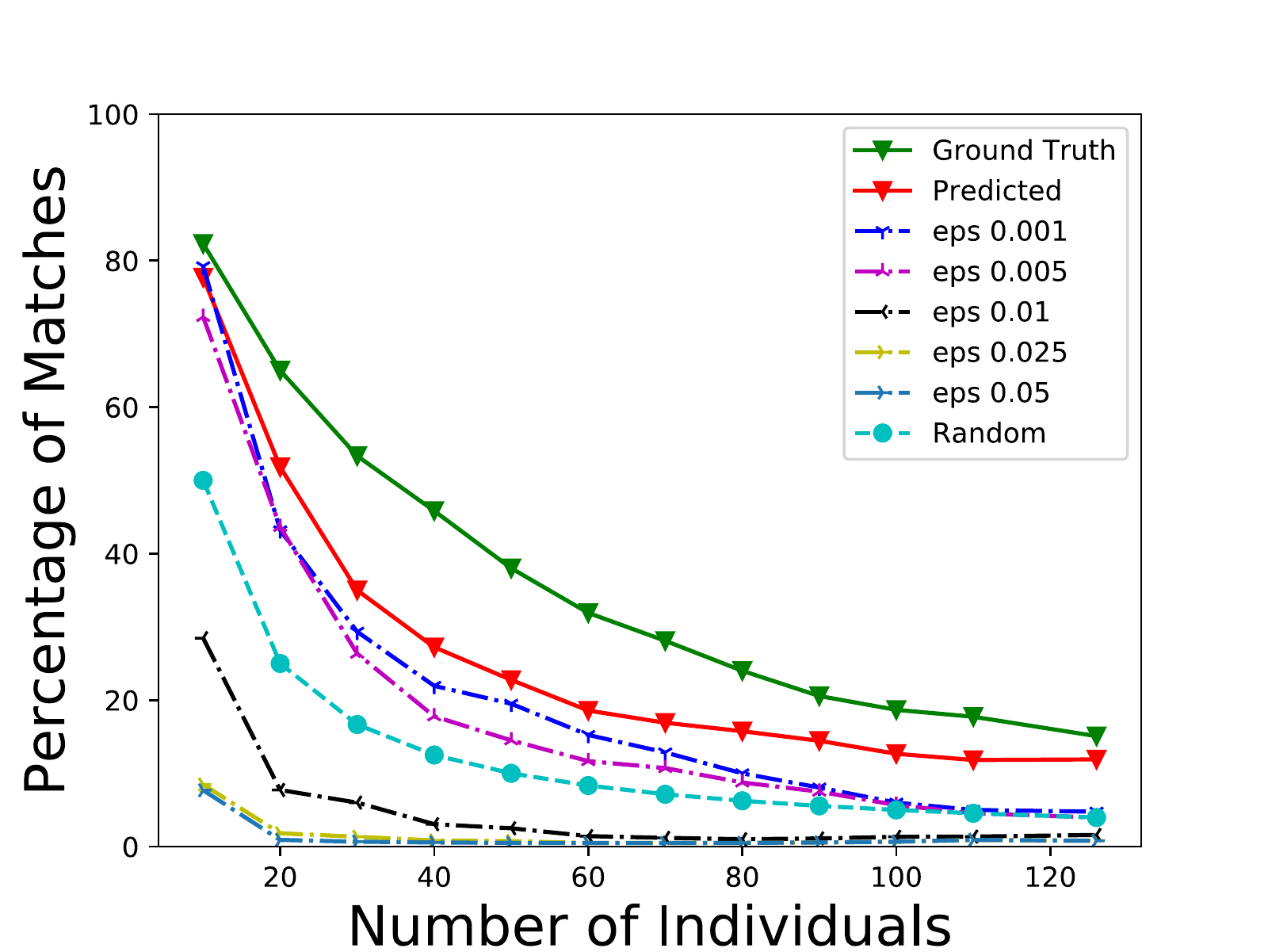}
    \caption{\emph{Universal Noise} attack - Top 5}
\end{subfigure}
\caption{\textbf{(a)-(e)}: Accuracy of Top-3 matching with adversarially perturbed facial images, at different strengths of attack, i.e. values of $\epsilon$, for a) Sex, b) Skin Color, c) Eye Color, d) Hair Color and e) \emph{Universal Noise}. Similar to the top-$1$ case, attacking sex is much more effective compared to attacking other phenotypes independently, and directly minimizing overall matching log-likelihood remains highly effective. \textbf{(f)-(j)}: Accuracy of Top-5 matching with adversarially perturbed facial images, at different strengths of attack, i.e. values of $\epsilon$, for f) Sex, g) Skin Color, h) Eye Color, i) Hair Color and j) \emph{Universal Noise}. Once again, while attacking sex manages to lower accuracy to below random for even small populations, attacking other phenotypes proves relatively ineffective. Yet again, direct minimization of the matching log-likelihood proves highly effective in preserving privacy.}
\label{fig:nonrobust_attacked_5}
\end{figure}

Next, we look at the top-$3$ and top-$5$ matching accuracy for attacking all phenotypes individually, as well as the \emph{Universal Noise} attack. The random baseline in these cases makes $3$ and $5$ random guesses out of the population in question respectively. This approach illustrates the scenario when a malicious actor is trying to narrow down an individuals genome down to a ssmall subset of the population as possible matches. Supplementary Fig.~\ref{fig:nonrobust_attacked_5} (a)-(e) present results in the top-$3$ case and Supplementary Fig.~\ref{fig:nonrobust_attacked_5} (f)-(j) present results in the top-$5$ case. Naturally, performance improves compared to the top-$1$, albeit at the cost of an increase in the number of false positives ($k$ or $k-1$ false positives per image). Similar to the results in the top-$1$ case, attacking sex, and the \emph{Universal Noise} approach prove to be highly effective, while independently attacking other phenotypes does not drop accuracy below random for any population size.

\section{Adversarial Training}
\begin{figure}[]
\centering
    \begin{subfigure}[t]{0.4\textwidth}
        \centering
        \includegraphics[width=5.5cm]{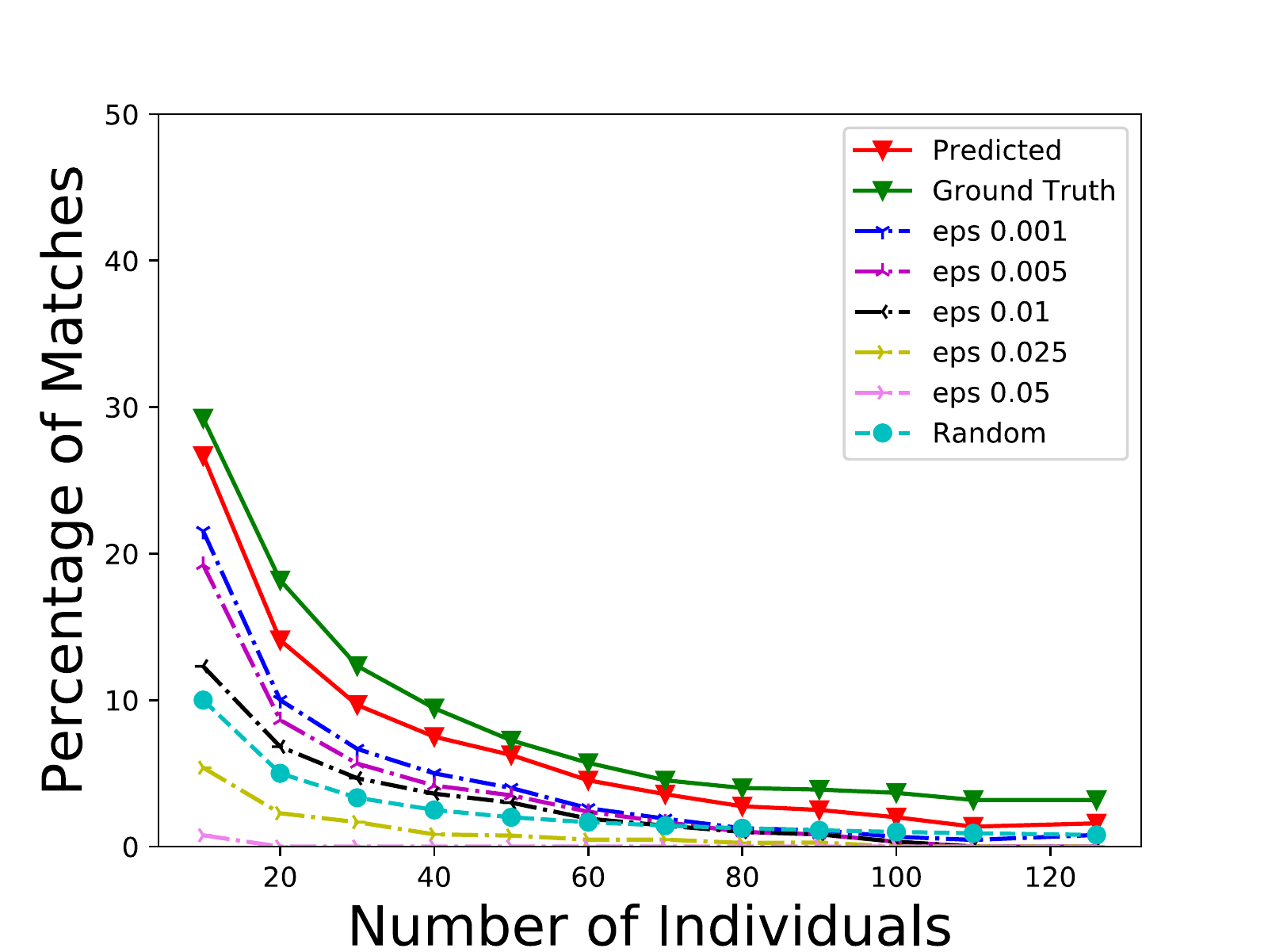}
        \caption{Sex - Robust, Attacked}
    \end{subfigure}
    \begin{subfigure}[t]{0.4\textwidth}
        \centering
        \includegraphics[width=5.5cm]{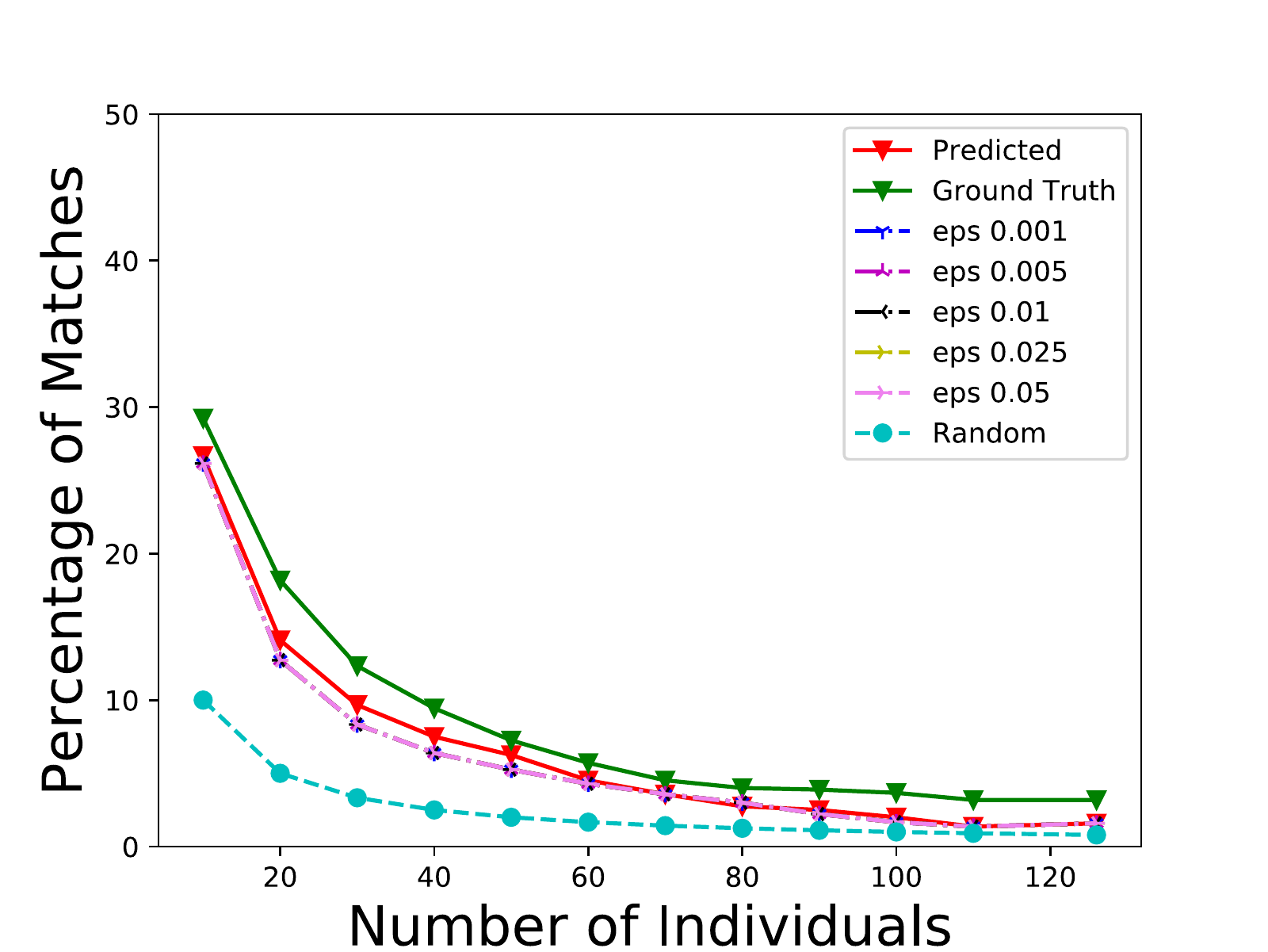}
        \caption{Skin Color - Robust, Attacked}
    \end{subfigure}
    \begin{subfigure}[t]{0.4\textwidth}
        \centering
        \includegraphics[width=5.5cm]{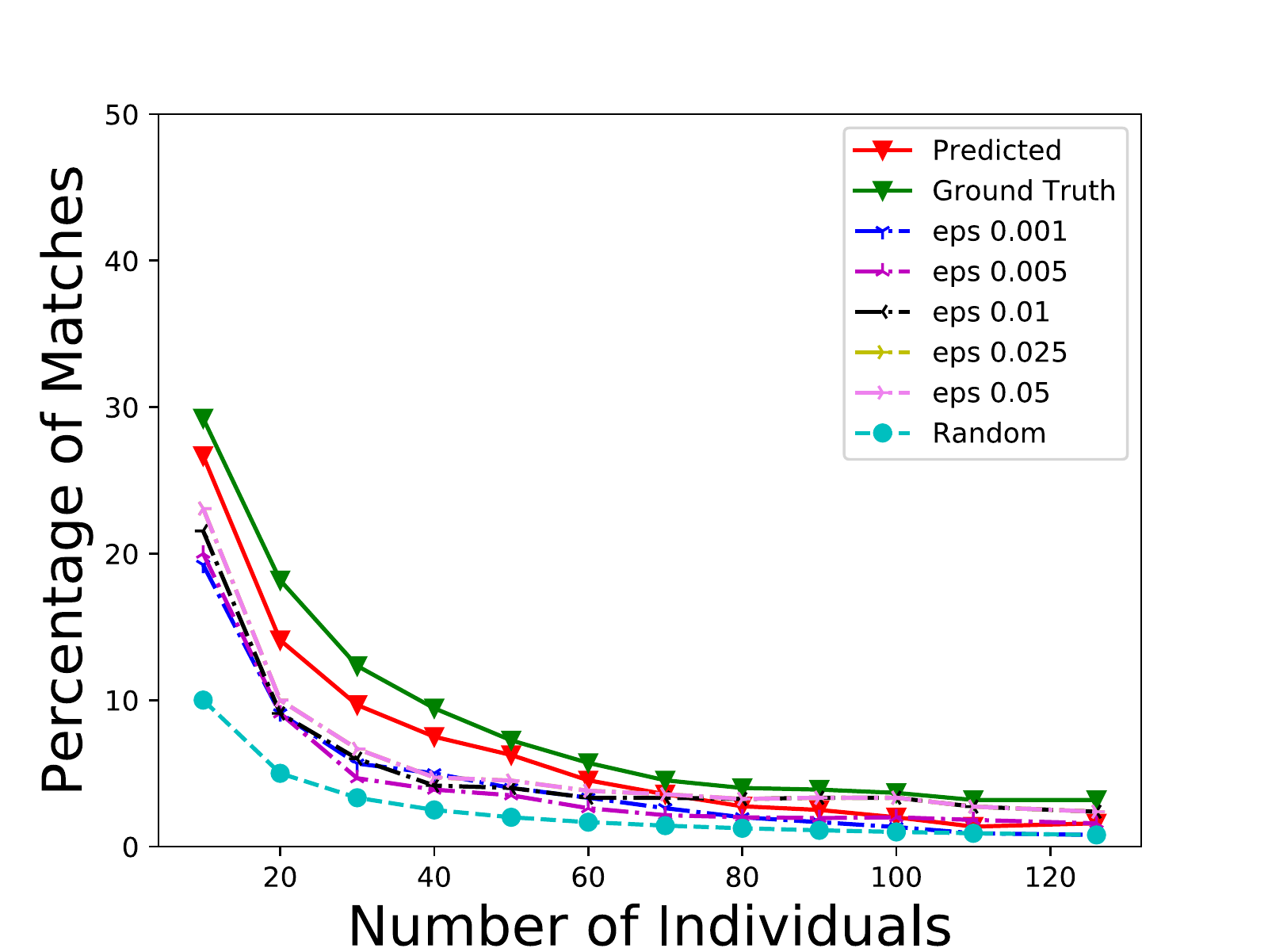}
        \caption{Eye Color - Robust, Attacked}
    \end{subfigure}
    \begin{subfigure}[t]{0.4\textwidth}
        \centering
        \includegraphics[width=5.5cm]{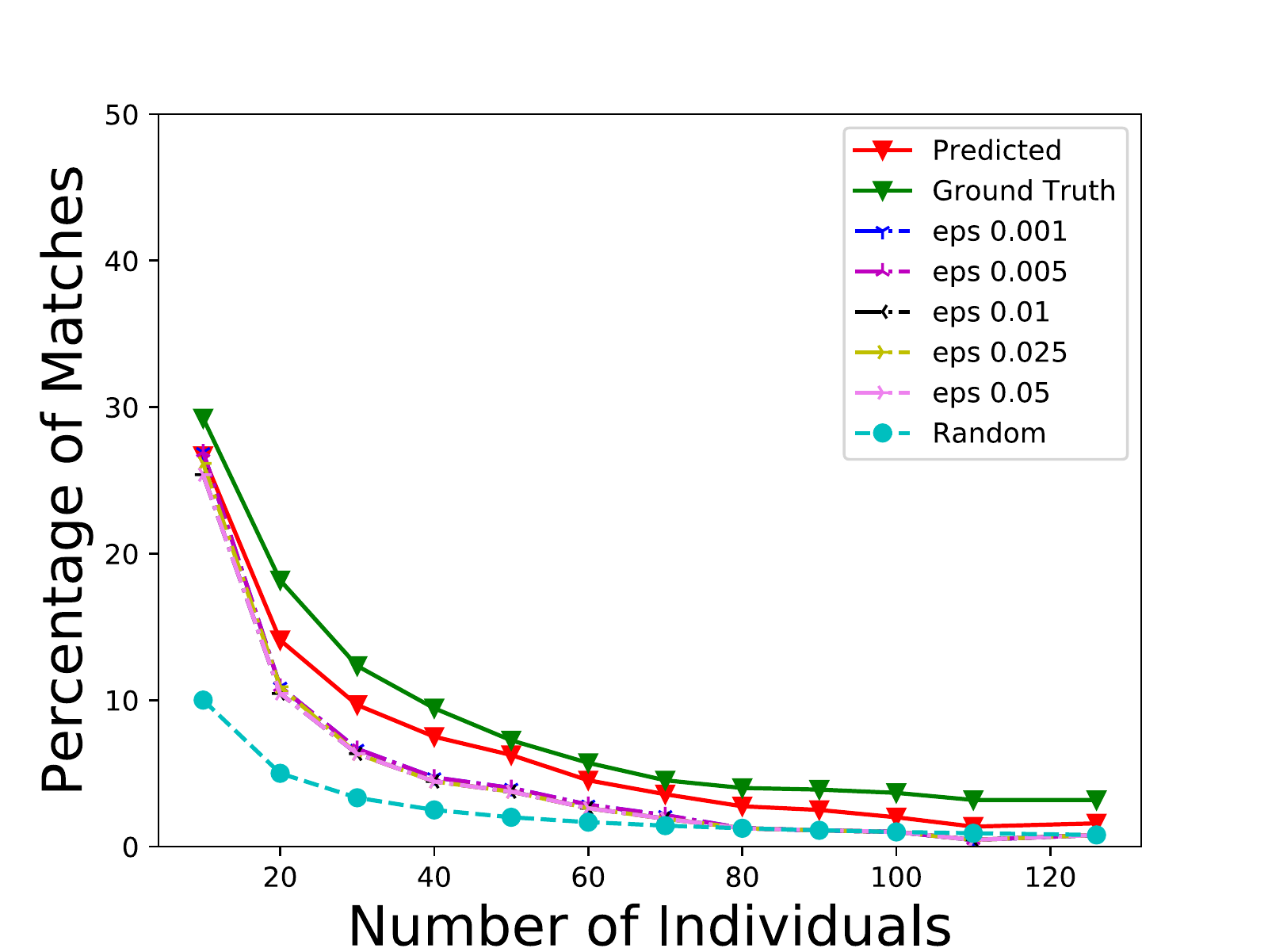}
        \caption{Hair Color - Robust, Attacked}
    \end{subfigure}
	\begin{subfigure}[t]{0.4\textwidth}
		\centering
	    \includegraphics[width=5.5cm]{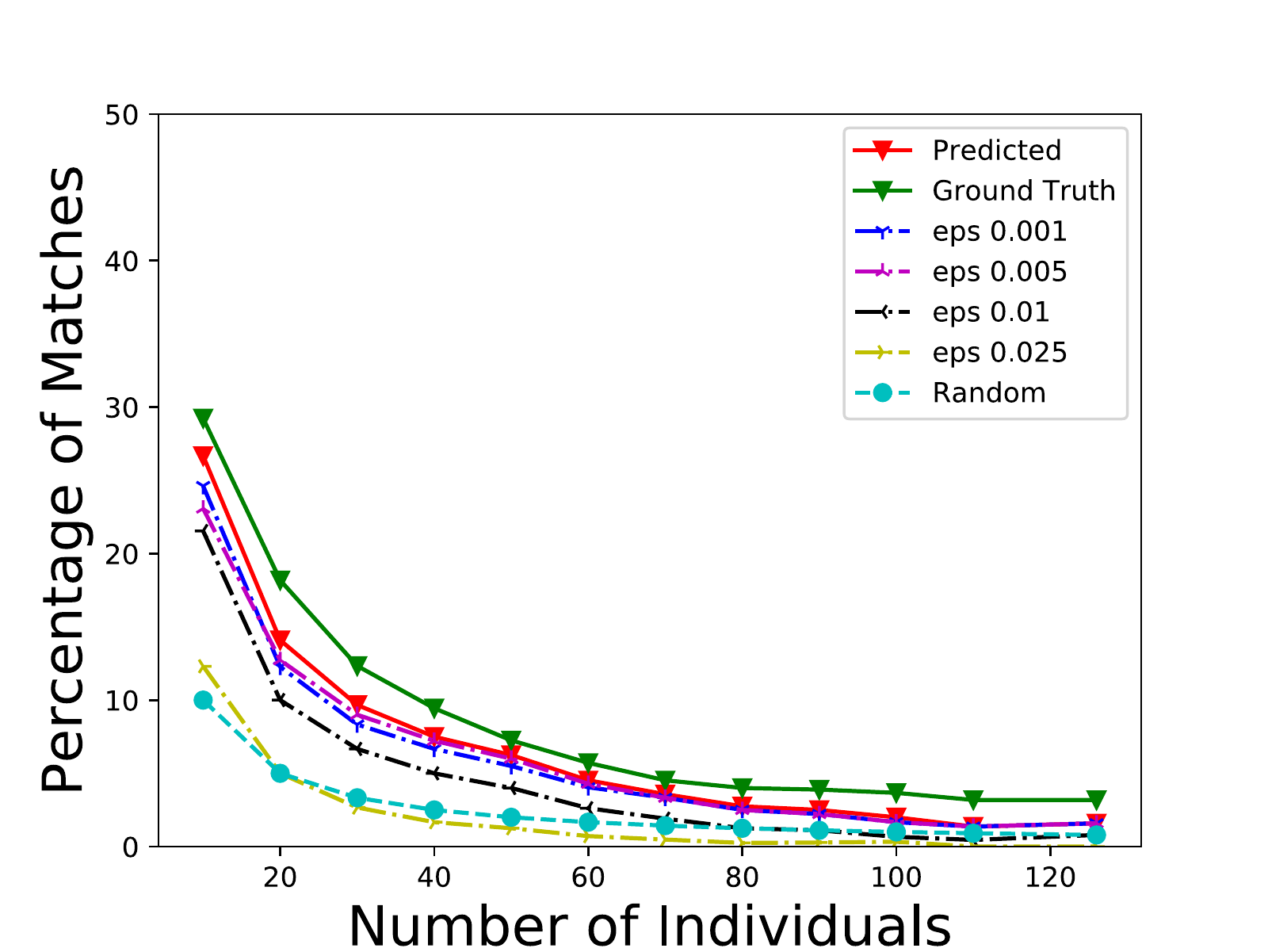}
	    \caption{Sex - Robust, Baseline}
	\end{subfigure}
	\begin{subfigure}[t]{0.4\textwidth}
		\centering
	    \includegraphics[width=5.5cm]{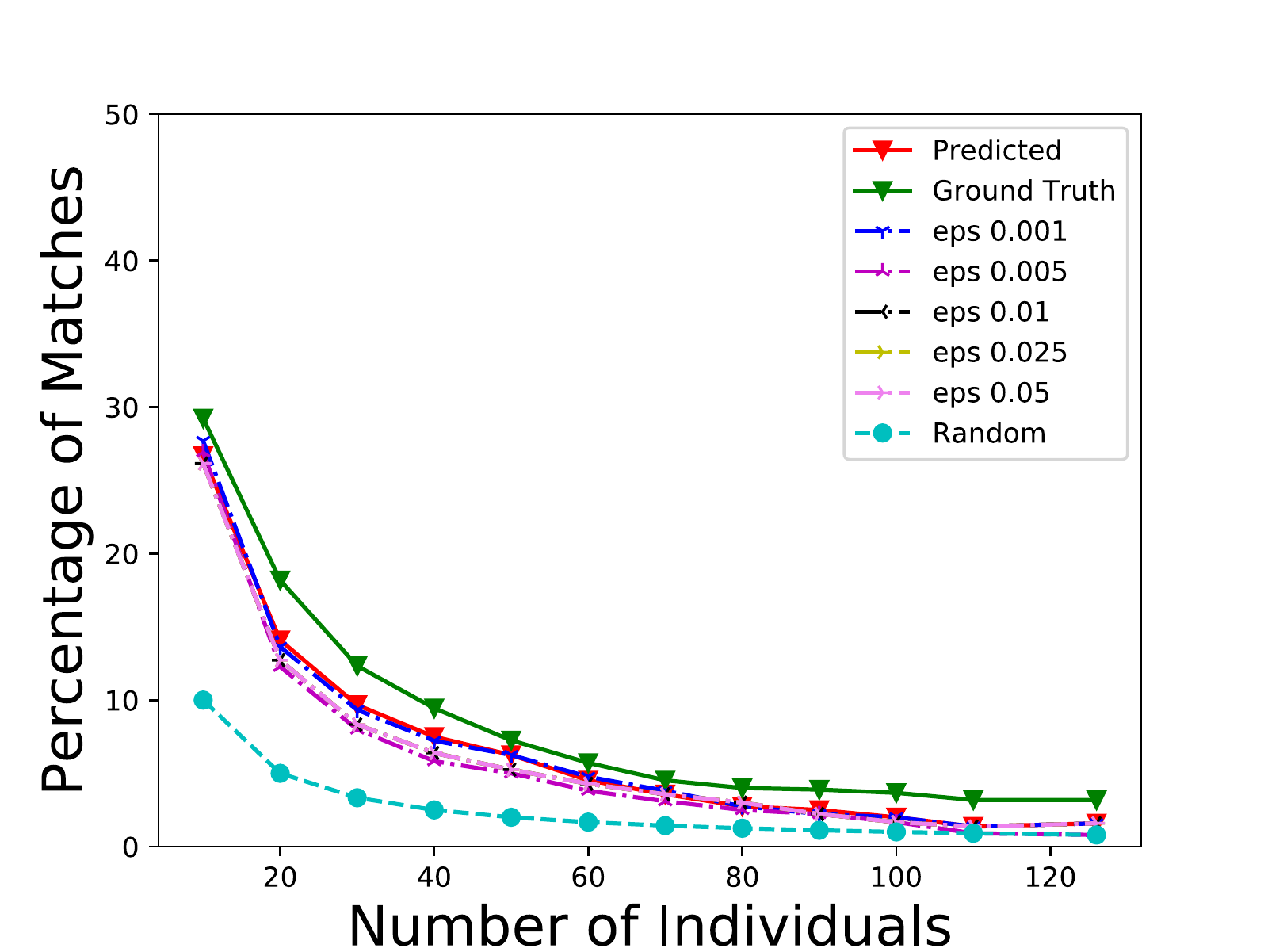}
	    \caption{Skin Color - Robust, Baseline}
	\end{subfigure}
	\begin{subfigure}[t]{0.4\textwidth}
		\centering
	    \includegraphics[width=5.5cm]{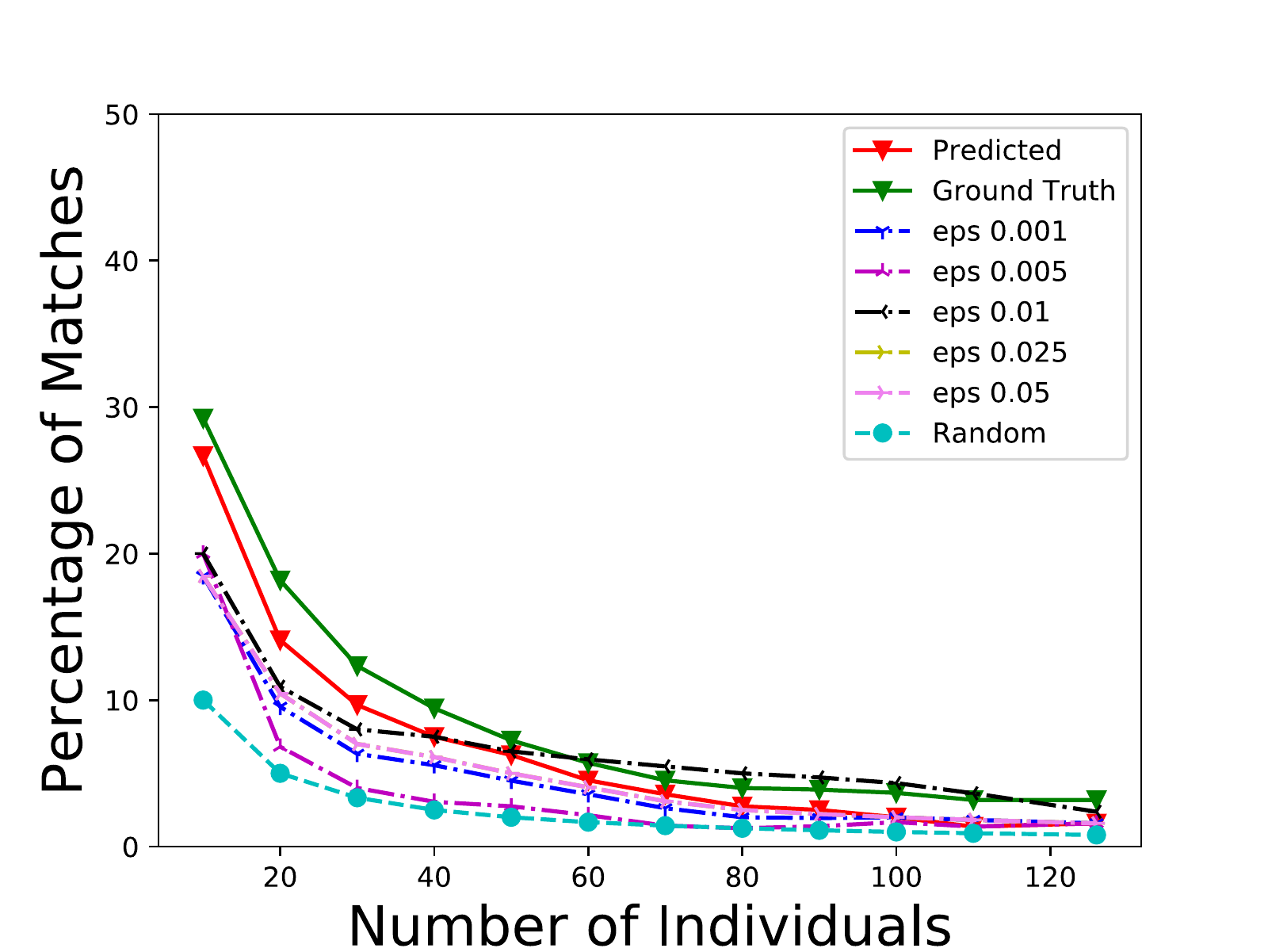}
	    \caption{Eye Color - Robust, Baseline}
	\end{subfigure}
	\begin{subfigure}[t]{0.4\textwidth}
		\centering
	    \includegraphics[width=5.5cm]{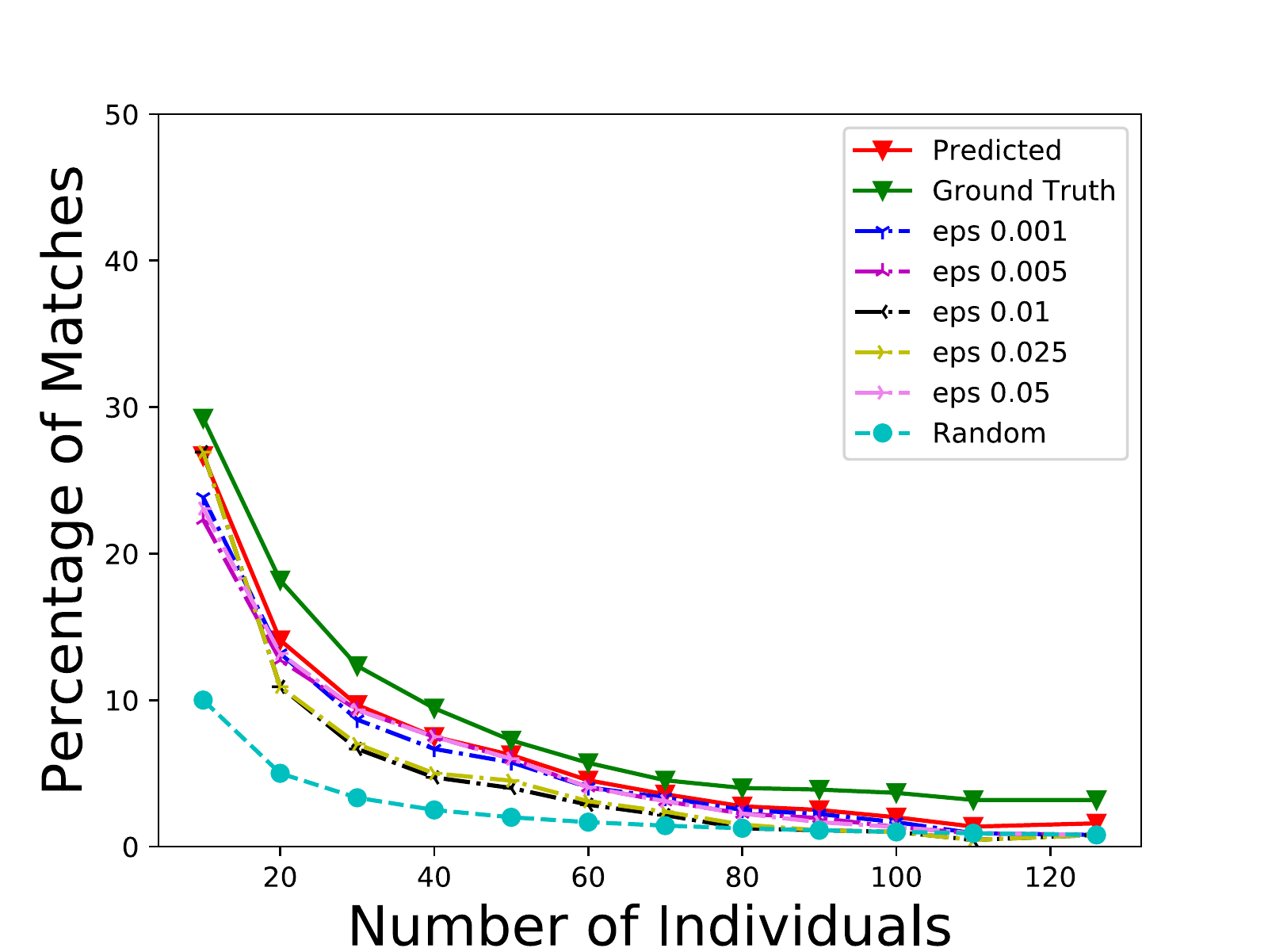}
	    \caption{Hair Color - Robust, Baseline}
	\end{subfigure}
	\caption{\textbf{(a)-(d)}: Top-1 Matching accuracy with robust classifiers, with images perturbed at $\epsilon$ between $0$ and $0.05$ for a) Sex, b) Skin Color, c) Eye Color, d) Hair Color. Each classifier was trained with adversarial examples at $\epsilon=0.01$. \textbf{(e)-(h)}: Top-1 Baseline accuracy of matching with adversarially trained classifiers, but clean images for a) Sex, b) Skin Color, c) Eye Color, d) Hair Color. Legends in each figure show the values of $\epsilon$ used for training the network.}
	\label{fig:robust_noattack_1}
\end{figure}

\begin{figure}[]
\centering
\captionsetup[subfigure]{width=0.85\textwidth, justification=centering}
\begin{subfigure}[t]{0.32\textwidth}
    \includegraphics[width=5.6cm]{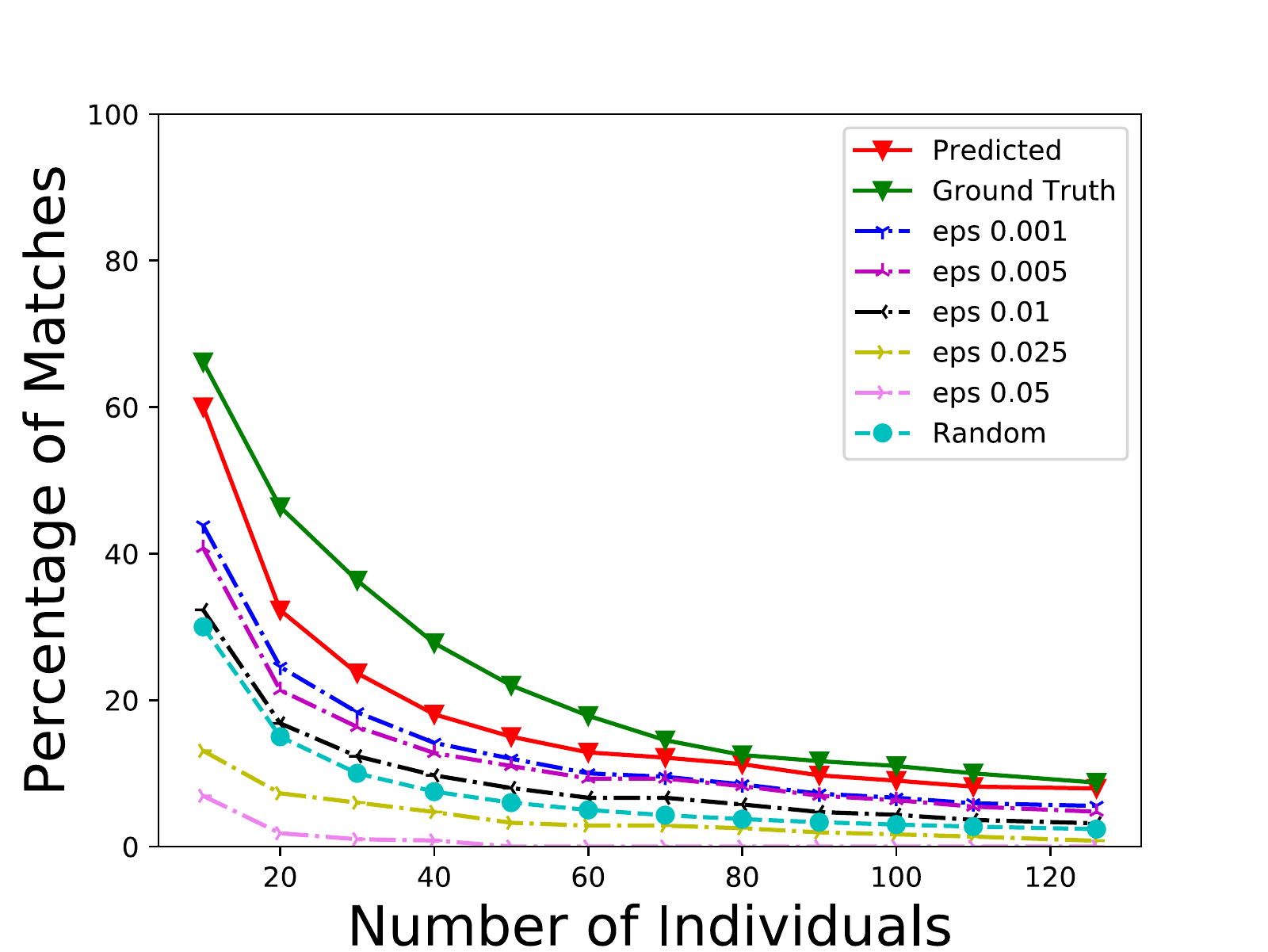}
    \caption{Sex - Top 3 - Robust, Attacked}
\end{subfigure}
\begin{subfigure}[t]{0.32\textwidth}
    \includegraphics[width=5.6cm]{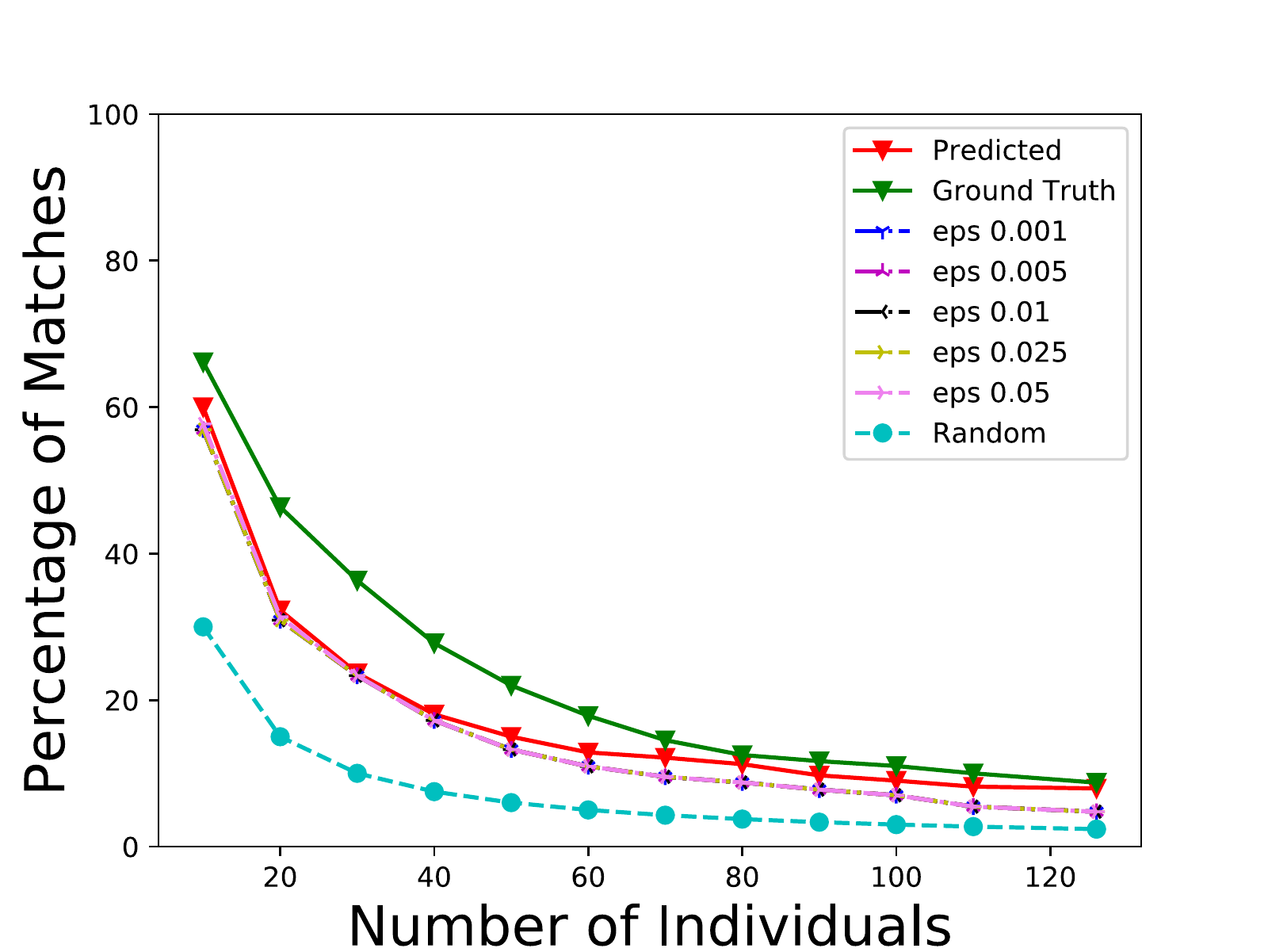}
    \caption{Skin Color - Top 3 - Robust, Attacked}
\end{subfigure}
\begin{subfigure}[t]{0.32\textwidth}
    \includegraphics[width=5.6cm]{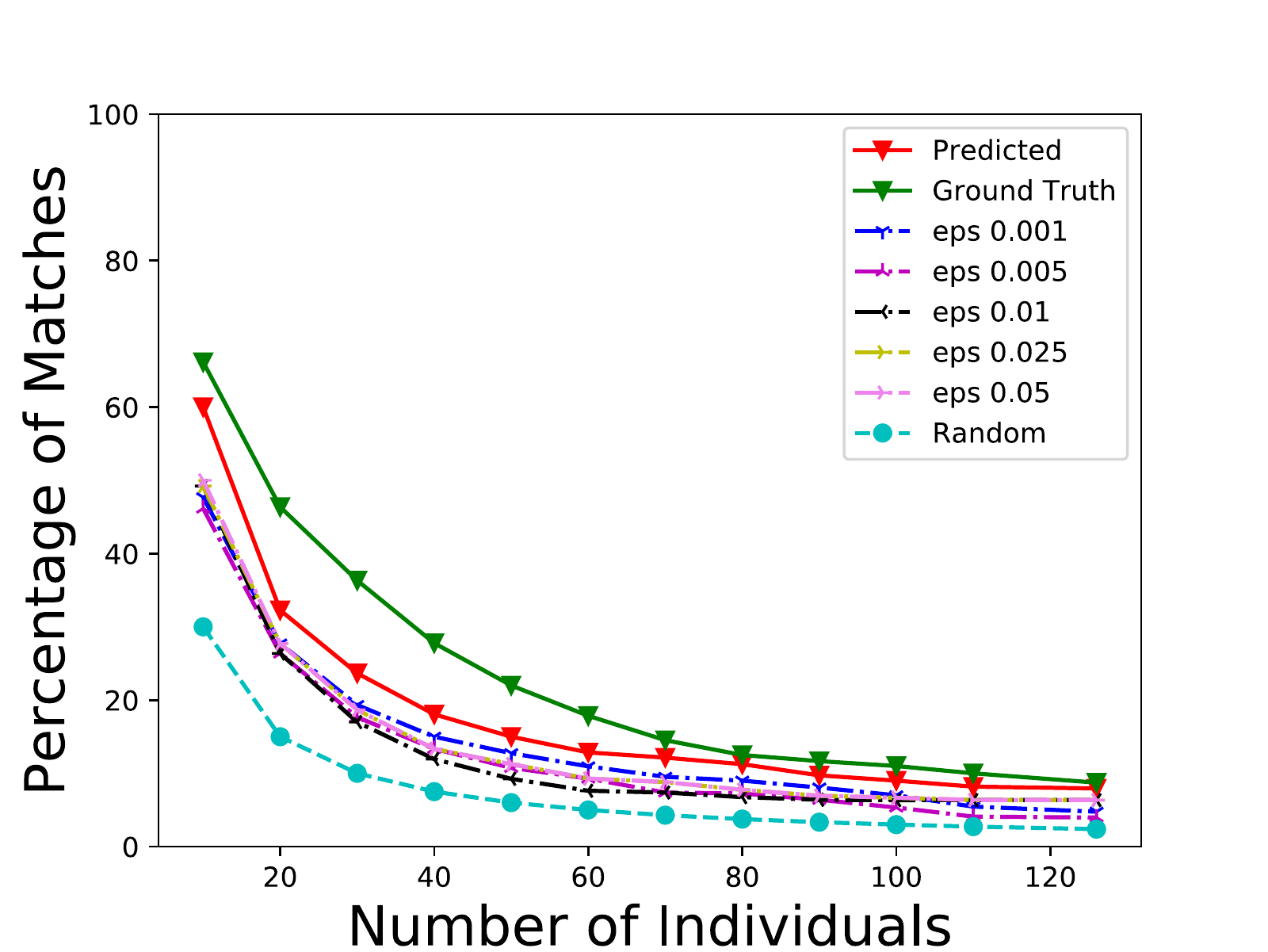}
    \caption{Eye Color - Top 3 - Robust, Attacked}
\end{subfigure}
\begin{subfigure}[t]{0.32\textwidth}
    \includegraphics[width=5.6cm]{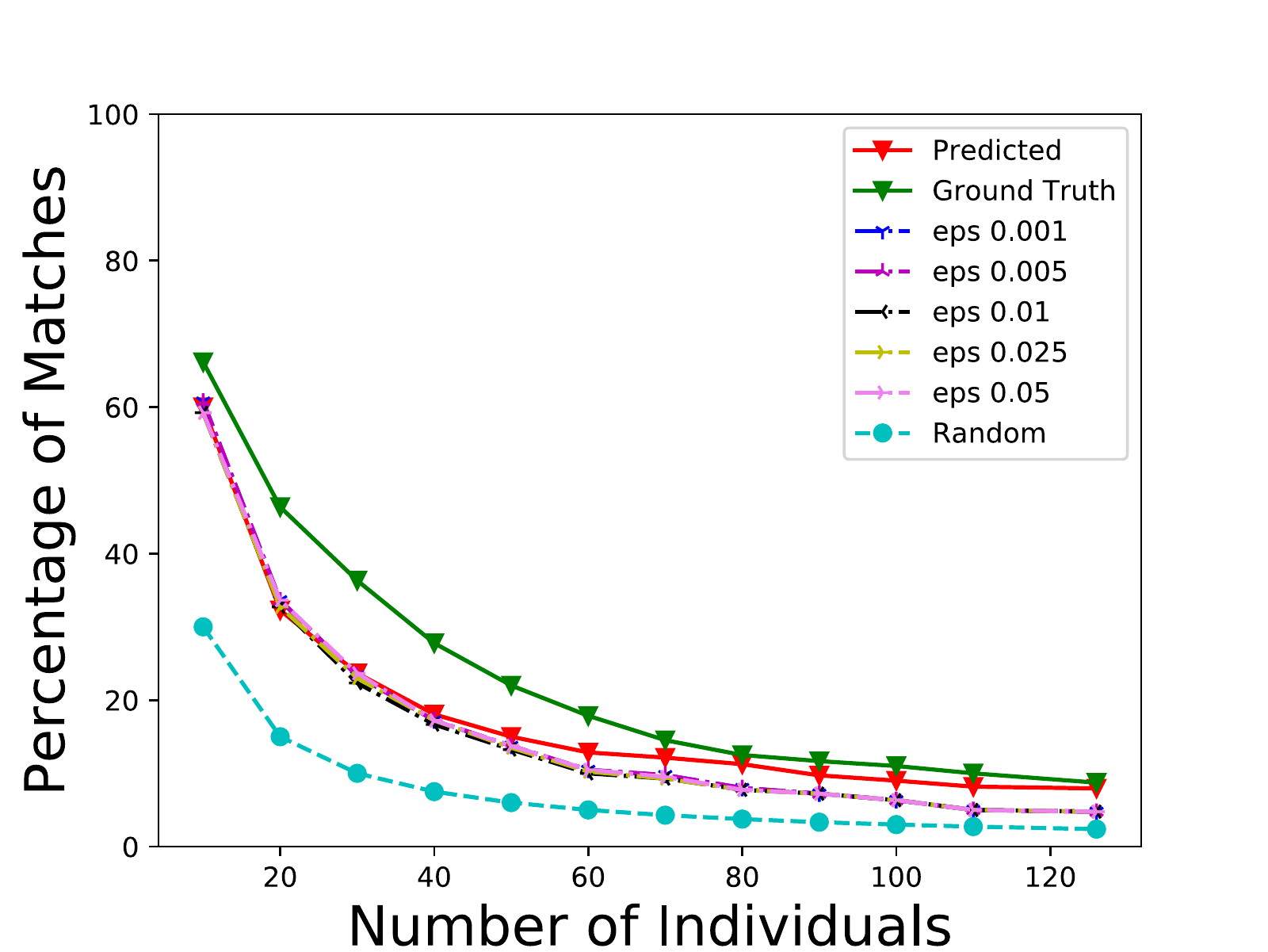}
    \caption{Hair Color - Top 3 - Robust, Attacked}
\end{subfigure}
\begin{subfigure}[t]{0.32\textwidth}
    \includegraphics[width=5.6cm]{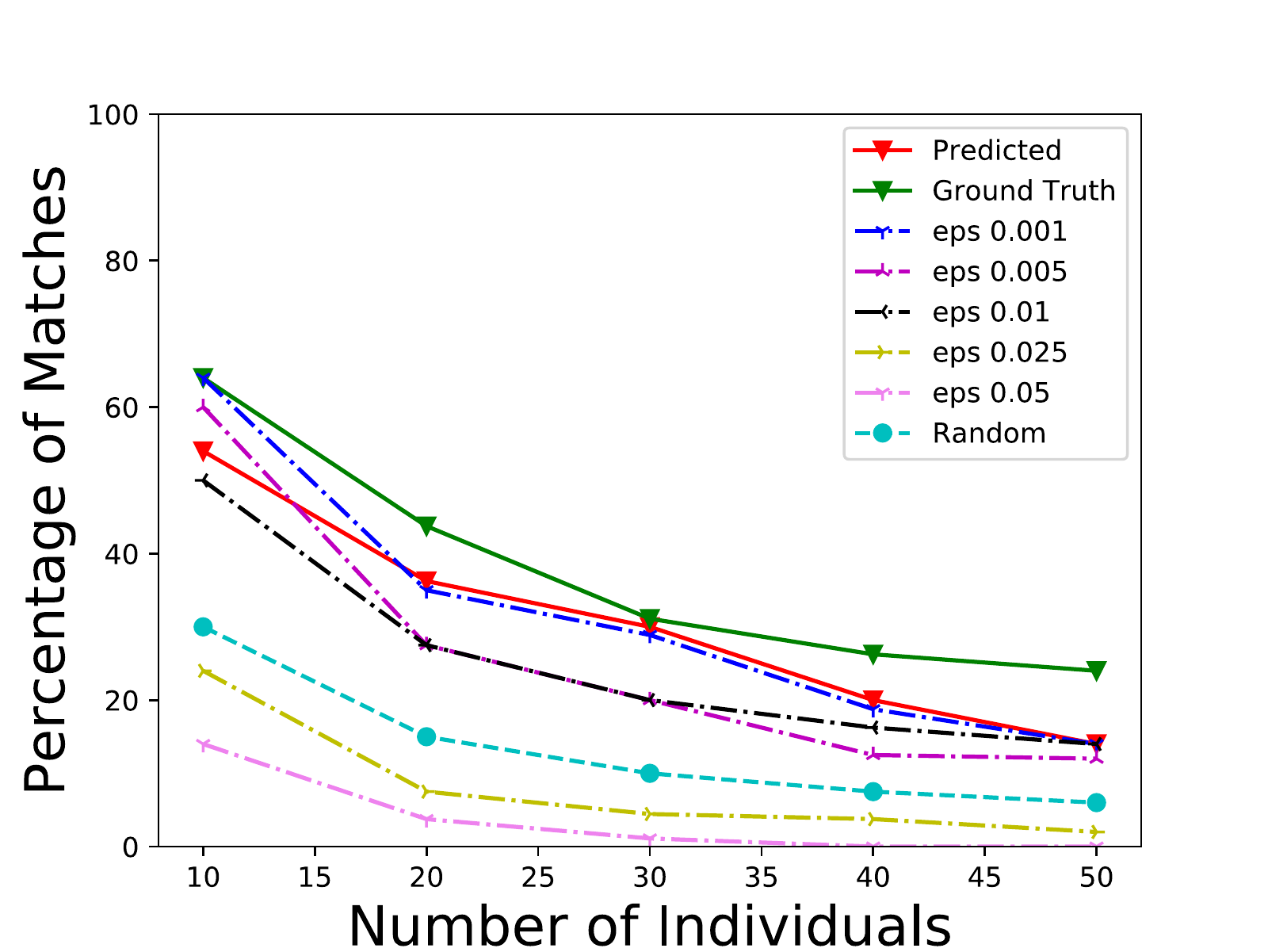}
    \caption{\emph{Universal Noise} - Top 3 - Robust, Attacked}
\end{subfigure}

\begin{subfigure}[t]{0.32\textwidth}
    \includegraphics[width=5.6cm]{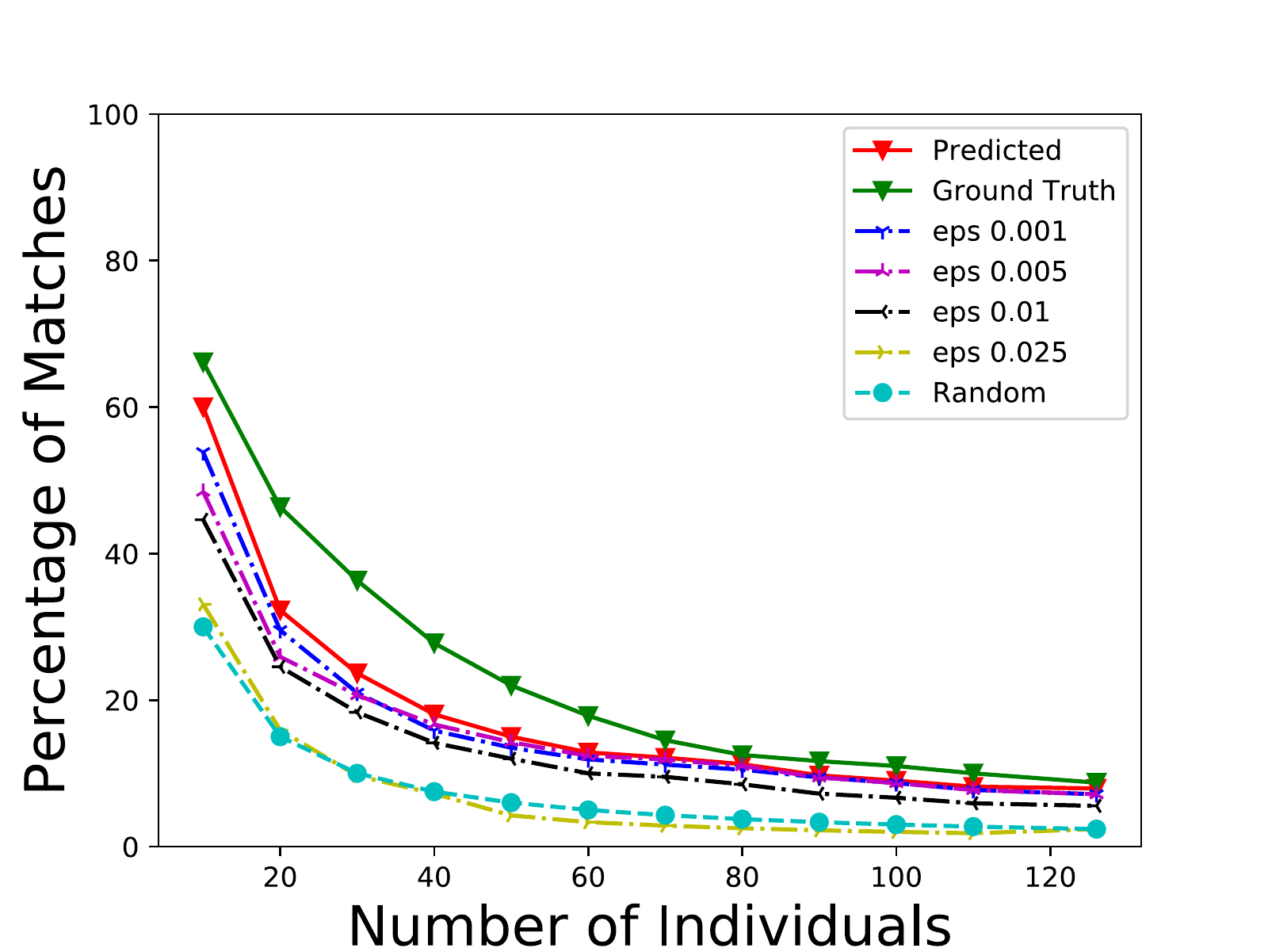}
    \caption{Sex - Top 3 - Robust, Baseline}
\end{subfigure}
\begin{subfigure}[t]{0.32\textwidth}
    \includegraphics[width=5.6cm]{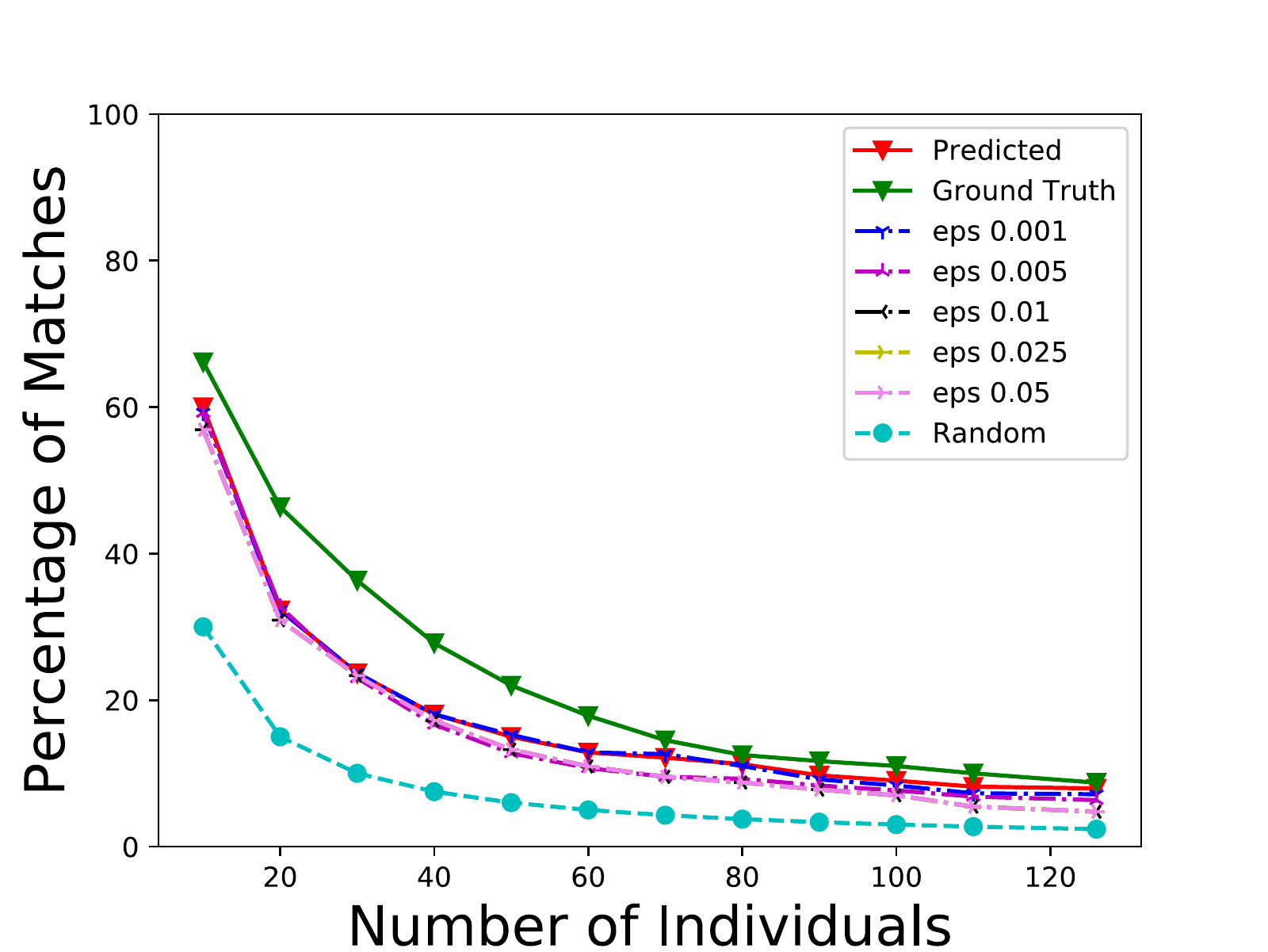}
    \caption{Skin Color - Top 3 - Robust, Baseline}
\end{subfigure}
\begin{subfigure}[t]{0.32\textwidth}
    \includegraphics[width=5.6cm]{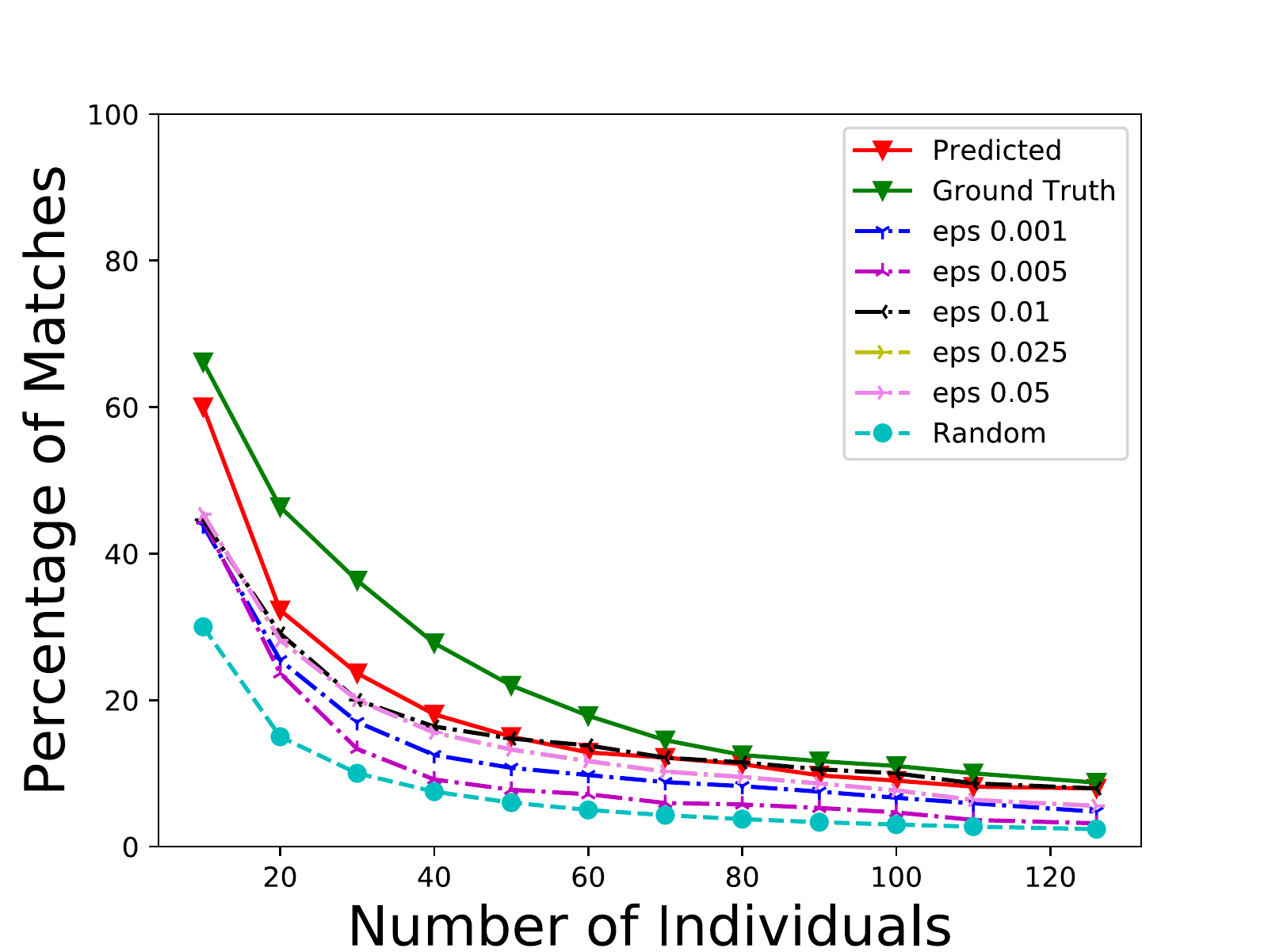}
    \caption{Eye Color - Top 3 - Robust, Baseline}
\end{subfigure}
\begin{subfigure}[t]{0.32\textwidth}
    \includegraphics[width=5.6cm]{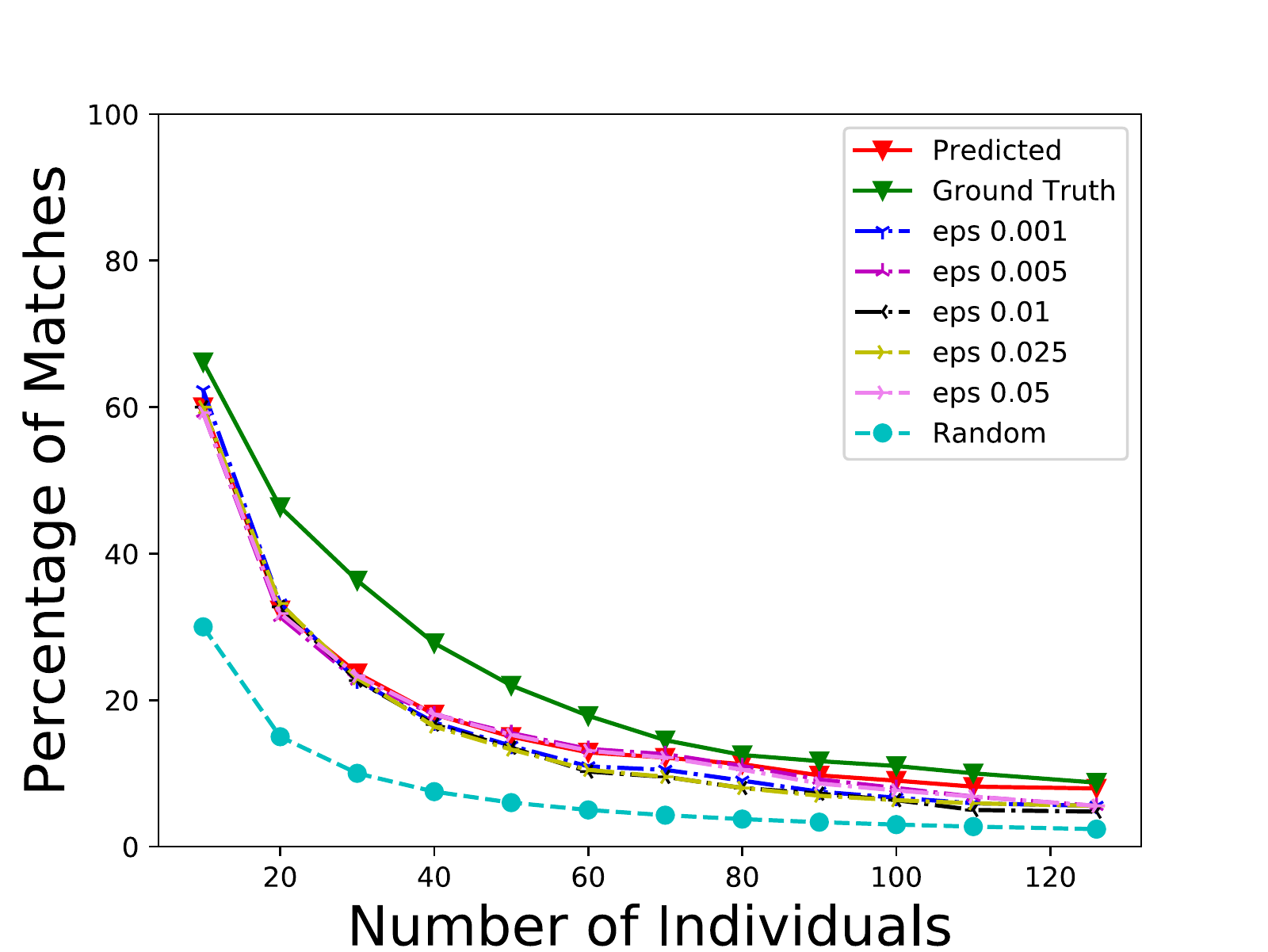}
    \caption{Hair Color - Top 3 - Robust, Baseline}
\end{subfigure}
\begin{subfigure}[t]{0.32\textwidth}
    \includegraphics[width=5.6cm]{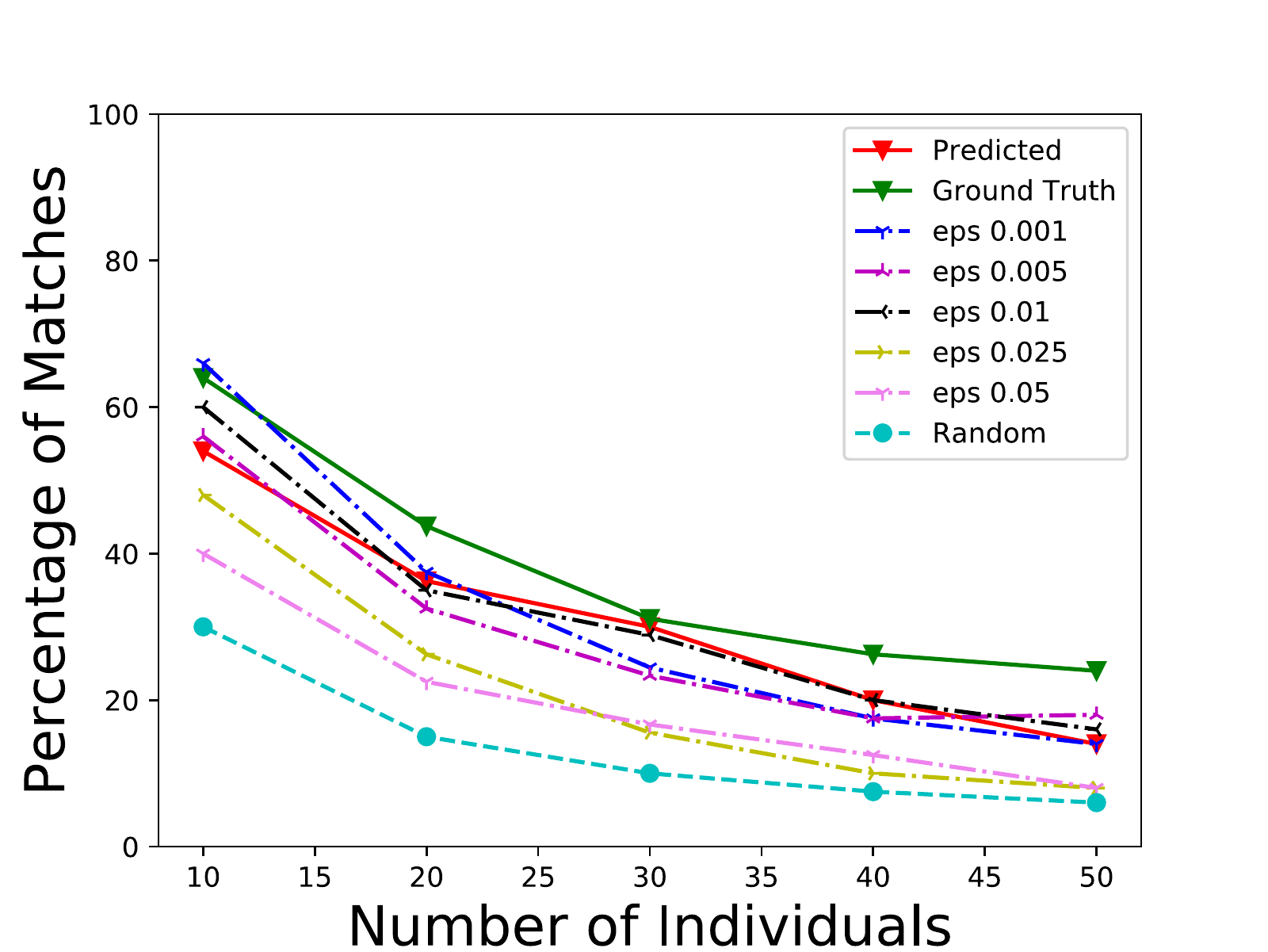}
    \caption{\emph{Universal Noise} - Top 3 - Robust, Baseline}
\end{subfigure}

\caption{\textbf{(a)-(e)}: Top-3 Matching accuracy with robust classifiers, with images perturbed at $\epsilon$ between $0$ and $0.05$ for a) Sex, b) Skin Color, c) Eye Color, d) Hair Color and e) \emph{Universal Noise}. Each classifier was trained with adversarial examples at $\epsilon=0.01$. \textbf{(f)-(j)}: Top-3 Baseline accuracy of matching with adversarially trained classifiers, but clean images for a) Sex, b) Skin Color, c) Eye Color, d) Hair Color and e) \emph{Universal Noise}. Legends in each figure show the values of $\epsilon$ used for training the network.}
\label{fig:robust_attacked_5}
\end{figure}

We next investigate if our adversarial noise defense may be impacted if the malicious actor in question trains phenotype-prediction models robust to such perturbations. A common way of doing this is adversarial training, where training is augmented with adversarial examples. First we look at the effect of retraining on adversarially perturbed images for various attack strengths. We use a value of $\epsilon=0.01$ to train each classifier, as our evaluation show it to be a somewhat optimal point, with regards to the tradeoff between effectiveness as a defense and perceptibility of image perturbation. Supplementary Fig.~\ref{fig:robust_noattack_1} (a)-(d) show top-$1$ matching results for phenotype classifiers adversarially trained at $\epsilon=0.01$ and attacked at various values of $\epsilon$ ranging from $0.001$ to $0.05$. Similarly, Supplementary Fig.~\ref{fig:robust_attacked_5} (a)-(d) show top-$3$ matching results. 

In case of sex and skin color, adversarial training boosts robustness to perturbed images as expected, it makes no difference, or even degrades performance slightly for eye color and hair color prediction. The slight loss in performance on adversarial examples is somewhat unusual, and most likely due to limited training data. Adversarial training against the \emph{Universal Noise} approach boosts accuracy for small populations, but quickly falls to zero as population size approaches 50.

Next we are interested in how adversarial training affects baseline performance, i.e. matching accuracy on clean images. It is typical for baseline performance to degrade for more robust models. Supplementary Fig.~\ref{fig:robust_noattack_1} (e)-(h) and Supplementary Fig.~\ref{fig:robust_attacked_5} (f)-(j) present top-$1$ and top-$3$ matching results respectively for various values of $\epsilon$ at which the model was adversarially trained, but using clean images. In all cases, we observe a performance decrease, compared to the original models' performance on the same clean images. We observe that performance degrades to the point where adversarial training may be more detrimental than beneficial to a malicious actor attempting re-identification on a genomic dataset, especially in the case of sex classification, where training against sufficiently strong adversarial noise ($\epsilon=0.025$) reduces accuracy to below random guessing. In case of the \emph{Universal Noise} approach, while accuracy remains above random, note that retraining was only run for a very small number of data-points, due to the lack of paired image-DNA data, which is not required when individually retraining phenotype classifiers against PGD.



\section{Evaluation on a Synthetic Dataset}
While our results on the OpenSNP dataset portray a realistic picture of what the risk of re-identification is, given publicly available data, we wish to also test how our approach performs on a larger dataset of higher quality - essentially controlling for model transfer error in the transfer-learning process, while accounting for much larger population sizes.
To accomplish this, we create a synthetic dataset using the CelebA dataset, by predicting genomes using existing OpenSNP data. We consider two settings - a) an `ideal' scenario where each image is assigned a genmoe that maximizes the probability of the phenotypes detected in the image (\emph{Synthetic-Ideal}), and b) a `realistic' scenario where genomes are randomly picked from a subset of OpenSNP individuals with the same set of phenotypes (\emph{Synthetic-Realistic}). Note that we do not require the OpenSNP individuals to be of the same sex in this synthesis process due to data sparsity - the considered facial phenotypes being independent of sex is a reasonable assumption. As not all phenotypes are labeled in the CelebA dataset, we start with $1000$ manually annotated images, and after cleaning and removing ambiguous cases, we are left with $456$ individuals in the synthetic dataset.

\begin{figure}[]
\centering
\begin{subfigure}[t]{0.32\textwidth}
    \includegraphics[width=5.6cm]{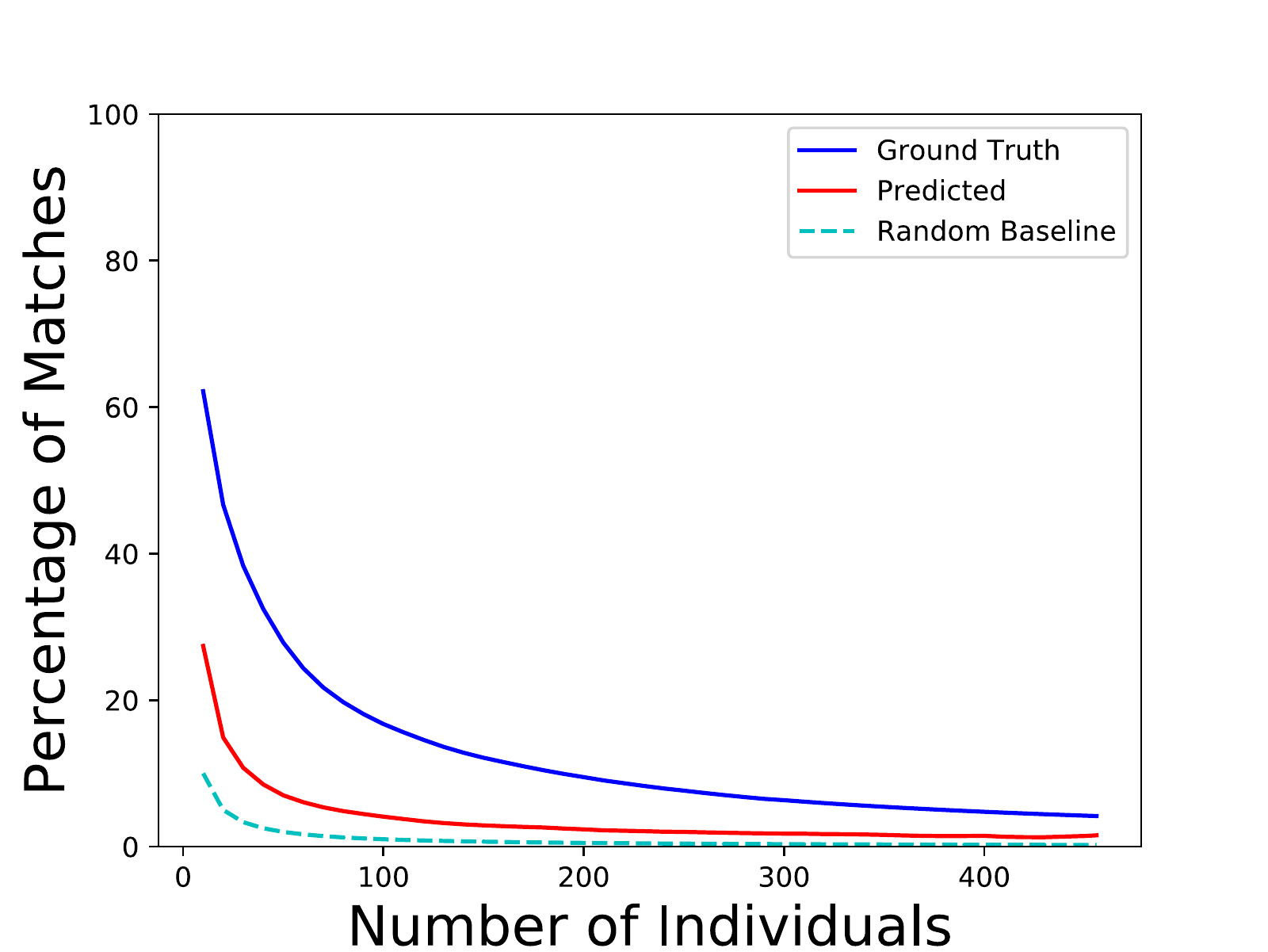}
    \caption{Top 1 - \emph{Synthetic - Ideal}}
\end{subfigure}
\begin{subfigure}[t]{0.32\textwidth}
    \includegraphics[width=5.6cm]{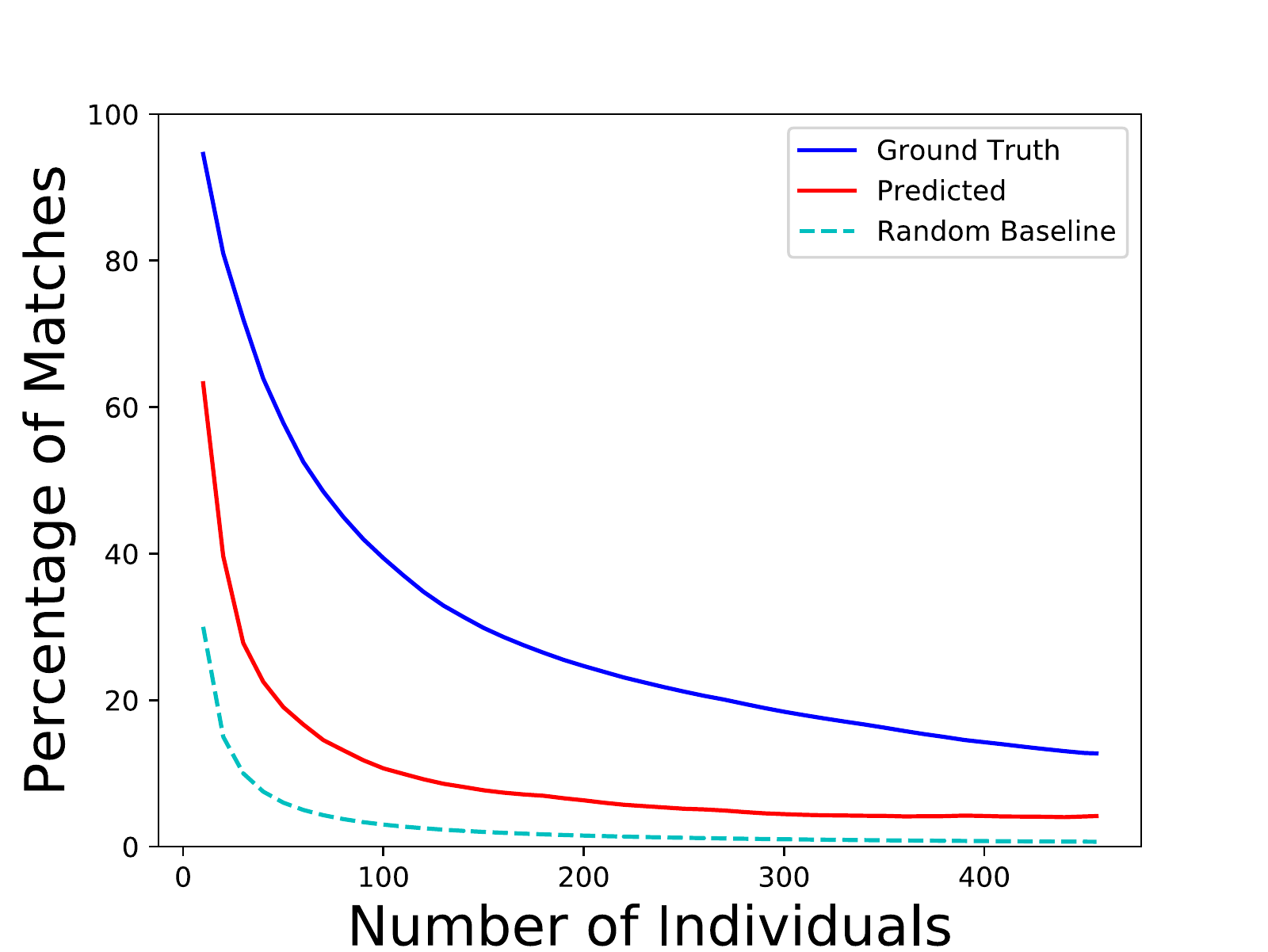}
    \caption{Top 3 - \emph{Synthetic - Ideal}}
\end{subfigure}
\begin{subfigure}[t]{0.32\textwidth}
    \includegraphics[width=5.6cm]{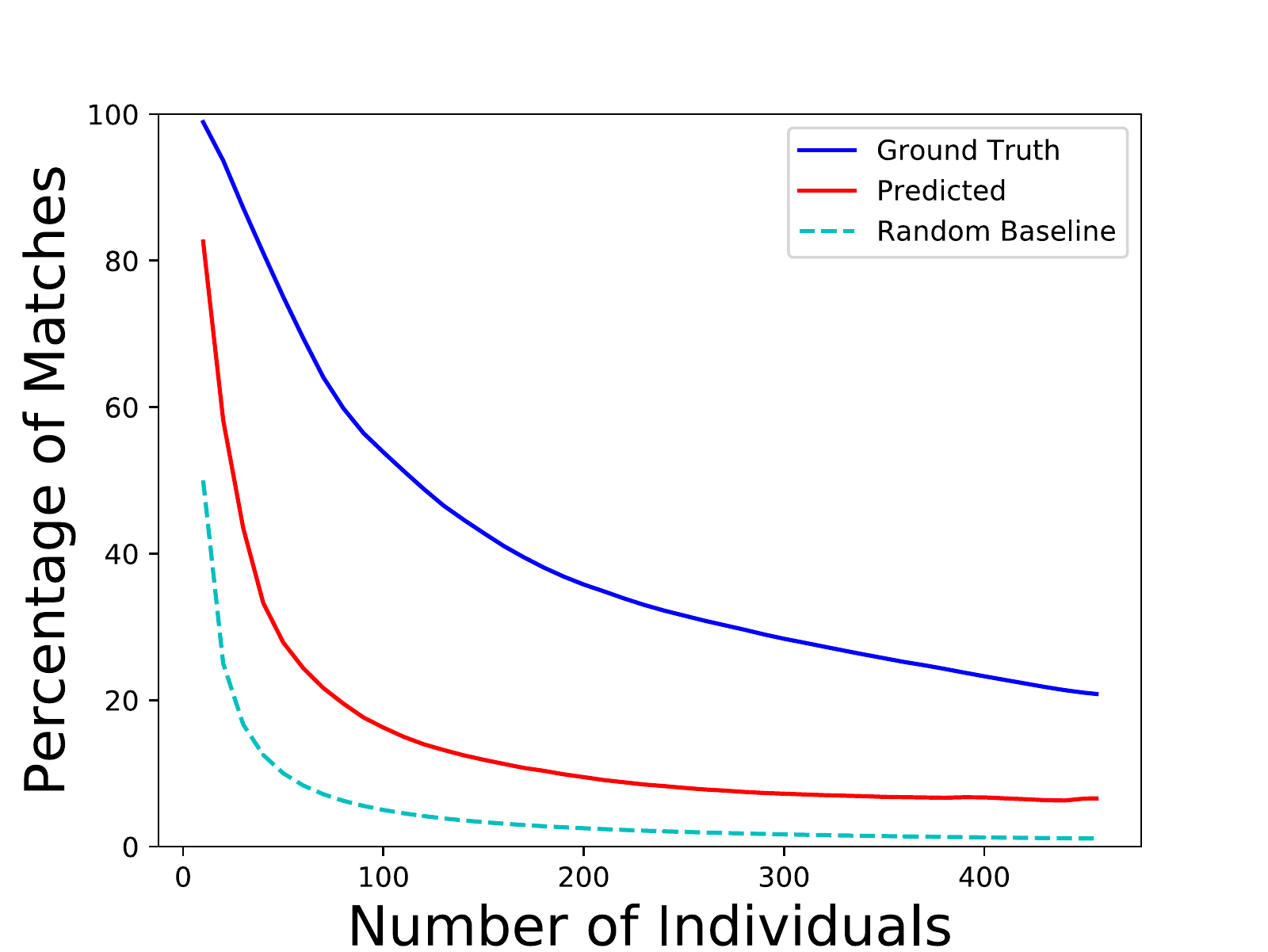}
    \caption{Top 5 - \emph{Synthetic - Ideal}}
\end{subfigure}
\begin{subfigure}[t]{0.32\textwidth}
    \includegraphics[width=5.6cm]{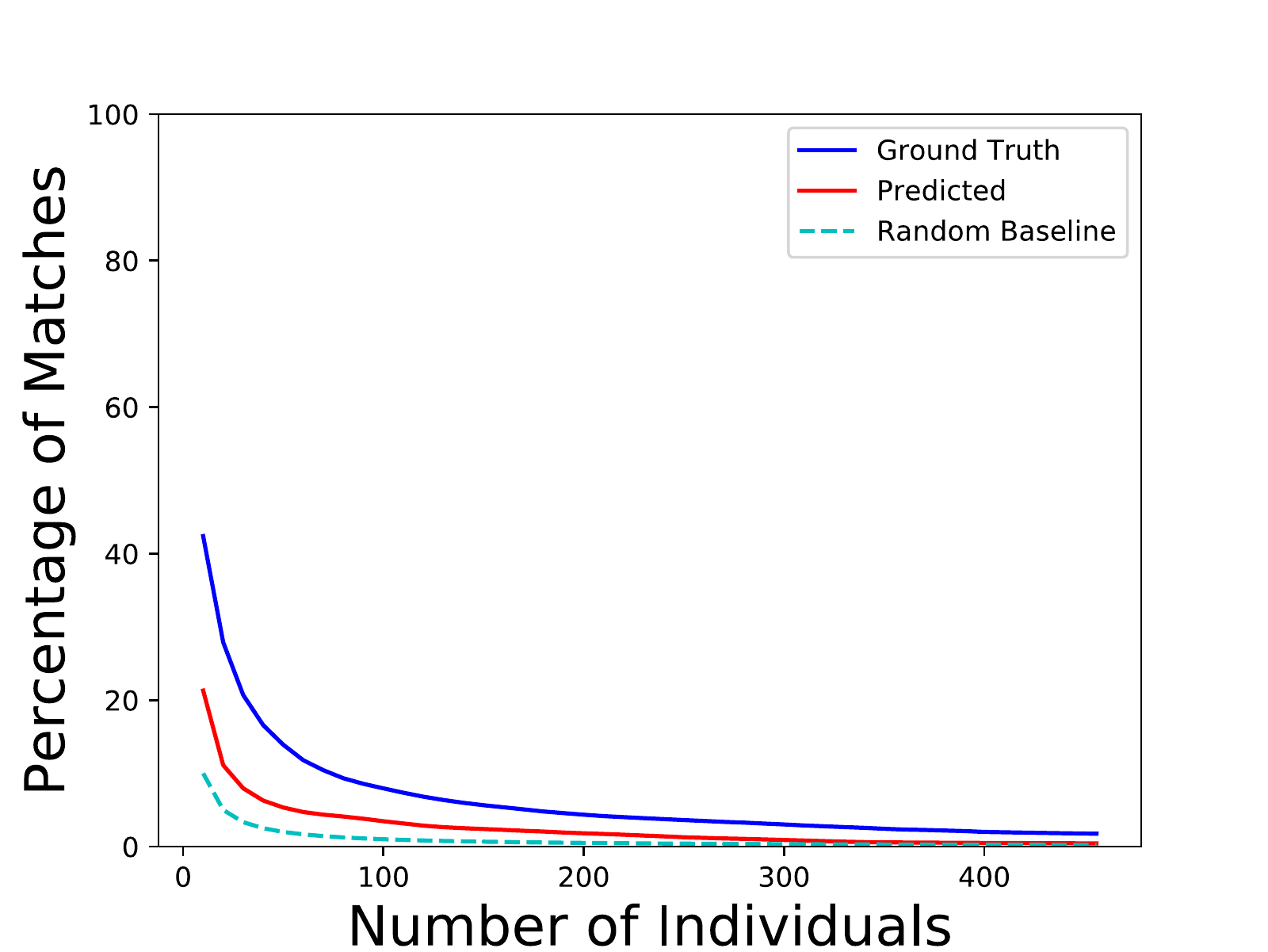}
    \caption{Top 1 - \emph{Synthetic - Realistic}}
\end{subfigure}
\begin{subfigure}[t]{0.32\textwidth}
    \includegraphics[width=5.6cm]{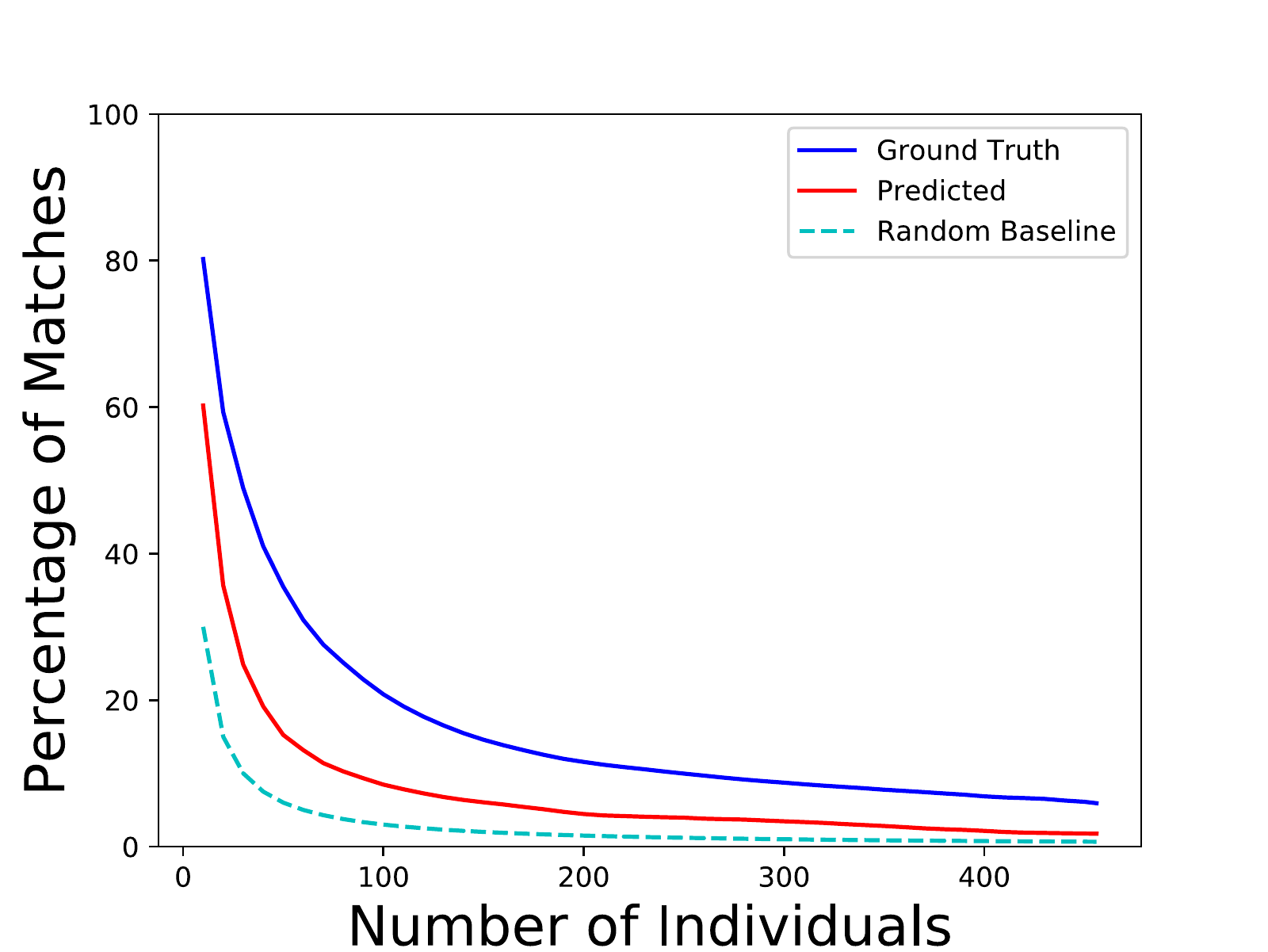}
    \caption{Top 3 - \emph{Synthetic - Realistic}}
\end{subfigure}
\begin{subfigure}[t]{0.32\textwidth}
    \includegraphics[width=5.6cm]{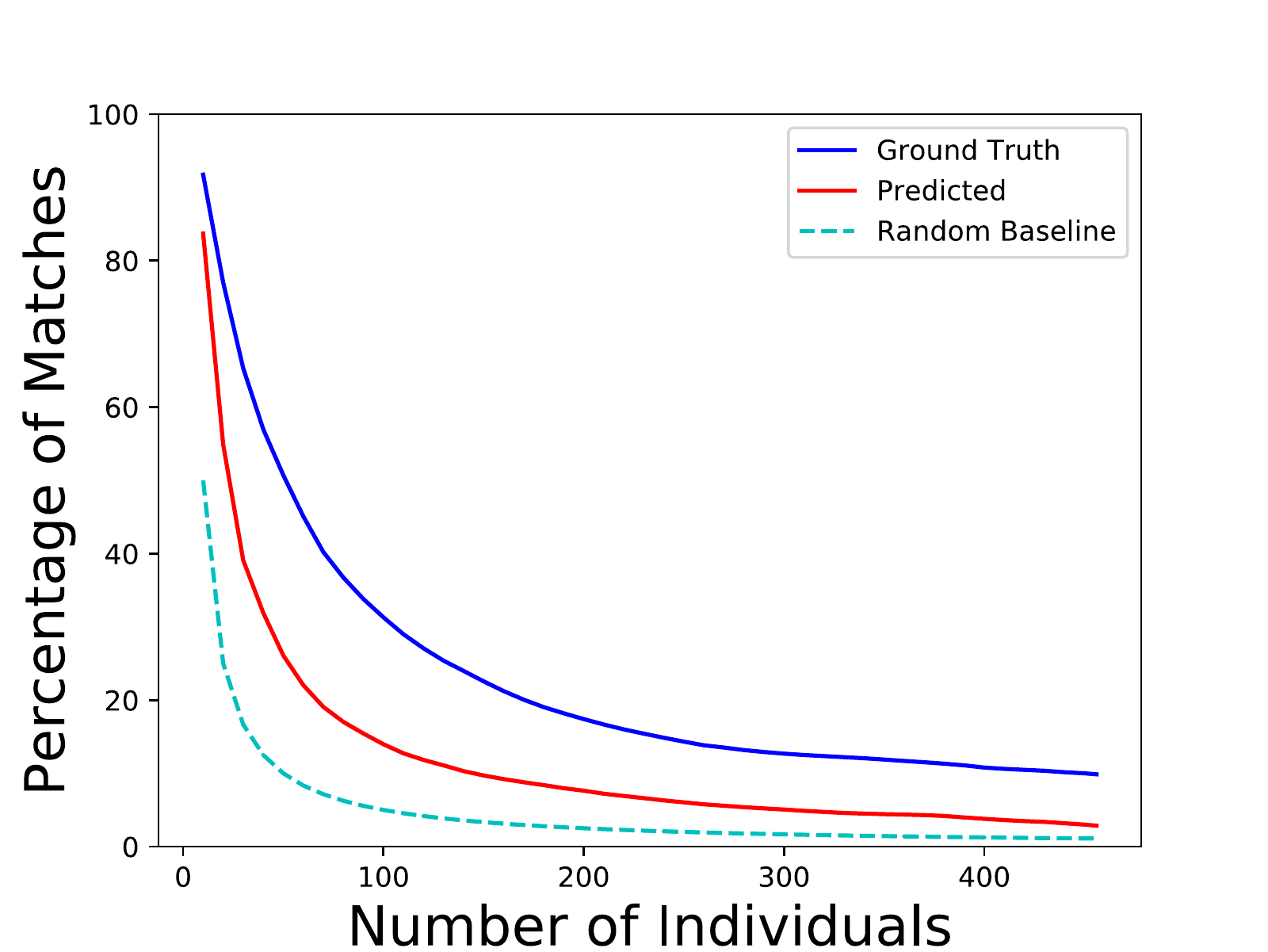}
    \caption{Top 5 - \emph{Synthetic - Realistic}}
\end{subfigure}

\begin{subfigure}[t]{0.32\textwidth}
    \includegraphics[width=5.6cm]{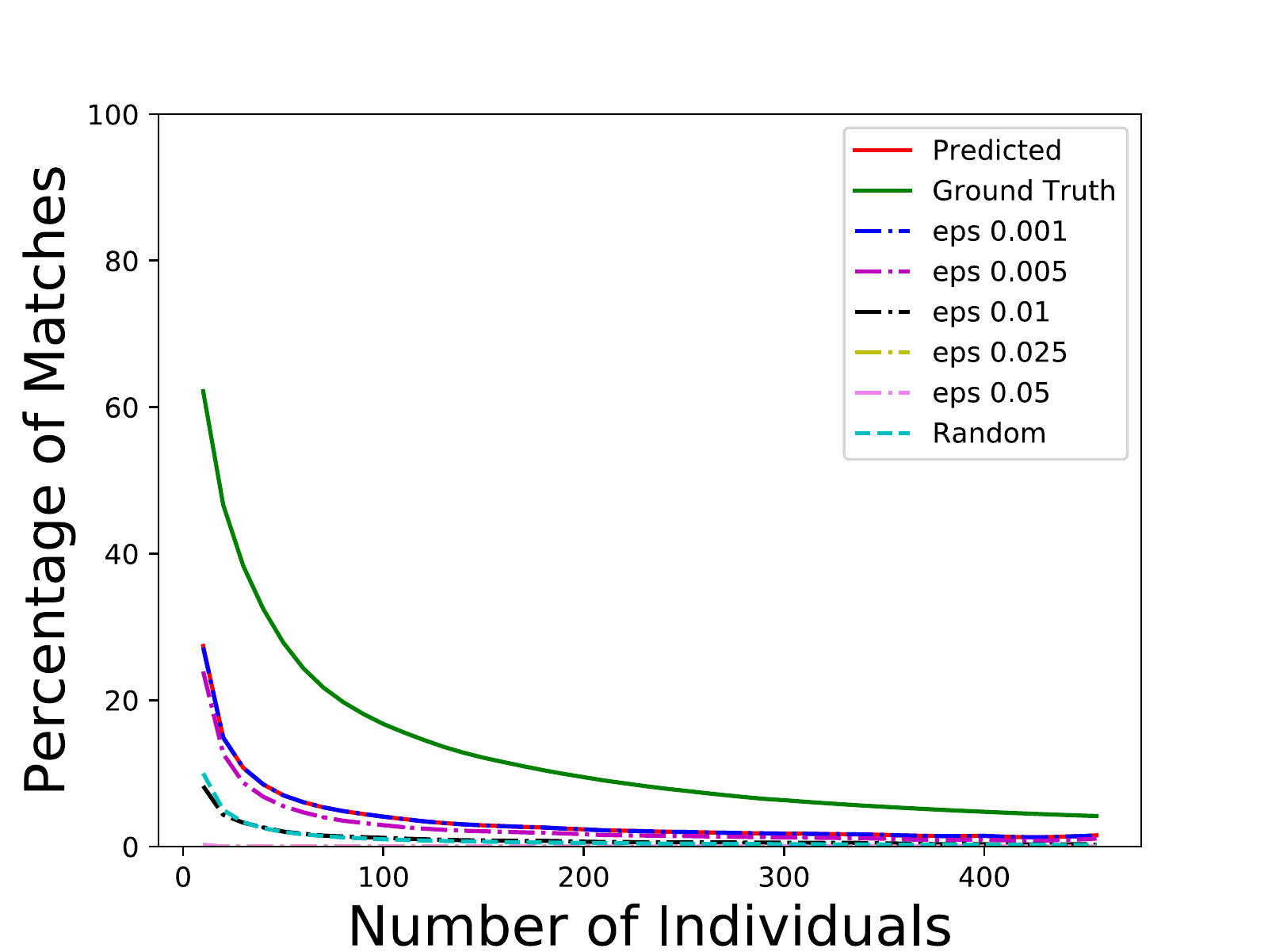}
    \caption{Top 1 - \emph{Synthetic - Ideal}, PGD}
\end{subfigure}
\begin{subfigure}[t]{0.32\textwidth}
    \includegraphics[width=5.6cm]{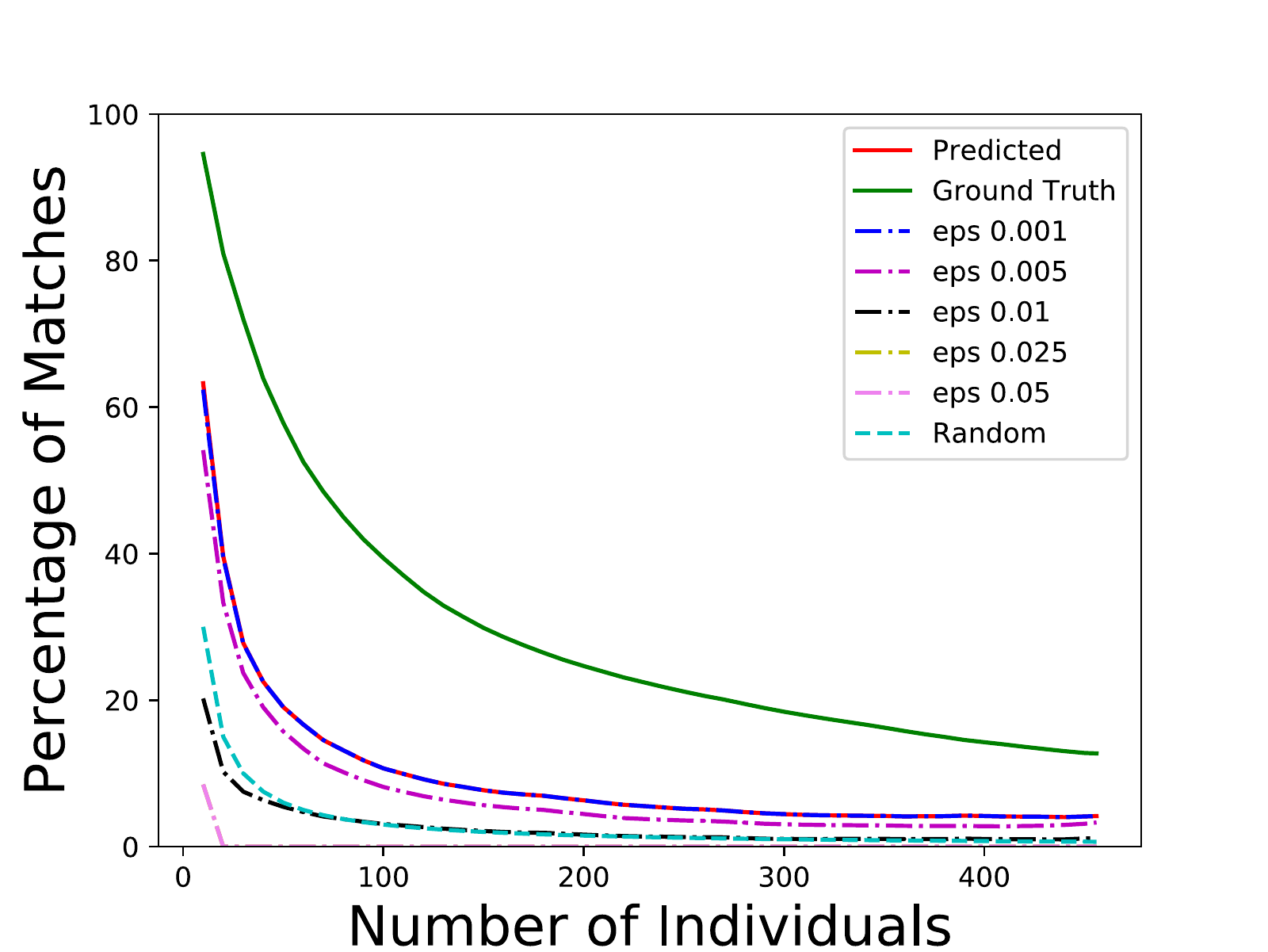}
    \caption{Top 3 - \emph{Synthetic - Ideal}, PGD}
\end{subfigure}
\begin{subfigure}[t]{0.32\textwidth}
    \includegraphics[width=5.6cm]{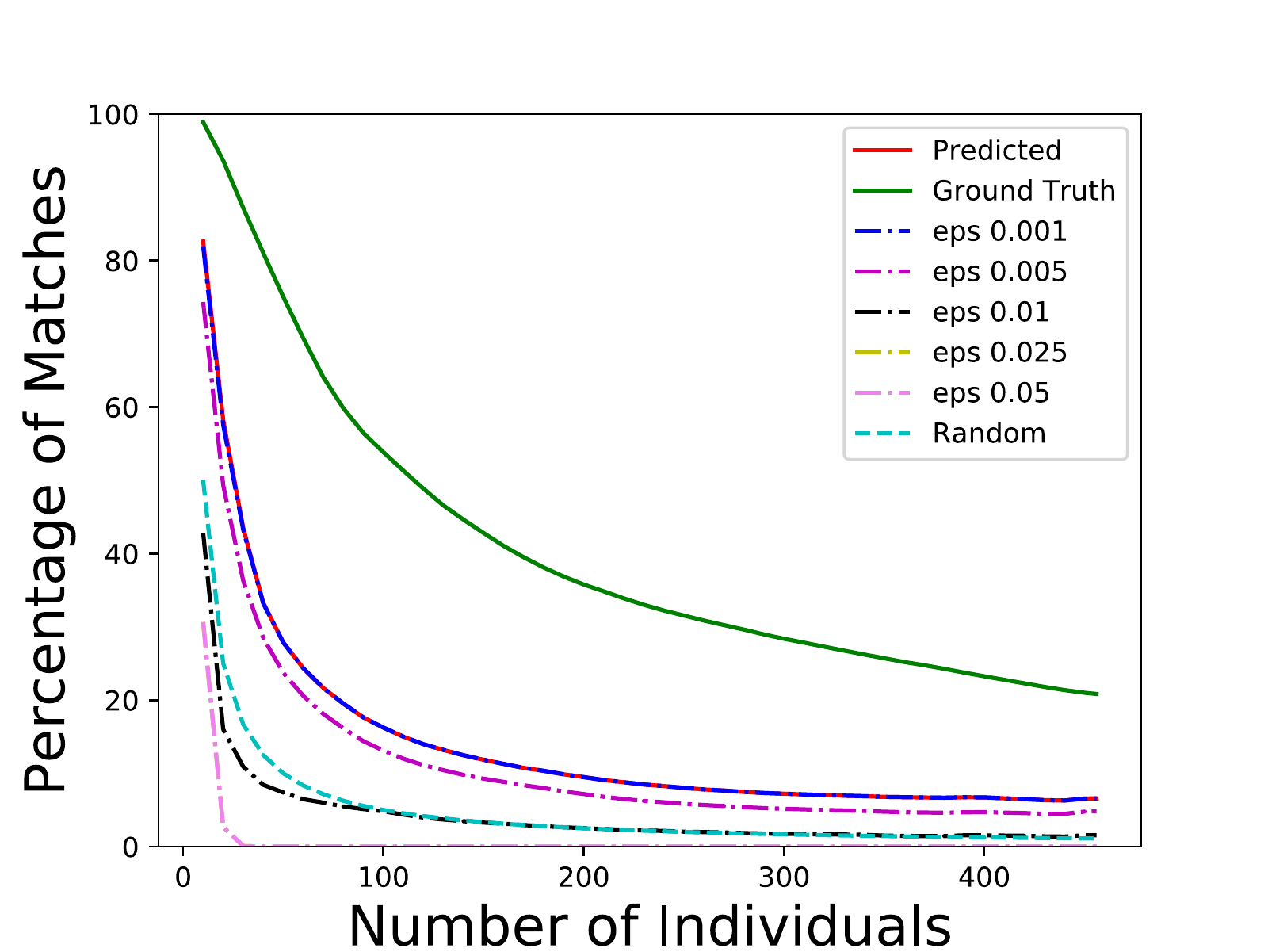}
    \caption{Top 5 - \emph{Synthetic - Ideal}, PGD}
\end{subfigure}
\begin{subfigure}[t]{0.32\textwidth}
    \includegraphics[width=5.6cm]{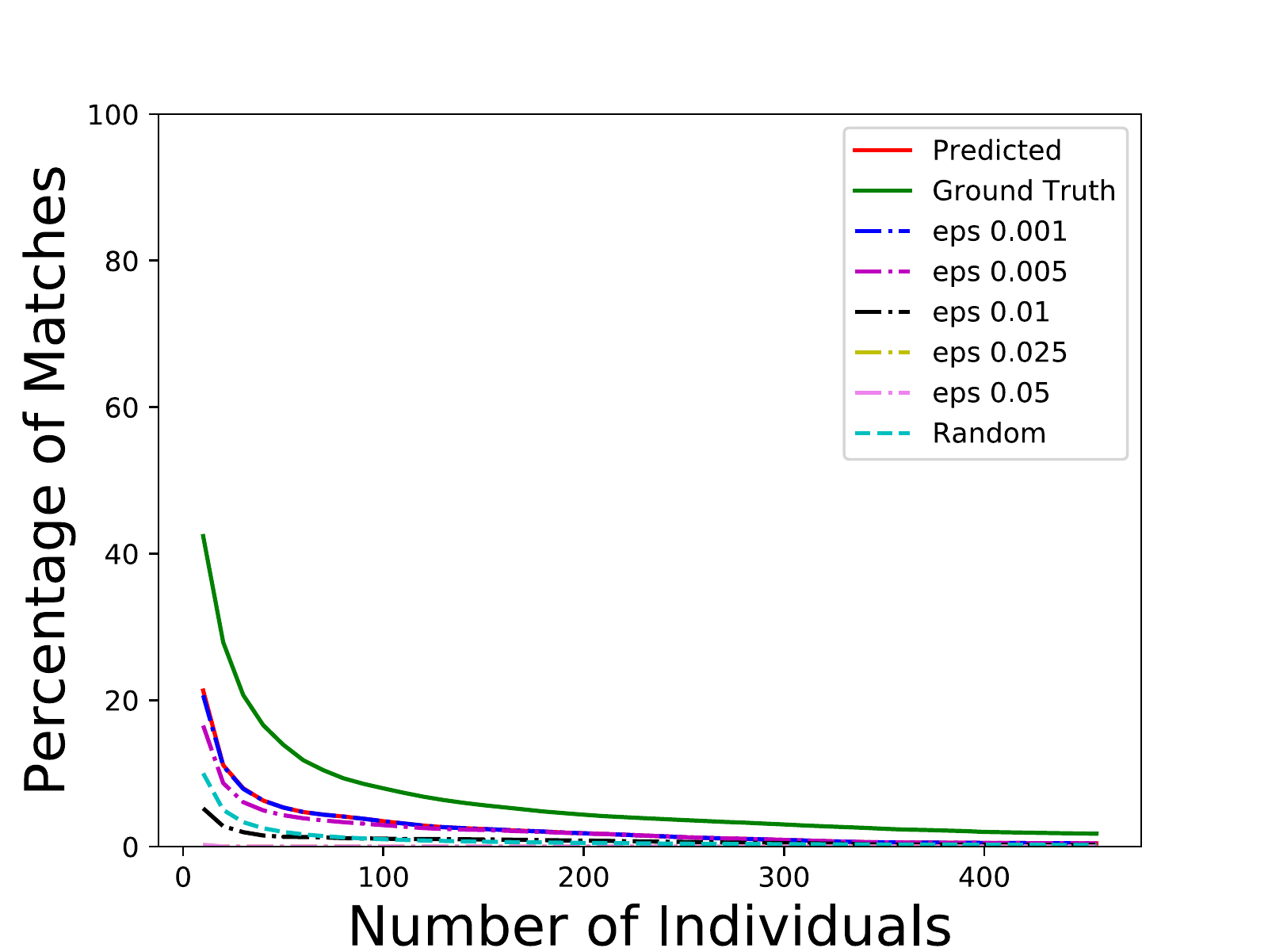}
    \caption{Top 1 - \emph{Synthetic - Realistic}, PGD}
\end{subfigure}
\begin{subfigure}[t]{0.32\textwidth}
    \includegraphics[width=5.6cm]{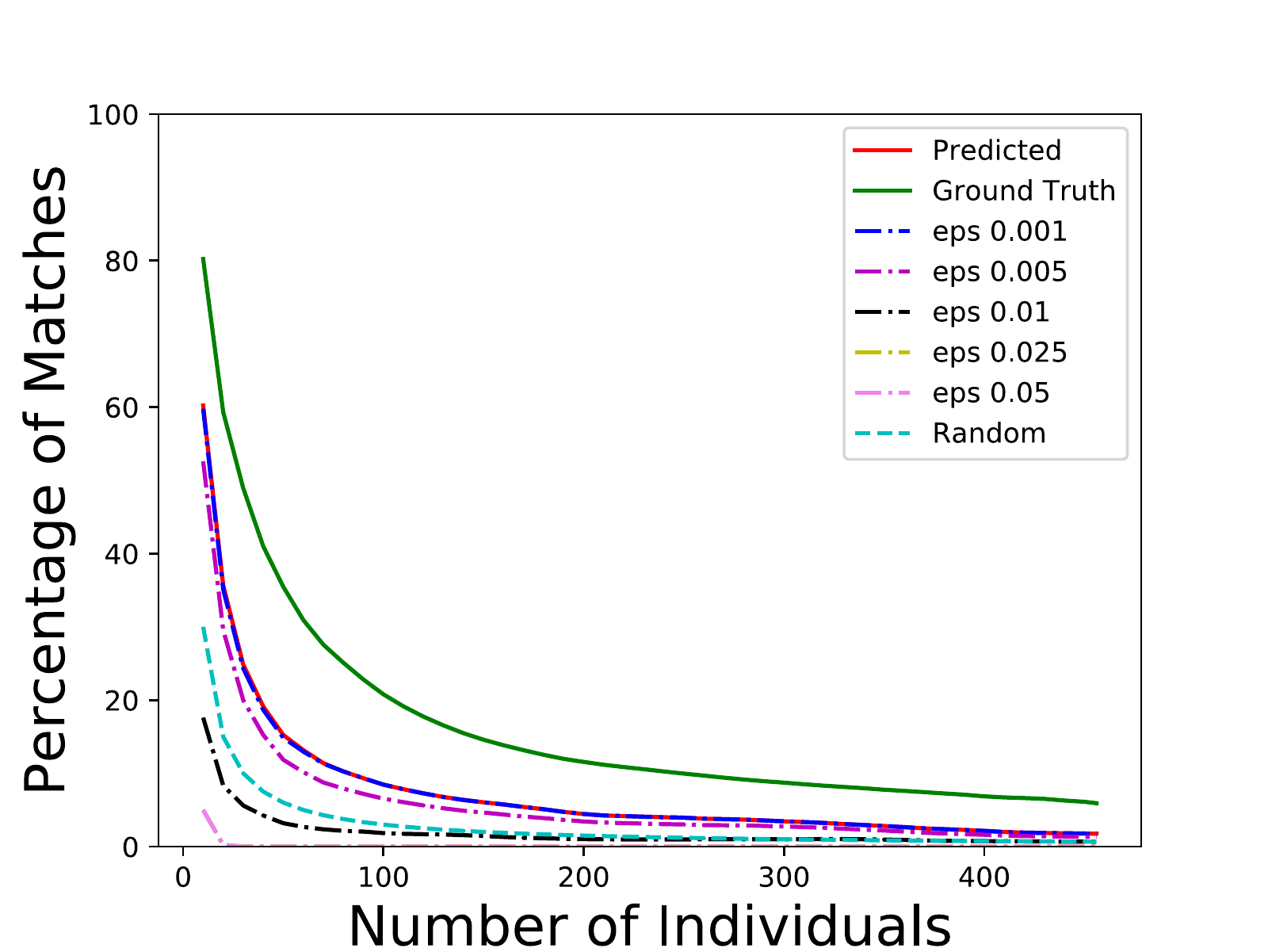}
    \caption{Top 3 - \emph{Synthetic - Realistic}, PGD}
\end{subfigure}
\begin{subfigure}[t]{0.32\textwidth}
    \includegraphics[width=5.6cm]{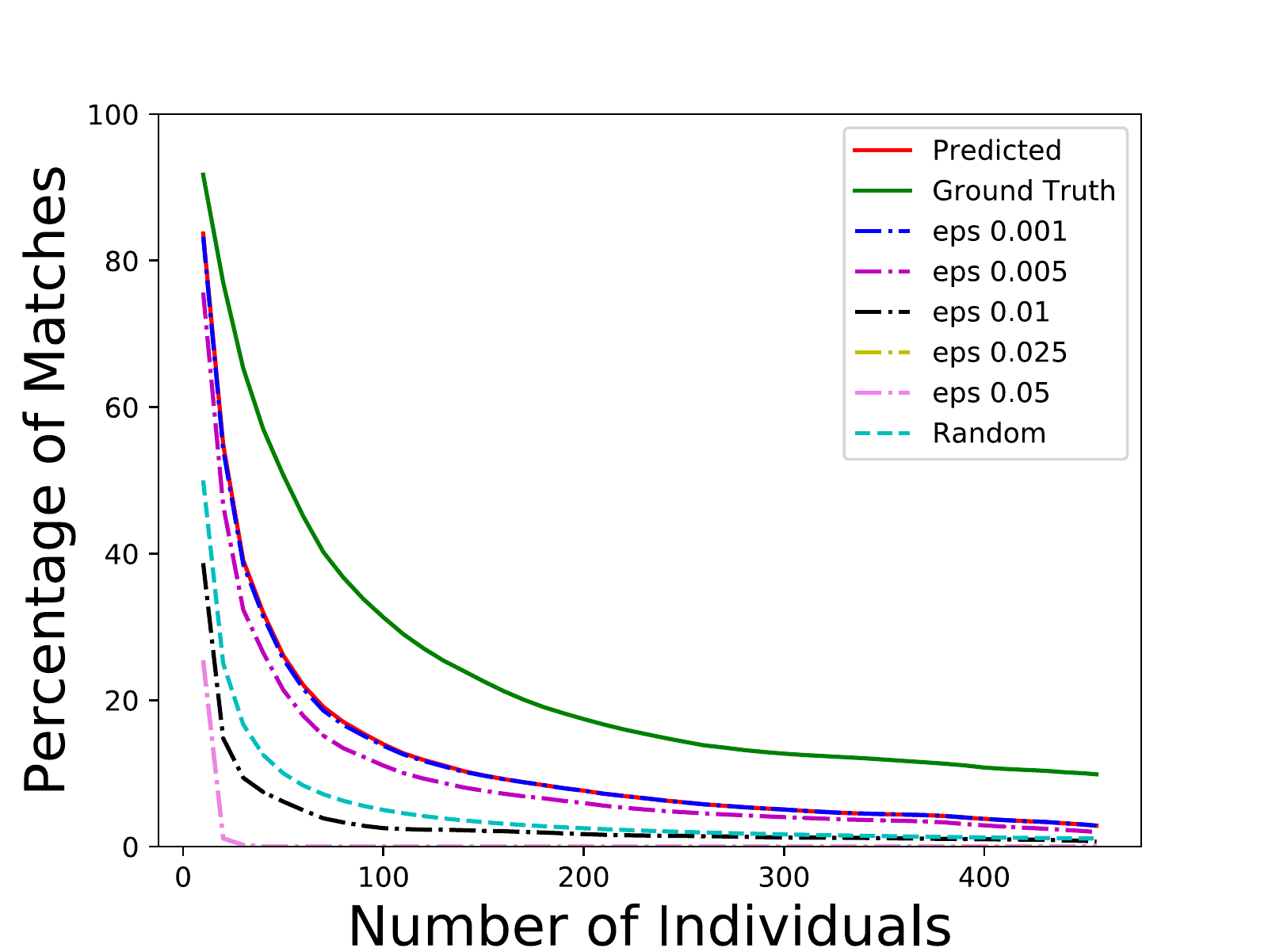}
    \caption{Top 5 - \emph{Synthetic - Realistic}, PGD}
\end{subfigure}
\caption{\textbf{(a)-(f)}: Matching Accuracy with the Ideal (top row) and Realistic (bottom row) synthetic datasets for top-$1$, top-$3$ and top-$5$. The ground truth accuracy is much higher in the ideal scenario - this is to be expected as genomes are explicitly picked to be the most representative of a corresponding image. The accuracy of predicted matches is much lower than results on the OpenSNP individuals would suggest. We narrow the cause of this down to the challenges in predicting eye-color from images. \textbf{(g)-(l)}: Matching Accuracy with PGD-perturbed images targeting the sex-classifier, for the synthetic datasets for top-$1$, top-$3$ and top-$5$. Much like the OpenSNP data, fooling the prediction of sex from images proves to be a highly effective defense for the synthetic datasets, although at a higher value of $\epsilon=0.025$.}
\label{fig:adv_synth}
\end{figure}

\begin{figure}[]
\centering
\begin{subfigure}[t]{0.32\textwidth}
    \includegraphics[width=5.6cm]{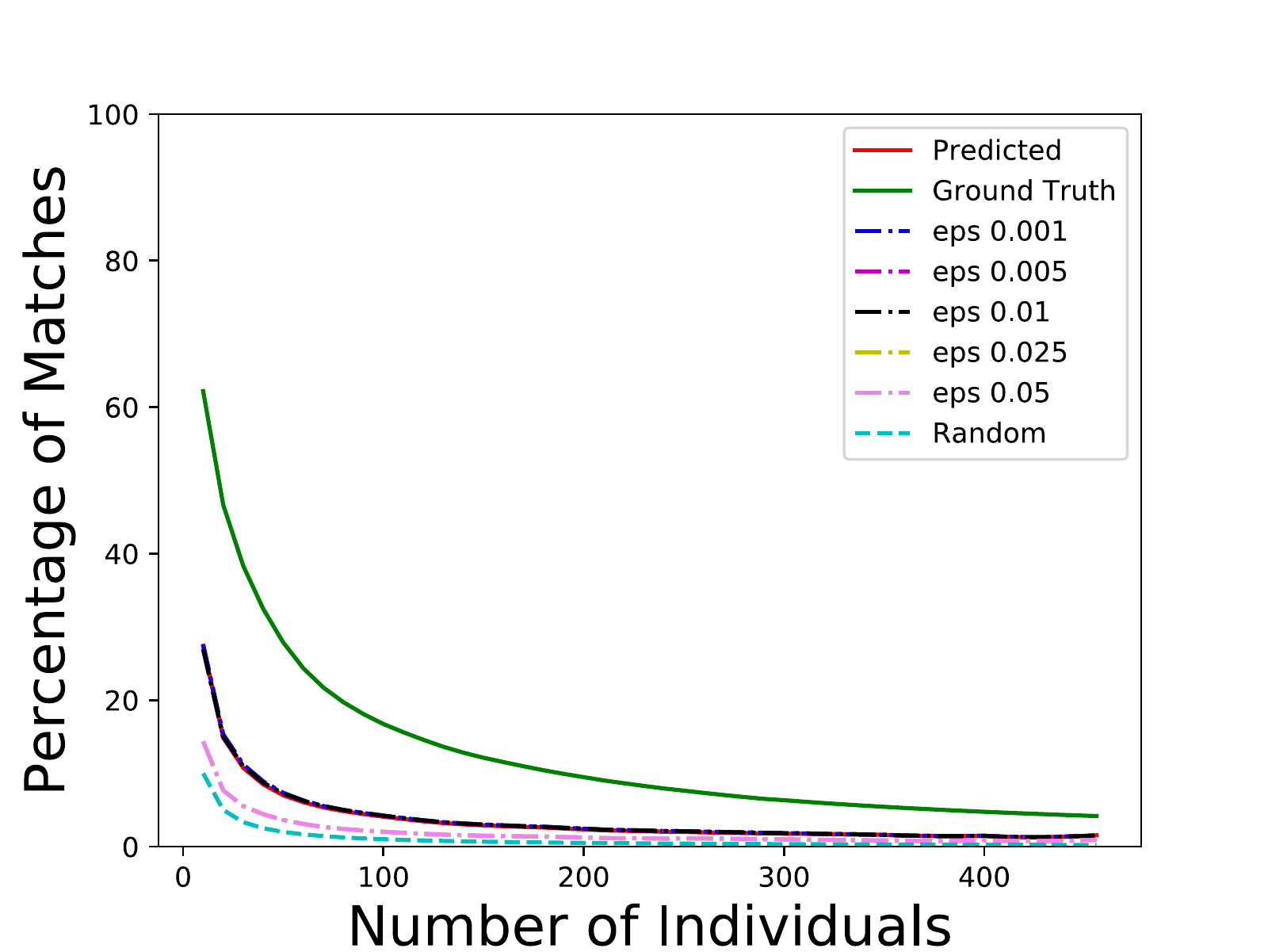}
    \caption{Top 1 - \emph{S-I}, Robust, Baseline}
\end{subfigure}
\begin{subfigure}[t]{0.32\textwidth}
    \includegraphics[width=5.6cm]{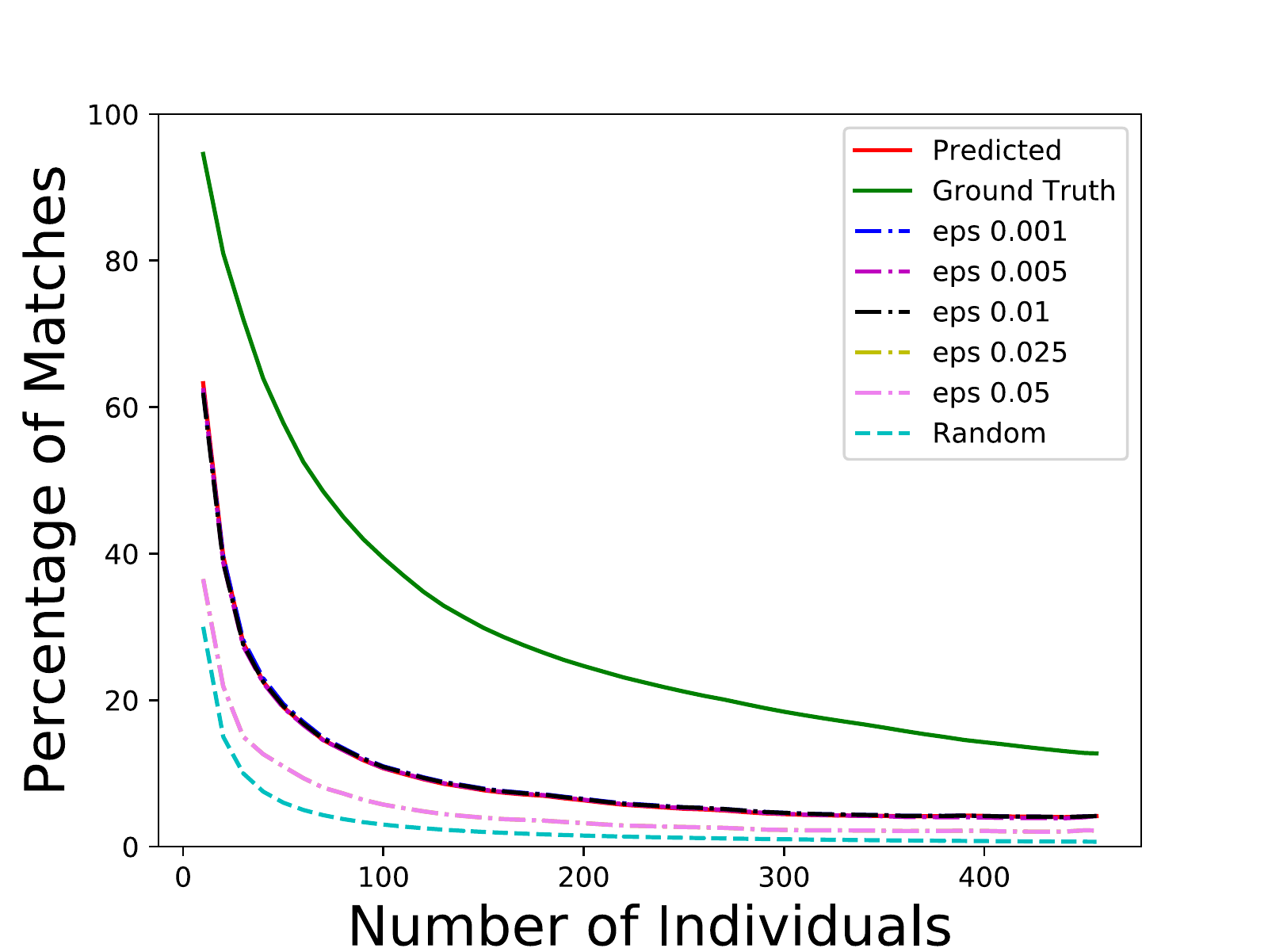}
    \caption{Top 3 - \emph{S-I}, Robust, Baseline}
\end{subfigure}
\begin{subfigure}[t]{0.32\textwidth}
    \includegraphics[width=5.6cm]{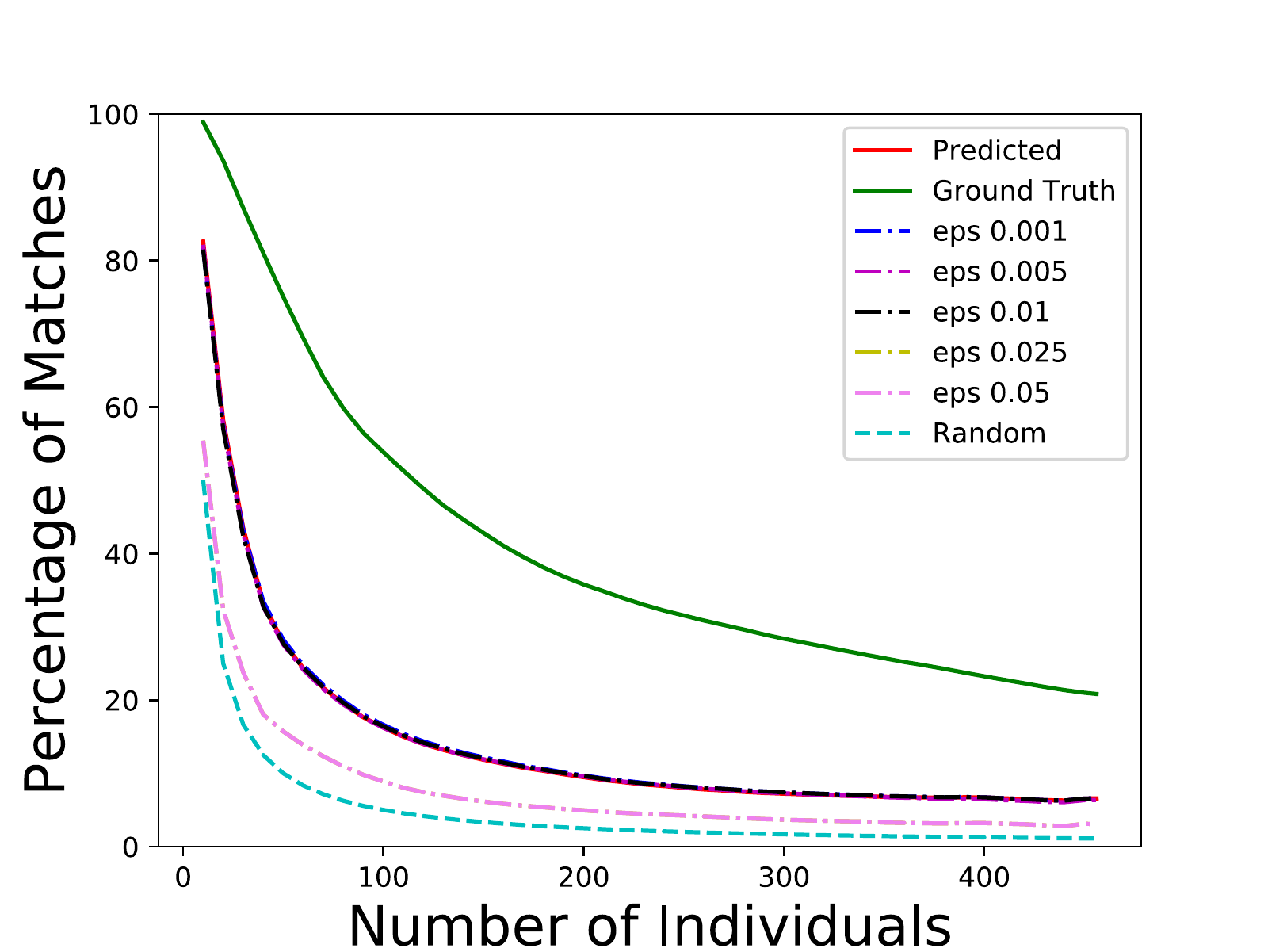}
    \caption{Top 5 - \emph{S-I}, Robust, Baseline}
\end{subfigure}
\begin{subfigure}[t]{0.32\textwidth}
    \includegraphics[width=5.6cm]{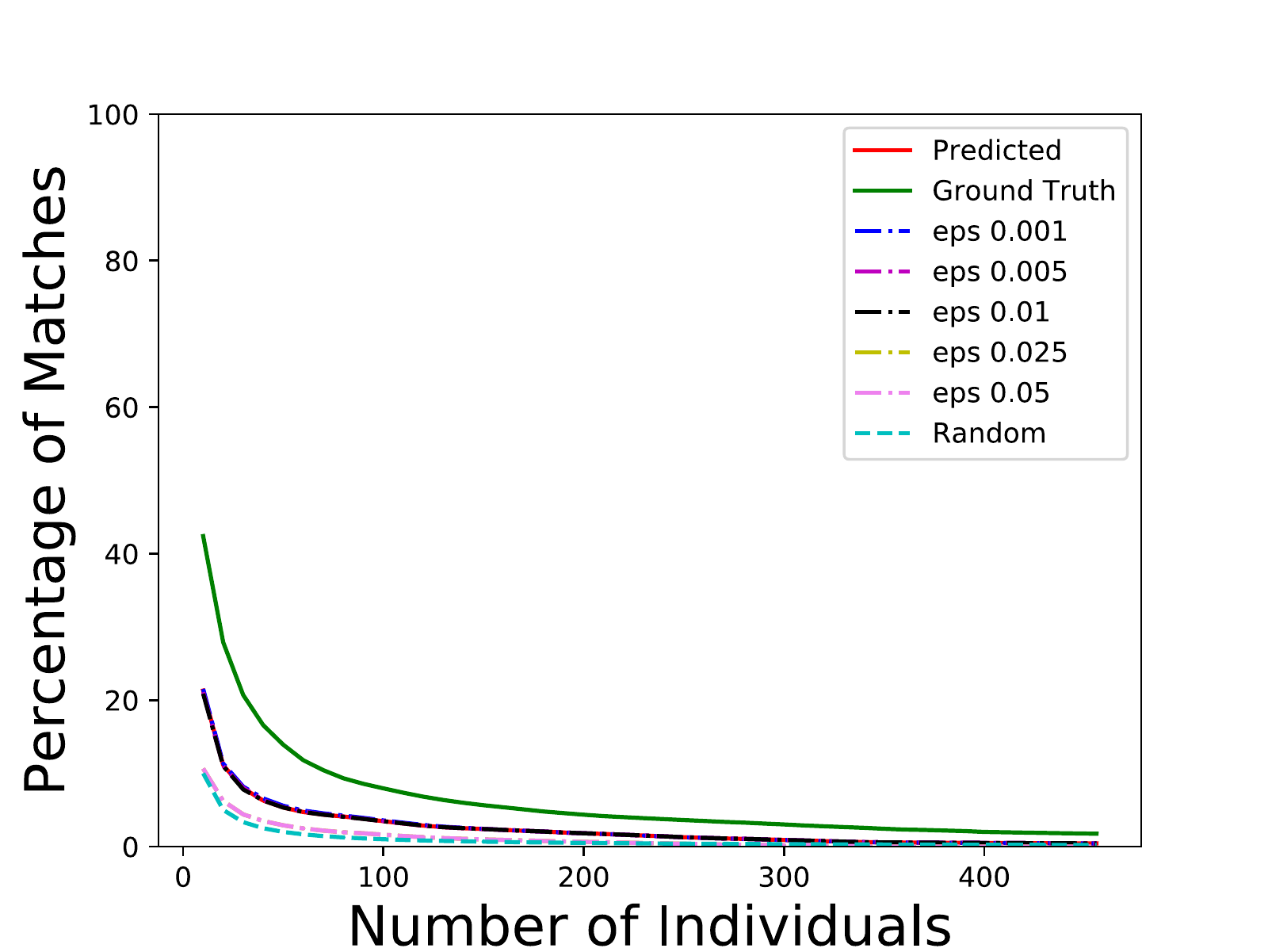}
    \caption{Top 1 - \emph{S-R}, Robust, Baseline}
\end{subfigure}
\begin{subfigure}[t]{0.32\textwidth}
    \includegraphics[width=5.6cm]{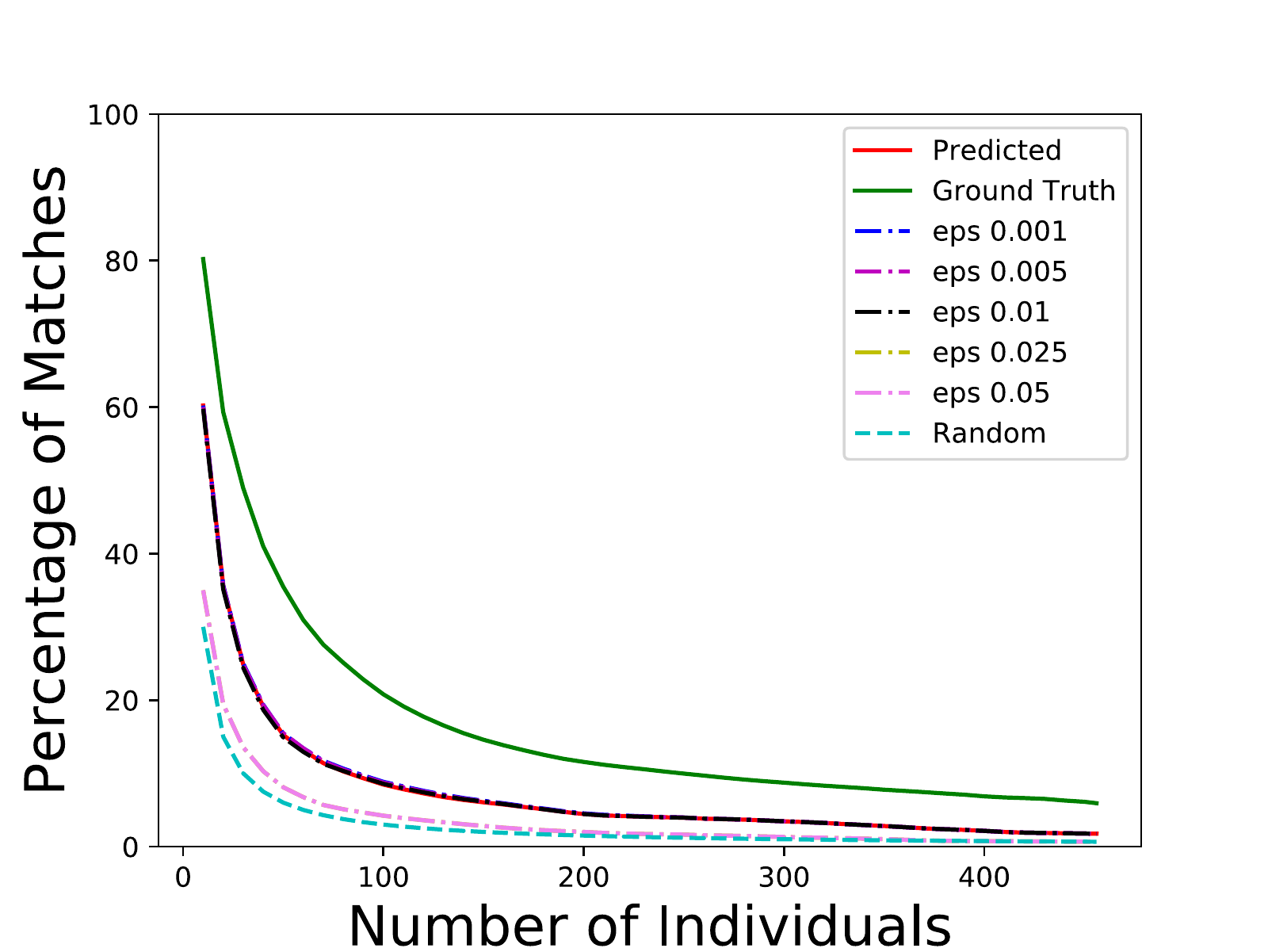}
    \caption{Top 3 - \emph{S-R}, Robust, Baseline}
\end{subfigure}
\begin{subfigure}[t]{0.32\textwidth}
    \includegraphics[width=5.6cm]{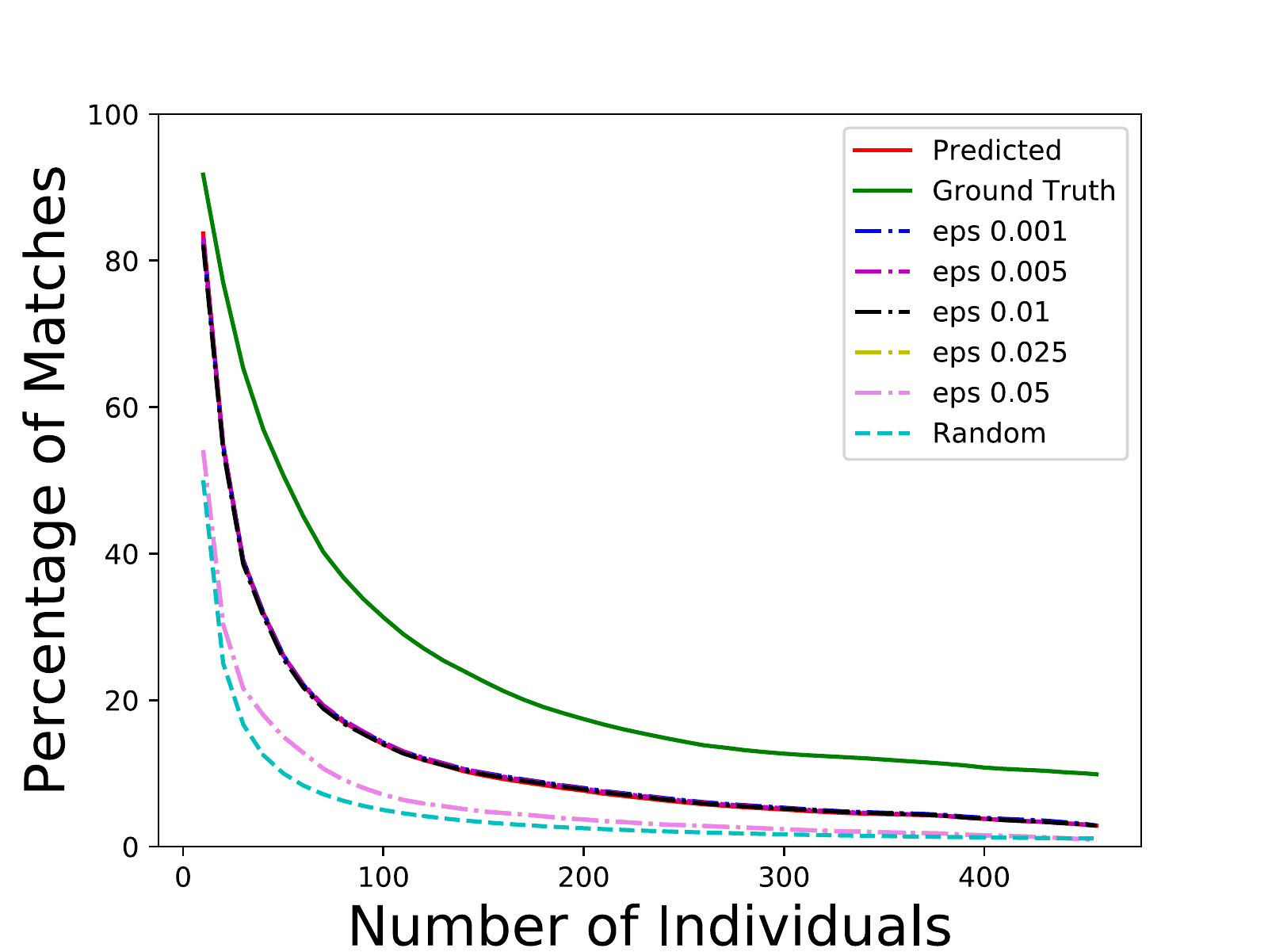}
    \caption{Top 5 - \emph{S-R}, Robust, Baseline}
\end{subfigure}

\begin{subfigure}[t]{0.32\textwidth}
    \includegraphics[width=5.6cm]{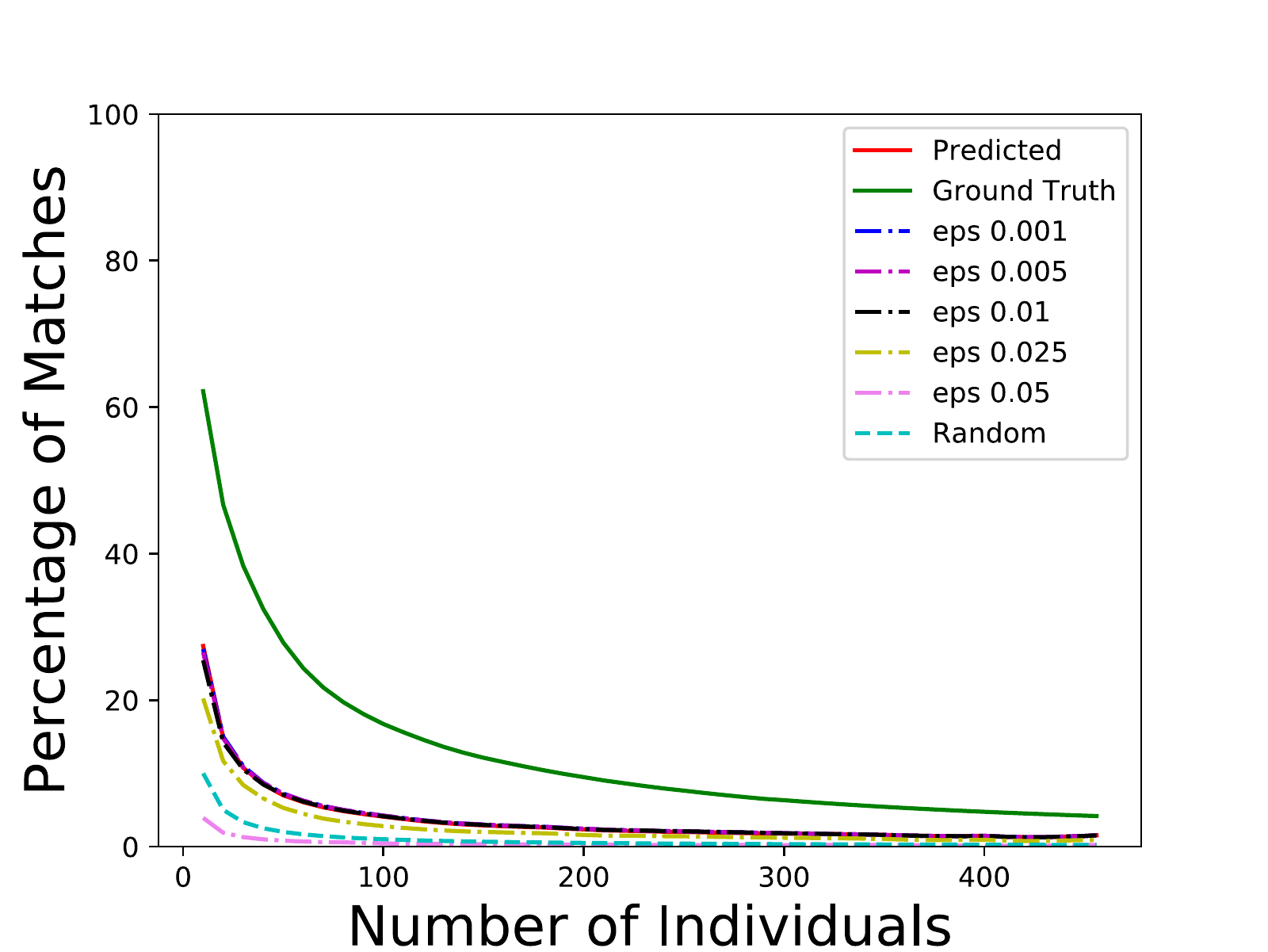}
    \caption{Top 1 - \emph{S-I}, Robust, Attacked}
\end{subfigure}
\begin{subfigure}[t]{0.32\textwidth}
    \includegraphics[width=5.6cm]{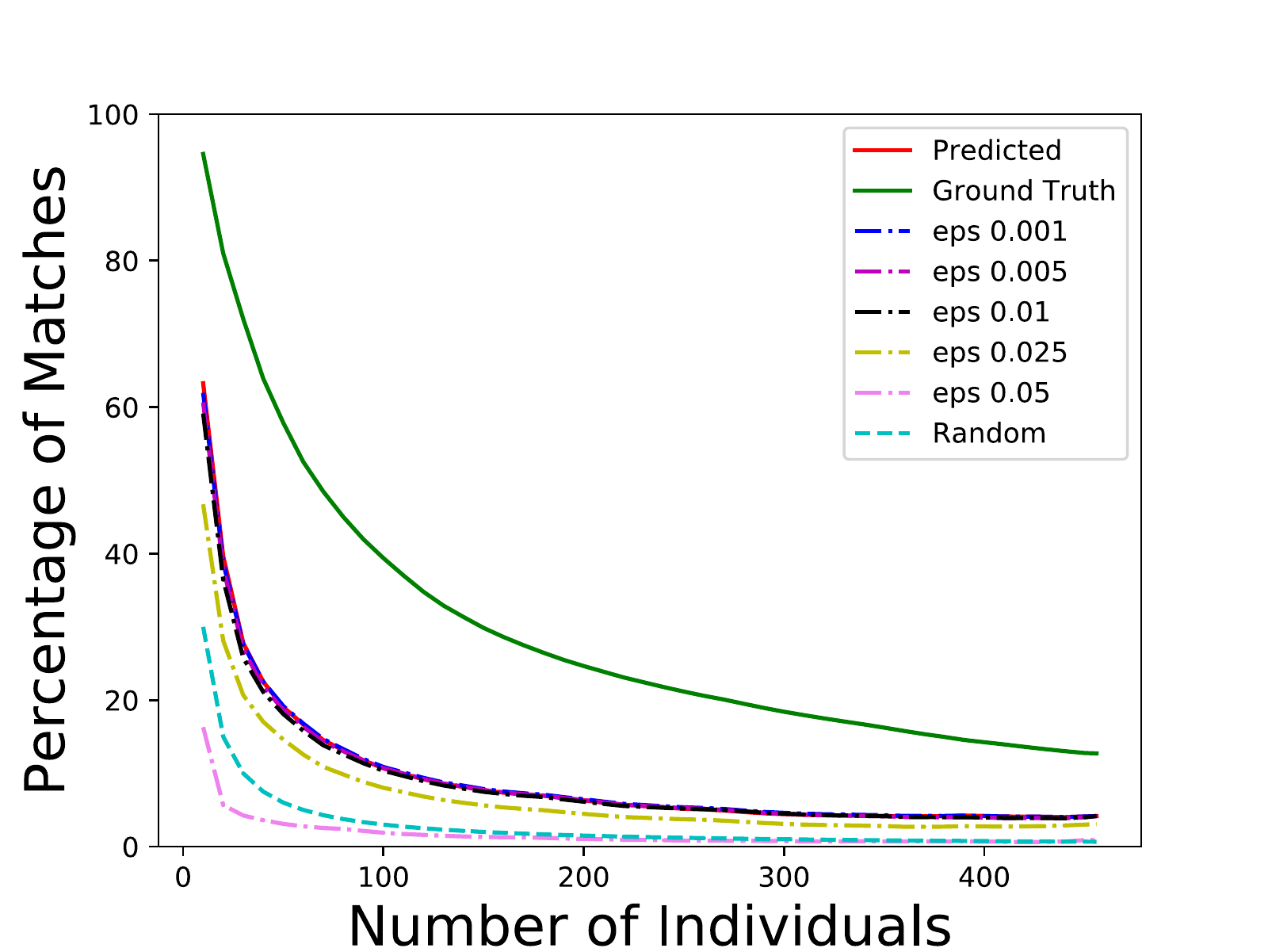}
    \caption{Top 3 - \emph{S-I}, Robust, Attacked}
\end{subfigure}
\begin{subfigure}[t]{0.32\textwidth}
    \includegraphics[width=5.6cm]{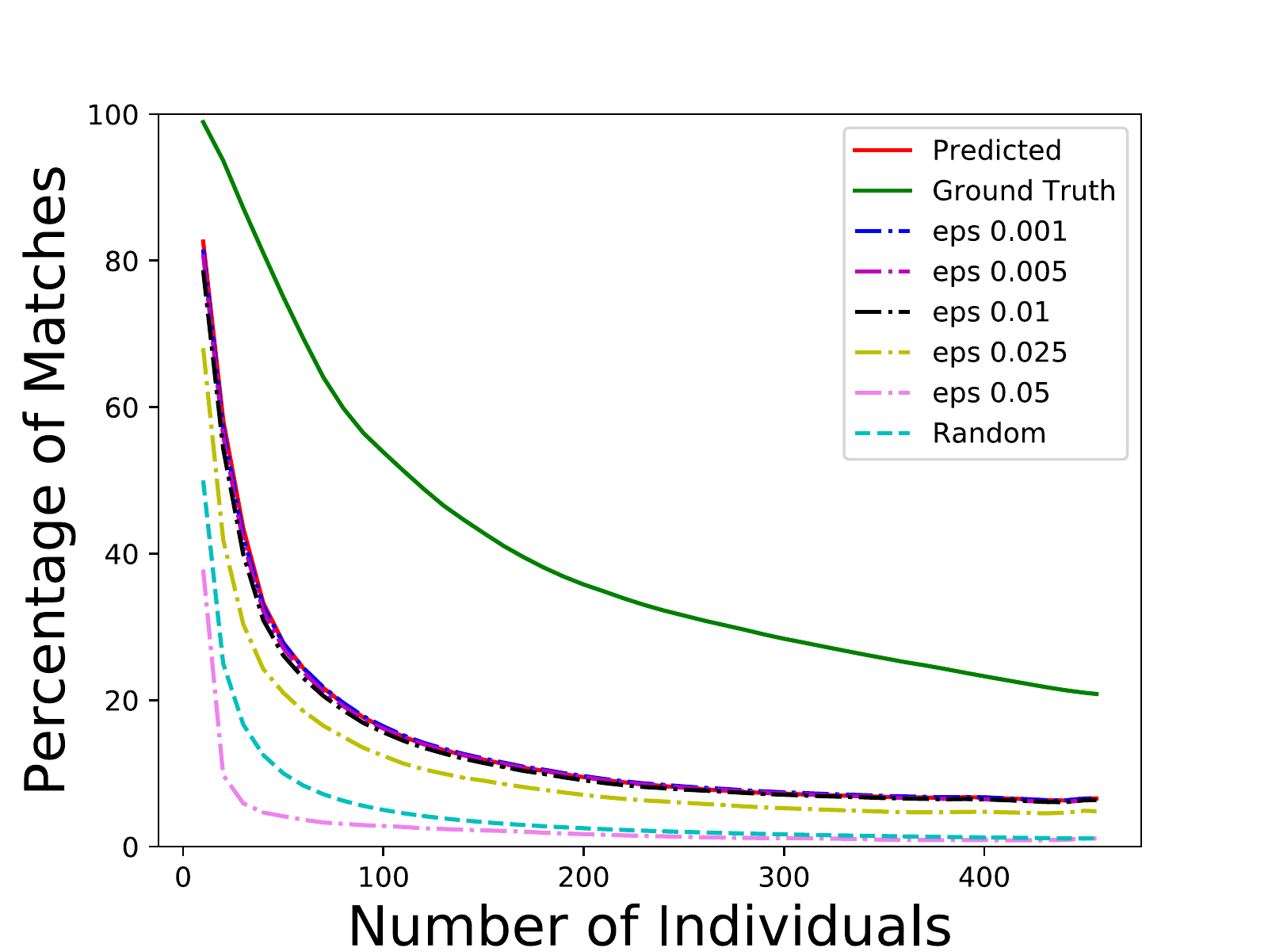}
    \caption{Top 5 - \emph{S-I}, Robust, Attacked}
\end{subfigure}
\begin{subfigure}[t]{0.32\textwidth}
    \includegraphics[width=5.6cm]{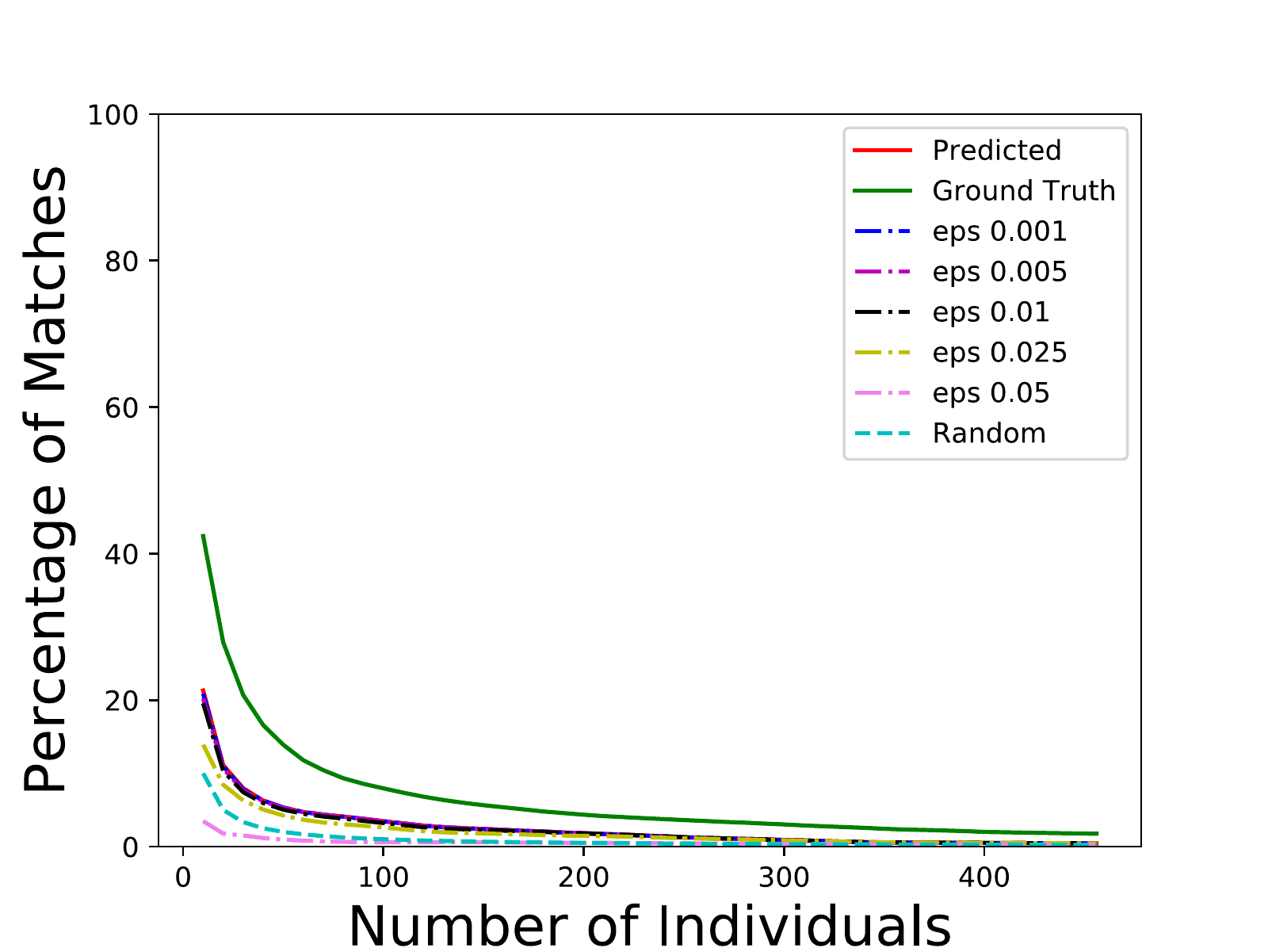}
    \caption{Top 1 - \emph{S-R}, Robust, Attacked}
\end{subfigure}
\begin{subfigure}[t]{0.32\textwidth}
    \includegraphics[width=5.6cm]{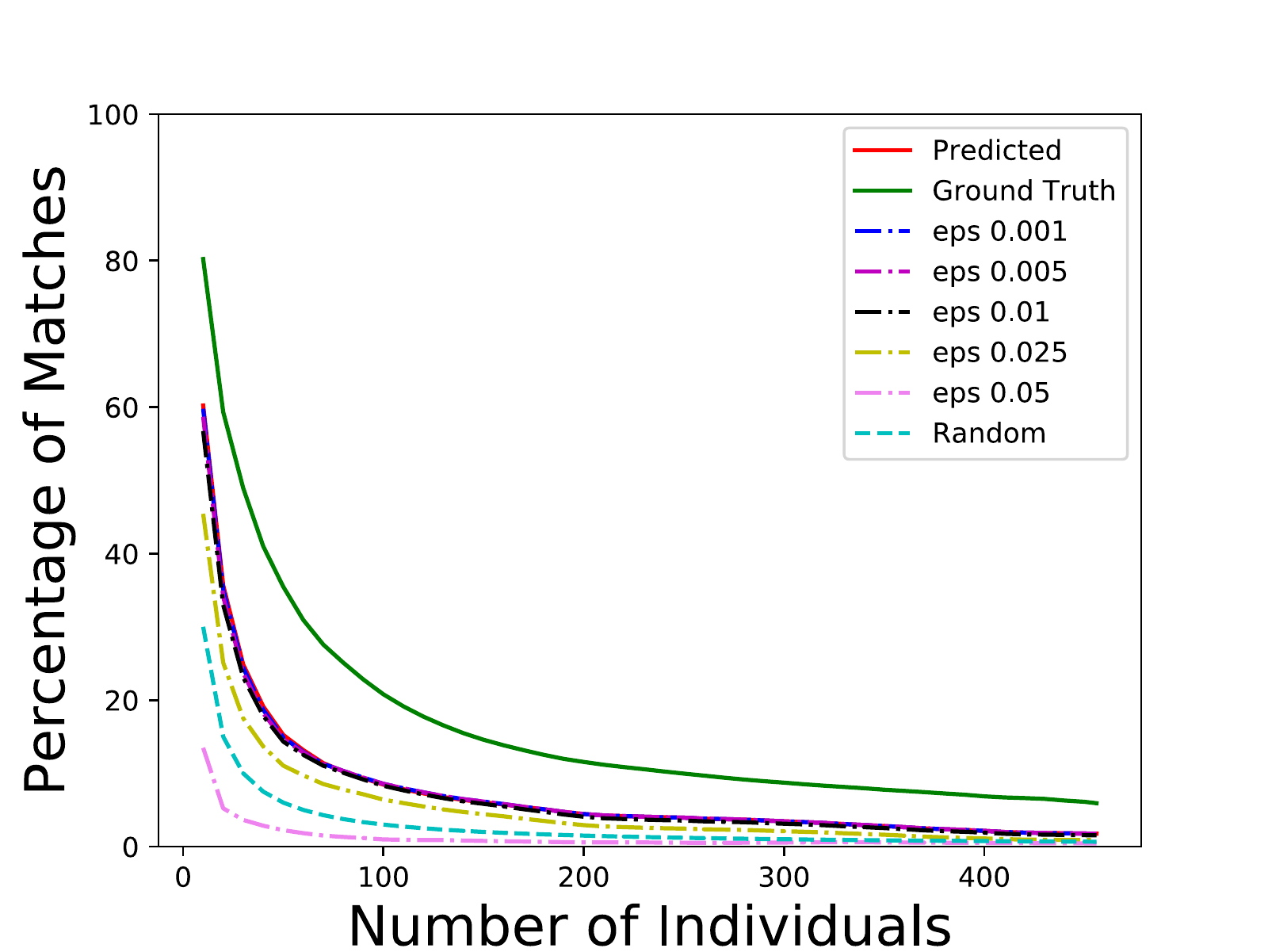}
    \caption{Top 3 - \emph{S-R}, Robust, Attacked}
\end{subfigure}
\begin{subfigure}[t]{0.32\textwidth}
    \includegraphics[width=5.6cm]{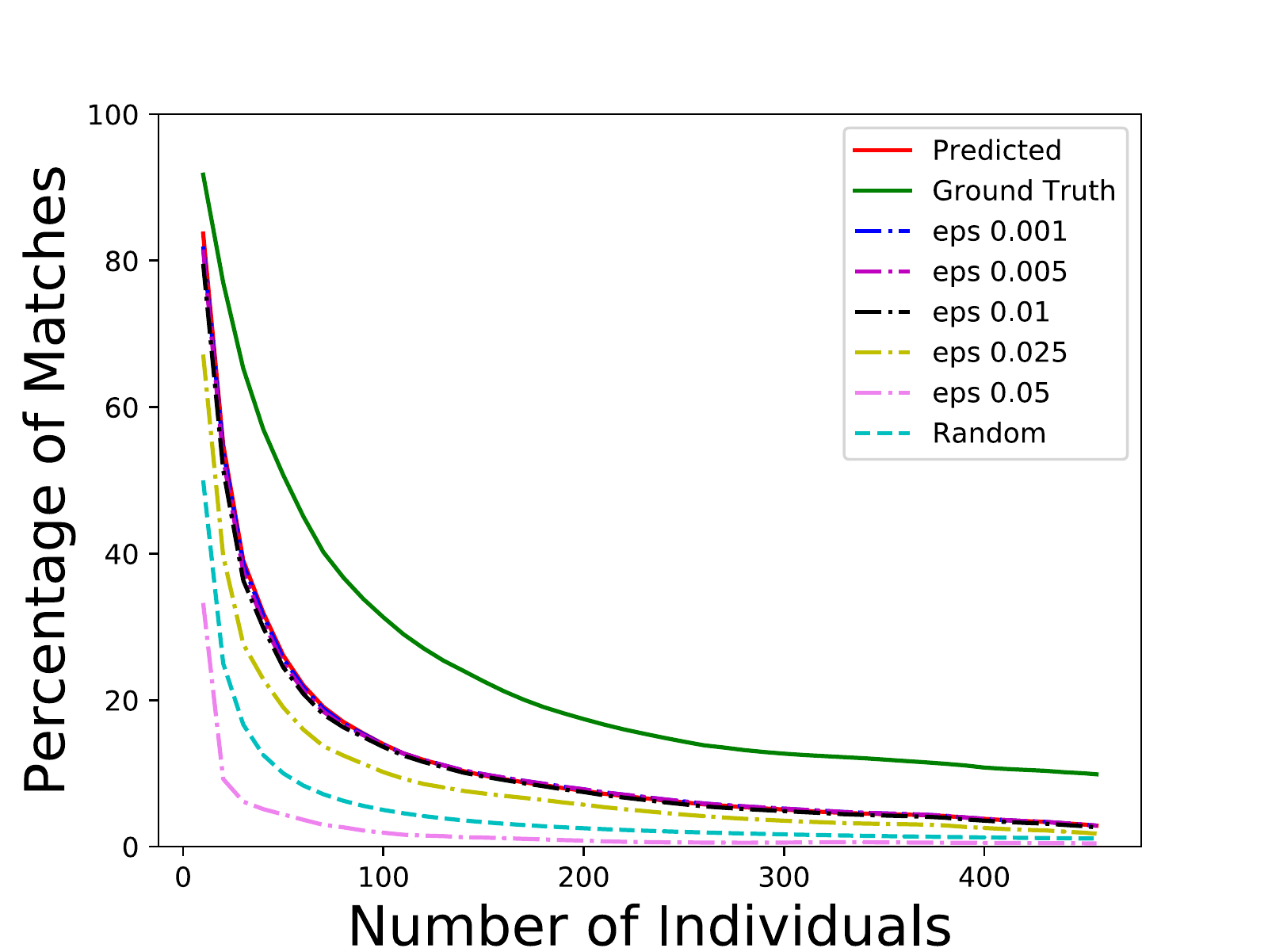}
    \caption{Top 5 - \emph{S-R}, Robust, Attacked}
\end{subfigure}
\caption{\textbf{(a)-(f)}: Matching accuracy with sex-classifiers adversarially trained at various values of $\epsilon$ but using clean images for the ideal (top row) and and realistic (bottom row) synthetic datasets, for top-$1$, top-$3$ and top-$5$. Yet again, we observe that training against a strong enough adversarial noise attack incurs a significant performance penalty on clean images, making retraining detrimental to the malicious actor attempting re-identification. \textbf{(g)-(l)}: Matching accuracy with a sex-classifier adversarially trained at $\epsilon=0.01$, attacked with adversarial images for various values of $\epsilon$, for the ideal (top row) and and realistic (bottom row) synthetic datasets, for top-$1$, top-$3$ and top-$5$. Similar to our results on the OpenSNP data, retraining boosts robustness to attacks of equal of lower strength as the retraining $\epsilon$, but fail to be robust to stronger attacks. However, in contrast to the OpenSNP data, to reduce accuracy below random in this scenario requires an attack with $\epsilon=0.05$, at which point adversarial noise starts to become visually evident.}
\label{fig:robust_adv_synth}
\end{figure}

\begin{figure}[]
\centering
\begin{subfigure}[t]{0.32\textwidth}
    \includegraphics[width=5.6cm]{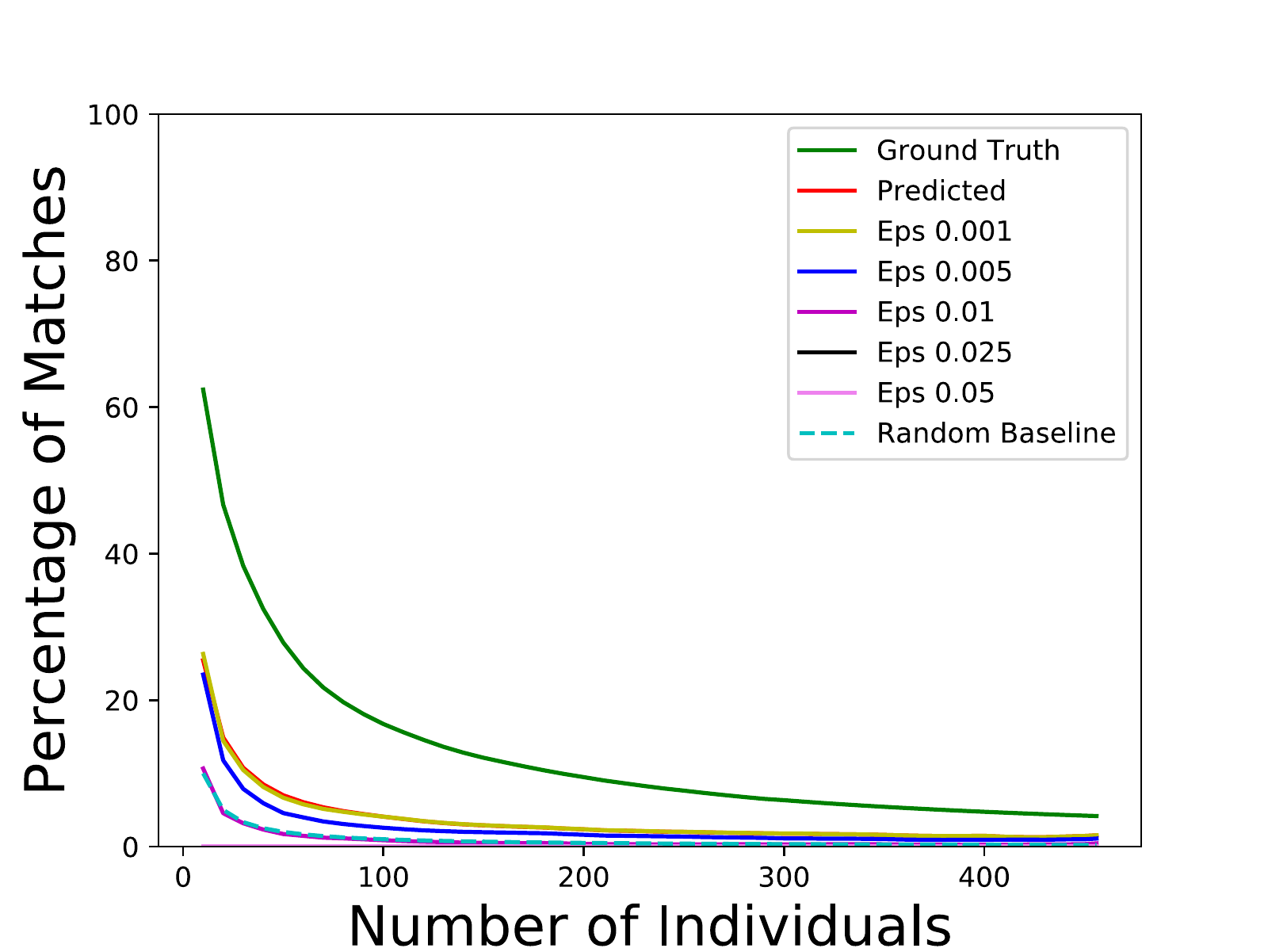}
    \caption{Top 1 - \emph{Synthetic - Ideal}}
\end{subfigure}
\begin{subfigure}[t]{0.32\textwidth}
    \includegraphics[width=5.6cm]{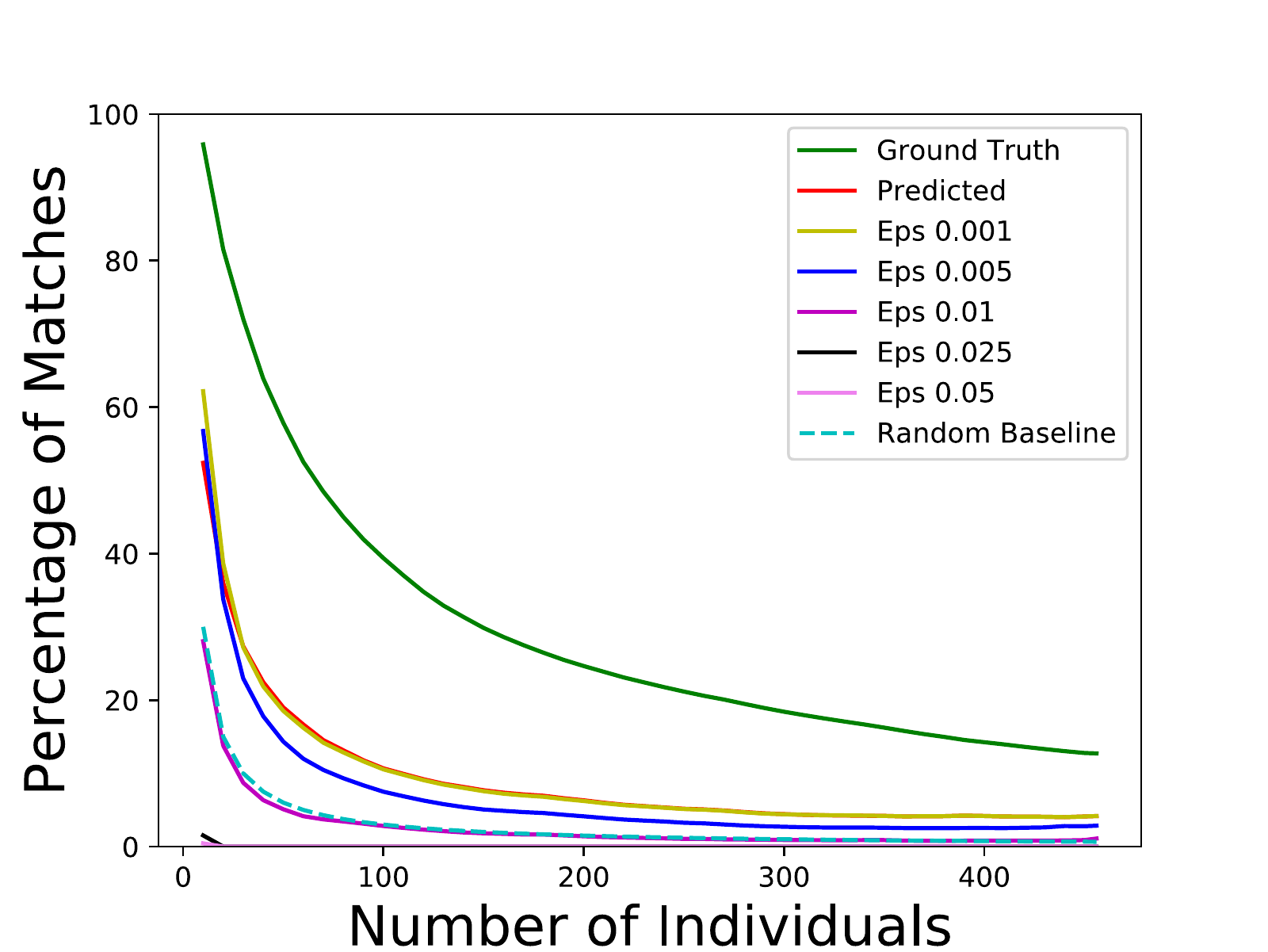}
    \caption{Top 3 - \emph{Synthetic - Ideal}}
\end{subfigure}
\begin{subfigure}[t]{0.32\textwidth}
    \includegraphics[width=5.6cm]{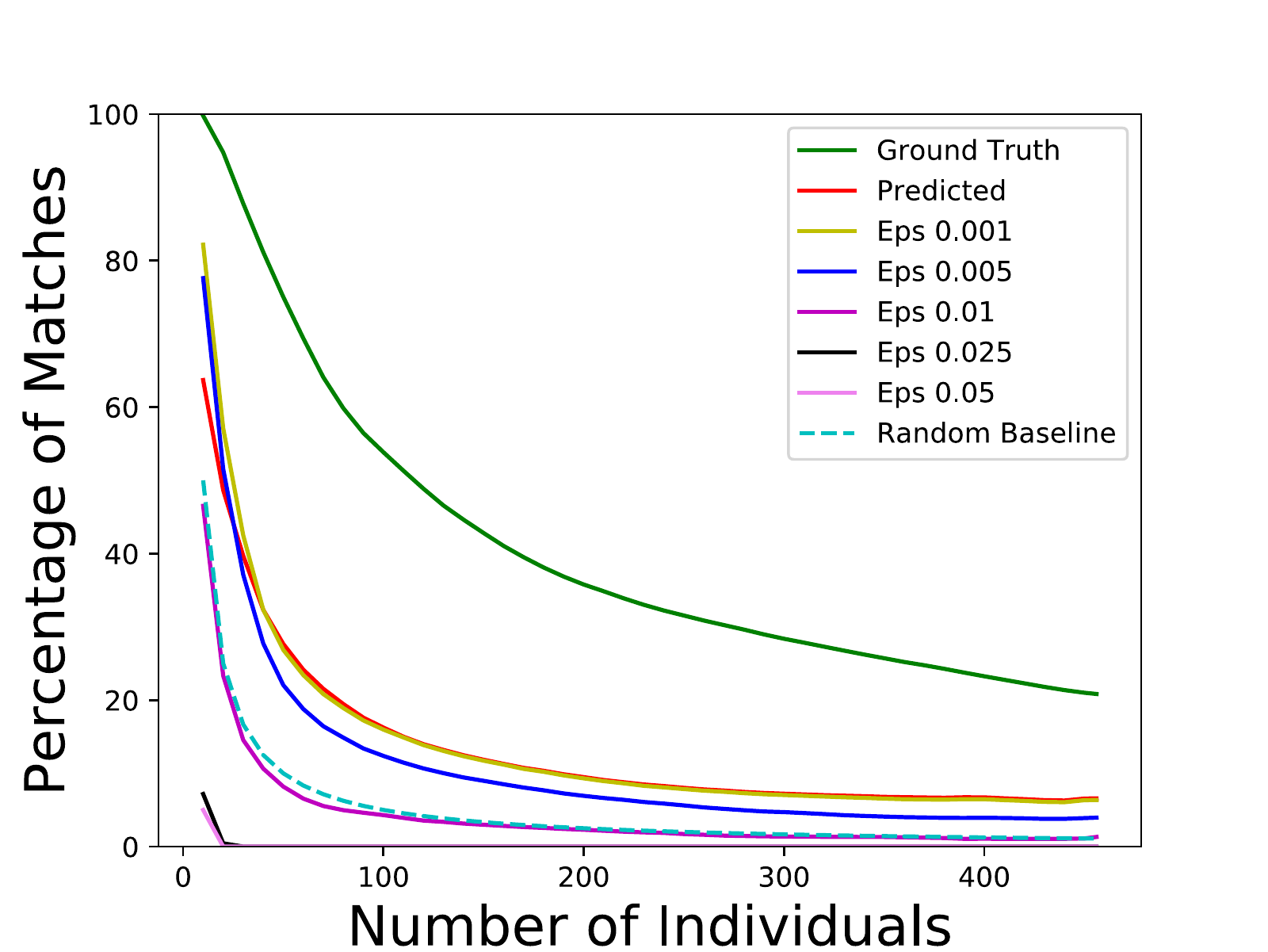}
    \caption{Top 5 - \emph{Synthetic - Ideal}}
\end{subfigure}
\begin{subfigure}[t]{0.32\textwidth}
    \includegraphics[width=5.6cm]{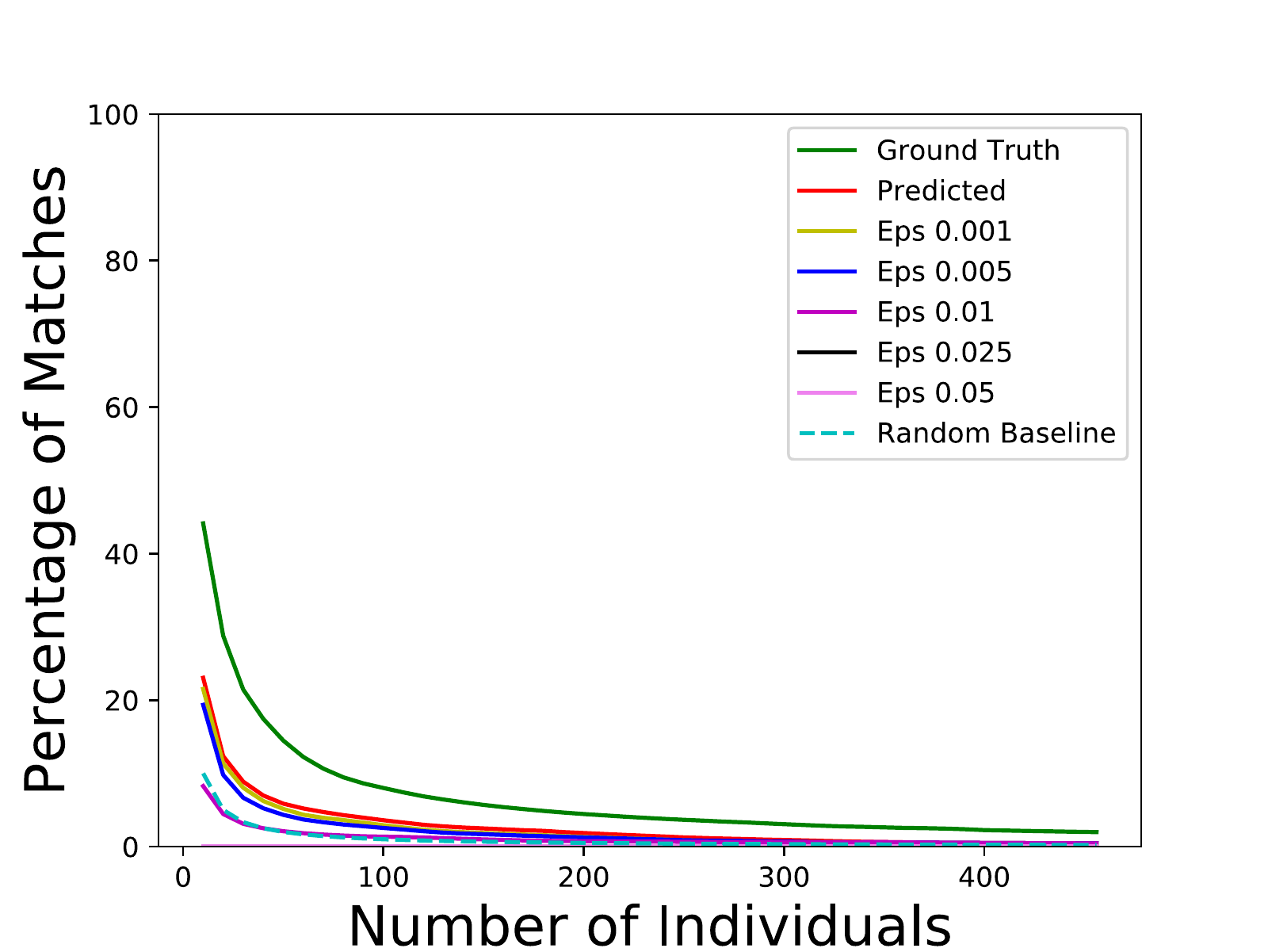}
    \caption{Top 1 - \emph{Synthetic - Realistic}}
\end{subfigure}
\begin{subfigure}[t]{0.32\textwidth}
    \includegraphics[width=5.6cm]{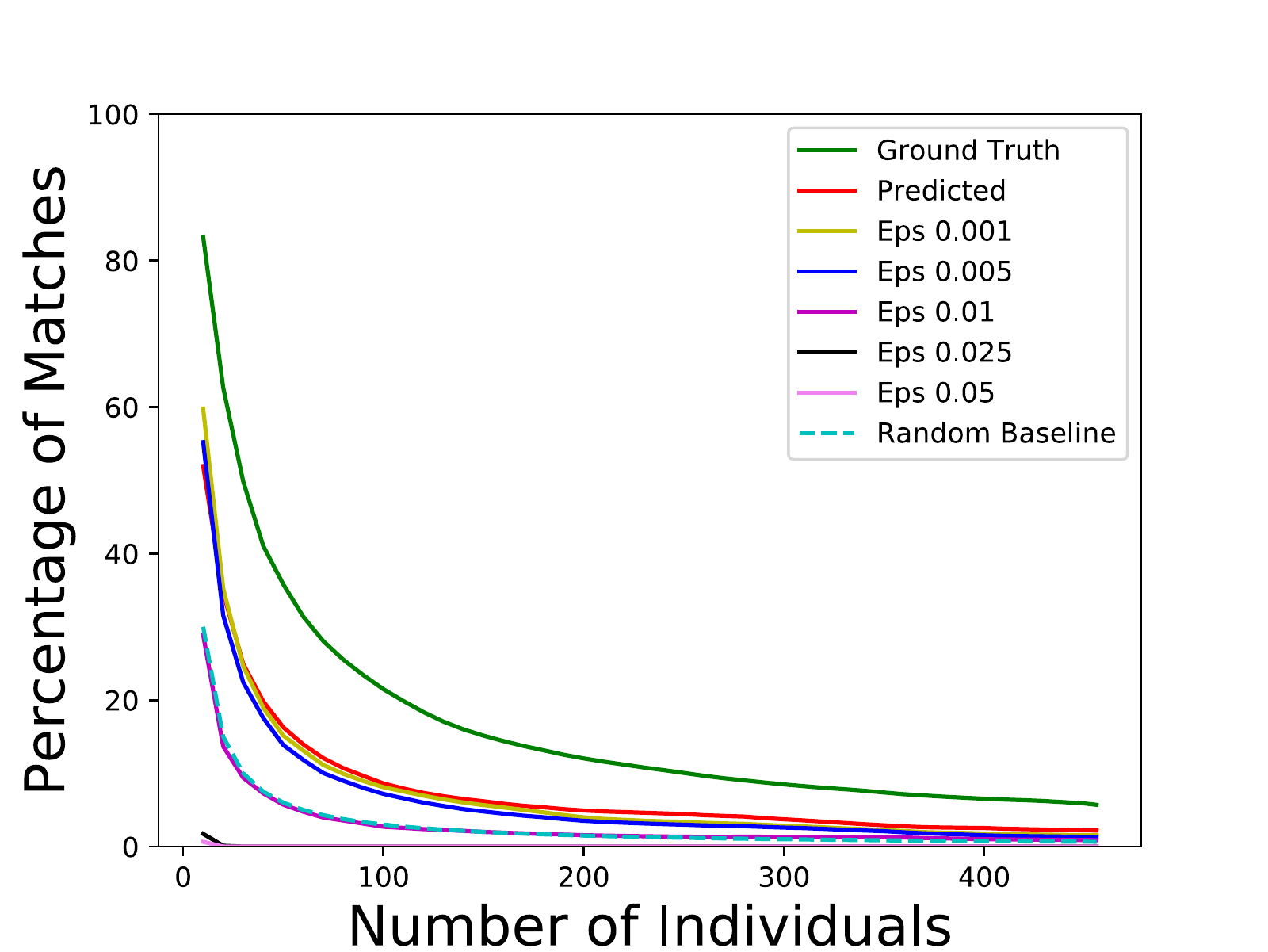}
    \caption{Top 3 - \emph{Synthetic - Realistic}}
\end{subfigure}
\begin{subfigure}[t]{0.32\textwidth}
    \includegraphics[width=5.6cm]{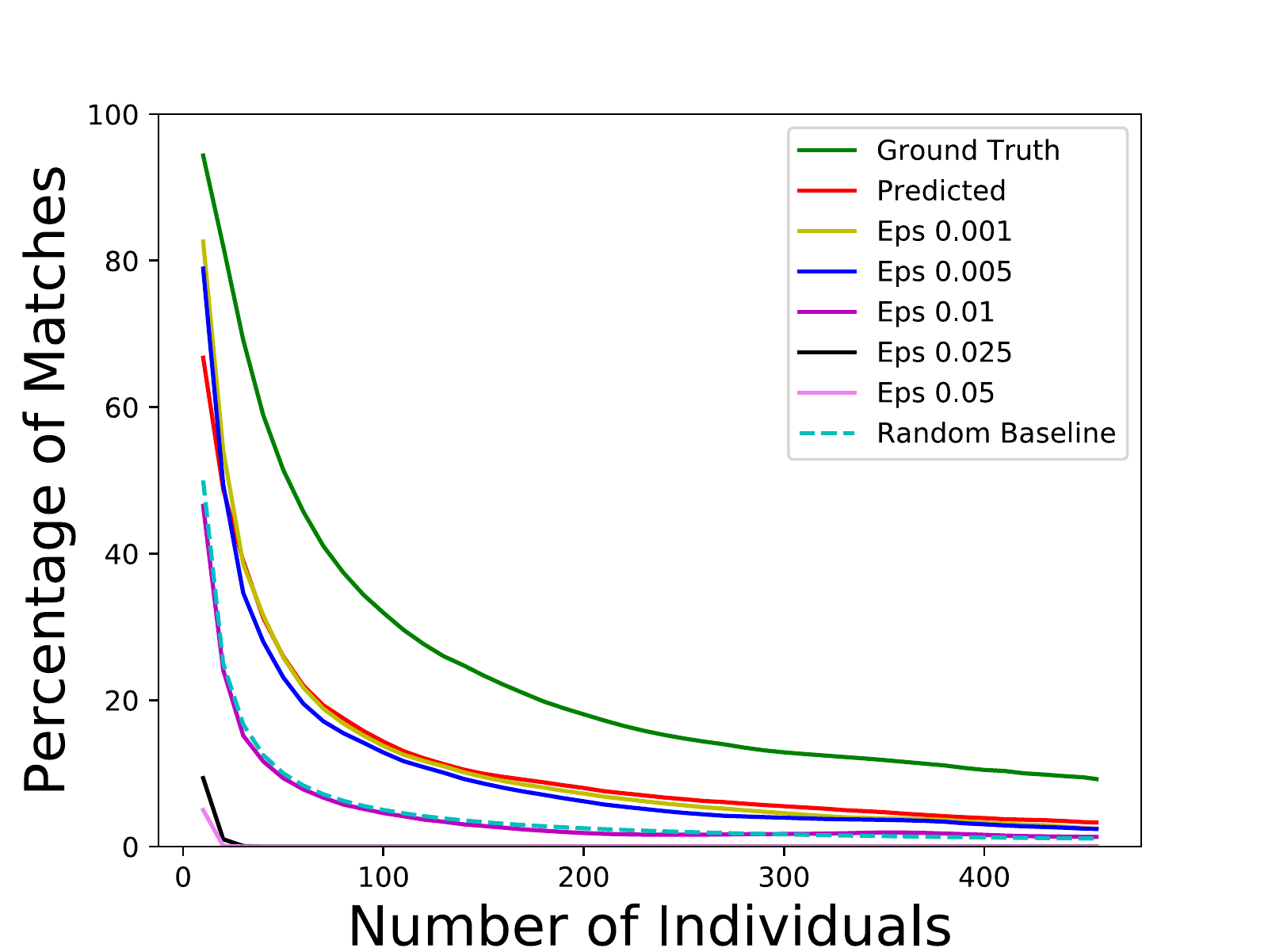}
    \caption{Top 5 - \emph{Synthetic - Realistic}}
\end{subfigure}
\caption{Matching Accuracy with images perturbed with the \emph{Universal Noise} attacking all phenotype-classifiers in parallel, for the Ideal (top row) and Realistic (bottom row) synthetic datasets for top-$1$, top-$3$ and top-$5$. Here, the effectiveness of this method relative to simply attacking sex is much more pronounced. For a very reasonable attack $\epsilon$ of $0.01$, the accuracy is reduced to near-random, and for attacks that are stronger, accuracy is reduced to zero, even for very small populations.}
\label{fig:univ_adv_synth}
\end{figure}

We run similar evaluations on these two synthetic datasets as we did on the $126$ OpenSNP individuals - matching accuracy (Supplementary Fig.~\ref{fig:adv_synth} (a)-(f)), attacking the sex classifier with PGD (Supplementary Fig.~\ref{fig:adv_synth} (g)-(l), we refrain from evaluating for other phenotypes as they proved to be largely ineffective on the OpenSNP data), baseline performance of an adversarially trained sex classifier (Supplementary Fig.~\ref{fig:robust_adv_synth} (a)-(f)), robustness of the adversarially trained sex classifier to perturbed images (Supplementary Fig.~\ref{fig:robust_adv_synth} (g)-(l)), and attacking with the \emph{Universal Noise} (Supplementary Fig.~\ref{fig:univ_adv_synth}). 

Next, we seek to explain the significantly larger gap between the accuracy of predicted matches and the accuracy with ground-truth phenotypes in case of the synthetic datasets. Experiments reveal that this behavior is rooted in the poor performance of our eye-color classification model, owing to both sparsity and low-quality of data available for training, and the fact that eye-color problem remains an open problem in computer vision - an especially hard one at that. Having evaluated multiple approaches (including segmentation of the eyes and classic machine learning methods with color histograms), we did not have much success in improving our predictions for this particular phenotype. The following are its implications to matching accuracy. We present as evidence two complementary sets of results - matching accuracy when eye color is entirely ignored, and matching accuracy when predicted variants from images are used for all phenotypes but for eye color, where we swap in favor of the ground-truth variants instead.

Results in Supplementary Fig.~\ref{fig:swapeye} (a)-(d) confirm our suspicions. Ground truth accuracy drops significantly when eye-color is entirely disregarded, signalling the importance of the phenotype in matching, while our predicted accuracy slightly increases upon disregarding eye-color, signalling high volume of noise in our predictions. Subsequently, replacing our eye-color predictions with ground truth values produce accuracies that are nearly the ground-truth upper bound, strongly indicating our eye-color prediction models act as a bottleneck in the matching pipeline. 

\begin{figure}[]
\centering
\begin{subfigure}[t]{0.36\textwidth}
    \includegraphics[width=5.6cm]{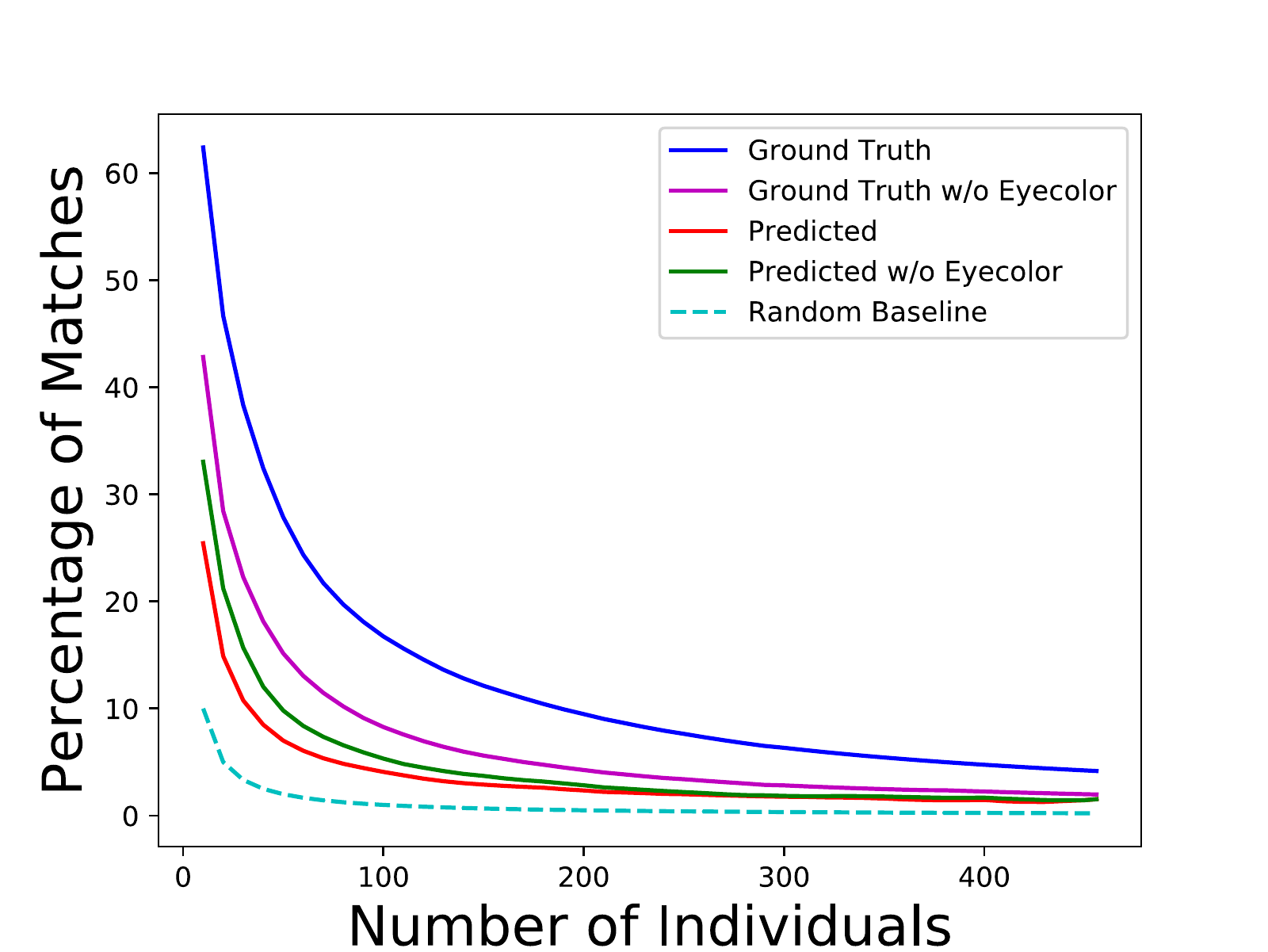}
    \caption{\emph{Synthetic-Ideal} without eye color}
\end{subfigure}
\begin{subfigure}[t]{0.36\textwidth}
    \includegraphics[width=5.6cm]{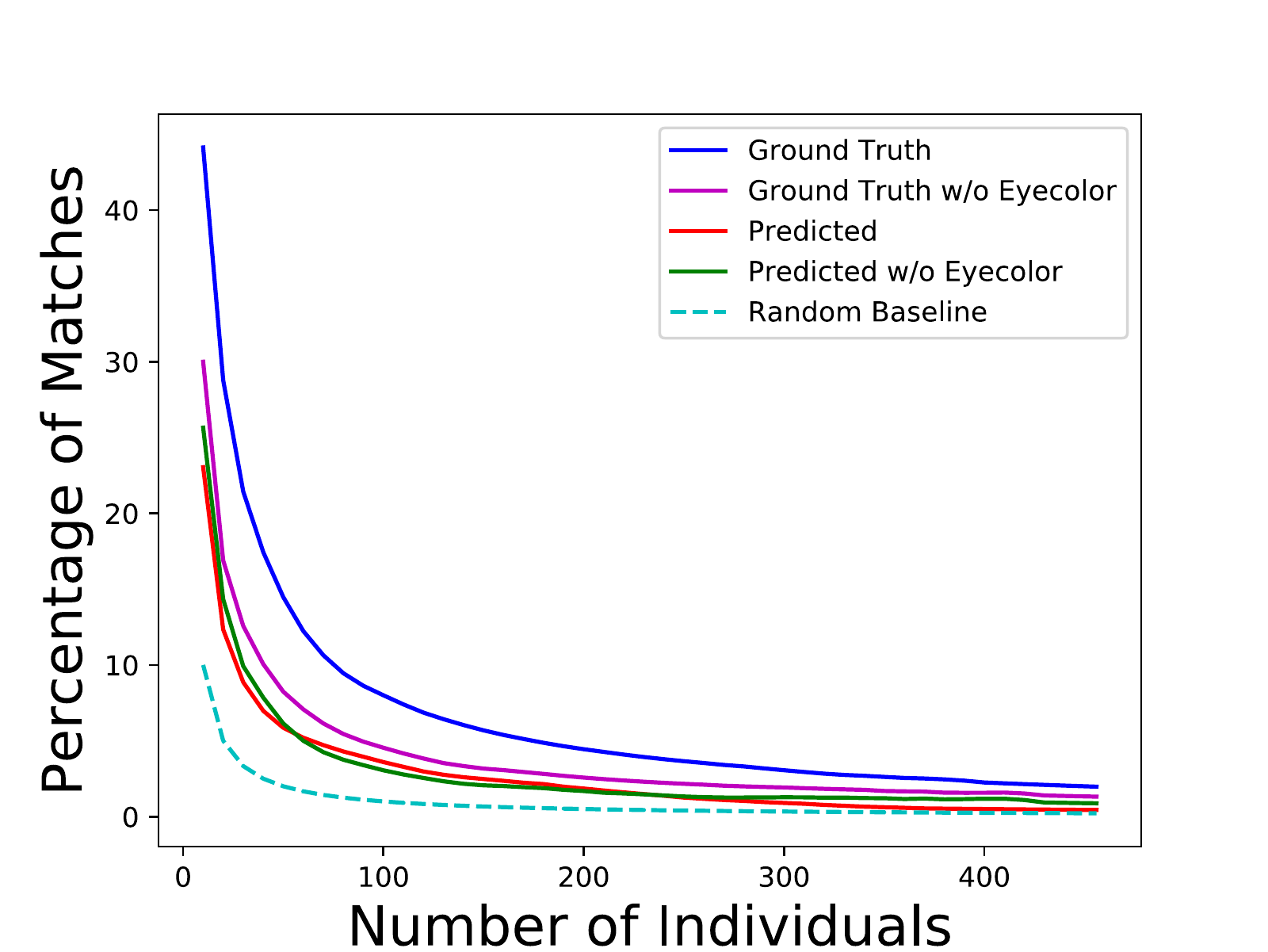}
    \caption{\emph{Synthetic-Real} without eye color}
\end{subfigure}

\begin{subfigure}[t]{0.36\textwidth}
    \includegraphics[width=5.6cm]{synthetic/ideal_swapeye_1_svg-tex.pdf}
    \caption{\emph{Synthetic-Ideal}, ground-truth eye color}
\end{subfigure}
\begin{subfigure}[t]{0.36\textwidth}
    \includegraphics[width=5.6cm]{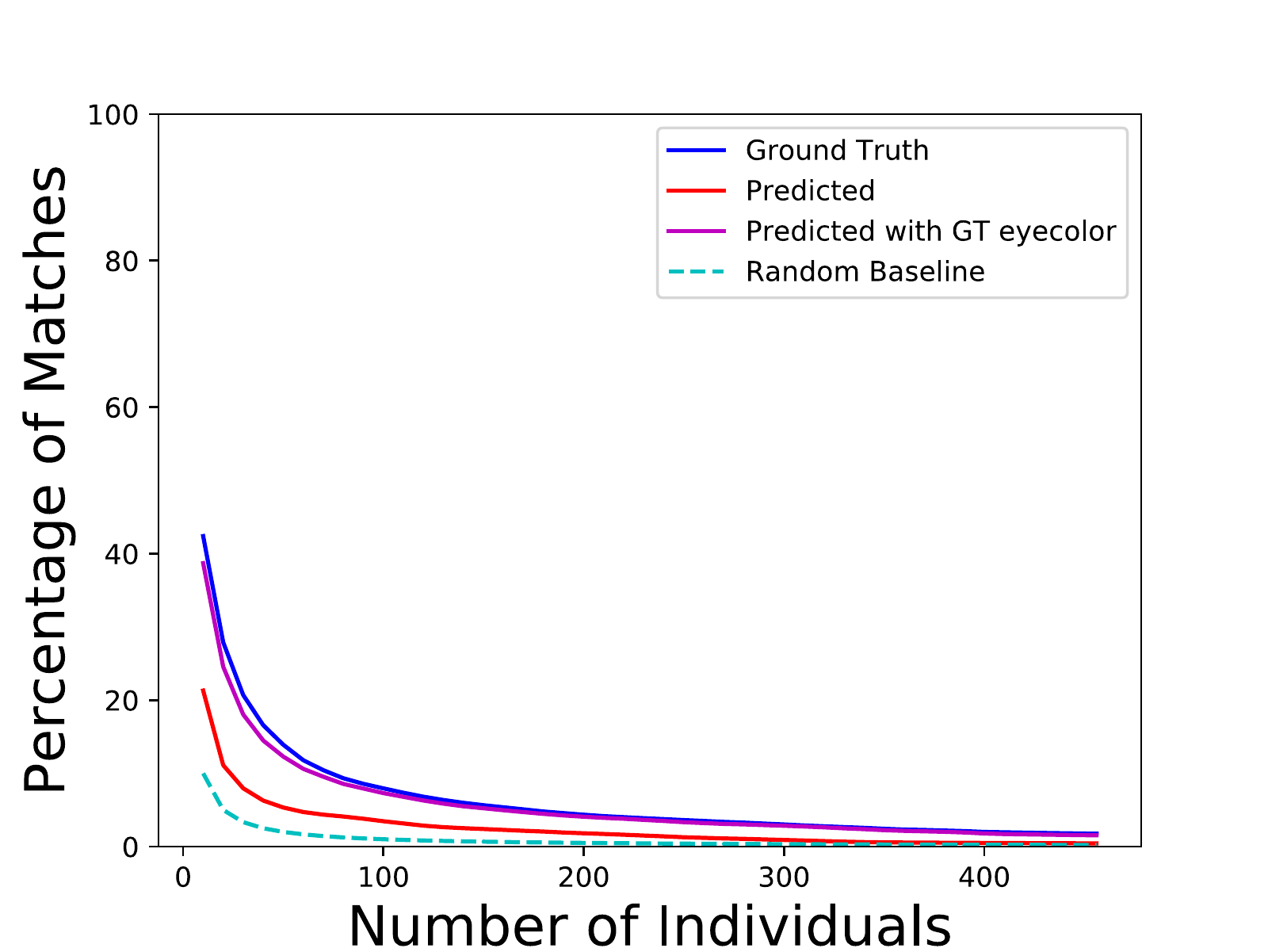}
    \caption{\emph{Synthetic-Real}, ground-truth eye color}
\end{subfigure}

\begin{subfigure}[t]{0.32\textwidth}
    \includegraphics[width=5.5cm]{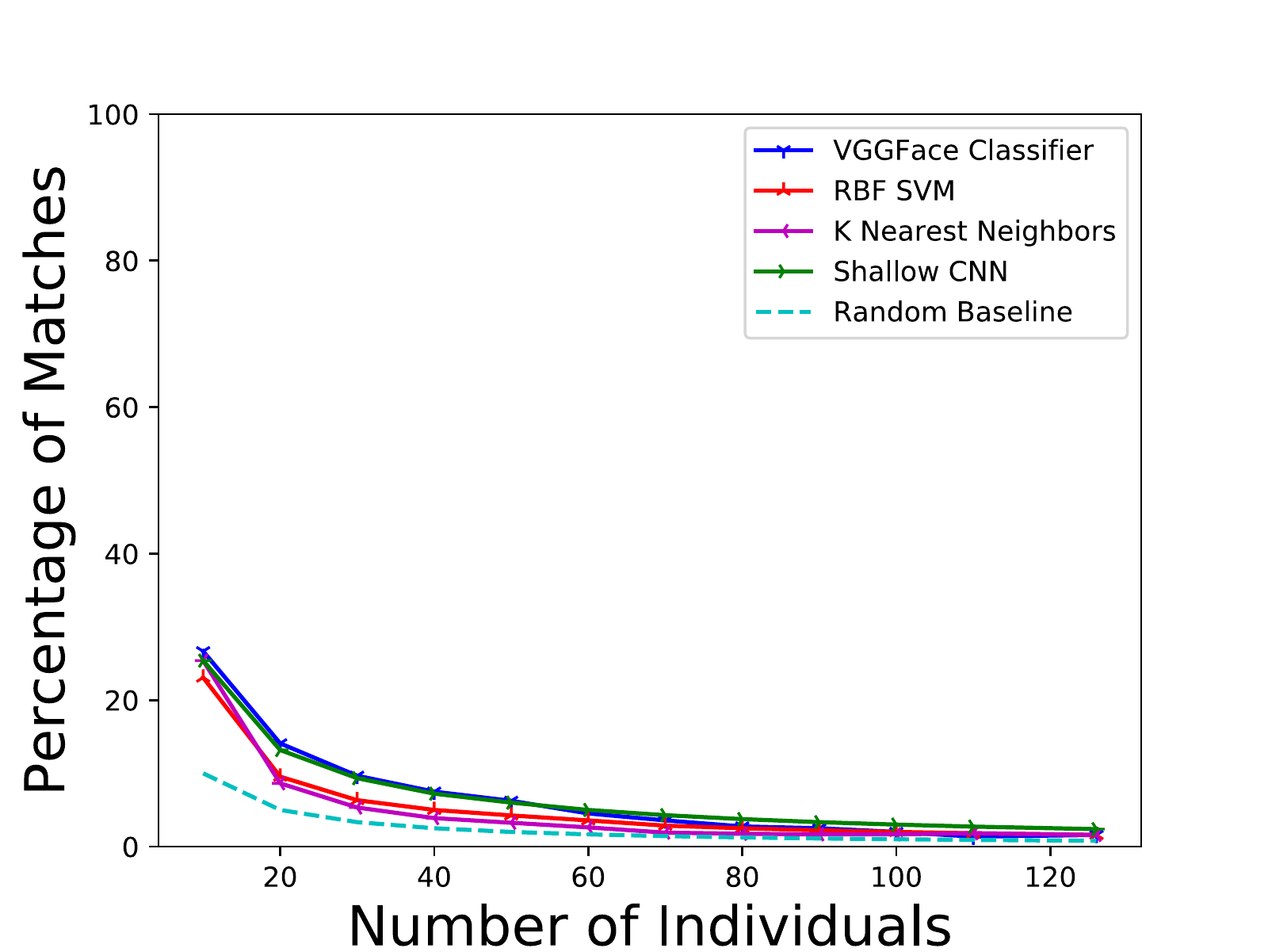}
    \caption{Top 1}
\end{subfigure}
\begin{subfigure}[t]{0.32\textwidth}
    \includegraphics[width=5.5cm]{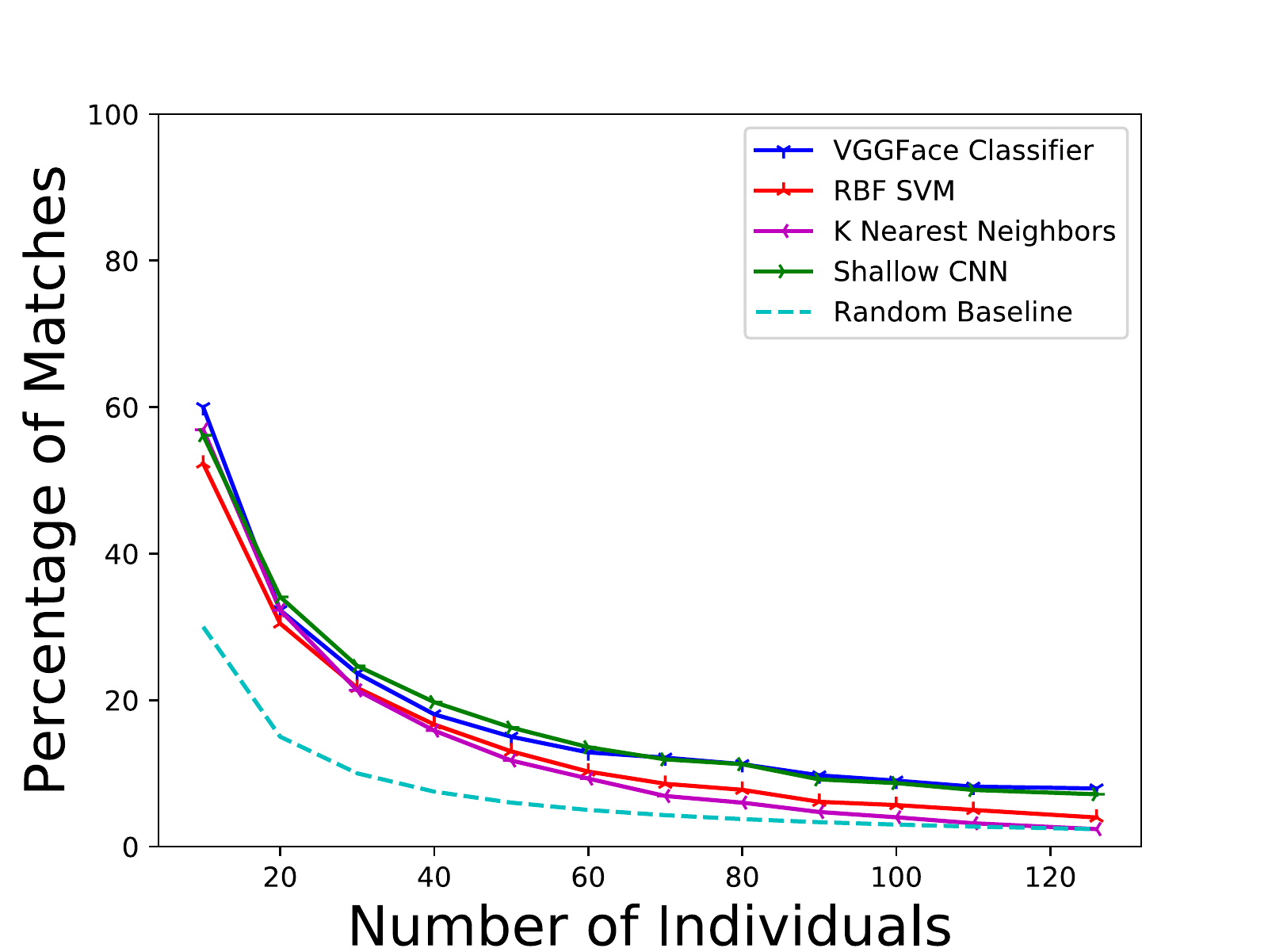}
    \caption{Top 3}
\end{subfigure}
\begin{subfigure}[t]{0.32\textwidth}
    \includegraphics[width=5.5cm]{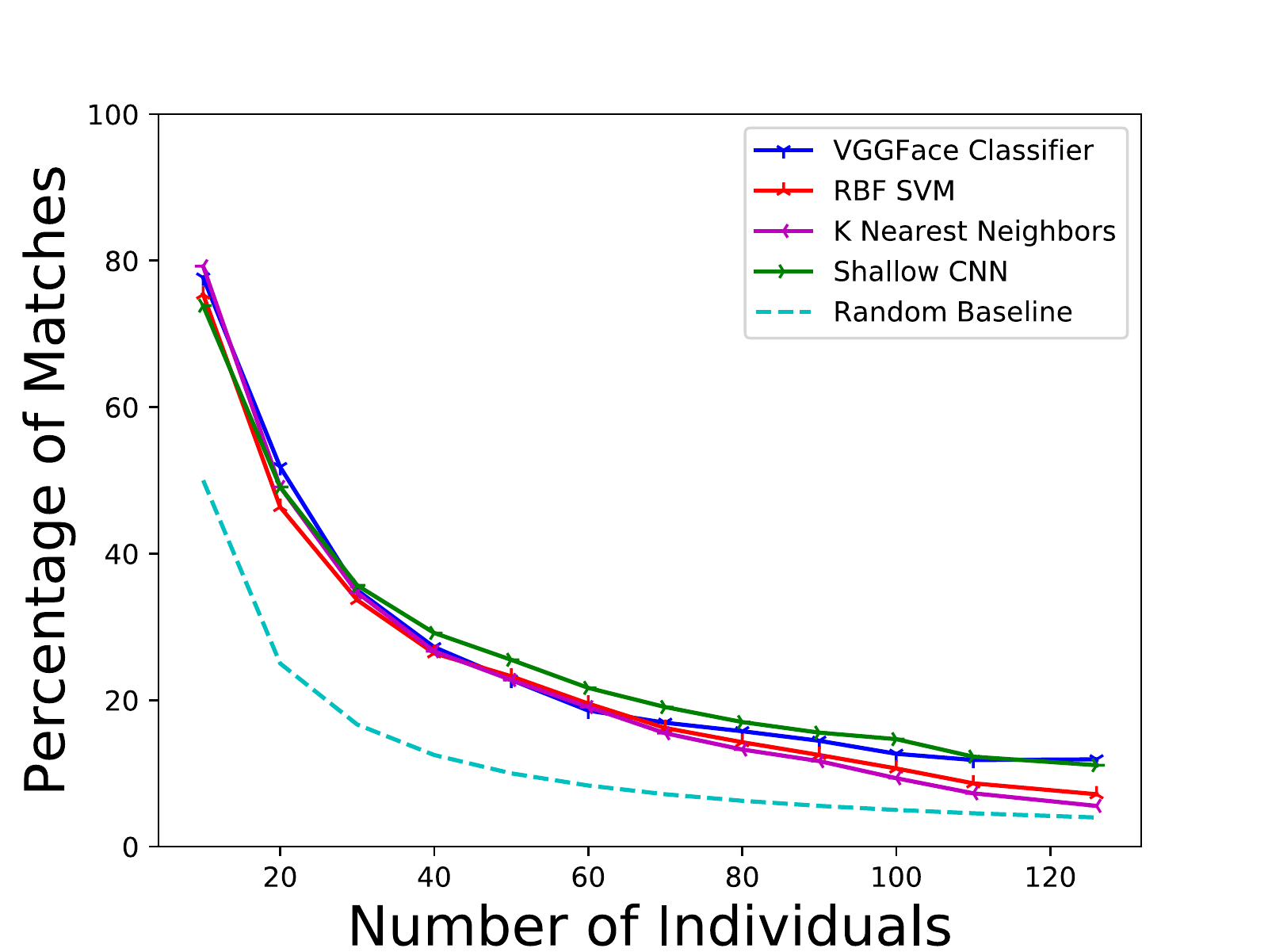}
    \caption{Top 5}
\end{subfigure}
\caption{\textbf{(a)-(b)}: Matching accuracies with and without considering the eye-color phenotype, both predicted and ground-truth for a) Ideal and b) Realistic synthetic datasets. Notice the significant drop in ground-truth accuracy when eye-color is disregarded entirely. This points to the high importance of eye-color in matching. At the same time, notice that in the ideal dataset, disregarding eye-color entirely produces \emph{better} matching accuracy than when including it, although it does not make a significant difference to the realistic dataset. This points to the presence of significant noise in our eye-color predictions. \textbf{(c)-(d)}: Matching accuracies when the eye-color predictions from images are replaced by ground-truth values, for c) Ideal and d) Realistic synthetic datasets. The significant increase in matching accuracy (nearly the upper bound) in both synthetic datasets strongly suggests that we are limited in our matching ability by the poor performance of eye-color prediction. \textbf{(e)-(g)}: Top-$1$, Top-$3$ and Top-$5$ matching accuracies with various eye-color prediction techniques. None of them are particularly effective.}
\label{fig:eyecomp}
\label{fig:swapeye}
\end{figure}

To be thorough, we try various techniques to try and solve this issue, ranging from conventional machine learning to neural networks. Because eyes are a small fraction of a face image, it could be very likely that the rest of the image makes eyecolor prediction harder. To understand the impact of this, we segment the eyes from each image using Multitask Cascade CNNs \cite{7553523}, and use each approach on the segmented eyes (with the exception of VGGFace, which was designed specifically to work on full face images). From conventional machine learning, we report results using K-Nearest-Neighbors and a Support Vector Machine with the rbf kernel, which outperform others with test accuracies of $57.89\%$ and $51.75\%$ respectively. We also build a shallow convolutional neural network, whose accuracy seems to be high at $60\%$. Unfortunately, this CNN converges to always predicting the majority class (Brown eyes), leading to its seemingly high performance. By contrast, the VGGFace classifier achieves a test accuracy of $59\%$.

Supplementary Fig.~\ref{fig:eyecomp} (e)-(g) show reidentification performance when using each approach to predict eye color. None of the approaches are particularly different from the VGGFace classifier. Future advances in computer vision could solve this problem, proposing an increased risk of re-identification in the wild.

\bibliography{sample}

\end{document}